\documentclass[]{academic_template}

\usepackage[toc,page,header]{appendix}

\usepackage{amssymb}
\usepackage{amsmath}
\usepackage{array}
\usepackage{pifont}
\usepackage{textcomp}
\usepackage{verbatim}

\usepackage{makecell}
\usepackage{algorithm}      
\usepackage{algpseudocode}  

\usepackage{graphicx}

\usepackage{listings}
\usepackage{url}
\usepackage{booktabs}
\usepackage{wrapfig}

\usepackage{colortbl}
\usepackage{xcolor}

\usepackage{caption}
\usepackage{subcaption}
\usepackage{marvosym}

\DeclareTextFontCommand{\textbf}{\bfseries\sffamily}

\newsavebox{\RISEScoreBox}
\newcommand{\RISEFitScore}[1]{%
  \sbox{\RISEScoreBox}{#1}%
  \ifdim\wd\RISEScoreBox>\linewidth
    \resizebox{\linewidth}{!}{\usebox{\RISEScoreBox}}%
  \else
    \usebox{\RISEScoreBox}%
  \fi
}

\definecolor{RISEGroup}{RGB}{239,239,239}
\definecolor{RISEGreen}{RGB}{31,128,79}
\definecolor{RISERed}{RGB}{187,65,65}
\newcommand{\RISEyes}{\textcolor{RISEGreen}{\ding{51}}}
\newcommand{\RISEno}{\textcolor{RISERed}{\ding{55}}}
\newcommand{\RISEnum}[1]{\makebox[2em][c]{#1}}

\newcommand{\IEEEPARstart}[2]{#1#2}
\usepackage{CJK}

\AtBeginEnvironment{figure*}{%
  \ifcsdef{RISECenterFigureCaption}{%
    \renewcommand{\RISECenterFigureCaption}{}%
  }{}%
}

\AddToHook{file/judge_spatial_cons.tex/before}{%
  \begingroup
  \patchcmd{\RISEJudgeBox}{\setlength{\parskip}{2pt}}%
    {\setlength{\parskip}{0.7pt}}{}%
    {\PackageError{RISE}{Could not adapt the spatial prompt spacing}{Check prompt_style.tex.}}%
}
\AddToHook{file/judge_spatial_cons.tex/after}{\endgroup}

\renewcommand{\affiliationformat}[2][]{%
  {\affiliationfont\sffamily $^{#1}$#2}%
}

\setlogoheight{14mm}   
\setlogospacing{5mm}
\setsjtublue 

\makeatletter
\fancypagestyle{firststyle}{
    \fancyhead[L]{%
        \raisebox{\dimexpr-\height+\logoverticalshift\relax}[0pt][0pt]{%
            \includegraphics[height=\logoheight]{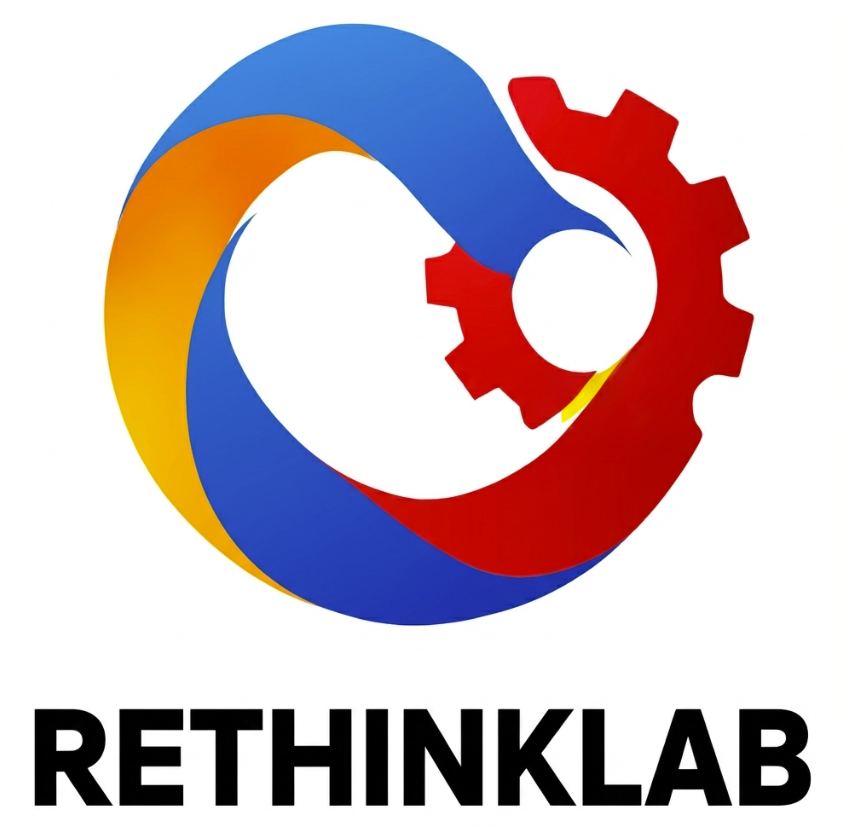}%
        }%
        \hspace{\logospacing}%
        \raisebox{\dimexpr-\height+\logoverticalshift\relax}[0pt][0pt]{%
            \includegraphics[height=\logoheight]{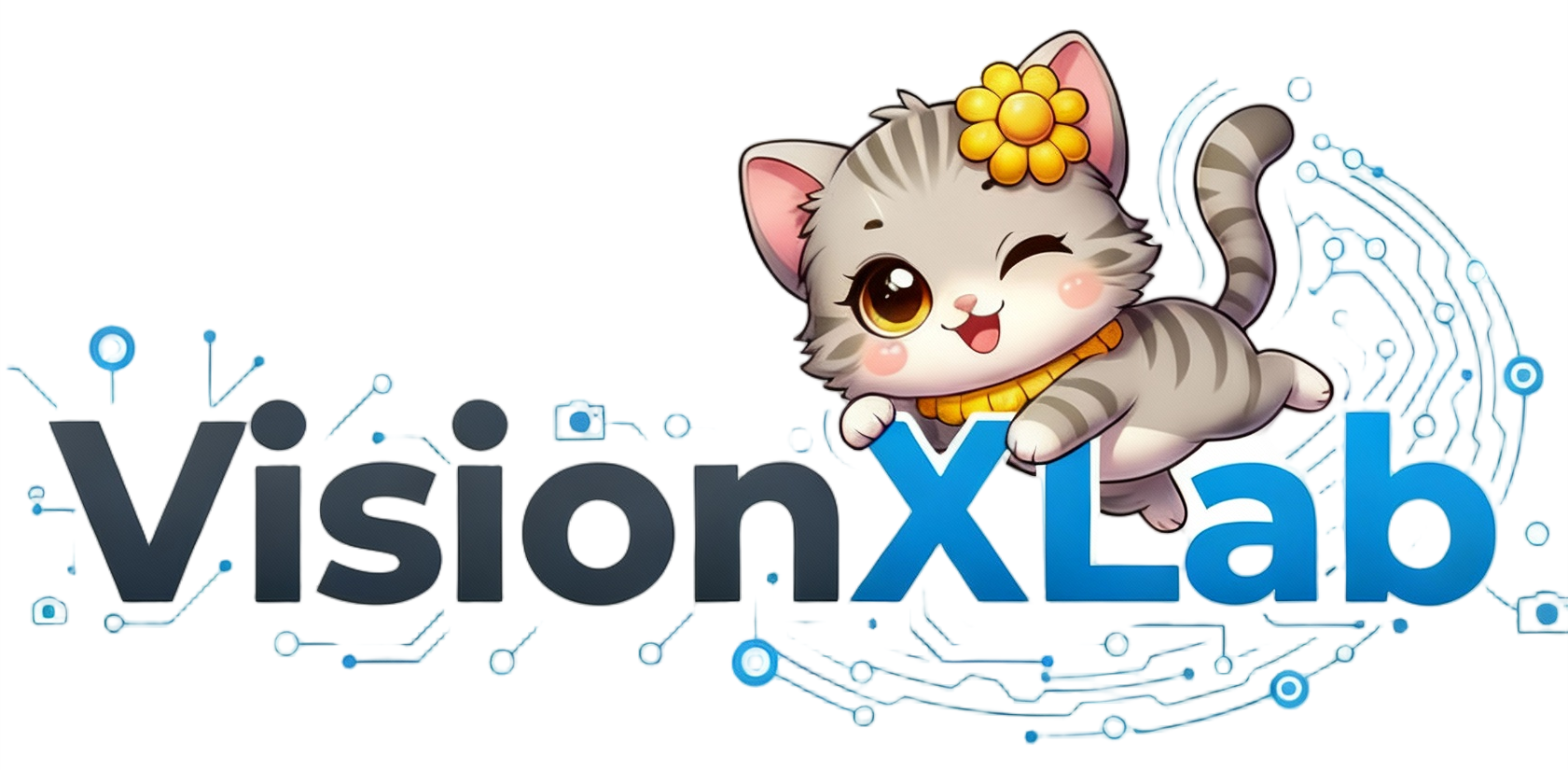}%
        }%
    }
    \fancyhead[R]{%
        \raisebox{\dimexpr-\height+\logoverticalshift\relax}[0pt][0pt]{%
            \if@sjtublue
                \includegraphics[height=\logoheight]{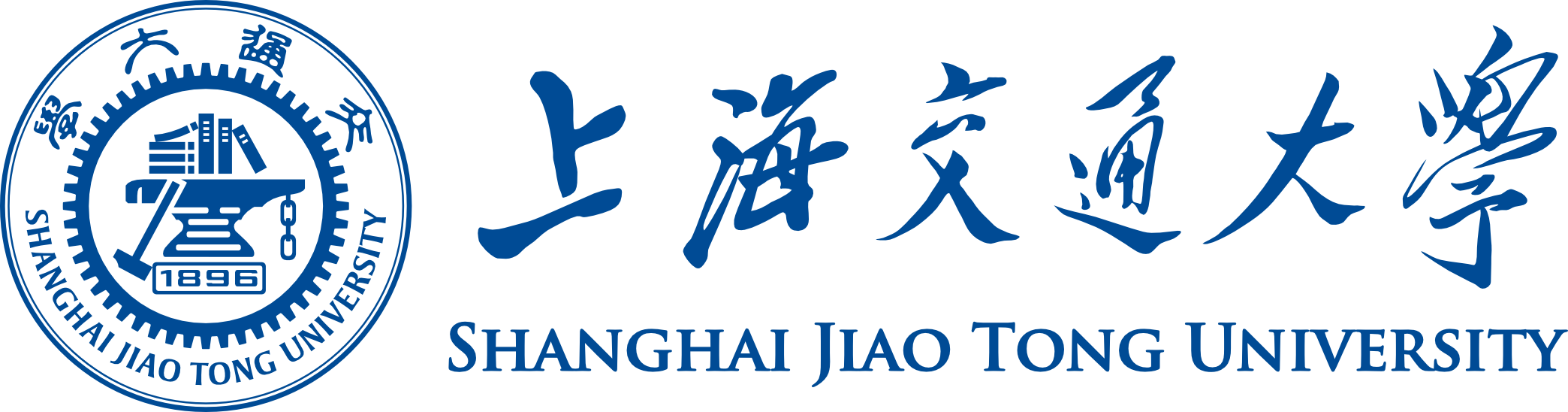}%
            \else
                \includegraphics[height=\logoheight]{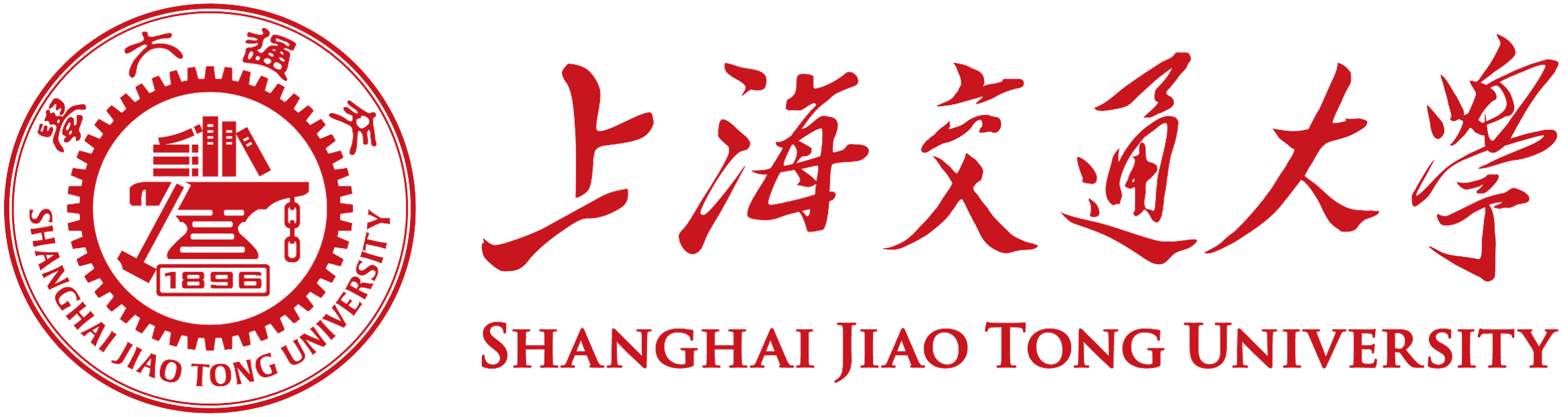}%
            \fi
        }%
    }
}
\makeatother

\settitlerulethickness{3pt}  

\abstractboxon 
\setabstractframecolor{gray} 
\setabstractbgcolor{gray!10} 
\setlogotolineshift{5mm} 

\settoprulethickness{2.5pt}  
\setbottomrulethickness{1.5pt}

\title{Reasoning-Informed Visual Editing}

\author[1,\mbox{\Letter}]{Xue Yang}
\author[1,*]{Peiyuan Zhang}
\author[2,*]{Yilun Zhu}
\author[1,*]{Qihao Yang}
\author[1]{Mingxin Liu}
\author[1]{Xiangyu Zhao}
\author[3]{Ziqian Fan}
\author[1]{Zhaokai Wang}
\author[1]{Yan Li}
\author[4,1]{Yifan Yang}
\author[2]{Xu Yang}
\author[5]{Xiaosong Jia}
\author[6]{Yue Zhou}
\author[1]{Zhihang Zhong}
\author[1]{Junchi Yan}

\affiliation[1]{Shanghai Jiao Tong University}
\affiliation[2]{Southeast University}
\affiliation[3]{South China University of Technology}
\affiliation[4]{Microsoft Research Asia}
\affiliation[5]{Fudan University}
\affiliation[6]{East China Normal University}

\contribution[\mbox{\Letter}]{Corresponding author}
\contribution[*]{Core student contributors}

\abstract{%
  \begingroup
  \renewenvironment{abstract}{}{}%
  \begin{abstract}
Large Multi-modality Models (LMMs) have made significant progress in visual understanding and generation, but they still face challenges in general visual editing, particularly in following complex instructions, preserving appearance consistency, and supporting flexible input formats. To study this gap, we introduce RISEBench, the first benchmark for evaluating Reasoning-Informed viSual Editing (RISE), and further extend it to RISEBench++, a more comprehensive and fine-grained benchmark for this emerging task. RISEBench++ broadens the taxonomy into a hierarchical scheme spanning six reasoning dimensions: \emph{Temporal}, \emph{Causal}, \emph{Spatial}, \emph{Logical}, and \emph{Counterfactual Reasoning}, together with \emph{Hybrid Reasoning} that integrates multiple reasoning types across multi-turn edits. These dimensions are further decomposed into 12 subcategories and 65 fine-grained task types. We expand the supported input formats to include multi-image conditioning and scale the benchmark to 1,000 carefully human-annotated test cases, each released bilingually in English and Chinese. We also improve our evaluation framework, which assesses \emph{Instruction Reasoning}, \emph{Appearance Consistency}, and \emph{Visual Plausibility} with both human judges and the LMM-as-a-judge approach, enabling more reliable and better-calibrated judgements. Beyond benchmarking, we introduce RISE-Agent, a training-free agentic framework integrating reasoning-driven planning, tool-augmented execution, and verifier-guided refinement, outperforming most strong existing approaches across diverse RISE tasks. We conduct extensive experiments on 58 visual editing approaches, covering 34 open-source models, 19 closed-source models, and 5 agentic methods. The evaluation results reveal substantial challenges in reasoning-based visual editing, with even the strongest evaluated approach, GPT-Image-2.5 Sunburst, achieving only 56.6\% accuracy. RISEBench++ effectively highlights the limitations of contemporary editing models, provides valuable insights, and indicates potential future directions for the field of reasoning-aware visual editing.
\end{abstract}%
  \endgroup
}

\date{\today}

\checkdata[Code]{\url{https://github.com/VisionXLab/RISEBench}}
\checkdata[Dataset]{\url{https://huggingface.co/collections/VisionXLab/rise}}
\begin{document}
\maketitle


\section{Introduction}
\IEEEPARstart{L}{arge} Multi-Modality Models (LMMs) have achieved remarkable progress in both visual understanding~\cite{liu2023visual,bai2023qwen,chen2024far,team2023gemini,luo2024mono} and visual generation~\cite{podell2023sdxl,rombach2022high,betker2023improving}. 
Meanwhile, significant efforts~\cite{team2024chameleon,xie2024show,wang2024emu3,wu2024janus,deng2025bagel} have been dedicated to unifying these two tasks, 
with the goal of enhancing overall performance through joint learning. 
Some open-source models have demonstrated decent capability in either visual understanding or image generation; however, they still exhibit substantial limitations in general visual editing (\textit{i.e.}, transforming an input image based on textual instructions). Specifically, current open-source methods struggle with: (1) accurately following complex editing instructions~\cite{sun2023emu2chat}; (2) preserving the original image's appearance during visual editing~\cite{koh2023generating}; and (3) accommodating flexible input formats~\cite{wang2024emu3,xie2024show} (\textit{e.g.}, supporting both single and multiple images with natural language instructions). These limitations severely hinder their practical utility, making them hardly worth rigorous evaluation in this task.

Recently, we observed that proprietary models such as the GPT-image series~\cite{openai2025gptimage} and the Nano Banana series~\cite{google2026nanobanana} have made significant advancements over open-source counterparts~\cite{wu2025qwen,labs2025flux1kontextflowmatching,cao2025hunyuanimage}. Notably, these models exhibit a remarkable capability in \textbf{R}easoning-\textbf{I}nformed vi\textbf{S}ual \textbf{E}diting (\textbf{RISE}) -- a sophisticated ability that enables models to make intelligent visual modifications based on contextual understanding and logical reasoning. This ability supports various real-world applications, such as context-aware image modification (\textit{e.g.}, adjusting lighting to match a scene's time of day), intelligent object insertion or removal with semantic consistency, and content adaptation based on inferred user intent. However, traditional image editing models~\cite{kawar2023imagic,hertz2022prompt,brooks2023instructpix2pix} that do not incorporate multi-modal reasoning lack these capabilities. While such a phenomenon is promising, we found that there is no well-established benchmark for systematically evaluating RISE task, making it difficult to quantitatively assess, compare, and further investigate this capability across existing approaches.

\begin{figure*}[t]
    \centering
    \includegraphics[width=0.98\textwidth]{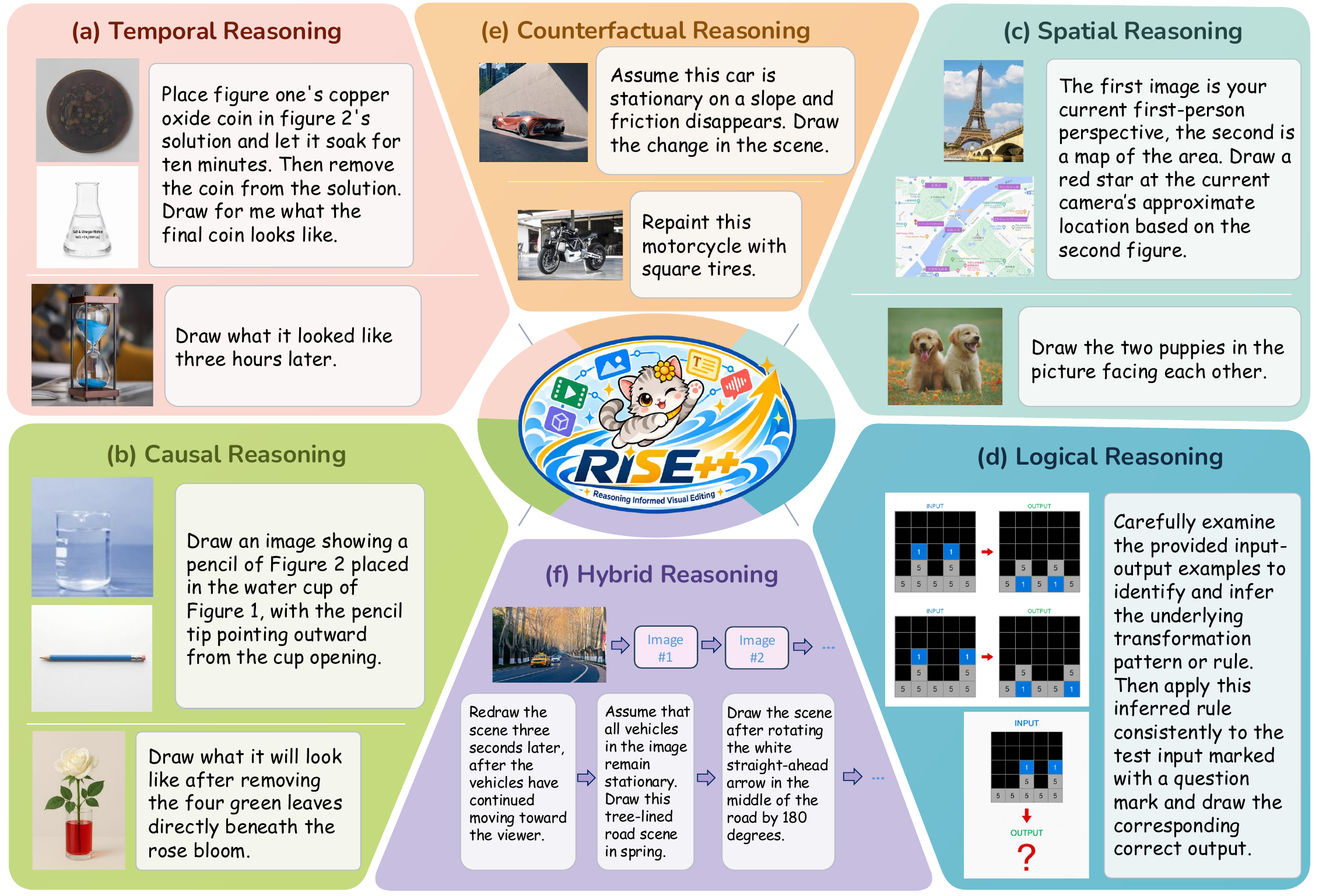}
    \caption{\textbf{Overview of RISEBench++.} We present illustrative example questions from each of the six major reasoning categories: temporal, causal, spatial, logical, counterfactual, and hybrid reasoning, each demanding profound image understanding and reasoning capabilities. }
    \label{fig:main}
\end{figure*}

To this end, we first introduce \textbf{RISEBench}, the first benchmark specifically designed to evaluate Reasoning-Informed viSual Editing (RISE), and further extend it to \textbf{RISEBench++}, a more comprehensive and fine-grained benchmark for this emerging task. RISEBench++ organizes reasoning-aware visual editing into a hierarchical taxonomy covering six fundamental reasoning dimensions: \textbf{temporal reasoning}, \textbf{causal reasoning}, \textbf{spatial reasoning}, \textbf{logical reasoning}, \textbf{counterfactual reasoning}, and \textbf{hybrid reasoning}. These dimensions are further decomposed into 12 subcategories and 65 fine-grained task types, providing broad coverage of diverse and reasoning-intensive editing scenarios. To enable comprehensive evaluation, we manually curate a diverse set of high-quality test cases across all six reasoning dimensions, including 200 for temporal reasoning, 200 for causal reasoning, 200 for spatial reasoning, 200 for logical reasoning, 100 for counterfactual reasoning, and 100 for hybrid reasoning, resulting in a total of \textbf{1,000 carefully curated and human-annotated samples}. Moreover, RISEBench++ supports single-image, multi-image,  and multi-turn editing, with each test case provided in both English and Chinese to facilitate bilingual evaluation.

For evaluation, we assess the quality of edited output images along three key dimensions: \textbf{instruction reasoning}, \textbf{appearance consistency}, and \textbf{visual plausibility}. Evaluations are conducted using both human judges and an LMM-as-a-judge framework. For the latter, we develop a rigorous evaluation pipeline that employs stronger judge models, incorporates dimension-specific visual and textual evidence, and adopts tailored scoring rubrics, yielding more reliable and well-calibrated assessments. Additionally, we conduct extensive experiments to quantify the correlation between the scores produced by the LMM and human experts, which verifies the reliability and effectiveness of our proposed framework.

More recently, agentic methods~\cite{wang2025imagent,he2026mind,zheng2026interleavethinker,bian2026rs} have emerged as a promising paradigm for augmenting the capabilities of generative models beyond direct generation. Building on this paradigm, we go beyond benchmarking and introduce \textbf{RISE-Agent}, an agentic framework with three collaborative roles: a \emph{planner} that interprets the editing instruction, reasons about the desired outcome, and invokes external tools when necessary to infer the target state; an \emph{executor} that realizes the target state using a generative model with a suite of programmatic editing tools; and a \emph{verifier} that assesses candidate results and decides whether to accept, refine, or trigger replanning. This closed-loop design enables RISE-Agent to reason, execute, and iteratively refine complex visual edits without extra training.

Using RISEBench++, we systematically evaluate 58 state-of-the-art visual editing approaches, spanning open-source and proprietary models as well as agentic methods. Our evaluation reveals a clear progression in visual editing capabilities. Early open-source models, including BAGEL~\cite{deng2025bagel}, Step1X-Edit~\cite{liu2025step1x}, FLUX~\cite{flux2024}, Emu2~\cite{sun2023emu2chat}, and OmniGen~\cite{xiao2024omnigen}, exhibit limited reasoning ability and consequently struggle across a broad range of test cases. More recent open-source models, such as Qwen-Image~\cite{wu2025qwen} and HunyuanImage~\cite{cao2025hunyuanimage}, substantially narrow this performance gap. Nevertheless, proprietary models, including the GPT-Image series~\cite{openai2025gptimage} and the Nano Banana series~\cite{google2026nanobanana}, generally achieve stronger overall performance, with GPT-Image-2.5 Sunburst achieving the best results among all evaluated models. RISE-Agent also achieves highly competitive performance, even outperforming most proprietary models and agentic methods across a broad range of RISE tasks. Notably, recent models already demonstrate strong appearance consistency and visual plausibility across diverse editing tasks, whereas instruction reasoning remains a major challenge, particularly for tasks involving logical and hybrid reasoning. These findings highlight reasoning ability as a key bottleneck and an important direction for future research.

The preliminary version of this work, RISEBench~\cite{zhao2026envisioning}\footnote{Compared with the preliminary conference version (NeurIPS'25, Oral, 0.35\%), this journal version introduces substantial extensions in several key aspects: (1) We reorganize the taxonomy from four reasoning categories into a three-level hierarchy of six dimensions, 12 subcategories and 65 fine-grained task types, and expand the benchmark from 360 to 1,000 test cases, newly supporting multi-image and multi-turn settings as well as bilingual instructions; (2) We improve the evaluation pipeline with stronger judge models, dimension-specific evidence and tailored scoring rubrics, and re-validate its agreement with human experts; (3) We develop RISE-Agent, a new agentic framework that augments an open-source executor with explicit planning, external tool use, and verifier-driven refinement, extending this work beyond benchmarking toward methodological exploration of effective solutions for RISE; (4) We expand our evaluation from 8 to 58 visual editing approaches, incorporating the latest open-source and proprietary models as well as agentic methods, and provide an in-depth, fine-grained analysis of failure modes across multiple reasoning dimensions.}, was presented as an Oral (Top 0.35\%) paper at the NeurIPS 2025 Datasets and Benchmarks Track. The main contributions of this extended work are summarized as follows:

(1) We present RISEBench++, a comprehensive and fine-grained benchmark for assessing \textbf{R}easoning-\textbf{I}nformed vi\textbf{S}ual \textbf{E}diting (RISE), comprising 1,000 carefully curated and human-annotated bilingual test cases across single-image, multi-image, and multi-turn settings, establishing a solid foundation for systematic assessment in this emerging area.

(2) We define a hierarchical taxonomy of RISE challenges, encompassing six reasoning dimensions that are further decomposed into 12 subcategories and 65 fine-grained task types, which allows model capabilities and failure modes to be diagnosed at the level of individual abilities rather than coarse category averages.

(3) We define three complementary evaluation dimensions and develop a robust LMM-as-a-judge framework that incorporates dimension-specific evidence and tailored scoring rubrics, enabling scalable and automated evaluation that is well aligned with human judgements.

(4) We develop RISE-Agent, a training-free agentic framework combining a reasoning planner, a tool-augmented executor, and a verifier-driven refinement loop, achieving strong performance across diverse RISE tasks and surpassing most proprietary models and existing agentic methods.

(5) We conduct a comprehensive evaluation and analysis of 58 prominent visual editing approaches, spanning open-source models, proprietary models, and agentic methods, revealing key strengths and limitations in their reasoning-informed visual editing capabilities and identifying important directions for future improvement.

\section{RELATED WORK}

\subsection{Image Editing Models}
\textbf{Diffusion-based Models}
Recent advances in image editing have been driven by the rapid development of large-scale generative models. Early diffusion-based editing methods~\cite{hertz2022prompt,meng2021sdedit,brooks2023instructpix2pix,ho2020denoising,dhariwal2021diffusion} demonstrate the feasibility of controllable image manipulation by leveraging pretrained diffusion priors. 
Building upon these foundations, recent approaches focus on improving instruction following, visual consistency, and editing generalization through larger-scale training and more diverse editing data. For example, MagicBrush~\cite{zhang2023magicbrush} introduces a large-scale human-annotated instruction-based editing dataset and enables more reliable text-guided image manipulation. FLUX.1 Kontext~\cite{labs2025flux1kontextflowmatching}, further enhance editing capability by integrating stronger generative backbones with context-aware conditioning. Despite their strong visual generation ability, diffusion-based editors are still primarily optimized for appearance-level transformations and lack explicit reasoning capabilities for complex semantic constraints.

\textbf{Unified Multimodal Models}
Beyond diffusion-based architectures, unified multimodal models (UMMs) have recently emerged as a promising direction for image editing by jointly modeling visual understanding, reasoning, and generation.
Open unified models~\cite{deng2025bagel,chen2025janusprounifiedmultimodalunderstanding,wang2025ovis} further explore the integration of multimodal understanding and image synthesis within a single framework. For example, OmniGen~\cite{xiao2024omnigen} focuses on unified and accessible image generation, providing a flexible framework for exploring multimodal generation capabilities.
Large-scale multimodal models, such as Gemini~\cite{team2023gemini}, demonstrate remarkable capabilities in conversational image generation and editing, where users can iteratively refine images through natural language interactions.

\textbf{Agentic Editing Methods}
Recent advances in agentic systems have demonstrated strong effectiveness in complex problem-solving scenarios, including code generation~\cite{yang2024sweagentagentcomputerinterfacesenable,wang2025openhandsopenplatformai,zhang2024autocoderoverautonomousprogramimprovement}, interactive environments~\cite{he2024webvoyagerbuildingendtoendweb,xie2024osworldbenchmarkingmultimodalagents,yao2023reactsynergizingreasoningacting}, and multimodal reasoning~\cite{kumar2024mmctagentmultimodalcriticalthinking,vlachos2025analyzepromptreasoncollaborativeagentbasedframework,ashraf2026agentxevaluatingdeepmultimodal}. 
Inspired by these advances, recent works have explored agentic reasoning for image editing, moving beyond conventional one-step instruction-based approaches~\cite{bian2026rs,zheng2026interleavethinker,wang2024genartistmultimodalllmagent,liang2025llmlvlmdrivenagentiterative,zeng2026miramultimodaliterativereasoning}. 
IntentEdit~\cite{zhang2026intentedit} and ImAgent~\cite{wang2025imagent} introduce multimodal agents that decompose user intentions into executable editing steps and iteratively refine generated results through perception-action feedback. Mind-Brush~\cite{he2026mind} further incorporates external multimodal knowledge retrieval to support editing under complex semantic constraints.

\subsection{Image Editing Benchmarks}
To systematically evaluate generative models, existing benchmarks~\cite{ye2025imgedit,huang2023smarteditexploringcomplexinstructionbased} typically measure two key aspects: visual fidelity~\cite{heusel2018ganstrainedtimescaleupdate,salimans2016improvedtechniquestraininggans,yu2022scalingautoregressivemodelscontentrich} and semantic consistency with the given instruction~\cite{hu2024ellaequipdiffusionmodels,ghosh2023genevalobjectfocusedframeworkevaluating}.
Earlier benchmarks~\cite{wang2023imagen,kawar2023imagic} mainly evaluate foundational editing capabilities, including inpainting, attribute modification, and layout adjustment. MagicBrush~\cite{zhang2023magicbrush} further extends evaluation to free-form instruction-guided editing and covers both single-turn and multi-turn scenarios. ImgEdit~\cite{ye2025imgedit} primarily targets traditional editing tasks where the required operations are explicitly specified and reasoning is not a central concern.
Subsequent benchmarks further expand task coverage and evaluation granularity. For example, AnyEdit~\cite{yu2025anyedit} introduces a broader taxonomy of editing operations, while I2I-Bench~\cite{wang2026i2i} evaluates diverse single-image and multi-image editing tasks with fine-grained dimensions covering editing accuracy, content preservation, and visual quality. More recent efforts continue to extend evaluation toward realistic and interactive settings. I2EBench2.0~\cite{ma2026extensive} incorporates both single-turn and multi-turn editing, enabling the assessment of consistency across successive interactions. 
However, these general benchmarks often rely on simple and explicit editing instructions. As model capabilities improve, more challenging evaluations are needed to assess whether models can handle complex and implicit editing requirements.

\begin{figure*}[t]
    \centering
    \includegraphics[width=\textwidth]{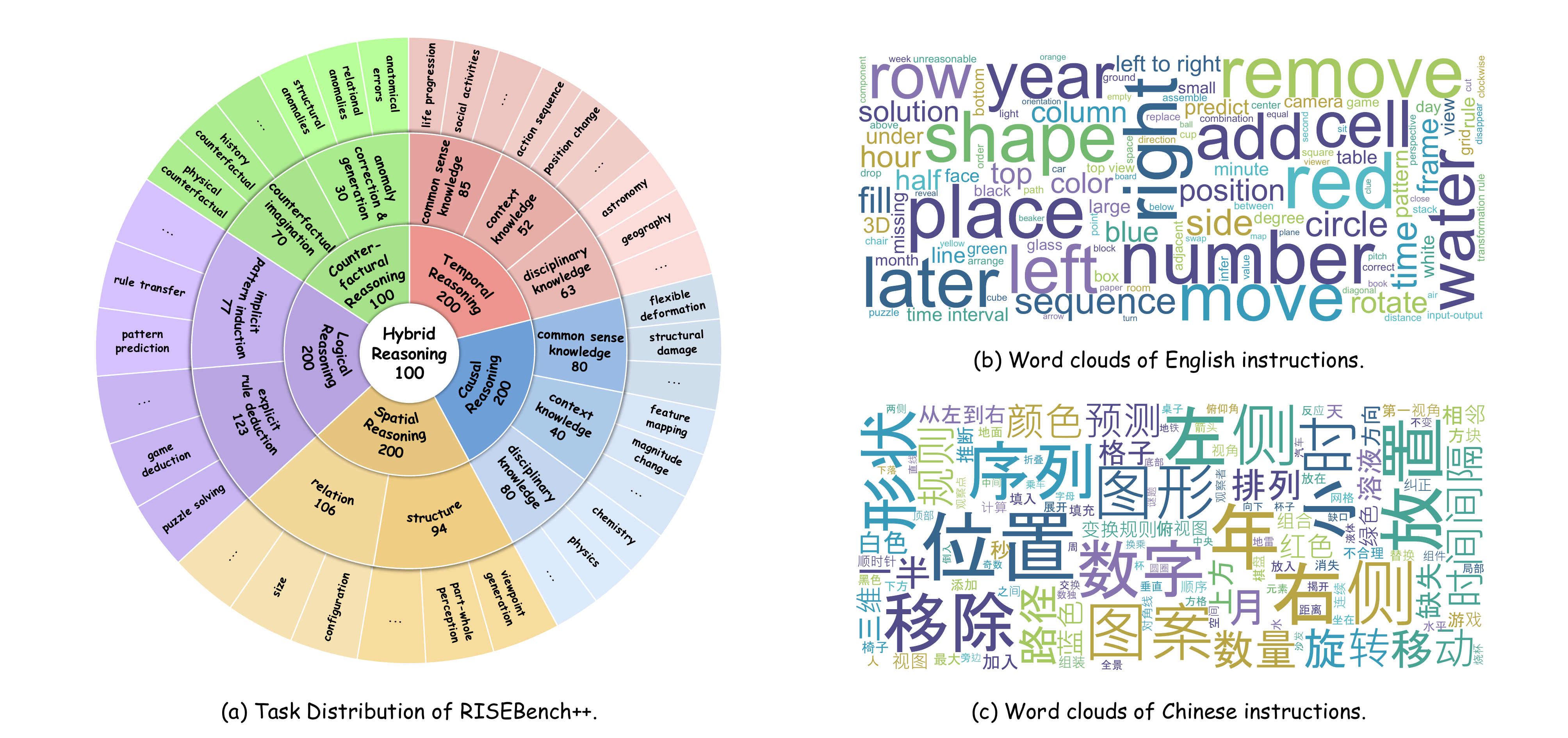}
    \caption{\textbf{Task distribution and word clouds of RISEBench++.} RISEBench++ comprises 1,000 tasks organized into a three-level taxonomy of six reasoning dimensions: \textit{Temporal} (200), \textit{Causal} (200), \textit{Spatial} (200), \textit{Logical} (200), \textit{Counterfactual} (100), and \textit{Hybrid} (100) Reasoning.}
    \label{fig:bench_statistics}
\end{figure*}

\begin{table*}[tp]
\centering
\caption{\textbf{Comparison with existing image editing benchmarks.}}
\label{tab:benchmark_comparison}
\resizebox{\textwidth}{!}{
\begin{tabular}{lcccccccc}
\toprule
\textbf{Benchmark} &
\textbf{Reasoning} &
\textbf{Multi-turn} &
\textbf{Multi-image} &
\textbf{Agent Support} &
\textbf{\# Task Types} &
\textbf{\# Samples} &
\textbf{\# Evaluated Models} &
\textbf{Language} \\
\midrule

MagicBrush~\cite{zhang2023magicbrush}     
& \RISEno 
& \RISEyes 
& \RISEno 
& \RISEno
& 9
& 1,053 
& 7
& EN \\

ImgEdit-Bench~\cite{ye2025imgedit}  
& \RISEno 
& \RISEyes 
& \RISEno 
& \RISEno
& 14
& 811   
& 7
& EN \\

AnyEdit~\cite{yu2025anyedit}        
& \RISEno 
& \RISEno 
& \RISEyes 
& \RISEno
& 25
& 1,250 
& 7
& EN \\

I2I-Bench~\cite{wang2026i2i}      
& \RISEno 
& \RISEno 
& \RISEyes 
& \RISEno
& 10
& 1,000 
& 11
& EN \\

UniREditBench~\cite{han2025unireditbench} 
& \RISEyes 
& \RISEno 
& \RISEno 
& \RISEno
& 18
& 2,700
& 16
& EN \\

WiseEdit~\cite{pan2026wiseedit} 
& \RISEyes 
& \RISEno 
& \RISEyes 
& \RISEno
& 4
& 1,220
& 22
& EN, ZH \\

KRIS-Bench~\cite{wu2026kris}     
& \RISEyes 
& \RISEno 
& \RISEyes 
& \RISEno
& 22
& 1,267 
& 10
& EN \\

RISE-Bench~\cite{zhao2026envisioning}     
& \RISEyes
& \RISEno 
& \RISEno
& \RISEno
& 16
& 360
& 9
& EN \\

\midrule

\textbf{RISEBench++ (Ours)} 
& \RISEyes 
& \RISEyes 
& \RISEyes 
& \textbf{\RISEyes}
& 65
& 1,000 
& 58
& EN, ZH \\

\bottomrule
\end{tabular}
}
\end{table*}

\subsection{Reasoning-Informed Benchmarks for Visual Generation}
\textbf{Image Generation Benchmarks}
Recent text-to-image benchmarks have increasingly shifted from surface-level prompt alignment toward evaluating whether generative models can reason over implicit knowledge and constraints. Early efforts such as WISE~\cite{niu2025wise} and Commonsense-T2I~\cite{fu2024commonsense} focus on world knowledge and everyday commonsense. They test whether models can move beyond literal word-to-pixel correspondence and generate images consistent with real-world knowledge. 
Beyond general commonsense, several benchmarks emphasize more structured and knowledge-intensive generation~\cite{luo2026mmmg,wang2025genexam,chang2026oneig}. They move toward discipline-oriented evaluation, covering academic knowledge and knowledge-intensive reasoning across multiple subjects. 

\textbf{Image Editing Benchmarks}
Image editing is also evolving toward a reasoning-oriented setting. KRIS-Bench~\cite{wu2026kris} takes an early step in this direction by organizing knowledge-based editing tasks around factual, conceptual, and procedural knowledge. Related efforts also investigate reasoning processes beyond direct image transformation. 
WiseEdit~\cite{pan2026wiseedit} extends reasoning-oriented editing to more diverse knowledge and interaction settings. It evaluates whether models can infer implicit editing goals from linguistic and visual context, including bilingual and multi-image inputs. UniREditBench~\cite{han2025unireditbench} covers a broad range of reasoning dimensions and evaluates both semantic correctness and visual fidelity with dual references.
Moving toward more domain-specific settings, GRADE~\cite{liu2026grade} evaluates discipline-informed image editing across multiple academic fields. It requires models to apply subject-specific knowledge while maintaining visual consistency and logical readability. 

\textbf{Video Generation Benchmarks}
Video generation is also an important component of visual generation. Compared with image generation, it places stronger demands on reasoning correctness because models must account for temporal evolution, motion dynamics, and cross-frame consistency. Physics-IQ~\cite{motamed2501generative} focuses on real-world physical processes and evaluates both image-to-video and video-to-video generation. RISE-Video~\cite{liu2026rise} studies a broader range of implicit rules in video generation. It evaluates whether models can infer hidden constraints from the input and preserve them consistently as the scene evolves over time across multiple frames.

\section{METHODOLOGY}

\subsection{Benchmark Overview}

Humans possess a deep, conceptual understanding of objects and scenes in the real world that goes far beyond superficial attributes such as color and shape, allowing them to effortlessly envision how a scene would appear after a described change. These capabilities, however, remain highly challenging for current generative models, which struggle to incorporate such reasoning into their visual outputs. As illustrated in Figure~\ref{fig:main}, RISEBench++ organizes this ability into six fundamental reasoning dimensions, which can be further grouped into three distinct types of reasoning demand. \emph{Temporal reasoning} and \emph{causal reasoning} are primarily \textbf{knowledge-driven}: once the relevant real-world knowledge is available, the target state can often be inferred directly with relatively few intermediate steps. \emph{Spatial reasoning} and \emph{logical reasoning} are \textbf{perception-driven}: a model must first infer the latent spatial structure or governing rules from visual observations, and then reason over this inferred representation to derive the target state. \emph{Counterfactual reasoning} is \textbf{imagination-driven}: a model must draw on real-world knowledge while deliberately overriding it to construct a coherent yet physically impossible state. Finally, \emph{hybrid reasoning} integrates multiple modes of reasoning across successive turns, evaluating a model's capacity for long-horizon reasoning by requiring it to correctly resolve each intermediate transformation while preserving the consistency of the evolving scene.

As shown in Figure~\ref{fig:bench_statistics}, the six reasoning dimensions are further decomposed into 12 subcategories and 65 fine-grained task types. Temporal and causal reasoning span common-sense, contextual, and disciplinary knowledge; Spatial reasoning captures structural and relational understanding; Logical reasoning examines explicit rule deduction and implicit pattern induction; and counterfactual reasoning encompasses counterfactual imagination and anomaly correction/generation. The bilingual word clouds further illustrate the diversity of editing concepts across these tasks. Overall, RISEBench++ comprises 1,000 diverse and carefully curated test cases, with 200 cases each for temporal, causal, spatial, and logical reasoning, and 100 cases each for counterfactual and hybrid reasoning.

Table~\ref{tab:benchmark_comparison} summarizes the comparison between RISEBench++ and existing image-editing benchmarks. Our benchmark offers several key advantages over existing studies: (i) it adopts a more comprehensive evaluation setting, jointly considering reasoning ability, multi-turn interaction, multi-image inputs, and agent support; (ii) it provides broader task coverage, with 65 fine-grained task types and 1,000 samples spanning diverse reasoning scenarios; (iii) it evaluates 58 representative models, offering a large-scale comparison across different model families; and (iv) it supports both English and Chinese, facilitating evaluation across multilingual instruction settings.

\subsection{Benchmark Construction}
RISEBench++ encompasses six major reasoning categories: Temporal, Causal, Spatial, Logical, Counterfactual, and Hybrid Reasoning. For each category, we curate a diverse set of high-quality, carefully designed test cases that require both deep visual understanding and precise reasoning.

\begin{figure*}[t]
    \centering
    \includegraphics[width=0.98\textwidth]{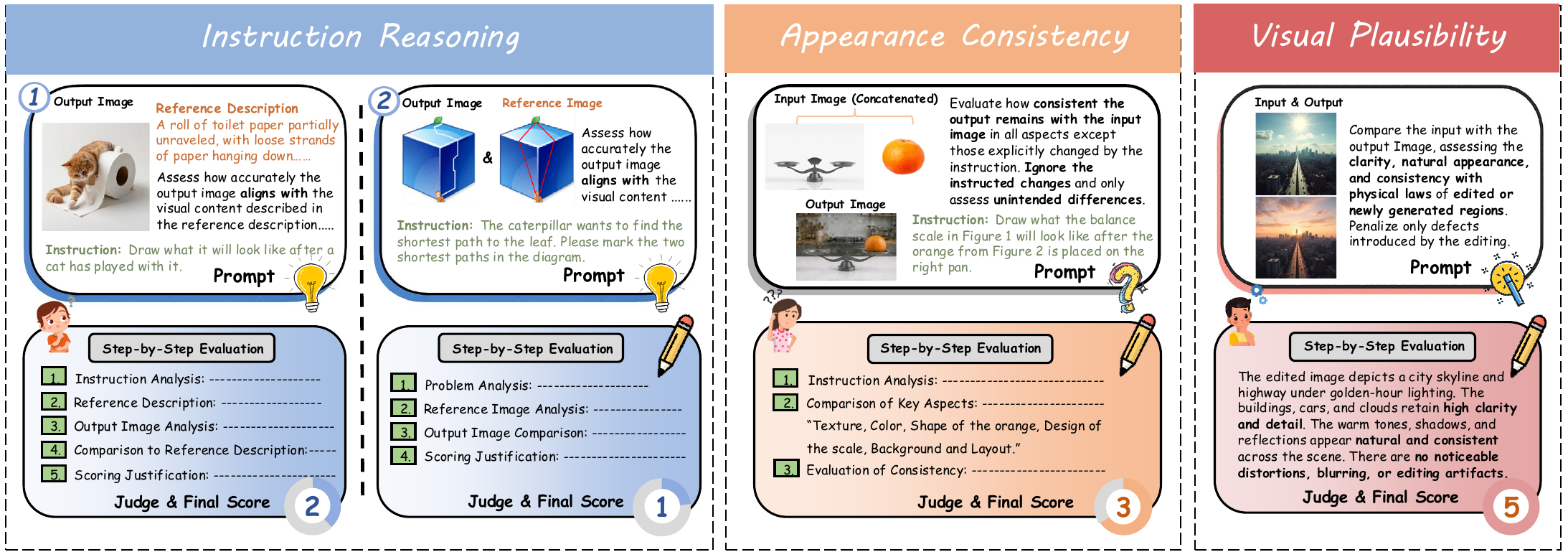}
    \caption{\textbf{Evaluation metrics and pipeline of RISEBench++.} RISEBench++ evaluates generated images across \textit{Instruction Reasoning}, \textit{Appearance Consistency}, and \textit{Visual Plausibility}. Gemini-3-Flash evaluates the first two using dimension-specific prompts, while Gemini-3.1-Flash-Lite evaluates the third.}
    \label{fig:judge_pipeline}
\end{figure*}

\noindent \textbf{Temporal Reasoning.}
Temporal reasoning tasks evaluate a model's ability to infer past, future, or intermediate states of objects and scenes. These tasks are primarily knowledge-driven, but require models to ground their inferences in the observed visual state and the specified time, duration, or sequence position. We construct 200 test cases spanning 19 fine-grained task types, organized into three complementary subcategories:
1) \textit{Common-Sense Knowledge} covers everyday temporal changes in natural and social settings, including life progression, material transformation, natural environmental changes, physical aging, social activities, tool evolution, and urban evolution. These tasks require models to infer how appearance and scene composition vary with growth, degradation, recurring events, or changes across eras.
2) \textit{Context Knowledge} focuses on image-specific temporal evidence, encompassing temporal state transitions, precise temporal localization, position changes, and action sequences. These tasks require models to interpret visible time indicators, such as clocks and calendars, or ongoing processes and, in multi-image cases, use the observed sequence to infer an earlier, intermediate, or subsequent frame.
3) \textit{Disciplinary Knowledge} draws on physics, chemistry, biology, astronomy, geography, history, medicine, and computer science. These tasks assess knowledge-based temporal inference, as in chemical reactions, biological regeneration, historical scene reconstruction, etc.
Overall, these subcategories assess the integration of temporal knowledge, visual grounding, and state inference, requiring edits to satisfy the requested temporal conditions while remaining visually coherent.

\noindent \textbf{Causal Reasoning.}
Causal reasoning concerns how actions, events, or operations change objects and scenes. These knowledge-driven tasks require predictions based on the input images. We construct 200 test cases covering 14 fine-grained task types in three subcategories:
1) \textit{Common-Sense Knowledge} covers physical state transition, structural damage, and flexible deformation, testing how models use everyday knowledge to predict the effects of actions or events on objects and scenes.
2) \textit{Context Knowledge} includes magnitude change, reading offset, order relation, and feature mapping. Models use the input images to determine which elements to change and by how much.
3) \textit{Disciplinary Knowledge} tests how models apply knowledge from physics, chemistry, biology, astronomy, medicine, music theory, and computer science to predict outcomes, with tasks involving physical laws, computer science concepts, chemical reactions and so on.
In summary, these tasks assess whether models can reason from a specified cause and the initial visual state to the resulting changes, including effects not explicitly stated in the instruction.

\noindent \textbf{Spatial Reasoning.}
Spatial reasoning tasks evaluate a model's ability to infer spatial relationships and scene structures from visual observations, and to translate this understanding into geometrically coherent edits. These tasks are primarily perception-driven because the required transformations must be grounded in the spatial evidence provided by the input images rather than in linguistic priors. To systematically assess this capability, we construct 200 test cases spanning 14 fine-grained task types, organized into two complementary subcategories: 1) \textit{Relation} focuses on understanding and manipulating spatial properties and relationships, covering adjacency, direction, orientation, projection, distance, size, configuration, viewing scale, and localization. These tasks require models to identify how objects are positioned, oriented, and related within a scene, and to preserve or modify these relationships according to the instruction. 2) \textit{Structure} examines the ability to infer coherent object and scene representations, encompassing part-whole perception, geometric regularities reasoning, viewpoint generation, environmental structure reasoning, and egocentric scene reconstruction. These tasks require models to reason beyond isolated visual attributes by understanding how components form a whole, recognizing geometric organization, and reconstructing scenes from different perspectives. These two subcategories evaluate both relational understanding and structural inference, providing a fine-grained assessment of spatially grounded visual editing.

\noindent \textbf{Logical Reasoning.}
Logical reasoning tasks evaluate a model's ability to interpret visual evidence, establish the relevant rules, and derive solutions that can be expressed through image editing. These tasks are also primarily perception-driven because successful editing requires reasoning over the specific visual configuration, rather than producing a plausible result from linguistic priors alone. We construct 200 test cases across seven fine-grained task types, distinguishing two complementary forms of inference: 1) \textit{Explicit Rule Deduction} requires models to apply given rules or task constraints to the observed visual state. It includes path searching, mathematical derivation, game deduction, and puzzle solving. These tasks assess whether a model can extract the relevant visual information, carry out structured reasoning, and accurately render a solution that satisfies the governing constraints. 2) \textit{Implicit Pattern Induction} requires models to infer the underlying regularities from visual observations before determining the appropriate edit. It comprises pattern prediction, visual oddity detection, and rule transfer, testing the ability to complete a pattern, identify an element that violates a shared regularity, or apply an inferred rule to a new visual instance. These two subcategories assess the connection between visual perception, rule-based inference, and precise visual realization.

\noindent \textbf{Counterfactual Reasoning.}
Counterfactual reasoning tests two abilities: using existing knowledge to identify and correct visual anomalies, and creating images that contradict common-sense or domain knowledge as instructed. Our 100 test cases cover 11 fine-grained task types in two subcategories:
1) \textit{Counterfactual Imagination} covers tasks in physics, chemistry, biology, astronomy, geography, history, music theory, and computer science. Models must predict and show what would happen if familiar rules or facts were changed.
2) \textit{Anomaly Correction and Generation} includes anatomical errors in human or animal bodies, structural anomalies within single non-living objects, and relational anomalies between objects or between objects and their environment.
Unlike the first four dimensions, counterfactual tasks require models to reason from existing knowledge while also imagining and creating images beyond real-world constraints.

\noindent \textbf{Hybrid Reasoning.}
Hybrid reasoning combines tasks from the five preceding dimensions through multi-turn image editing. We construct 100 test cases with three to five instructions each, linking tasks within or across these dimensions. These tasks require models to combine temporal reasoning for tracking event order and inferring successive states with causal reasoning for predicting the consequences of actions. They also draw on spatial reasoning to maintain coherent scene relationships, logical reasoning to apply task rules, and counterfactual reasoning to imagine alternative conditions and identify or correct visual anomalies. This combination demands long-horizon reasoning and sustained visual consistency: models must track the evolving scene, use intermediate results to guide subsequent edits, and preserve relevant prior edits and visual details across turns. The cases span a range of complex tasks, including chemical experiments, numerical logic puzzles, and everyday scenarios, placing high demands on models' knowledge across fields and their reasoning, visual understanding, and editing skills.

\begin{figure*}[t]
    \centering
    \includegraphics[width=\textwidth]{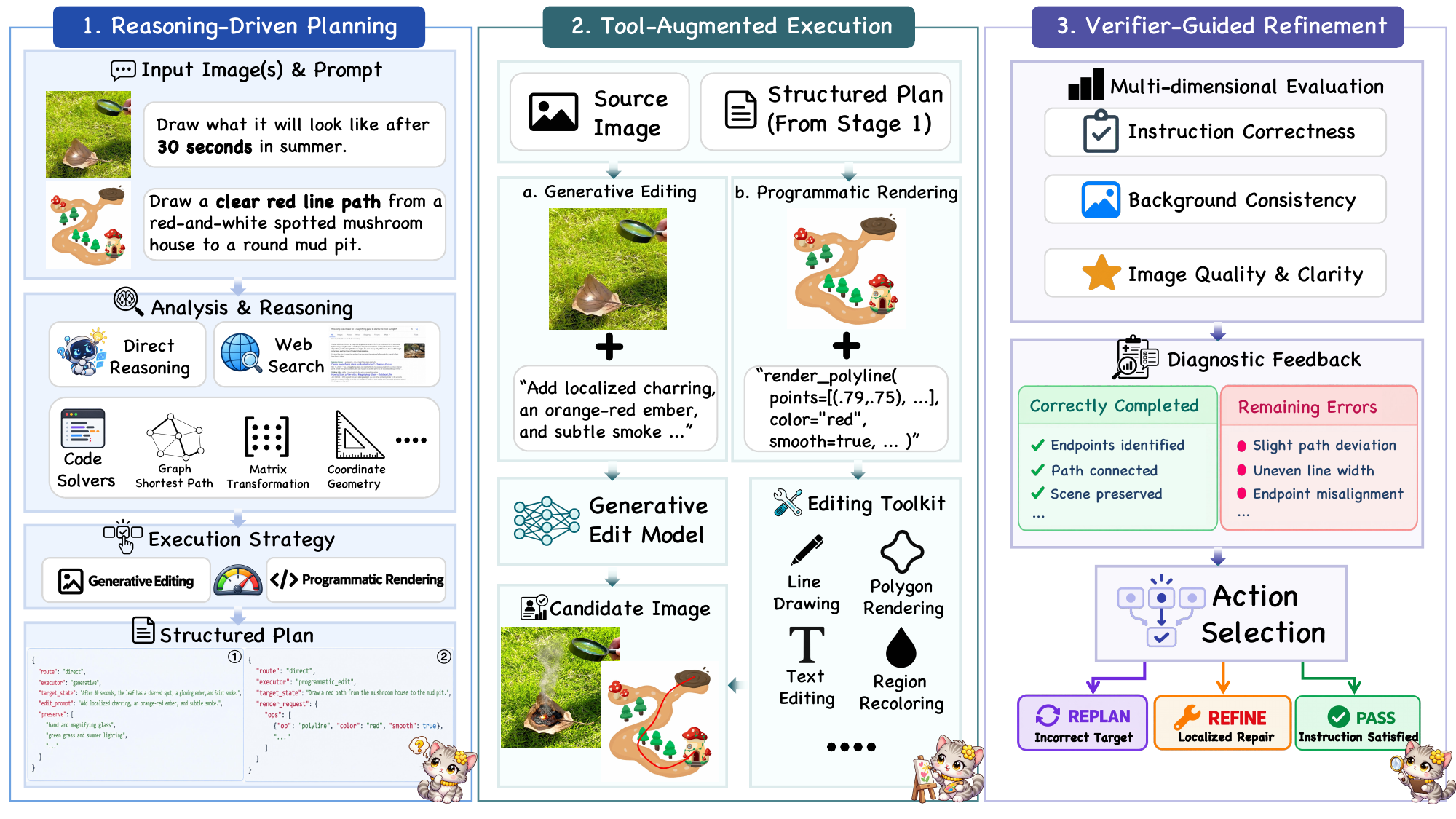}
    \caption{\textbf{Overview of RISE-Agent.} RISE-Agent follows a closed-loop workflow of reasoning-driven planning, tool-augmented execution, and verifier-guided refinement, bridging target-state inference with precise visual realization.}
    \label{fig:rise-edit}
\end{figure*}

\subsection{Evaluation Pipeline}

Evaluating the quality of reasoning-informed visual editing remains a challenging task. To address this, we first establish detailed scoring guidelines and conduct comprehensive human evaluations along three key dimensions: 
\textbf{1. Instruction Reasoning},
assessing whether the model correctly interprets and follows the editing instruction;
\textbf{2. Appearance Consistency}, 
evaluating preservation of relevant visual attributes from the original image;
\textbf{3. Visual Plausibility}, 
determining whether the edited regions of the output are coherent, realistic, and physically or logically plausible within context.
Since human evaluation is resource-intensive and difficult to scale, we further adopt an LMM-as-a-Judge approach to overcome these limitations.  Given their strong visual understanding and reasoning abilities~\cite{peng2025lmm,wu2026generalization}, state-of-the-art LMMs offer a promising alternative for automatic and human-aligned evaluation~\cite{ku2024viescore,chen2024mllm,zhao2026trust,zhang2026firm}. We develop a robust evaluation pipeline (Figure~\ref{fig:judge_pipeline}) leveraging these models to produce scalable assessments.
In the following part, we detail each evaluation dimension:

\noindent \textit{Dimension 1: Instruction Reasoning.}
This dimension assesses the model's ability to accurately understand and execute the given instruction, with particular attention to both explicit directives and implicit requirements embedded within the prompt. A high-quality response not only performs the literal task specified but also captures the underlying reasoning or intended visual effect implied by the instruction. To improve the accuracy of LMMs in assessing instruction reasoning scores, we propose two evaluation methods. First, for samples with simple scenes that are easily describable through comprehensive text, we annotate a \textit{reference text}, which serves as the ground truth and is used by the LMM to determine if the output image aligns with this description. For samples involving more complex scenes or unique shapes that are difficult to describe in text, particularly in Logical Reasoning and Spatial Reasoning tasks, we provide a \textit{reference image} that fully matches the desired output. The judging LMM then compares the output image with the reference image to assess whether the instruction has been correctly executed. This approach expands the range of instruction types and ensures that the LMM judgment scores are relatively accurate, with further details illustrated in Figure~\ref{fig:judge_pipeline} (left). A full score (e.g., 5) is reserved for outputs that satisfy all explicit requirements and reasoning-dependent conditions in the instruction, indicating robust instruction comprehension and reasoning.

\noindent \textit{Dimension 2: Appearance Consistency.}
Appearance consistency measures how well the visual elements unrelated to the instruction are preserved between the input and output images. This is particularly important in visual editing tasks, as it distinguishes between models that perform grounded edits based on the original image (\textit{e.g.}, native generation models) and those that regenerate scenes from scratch (\textit{e.g.}, cascade-based models). The LMM evaluates this metric by comparing the output image with the input image in accordance with the given instruction, as illustrated in Figure~\ref{fig:judge_pipeline} (middle). For multi-image inputs, all source images are concatenated into a single composite image, which is then provided to the LMM for consistency evaluation. For tasks involving temporal, causal, or spatial reasoning -- where the input is typically a natural image rich in visual complexity -- appearance consistency is scored on a continuous scale from 1 to 5, allowing for nuanced evaluation of how well the core scene is preserved post-editing. In contrast, logical reasoning tasks often involve stylized or synthetic inputs with simple layouts. Given their minimalistic structure, consistency in these cases is evaluated using a binary scheme: a score of 5 indicates full preservation of visual properties, while 1 reflects major deviations. This dimension ensures that models not only generate correct content but also do so in a way that respects the visual fidelity of the original input, which is essential for coherent and context-preserving visual editing.

\noindent \textit{Dimension 3: Visual Plausibility.}
The visual quality and realism of the generated image are critical factors in evaluating the performance of generative models. This dimension assesses whether the model-introduced modifications are free from common generation artifacts such as blurriness, unnatural distortions, structural incoherence, or other visual defects. We focus on the modified regions rather than the entire output image to prevent the judge's bias toward the original image from affecting the evaluation, thereby avoiding unreasonable penalties when the model faithfully preserves its visual characteristics.
To this end, we provide the LMM with both the original and generated images and ask it to assess the visual quality of the modified content, as depicted in Figure~\ref{fig:judge_pipeline} (right). For temporal, causal, and spatial reasoning, visual plausibility further considers whether the modified content remains physically coherent and realistic. For counterfactual reasoning, where physically implausible outcomes may be intentional, we only assess perceptual quality and do not penalize violations of physical laws. Visual plausibility is evaluated on a graded scale from 1 to 5, allowing for nuanced differentiation between high-quality and flawed generations. This dimension ensures that, beyond correctness and consistency, the model-induced modifications meet a basic threshold of visual fidelity while avoiding unreasonable penalties for visual characteristics inherited from the original image.

The evaluation details, such as the specific instructions provided to judges (human evaluators and LMM-based assessors), carefully selected in-context examples, and the detailed configuration of the LMM judgement, are provided in the supplementary materials.

During evaluation, all \textit{Dimension Scores} are normalized to the range [1, 5] and subsequently rescaled to a 100-point scale for reporting. A sample is considered successfully solved only if it achieves scores of 5 on the three metrics, indicating full satisfaction of all applicable evaluation dimensions. \textit{Accuracy} is then defined as the percentage of samples that are successfully solved out of the total number of test cases. The two complementary metrics offer both fine-grained performance measurement and an interpretable success rate across tasks.

\subsection{RISE-Agent}

Our evaluation reveals a clear gap between determining what an edited image should look like and actually generating it. Existing editing models can often produce visually plausible and appearance-consistent results when the desired change is explicitly specified, but struggle when the target state must first be inferred through visual understanding, external knowledge, or multi-step reasoning. This suggests that the main bottleneck in RISE lies in reasoning before generation rather than image synthesis itself. 
Motivated by this observation, we introduce RISE-Agent, a training-free multi-agent framework with three collaborative roles: a planner, an executor, and a verifier, as illustrated in Figure~\ref{fig:rise-edit}. The planner determines the desired target state, the executor realizes it using a generative model or programmatic editing tools, and the verifier evaluates the result and provides feedback for refinement or replanning. Together, these components form a closed loop that enables RISE-Agent to tackle diverse reasoning-informed image editing tasks.

\noindent \textbf{Reasoning-Driven Planning.}
Given the source image(s) and an editing instruction, the planner operates in two stages. First, it analyzes the visual content and instruction to infer the intended modification. When the task can be resolved using visual evidence and commonsense knowledge, the planner reasons directly. If additional domain knowledge is needed, it retrieves relevant information through web search. For problems that can be formalized as structured computational tasks, it can invoke one of seven carefully designed code-based tools, covering arithmetic, date calculation, grid and graph shortest-path search, Sudoku solving, coordinate geometry, and matrix transformation. These solvers provide reliable intermediate results for formally structured reasoning tasks.
Second, the planner integrates its visual analysis with any retrieved or tool-derived results to construct an explicit target state and determine how the edit should be realized. It then outputs either a standalone prompt for generative editing or a declarative program for programmatic image manipulation.

\noindent \textbf{Tool-Augmented Execution.}
The executor realizes the target state by invoking either a generative editor or programmatic editing tools according to the plan produced by the planner. The generative editor handles tasks that require synthesizing new content or substantially modifying existing content, such as object insertion, state transformation, and appearance adjustment. However, generative models often lack the fine-grained control needed for precise local edits and may inadvertently modify irrelevant regions. This limitation is particularly evident in logical reasoning tasks involving diagrams, grids, charts, and puzzles, where small geometric or symbolic errors can invalidate the result. We therefore equip the executor with a programmatic editing toolkit consisting of 12 deterministic operations, including but not limited to line drawing, polygon rendering, text editing, region recoloring, and content relocation. These operations enable precise drawing, marking, label replacement, cell modification, and element rearrangement. Since the programmatic edits are applied directly to the selected canvas, all untouched pixels remain unchanged. For multi-image inputs, other images can serve as visual references or content sources. Together, generative editing and deterministic programmatic manipulation provide the executor with both the flexibility required for content synthesis and the fine-grained control needed for precise visual editing.

\noindent \textbf{Verifier-Guided Refinement.}
After each execution, the verifier evaluates the candidate in terms of instruction following, background consistency, and visual quality, while identifying what has been correctly completed and what remains to be corrected. Based on this assessment, it selects one of three actions: \emph{pass}, \emph{refine}, or \emph{replan}. \emph{Pass} is selected when the candidate adequately fulfills the instruction without significant consistency or quality issues. \emph{Refine} is selected when the overall result is correct but contains localized errors that can be repaired without discarding successful content. The verifier then provides a concise correction instruction and determines whether it should be executed generatively or with programmatic editing tools. \emph{Replan} is selected when the target state has been incorrectly inferred or the current execution strategy is unsuitable. In this case, the failed result and diagnostic feedback are returned to the planner, which produces a new plan from the original input. The loop operates under predefined replanning and refinement budgets to prevent unbounded iteration. Upon termination, the best candidate produced throughout the process is selected as the final output.

\begin{figure*}[t]
    \centering
    \includegraphics[width=0.98\textwidth]{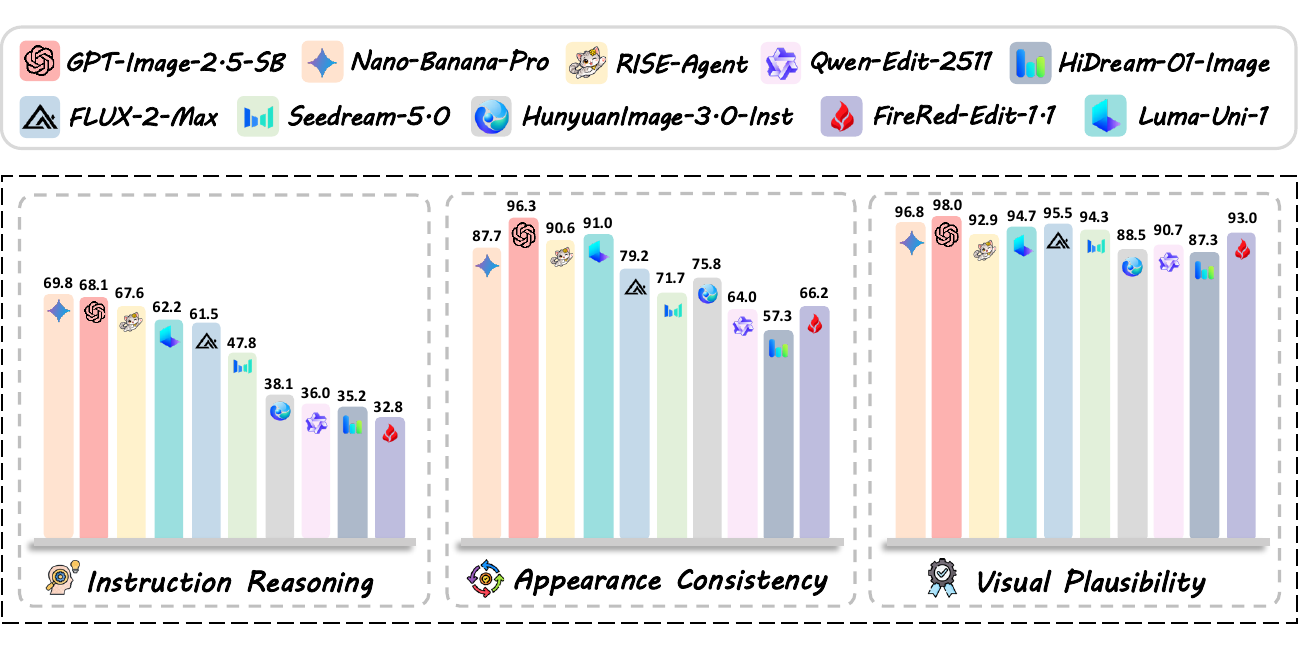}
    \caption{\textbf{Comparison across three evaluation dimensions on RISEBench++ (EN).} Nano Banana Pro achieves the highest instruction reasoning score, while GPT-Image-2.5 Sunburst leads in appearance consistency and visual plausibility.}
    \label{fig:metric_score}
\end{figure*}

\section{EXPERIMENT}

\subsection{Implementation Details}
\paragraph{Benchmarking Models}
We evaluate 58 visual editing approaches, including 19 closed-source models, 34 open-source models, and 5 agentic methods.
\textbf{Closed-source Models.}
GPT-Image-1/1.5/2/2.5 Flare/2.5 Sunburst~\cite{openai2025gptimage}, Nano Banana/Nano Banana Pro/Nano Banana 2/Nano Banana 2 Lite~\cite{google2026nanobanana}, Luma Uni-1~\cite{luma2026uni1}, FLUX.2 Pro/Max~\cite{bfl2025flux2}, Wan2.7-Image-Pro~\cite{mao2026wan}, Qwen-Image-2.0/3.0~\cite{zhao2026qwen}, HunyuanImage-3.5-Preview~\cite{cao2025hunyuanimage}, Seedream 4.5/5.0~\cite{seedream2025seedream}, and Step Image Edit 2~\cite{liu2025step1x}.
\textbf{Open-source Models.}
HunyuanImage-3.0-Instruct~\cite{cao2025hunyuanimage}, Qwen-Image-Edit-2509/2511 and Qwen-Image-2.1~\cite{wu2025qwen}, FireRed-Image-Edit-1.0/1.1~\cite{team2026firered}, JoyAI-Image-Edit/Plus~\cite{song2026joyai}, HiDream-O1-Image~\cite{cai2026hidream}, FLUX.1 Kontext Dev~\cite{labs2025flux1kontextflowmatching}, FLUX.2 Dev/Klein 4B/Klein 9B~\cite{bfl2025flux2}, Step1X-Edit~\cite{liu2025step1x}, Boogu-Image-0.1-Edit~\cite{chen2026boogu}, SenseNova-U1~\cite{diao2026sensenova}, UniPic 2.0/3.0~\cite{wei2025skywork,wei2026skywork}, BAGEL (\textpm CoT)~\cite{deng2025bagel}, LongCat-Image-Edit~\cite{team2025longcat}, UniWorld-V1/V2~\cite{lin2025uniworld,li2025uniworld}, InternVL-U (\textpm CoT)~\cite{tian2026internvl},  OmniGen/OmniGen2~\cite{xiao2024omnigen,wu2025omnigen2}, LLaDA-Image~\cite{chen2026llada}, Ovis-U1~\cite{wang2025ovis}, Emu2~\cite{sun2023emu2chat}, and Lumina-DiMOO~\cite{xin2025lumina}.
\textbf{Agentic Methods.}
ImAgent~\cite{wang2025imagent}, MindBrush~\cite{he2026mind}, IntentEdit~\cite{zhang2026intentedit}, InterleaveThinker~\cite{zheng2026interleavethinker} and our RISE-Agent.
For all of the proprietary models, we obtained their outputs directly via their respective official API services.

\paragraph{Judge Models}
For instruction reasoning and appearance consistency, we use Gemini-3-Flash as the judge model. For visual plausibility, considering the trade-off between evaluation quality and cost efficiency, we use Gemini-3.1-Flash-Lite as the judge model.

\paragraph{RISE-Agent Configuration}
We use GPT-5.1 as the planner and Gemini-3-Flash as the verifier. For generative image editing, we adopt the open-source FLUX.2 Klein 9B model. We set the maximum number of replanning rounds to one and the maximum number of refinement rounds to two for each input. For hybrid reasoning, we decompose the multi-turn process into a sequence of single-turn edits, where the output of each turn serves as the input to the subsequent turn.

\subsection{Main Results (LMM-as-a-Judge)}

\begin{table*}[!tp]
\caption{\textbf{Overall performance on RISEBench++ (EN).} Models are grouped into closed-source, open-source, and agentic methods. \textbf{Bold} and \underline{underlined} values indicate the best and second-best results within each group, respectively.
}
\label{tab:main_result_en}
\centering
\begingroup
\fontsize{7}{8.6}\selectfont
\setlength{\tabcolsep}{2pt}
\renewcommand{\arraystretch}{0.98}
\def\RISEnum#1{\makebox[2em][c]{#1}}

\begin{tabular}{
>{\raggedright\arraybackslash}p{\dimexpr .299\textwidth-2\tabcolsep\relax}
>{\centering\arraybackslash}p{\dimexpr .060\textwidth-2\tabcolsep\relax}
>{\centering\arraybackslash}p{\dimexpr .073\textwidth-2\tabcolsep\relax}
>{\centering\arraybackslash}p{\dimexpr .065\textwidth-2\tabcolsep\relax}
>{\centering\arraybackslash}p{\dimexpr .065\textwidth-2\tabcolsep\relax}
>{\centering\arraybackslash}p{\dimexpr .065\textwidth-2\tabcolsep\relax}
>{\centering\arraybackslash}p{\dimexpr .120\textwidth-2\tabcolsep\relax}
>{\centering\arraybackslash}p{\dimexpr .068\textwidth-2\tabcolsep\relax}
>{\centering\arraybackslash}p{\dimexpr .062\textwidth-2\tabcolsep\relax}
>{\centering\arraybackslash}p{\dimexpr .060\textwidth-2\tabcolsep\relax}
>{\centering\arraybackslash}p{\dimexpr .063\textwidth-2\tabcolsep\relax}
}
\toprule[0.8pt]
\multirow{2}{*}[-2pt]{\textbf{Model}} &
\multicolumn{1}{c}{\multirow{2}{*}[-2pt]{\shortstack{\textbf{Multi-}\\\textbf{img}}}} &
\multicolumn{6}{c}{\textbf{Reasoning Tasks}} &
\multicolumn{3}{c}{\textbf{Overall}} \\
\cmidrule(lr){3-8}\cmidrule(lr){9-11}
 & &
\multicolumn{1}{c}{\textbf{Temporal}} &
\multicolumn{1}{c}{\textbf{Causal}} &
\multicolumn{1}{c}{\textbf{Spatial}} &
\multicolumn{1}{c}{\textbf{Logical}} &
\multicolumn{1}{c}{\textbf{Counterfactual}} &
\multicolumn{1}{c}{\textbf{Hybrid}} &
\multicolumn{1}{c}{\textbf{Single}} &
\multicolumn{1}{c}{\textbf{Multi}} &
\multicolumn{1}{c}{\textbf{Total}} \\
\midrule[0.5pt]
\rowcolor{RISEGroup}
\multicolumn{11}{l}{\rule{0pt}{10pt}\raisebox{1pt}[0pt][0pt]{\textbf{Closed-source Models}}\rule[-2pt]{0pt}{0pt}} \\
\midrule[0.4pt]
GPT-Image-2.5 Sunburst~\cite{openai2025gptimage} & \RISEyes & \RISEnum{\textbf{67.5}} & \RISEnum{\underline{69.0}} & \RISEnum{\textbf{64.0}} & \RISEnum{27.0} & \RISEnum{\textbf{72.0}} & \RISEnum{\textbf{39.0}} & \RISEnum{\textbf{56.9}} & \RISEnum{\textbf{52.6}} & \RISEnum{\textbf{56.6}} \\
Nano Banana Pro~\cite{google2026nanobanana} & \RISEyes & \RISEnum{\underline{63.0}} & \RISEnum{\textbf{71.0}} & \RISEnum{51.5} & \RISEnum{\textbf{36.0}} & \RISEnum{\underline{65.0}} & \RISEnum{\underline{38.0}} & \RISEnum{\underline{55.4}} & \RISEnum{44.9} & \RISEnum{\underline{54.6}} \\
Nano Banana 2~\cite{google2026nanobanana} & \RISEyes & \RISEnum{\textbf{67.5}} & \RISEnum{67.0} & \RISEnum{45.0} & \RISEnum{\underline{28.5}} & \RISEnum{62.0} & \RISEnum{30.0} & \RISEnum{51.3} & \RISEnum{44.9} & \RISEnum{50.8} \\
GPT-Image-2~\cite{openai2025gptimage} & \RISEyes & \RISEnum{59.5} & \RISEnum{64.5} & \RISEnum{\underline{60.5}} & \RISEnum{15.0} & \RISEnum{58.0} & \RISEnum{21.0} & \RISEnum{48.2} & \RISEnum{43.6} & \RISEnum{47.8} \\
GPT-Image-2.5 Flare~\cite{openai2025gptimage} & \RISEyes & \RISEnum{61.0} & \RISEnum{63.0} & \RISEnum{56.5} & \RISEnum{11.0} & \RISEnum{57.0} & \RISEnum{20.0} & \RISEnum{46.6} & \RISEnum{38.5} & \RISEnum{46.0} \\
Luma Uni-1~\cite{luma2026uni1} & \RISEyes & \RISEnum{61.0} & \RISEnum{57.0} & \RISEnum{50.5} & \RISEnum{18.0} & \RISEnum{55.0} & \RISEnum{24.0} & \RISEnum{45.7} & \RISEnum{39.7} & \RISEnum{45.2} \\
Qwen-Image-3.0~\cite{zhao2026qwen} & \RISEyes & \RISEnum{58.5} & \RISEnum{47.5} & \RISEnum{41.0} & \RISEnum{\textbf{36.0}} & \RISEnum{49.0} & \RISEnum{34.0} & \RISEnum{44.5} & \RISEnum{\underline{50.0}} & \RISEnum{44.9} \\
Nano Banana 2 Lite~\cite{google2026nanobanana} & \RISEyes & \RISEnum{60.5} & \RISEnum{61.5} & \RISEnum{46.5} & \RISEnum{15.0} & \RISEnum{62.0} & \RISEnum{12.0} & \RISEnum{44.5} & \RISEnum{39.7} & \RISEnum{44.1} \\
FLUX.2 Max~\cite{bfl2025flux2} & \RISEyes & \RISEnum{57.0} & \RISEnum{57.0} & \RISEnum{42.5} & \RISEnum{12.5} & \RISEnum{61.0} & \RISEnum{18.0} & \RISEnum{41.5} & \RISEnum{43.6} & \RISEnum{41.7} \\
GPT-Image-1.5~\cite{openai2025gptimage} & \RISEyes & \RISEnum{57.5} & \RISEnum{62.5} & \RISEnum{39.5} & \RISEnum{8.0} & \RISEnum{59.0} & \RISEnum{12.0} & \RISEnum{41.2} & \RISEnum{33.3} & \RISEnum{40.6} \\
FLUX.2 Pro~\cite{bfl2025flux2} & \RISEyes & \RISEnum{53.5} & \RISEnum{48.5} & \RISEnum{36.5} & \RISEnum{10.5} & \RISEnum{61.0} & \RISEnum{18.0} & \RISEnum{37.3} & \RISEnum{42.3} & \RISEnum{37.7} \\
Nano Banana~\cite{google2026nanobanana} & \RISEyes & \RISEnum{52.0} & \RISEnum{52.5} & \RISEnum{39.5} & \RISEnum{11.0} & \RISEnum{47.0} & \RISEnum{11.0} & \RISEnum{36.8} & \RISEnum{37.2} & \RISEnum{36.8} \\
HunyuanImage-3.5~\cite{cao2025hunyuanimage} & \RISEyes & \RISEnum{43.5} & \RISEnum{46.0} & \RISEnum{43.5} & \RISEnum{14.0} & \RISEnum{37.0} & \RISEnum{17.0} & \RISEnum{34.5} & \RISEnum{38.5} & \RISEnum{34.8} \\
Wan2.7-Image-Pro~\cite{mao2026wan} & \RISEyes & \RISEnum{41.5} & \RISEnum{52.0} & \RISEnum{32.5} & \RISEnum{10.5} & \RISEnum{49.0} & \RISEnum{13.0} & \RISEnum{36.3} & \RISEnum{0.0} & \RISEnum{33.5} \\
GPT-Image-1~\cite{openai2025gptimage} & \RISEyes & \RISEnum{51.0} & \RISEnum{49.0} & \RISEnum{29.5} & \RISEnum{5.5} & \RISEnum{41.0} & \RISEnum{7.0} & \RISEnum{32.3} & \RISEnum{25.6} & \RISEnum{31.8} \\
Qwen-Image-2.0~\cite{zhao2026qwen} & \RISEyes & \RISEnum{44.0} & \RISEnum{47.0} & \RISEnum{36.0} & \RISEnum{6.5} & \RISEnum{36.0} & \RISEnum{0.0} & \RISEnum{30.2} & \RISEnum{32.1} & \RISEnum{30.3} \\
Seedream 4.5~\cite{seedream2025seedream} & \RISEyes & \RISEnum{42.5} & \RISEnum{42.0} & \RISEnum{26.5} & \RISEnum{7.5} & \RISEnum{36.0} & \RISEnum{6.0} & \RISEnum{28.0} & \RISEnum{26.9} & \RISEnum{27.9} \\
Seedream 5.0~\cite{seedream2025seedream} & \RISEyes & \RISEnum{42.5} & \RISEnum{43.0} & \RISEnum{26.0} & \RISEnum{5.5} & \RISEnum{30.0} & \RISEnum{3.0} & \RISEnum{27.2} & \RISEnum{20.5} & \RISEnum{26.7} \\
Step Image Edit 2~\cite{liu2025step1x} & \RISEno & \RISEnum{18.5} & \RISEnum{22.9} & \RISEnum{26.4} & \RISEnum{1.7} & \RISEnum{10.0} & \RISEnum{1.0} & \RISEnum{14.9} & \RISEnum{\textemdash} & \RISEnum{\textemdash} \\
\midrule[0.4pt]
\rowcolor{RISEGroup}
\multicolumn{11}{l}{\rule{0pt}{10pt}\raisebox{1pt}[0pt][0pt]{\textbf{Open-source Models}}\rule[-2pt]{0pt}{0pt}} \\
\midrule[0.4pt]
HunyuanImage-3.0-Instruct~\cite{cao2025hunyuanimage} & \RISEyes & \RISEnum{\textbf{30.5}} & \RISEnum{\textbf{37.5}} & \RISEnum{\textbf{29.5}} & \RISEnum{\textbf{7.0}} & \RISEnum{\textbf{21.0}} & \RISEnum{0.0} & \RISEnum{\textbf{22.5}} & \RISEnum{29.5} & \RISEnum{\textbf{23.0}} \\
Qwen-Image-Edit-2511~\cite{wu2025qwen} & \RISEyes & \RISEnum{22.0} & \RISEnum{24.5} & \RISEnum{23.5} & \RISEnum{4.5} & \RISEnum{17.0} & \RISEnum{\textbf{5.0}} & \RISEnum{\underline{16.9}} & \RISEnum{19.2} & \RISEnum{\underline{17.1}} \\
Qwen-Image-2.1~\cite{zhao2026qwen} & \RISEyes & \RISEnum{21.5} & \RISEnum{25.5} & \RISEnum{24.5} & \RISEnum{2.0} & \RISEnum{16.0} & \RISEnum{0.0} & \RISEnum{15.0} & \RISEnum{\underline{32.1}} & \RISEnum{16.3} \\
FireRed-Image-Edit-1.0~\cite{team2026firered} & \RISEyes & \RISEnum{22.0} & \RISEnum{18.0} & \RISEnum{\underline{28.0}} & \RISEnum{2.0} & \RISEnum{\textbf{21.0}} & \RISEnum{2.0} & \RISEnum{16.1} & \RISEnum{19.2} & \RISEnum{16.3} \\
FireRed-Image-Edit-1.1~\cite{team2026firered} & \RISEyes & \RISEnum{22.5} & \RISEnum{23.5} & \RISEnum{22.5} & \RISEnum{1.5} & \RISEnum{16.0} & \RISEnum{2.0} & \RISEnum{14.8} & \RISEnum{28.2} & \RISEnum{15.8} \\
Qwen-Image-Edit-2509~\cite{wu2025qwen} & \RISEyes & \RISEnum{21.0} & \RISEnum{24.5} & \RISEnum{20.0} & \RISEnum{0.5} & \RISEnum{16.0} & \RISEnum{0.0} & \RISEnum{15.0} & \RISEnum{12.8} & \RISEnum{14.8} \\
HiDream-O1-Image~\cite{cai2026hidream} & \RISEyes & \RISEnum{\underline{27.5}} & \RISEnum{18.0} & \RISEnum{20.5} & \RISEnum{3.0} & \RISEnum{5.0} & \RISEnum{0.0} & \RISEnum{13.6} & \RISEnum{23.1} & \RISEnum{14.3} \\
FLUX.2 Dev~\cite{bfl2025flux2} & \RISEyes & \RISEnum{19.5} & \RISEnum{\underline{27.5}} & \RISEnum{17.0} & \RISEnum{1.0} & \RISEnum{10.0} & \RISEnum{1.0} & \RISEnum{12.8} & \RISEnum{29.5} & \RISEnum{14.1} \\
FLUX.2 Klein 9B~\cite{bfl2025flux2} & \RISEyes & \RISEnum{16.0} & \RISEnum{16.5} & \RISEnum{22.0} & \RISEnum{0.5} & \RISEnum{17.0} & \RISEnum{2.0} & \RISEnum{11.4} & \RISEnum{30.8} & \RISEnum{12.9} \\
JoyAI-Image-Edit-Plus~\cite{song2026joyai} & \RISEyes & \RISEnum{16.5} & \RISEnum{16.0} & \RISEnum{20.0} & \RISEnum{4.0} & \RISEnum{13.0} & \RISEnum{\underline{3.0}} & \RISEnum{11.7} & \RISEnum{26.9} & \RISEnum{12.9} \\
SenseNova-U1-A3B-MoT-SFT~\cite{diao2026sensenova} & \RISEyes & \RISEnum{20.5} & \RISEnum{25.5} & \RISEnum{2.5} & \RISEnum{3.0} & \RISEnum{17.0} & \RISEnum{0.0} & \RISEnum{12.9} & \RISEnum{1.3} & \RISEnum{12.0} \\
UniPic 3.0~\cite{wei2026skywork} & \RISEyes & \RISEnum{18.5} & \RISEnum{17.5} & \RISEnum{18.0} & \RISEnum{0.5} & \RISEnum{9.0} & \RISEnum{1.0} & \RISEnum{11.1} & \RISEnum{21.8} & \RISEnum{11.9} \\
BAGEL (w/ CoT)~\cite{deng2025bagel} & \RISEyes & \RISEnum{18.5} & \RISEnum{16.0} & \RISEnum{16.0} & \RISEnum{1.5} & \RISEnum{10.0} & \RISEnum{0.0} & \RISEnum{10.4} & \RISEnum{23.1} & \RISEnum{11.4} \\
SenseNova-U1-8B-MoT-SFT~\cite{diao2026sensenova} & \RISEyes & \RISEnum{16.5} & \RISEnum{22.5} & \RISEnum{7.5} & \RISEnum{3.5} & \RISEnum{13.0} & \RISEnum{0.0} & \RISEnum{12.3} & \RISEnum{0.0} & \RISEnum{11.3} \\
FLUX.2 Klein 4B~\cite{bfl2025flux2} & \RISEyes & \RISEnum{15.5} & \RISEnum{14.5} & \RISEnum{17.0} & \RISEnum{2.5} & \RISEnum{8.0} & \RISEnum{1.0} & \RISEnum{8.7} & \RISEnum{\textbf{35.9}} & \RISEnum{10.8} \\
UniWorld-V2~\cite{li2025uniworld} & \RISEyes & \RISEnum{10.0} & \RISEnum{18.0} & \RISEnum{12.0} & \RISEnum{0.5} & \RISEnum{13.0} & \RISEnum{1.0} & \RISEnum{9.9} & \RISEnum{5.1} & \RISEnum{9.5} \\
BAGEL~\cite{deng2025bagel} & \RISEyes & \RISEnum{6.5} & \RISEnum{6.0} & \RISEnum{8.0} & \RISEnum{1.0} & \RISEnum{2.0} & \RISEnum{0.0} & \RISEnum{3.1} & \RISEnum{20.5} & \RISEnum{4.5} \\
OmniGen2~\cite{wu2025omnigen2} & \RISEyes & \RISEnum{7.0} & \RISEnum{6.0} & \RISEnum{5.5} & \RISEnum{0.0} & \RISEnum{8.0} & \RISEnum{0.0} & \RISEnum{3.7} & \RISEnum{14.1} & \RISEnum{4.5} \\
OmniGen~\cite{xiao2024omnigen} & \RISEyes & \RISEnum{4.5} & \RISEnum{4.0} & \RISEnum{5.0} & \RISEnum{1.5} & \RISEnum{1.0} & \RISEnum{0.0} & \RISEnum{2.7} & \RISEnum{7.7} & \RISEnum{3.1} \\
Emu2-Gen~\cite{sun2023emu2chat} & \RISEyes & \RISEnum{4.0} & \RISEnum{1.5} & \RISEnum{0.5} & \RISEnum{0.5} & \RISEnum{1.0} & \RISEnum{0.0} & \RISEnum{1.4} & \RISEnum{1.3} & \RISEnum{1.4} \\
JoyAI-Image-Edit~\cite{song2026joyai} & \RISEno & \RISEnum{17.9} & \RISEnum{24.0} & \RISEnum{22.0} & \RISEnum{2.2} & \RISEnum{16.0} & \RISEnum{2.0} & \RISEnum{15.0} & \RISEnum{\textemdash} & \RISEnum{\textemdash} \\
Step1X-Edit (thinking)~\cite{liu2025step1x} & \RISEno & \RISEnum{19.6} & \RISEnum{27.1} & \RISEnum{14.3} & \RISEnum{0.6} & \RISEnum{16.0} & \RISEnum{2.0} & \RISEnum{14.1} & \RISEnum{\textemdash} & \RISEnum{\textemdash} \\
Boogu-Image-0.1-Edit~\cite{chen2026boogu} & \RISEno & \RISEnum{15.5} & \RISEnum{21.9} & \RISEnum{17.6} & \RISEnum{\underline{6.1}} & \RISEnum{11.0} & \RISEnum{\underline{3.0}} & \RISEnum{13.6} & \RISEnum{\textemdash} & \RISEnum{\textemdash} \\
Step1X-Edit (thinking + reflection)~\cite{liu2025step1x} & \RISEno & \RISEnum{21.4} & \RISEnum{22.9} & \RISEnum{9.9} & \RISEnum{1.7} & \RISEnum{\underline{19.0}} & \RISEnum{1.0} & \RISEnum{13.1} & \RISEnum{\textemdash} & \RISEnum{\textemdash} \\
LongCat-Image-Edit~\cite{team2025longcat} & \RISEno & \RISEnum{10.1} & \RISEnum{20.8} & \RISEnum{15.4} & \RISEnum{1.1} & \RISEnum{9.0} & \RISEnum{0.0} & \RISEnum{10.4} & \RISEnum{\textemdash} & \RISEnum{\textemdash} \\
Step1X-Edit (base)~\cite{liu2025step1x} & \RISEno & \RISEnum{13.1} & \RISEnum{18.8} & \RISEnum{10.4} & \RISEnum{1.1} & \RISEnum{14.0} & \RISEnum{1.0} & \RISEnum{10.2} & \RISEnum{\textemdash} & \RISEnum{\textemdash} \\
InternVL-U (w/ CoT)~\cite{tian2026internvl} & \RISEno & \RISEnum{8.9} & \RISEnum{14.1} & \RISEnum{4.9} & \RISEnum{1.1} & \RISEnum{16.0} & \RISEnum{0.0} & \RISEnum{7.5} & \RISEnum{\textemdash} & \RISEnum{\textemdash} \\
FLUX.1 Kontext Dev~\cite{labs2025flux1kontextflowmatching} & \RISEno & \RISEnum{5.4} & \RISEnum{9.4} & \RISEnum{12.1} & \RISEnum{1.7} & \RISEnum{7.0} & \RISEnum{0.0} & \RISEnum{6.4} & \RISEnum{\textemdash} & \RISEnum{\textemdash} \\
LLaDA-Image~\cite{chen2026llada} & \RISEno & \RISEnum{7.1} & \RISEnum{4.7} & \RISEnum{3.8} & \RISEnum{0.0} & \RISEnum{4.0} & \RISEnum{0.0} & \RISEnum{3.5} & \RISEnum{\textemdash} & \RISEnum{\textemdash} \\
Ovis-U1~\cite{wang2025ovis} & \RISEno & \RISEnum{4.8} & \RISEnum{6.3} & \RISEnum{2.2} & \RISEnum{1.7} & \RISEnum{2.0} & \RISEnum{0.0} & \RISEnum{3.1} & \RISEnum{\textemdash} & \RISEnum{\textemdash} \\
InternVL-U~\cite{tian2026internvl} & \RISEno & \RISEnum{3.6} & \RISEnum{2.6} & \RISEnum{3.8} & \RISEnum{0.6} & \RISEnum{5.0} & \RISEnum{0.0} & \RISEnum{2.6} & \RISEnum{\textemdash} & \RISEnum{\textemdash} \\
UniPic2-SD3.5M-Kontext-2B~\cite{wei2025skywork} & \RISEno & \RISEnum{1.2} & \RISEnum{2.1} & \RISEnum{0.5} & \RISEnum{0.0} & \RISEnum{3.0} & \RISEnum{0.0} & \RISEnum{1.1} & \RISEnum{\textemdash} & \RISEnum{\textemdash} \\
Lumina-DiMOO~\cite{xin2025lumina} & \RISEno & \RISEnum{0.6} & \RISEnum{1.6} & \RISEnum{0.0} & \RISEnum{0.6} & \RISEnum{4.0} & \RISEnum{0.0} & \RISEnum{1.0} & \RISEnum{\textemdash} & \RISEnum{\textemdash} \\
UniWorld-V1~\cite{lin2025uniworld} & \RISEno & \RISEnum{1.2} & \RISEnum{1.0} & \RISEnum{0.5} & \RISEnum{0.6} & \RISEnum{1.0} & \RISEnum{0.0} & \RISEnum{0.8} & \RISEnum{\textemdash} & \RISEnum{\textemdash} \\
\midrule[0.4pt]
\rowcolor{RISEGroup}
\multicolumn{11}{l}{\rule{0pt}{10pt}\raisebox{1pt}[0pt][0pt]{\textbf{Agentic Methods}}\rule[-2pt]{0pt}{0pt}} \\
\midrule[0.4pt]
\textbf{RISE-Agent (Ours)} & \RISEyes & \RISEnum{\textbf{58.5}} & \RISEnum{\textbf{58.0}} & \RISEnum{\textbf{37.5}} & \RISEnum{\textbf{39.0}} & \RISEnum{\textbf{66.0}} & \RISEnum{\textbf{28.0}} & \RISEnum{\textbf{48.3}} & \RISEnum{\textbf{44.9}} & \RISEnum{\textbf{48.0}} \\
MindBrush~\cite{he2026mind} & \RISEno & \RISEnum{35.1} & \RISEnum{\underline{33.3}} & \RISEnum{\underline{16.5}} & \RISEnum{\underline{8.9}} & \RISEnum{\underline{32.0}} & \RISEnum{0.0} & \RISEnum{\underline{21.8}} & \RISEnum{\textemdash} & \RISEnum{\textemdash} \\
InterleaveThinker~\cite{zheng2026interleavethinker} & \RISEno & \RISEnum{\underline{38.7}} & \RISEnum{30.7} & \RISEnum{15.9} & \RISEnum{1.7} & \RISEnum{26.0} & \RISEnum{\underline{8.0}} & \RISEnum{20.6} & \RISEnum{\textemdash} & \RISEnum{\textemdash} \\
IntentEdit~\cite{zhang2026intentedit} & \RISEno & \RISEnum{23.2} & \RISEnum{24.0} & \RISEnum{11.5} & \RISEnum{1.1} & \RISEnum{14.0} & \RISEnum{0.0} & \RISEnum{13.2} & \RISEnum{\textemdash} & \RISEnum{\textemdash} \\
ImAgent~\cite{wang2025imagent} & \RISEno & \RISEnum{5.4} & \RISEnum{4.7} & \RISEnum{4.4} & \RISEnum{0.6} & \RISEnum{5.0} & \RISEnum{0.0} & \RISEnum{3.5} & \RISEnum{\textemdash} & \RISEnum{\textemdash} \\
\bottomrule[0.8pt]
\end{tabular}
\par\vspace{4pt}
\parbox{\textwidth}{\scriptsize
Some models do not support multi-image inputs and are evaluated only on single-image cases. \RISEyes{} denotes evaluation in both single- and multi-image settings, while \RISEno{} denotes single-image-only evaluation. Overall results are reported for single-image cases (\textbf{Single}), multi-image cases (\textbf{Multi}), and the full set (\textbf{Total}); \textemdash{} denotes settings not evaluated.}
\endgroup
\end{table*}

\begin{table*}[!tp]
\caption{\textbf{Overall performance on RISEBench++ (CN).} Models are grouped into closed-source, open-source, and agentic methods. \textbf{Bold} and \underline{underlined} values indicate the best and second-best results within each group, respectively.}
\label{tab:main_result_cn}
\centering
\begingroup
\fontsize{7}{8.6}\selectfont
\setlength{\tabcolsep}{2pt}
\renewcommand{\arraystretch}{0.98}

\definecolor{RISECNGroup}{RGB}{239,239,239}
\def\RISECNyes{\textcolor{green!50!black}{\ding{51}}}
\def\RISECNno{\textcolor{red!65!black}{\ding{55}}}
\def\RISECNNum#1{\makebox[2em][c]{#1}}
\begin{tabular}{
>{\raggedright\arraybackslash}p{\dimexpr .299\textwidth-2\tabcolsep\relax}
>{\centering\arraybackslash}p{\dimexpr .060\textwidth-2\tabcolsep\relax}
>{\centering\arraybackslash}p{\dimexpr .073\textwidth-2\tabcolsep\relax}
>{\centering\arraybackslash}p{\dimexpr .065\textwidth-2\tabcolsep\relax}
>{\centering\arraybackslash}p{\dimexpr .065\textwidth-2\tabcolsep\relax}
>{\centering\arraybackslash}p{\dimexpr .065\textwidth-2\tabcolsep\relax}
>{\centering\arraybackslash}p{\dimexpr .120\textwidth-2\tabcolsep\relax}
>{\centering\arraybackslash}p{\dimexpr .068\textwidth-2\tabcolsep\relax}
>{\centering\arraybackslash}p{\dimexpr .062\textwidth-2\tabcolsep\relax}
>{\centering\arraybackslash}p{\dimexpr .060\textwidth-2\tabcolsep\relax}
>{\centering\arraybackslash}p{\dimexpr .063\textwidth-2\tabcolsep\relax}
}
\toprule[0.8pt]
\multirow{2}{*}[-2pt]{\textbf{Model}} &
\multicolumn{1}{c}{\multirow{2}{*}[-2pt]{\shortstack{\textbf{Multi-}\\\textbf{img}}}} &
\multicolumn{6}{c}{\textbf{Reasoning Tasks}} &
\multicolumn{3}{c}{\textbf{Overall}} \\
\cmidrule(lr){3-8}\cmidrule(lr){9-11}
 & &
\multicolumn{1}{c}{\textbf{Temporal}} &
\multicolumn{1}{c}{\textbf{Causal}} &
\multicolumn{1}{c}{\textbf{Spatial}} &
\multicolumn{1}{c}{\textbf{Logical}} &
\multicolumn{1}{c}{\textbf{Counterfactual}} &
\multicolumn{1}{c}{\textbf{Hybrid}} &
\multicolumn{1}{c}{\textbf{Single}} &
\multicolumn{1}{c}{\textbf{Multi}} &
\multicolumn{1}{c}{\textbf{Total}} \\
\midrule[0.5pt]
\rowcolor{RISECNGroup}
\multicolumn{11}{l}{\rule{0pt}{10pt}\raisebox{1pt}[0pt][0pt]{\textbf{Closed-source Models}}\rule[-2pt]{0pt}{0pt}} \\
\midrule[0.4pt]
GPT-Image-2.5 Sunburst~\cite{openai2025gptimage} & \RISECNyes & \RISECNNum{\underline{62.5}} & \RISECNNum{\textbf{70.5}} & \RISECNNum{\textbf{63.0}} & \RISECNNum{29.0} & \RISECNNum{\underline{69.0}} & \RISECNNum{\textbf{36.0}} & \RISECNNum{\textbf{56.0}} & \RISECNNum{\textbf{50.0}} & \RISECNNum{\textbf{55.5}} \\
Nano Banana Pro~\cite{google2026nanobanana} & \RISECNyes & \RISECNNum{\underline{62.5}} & \RISECNNum{67.0} & \RISECNNum{50.5} & \RISECNNum{\underline{38.5}} & \RISECNNum{50.0} & \RISECNNum{27.0} & \RISECNNum{\underline{52.4}} & \RISECNNum{39.7} & \RISECNNum{\underline{51.4}} \\
Nano Banana 2~\cite{google2026nanobanana} & \RISECNyes & \RISECNNum{62.0} & \RISECNNum{\underline{69.0}} & \RISECNNum{50.0} & \RISECNNum{31.5} & \RISECNNum{57.0} & \RISECNNum{28.0} & \RISECNNum{51.7} & \RISECNNum{42.3} & \RISECNNum{51.0} \\
Luma Uni-1~\cite{luma2026uni1} & \RISECNyes & \RISECNNum{\textbf{65.5}} & \RISECNNum{67.0} & \RISECNNum{53.5} & \RISECNNum{17.0} & \RISECNNum{56.0} & \RISECNNum{24.0} & \RISECNNum{49.1} & \RISECNNum{42.3} & \RISECNNum{48.6} \\
GPT-Image-2~\cite{openai2025gptimage} & \RISECNyes & \RISECNNum{57.0} & \RISECNNum{62.0} & \RISECNNum{\underline{54.5}} & \RISECNNum{20.5} & \RISECNNum{61.0} & \RISECNNum{20.0} & \RISECNNum{46.9} & \RISECNNum{47.4} & \RISECNNum{46.9} \\
Qwen-Image-3.0~\cite{zhao2026qwen} & \RISECNyes & \RISECNNum{58.0} & \RISECNNum{50.0} & \RISECNNum{43.0} & \RISECNNum{\textbf{39.0}} & \RISECNNum{49.0} & \RISECNNum{\underline{34.0}} & \RISECNNum{46.1} & \RISECNNum{\underline{48.7}} & \RISECNNum{46.3} \\
GPT-Image-2.5 Flare~\cite{openai2025gptimage} & \RISECNyes & \RISECNNum{52.0} & \RISECNNum{66.0} & \RISECNNum{54.0} & \RISECNNum{14.0} & \RISECNNum{61.0} & \RISECNNum{19.0} & \RISECNNum{46.3} & \RISECNNum{32.1} & \RISECNNum{45.2} \\
FLUX.2 Max~\cite{bfl2025flux2} & \RISECNyes & \RISECNNum{59.5} & \RISECNNum{59.0} & \RISECNNum{42.0} & \RISECNNum{14.0} & \RISECNNum{\textbf{72.0}} & \RISECNNum{18.0} & \RISECNNum{44.4} & \RISECNNum{38.5} & \RISECNNum{43.9} \\
Nano Banana 2 Lite~\cite{google2026nanobanana} & \RISECNyes & \RISECNNum{58.5} & \RISECNNum{60.5} & \RISECNNum{41.5} & \RISECNNum{12.0} & \RISECNNum{59.0} & \RISECNNum{18.0} & \RISECNNum{42.5} & \RISECNNum{38.5} & \RISECNNum{42.2} \\
GPT-Image-1.5~\cite{openai2025gptimage} & \RISECNyes & \RISECNNum{55.0} & \RISECNNum{57.0} & \RISECNNum{35.5} & \RISECNNum{4.5} & \RISECNNum{57.0} & \RISECNNum{10.0} & \RISECNNum{37.9} & \RISECNNum{28.2} & \RISECNNum{37.1} \\
Nano Banana~\cite{google2026nanobanana} & \RISECNyes & \RISECNNum{49.5} & \RISECNNum{53.0} & \RISECNNum{40.0} & \RISECNNum{12.0} & \RISECNNum{52.0} & \RISECNNum{5.0} & \RISECNNum{36.6} & \RISECNNum{37.2} & \RISECNNum{36.6} \\
FLUX.2 Pro~\cite{bfl2025flux2} & \RISECNyes & \RISECNNum{51.0} & \RISECNNum{44.5} & \RISECNNum{38.5} & \RISECNNum{8.0} & \RISECNNum{51.0} & \RISECNNum{11.0} & \RISECNNum{33.5} & \RISECNNum{47.4} & \RISECNNum{34.6} \\
HunyuanImage-3.5~\cite{cao2025hunyuanimage} & \RISECNyes & \RISECNNum{44.0} & \RISECNNum{39.5} & \RISECNNum{45.0} & \RISECNNum{14.0} & \RISECNNum{42.0} & \RISECNNum{13.0} & \RISECNNum{33.3} & \RISECNNum{42.3} & \RISECNNum{34.0} \\
Wan2.7-Image-Pro~\cite{mao2026wan} & \RISECNyes & \RISECNNum{40.5} & \RISECNNum{44.5} & \RISECNNum{37.5} & \RISECNNum{9.5} & \RISECNNum{47.0} & \RISECNNum{13.0} & \RISECNNum{32.4} & \RISECNNum{32.1} & \RISECNNum{32.4} \\
Qwen-Image-2.0~\cite{zhao2026qwen} & \RISECNyes & \RISECNNum{51.5} & \RISECNNum{37.0} & \RISECNNum{30.5} & \RISECNNum{9.0} & \RISECNNum{37.0} & \RISECNNum{1.0} & \RISECNNum{28.7} & \RISECNNum{37.2} & \RISECNNum{29.4} \\
Seedream 4.5~\cite{seedream2025seedream} & \RISECNyes & \RISECNNum{44.0} & \RISECNNum{45.0} & \RISECNNum{26.5} & \RISECNNum{8.5} & \RISECNNum{35.0} & \RISECNNum{10.0} & \RISECNNum{29.9} & \RISECNNum{21.8} & \RISECNNum{29.3} \\
Seedream 5.0~\cite{seedream2025seedream} & \RISECNyes & \RISECNNum{40.5} & \RISECNNum{39.5} & \RISECNNum{28.0} & \RISECNNum{11.5} & \RISECNNum{30.0} & \RISECNNum{7.0} & \RISECNNum{27.7} & \RISECNNum{26.9} & \RISECNNum{27.6} \\
GPT-Image-1~\cite{openai2025gptimage} & \RISECNyes & \RISECNNum{43.5} & \RISECNNum{40.0} & \RISECNNum{24.5} & \RISECNNum{5.5} & \RISECNNum{37.0} & \RISECNNum{8.0} & \RISECNNum{28.1} & \RISECNNum{16.7} & \RISECNNum{27.2} \\
Step Image Edit 2~\cite{liu2025step1x} & \RISECNno & \RISECNNum{14.9} & \RISECNNum{19.3} & \RISECNNum{22.5} & \RISECNNum{2.2} & \RISECNNum{12.0} & \RISECNNum{0.0} & \RISECNNum{12.9} & \RISECNNum{\textemdash} & \RISECNNum{\textemdash} \\
\midrule[0.4pt]
\rowcolor{RISECNGroup}
\multicolumn{11}{l}{\rule{0pt}{10pt}\raisebox{1pt}[0pt][0pt]{\textbf{Open-source Models}}\rule[-2pt]{0pt}{0pt}} \\
\midrule[0.4pt]
HunyuanImage-3.0-Instruct~\cite{cao2025hunyuanimage} & \RISECNyes & \RISECNNum{\textbf{28.0}} & \RISECNNum{\textbf{32.0}} & \RISECNNum{\textbf{35.0}} & \RISECNNum{\textbf{8.5}} & \RISECNNum{\underline{20.0}} & \RISECNNum{0.0} & \RISECNNum{\textbf{22.3}} & \RISECNNum{26.9} & \RISECNNum{\textbf{22.7}} \\
Qwen-Image-2.1~\cite{zhao2026qwen} & \RISECNyes & \RISECNNum{\underline{22.5}} & \RISECNNum{23.0} & \RISECNNum{\underline{25.0}} & \RISECNNum{\underline{6.5}} & \RISECNNum{16.0} & \RISECNNum{0.0} & \RISECNNum{\underline{15.8}} & \RISECNNum{30.8} & \RISECNNum{\underline{17.0}} \\
FireRed-Image-Edit-1.1~\cite{team2026firered} & \RISECNyes & \RISECNNum{21.0} & \RISECNNum{24.5} & \RISECNNum{\underline{25.0}} & \RISECNNum{1.0} & \RISECNNum{15.0} & \RISECNNum{\textbf{4.0}} & \RISECNNum{15.6} & \RISECNNum{23.1} & \RISECNNum{16.2} \\
FLUX.2 Dev~\cite{bfl2025flux2} & \RISECNyes & \RISECNNum{21.0} & \RISECNNum{26.0} & \RISECNNum{16.5} & \RISECNNum{1.5} & \RISECNNum{13.0} & \RISECNNum{\textbf{4.0}} & \RISECNNum{13.3} & \RISECNNum{30.8} & \RISECNNum{14.7} \\
Qwen-Image-Edit-2511~\cite{wu2025qwen} & \RISECNyes & \RISECNNum{19.5} & \RISECNNum{19.5} & \RISECNNum{23.0} & \RISECNNum{2.0} & \RISECNNum{10.0} & \RISECNNum{0.0} & \RISECNNum{13.7} & \RISECNNum{15.4} & \RISECNNum{13.8} \\
Qwen-Image-Edit-2509~\cite{wu2025qwen} & \RISECNyes & \RISECNNum{17.5} & \RISECNNum{19.0} & \RISECNNum{21.5} & \RISECNNum{1.5} & \RISECNNum{15.0} & \RISECNNum{\underline{3.0}} & \RISECNNum{13.7} & \RISECNNum{14.1} & \RISECNNum{13.7} \\
FireRed-Image-Edit-1.0~\cite{team2026firered} & \RISECNyes & \RISECNNum{19.5} & \RISECNNum{16.5} & \RISECNNum{\underline{25.0}} & \RISECNNum{0.5} & \RISECNNum{11.0} & \RISECNNum{2.0} & \RISECNNum{13.2} & \RISECNNum{17.9} & \RISECNNum{13.6} \\
SenseNova-U1-A3B-MoT-SFT~\cite{diao2026sensenova} & \RISECNyes & \RISECNNum{21.5} & \RISECNNum{\underline{28.0}} & \RISECNNum{7.0} & \RISECNNum{3.0} & \RISECNNum{11.0} & \RISECNNum{0.0} & \RISECNNum{14.0} & \RISECNNum{1.3} & \RISECNNum{13.0} \\
JoyAI-Image-Edit-Plus~\cite{song2026joyai} & \RISECNyes & \RISECNNum{14.5} & \RISECNNum{19.5} & \RISECNNum{19.0} & \RISECNNum{5.0} & \RISECNNum{9.0} & \RISECNNum{\textbf{4.0}} & \RISECNNum{11.2} & \RISECNNum{\underline{33.3}} & \RISECNNum{12.9} \\
FLUX.2 Klein 9B~\cite{bfl2025flux2} & \RISECNyes & \RISECNNum{22.0} & \RISECNNum{16.0} & \RISECNNum{16.5} & \RISECNNum{3.5} & \RISECNNum{9.0} & \RISECNNum{\underline{3.0}} & \RISECNNum{11.1} & \RISECNNum{\underline{33.3}} & \RISECNNum{12.8} \\
BAGEL (w/ CoT)~\cite{deng2025bagel} & \RISECNyes & \RISECNNum{19.0} & \RISECNNum{18.0} & \RISECNNum{16.5} & \RISECNNum{0.0} & \RISECNNum{6.0} & \RISECNNum{0.0} & \RISECNNum{9.8} & \RISECNNum{29.5} & \RISECNNum{11.3} \\
HiDream-O1-Image~\cite{cai2026hidream} & \RISECNyes & \RISECNNum{18.5} & \RISECNNum{13.5} & \RISECNNum{17.5} & \RISECNNum{0.5} & \RISECNNum{7.0} & \RISECNNum{0.0} & \RISECNNum{8.8} & \RISECNNum{\underline{33.3}} & \RISECNNum{10.7} \\
FLUX.2 Klein 4B~\cite{bfl2025flux2} & \RISECNyes & \RISECNNum{16.5} & \RISECNNum{12.5} & \RISECNNum{14.5} & \RISECNNum{4.0} & \RISECNNum{10.0} & \RISECNNum{1.0} & \RISECNNum{8.0} & \RISECNNum{\textbf{41.0}} & \RISECNNum{10.6} \\
UniWorld-V2~\cite{li2025uniworld} & \RISECNyes & \RISECNNum{13.5} & \RISECNNum{17.0} & \RISECNNum{15.5} & \RISECNNum{0.0} & \RISECNNum{12.0} & \RISECNNum{1.0} & \RISECNNum{10.7} & \RISECNNum{7.7} & \RISECNNum{10.5} \\
SenseNova-U1-8B-MoT-SFT~\cite{diao2026sensenova} & \RISECNyes & \RISECNNum{16.0} & \RISECNNum{21.0} & \RISECNNum{3.5} & \RISECNNum{3.5} & \RISECNNum{15.0} & \RISECNNum{0.0} & \RISECNNum{11.1} & \RISECNNum{1.3} & \RISECNNum{10.3} \\
UniPic 3.0~\cite{wei2026skywork} & \RISECNyes & \RISECNNum{7.0} & \RISECNNum{14.0} & \RISECNNum{12.5} & \RISECNNum{0.0} & \RISECNNum{7.0} & \RISECNNum{0.0} & \RISECNNum{7.5} & \RISECNNum{6.4} & \RISECNNum{7.4} \\
BAGEL~\cite{deng2025bagel} & \RISECNyes & \RISECNNum{10.0} & \RISECNNum{7.0} & \RISECNNum{10.0} & \RISECNNum{1.0} & \RISECNNum{8.0} & \RISECNNum{0.0} & \RISECNNum{4.8} & \RISECNNum{25.6} & \RISECNNum{6.4} \\
OmniGen2~\cite{wu2025omnigen2} & \RISECNyes & \RISECNNum{6.0} & \RISECNNum{5.5} & \RISECNNum{6.0} & \RISECNNum{0.0} & \RISECNNum{6.0} & \RISECNNum{0.0} & \RISECNNum{2.9} & \RISECNNum{17.9} & \RISECNNum{4.1} \\
OmniGen~\cite{xiao2024omnigen} & \RISECNyes & \RISECNNum{7.0} & \RISECNNum{2.5} & \RISECNNum{5.0} & \RISECNNum{3.0} & \RISECNNum{4.0} & \RISECNNum{0.0} & \RISECNNum{2.7} & \RISECNNum{17.9} & \RISECNNum{3.9} \\
Emu2-Gen~\cite{sun2023emu2chat} & \RISECNyes & \RISECNNum{2.5} & \RISECNNum{1.0} & \RISECNNum{1.0} & \RISECNNum{0.0} & \RISECNNum{2.0} & \RISECNNum{0.0} & \RISECNNum{1.0} & \RISECNNum{2.6} & \RISECNNum{1.1} \\
JoyAI-Image-Edit~\cite{song2026joyai} & \RISECNno & \RISECNNum{13.7} & \RISECNNum{24.0} & \RISECNNum{15.4} & \RISECNNum{1.1} & \RISECNNum{\textbf{21.0}} & \RISECNNum{\textbf{4.0}} & \RISECNNum{13.4} & \RISECNNum{\textemdash} & \RISECNNum{\textemdash} \\
Boogu-Image-0.1-Edit~\cite{chen2026boogu} & \RISECNno & \RISECNNum{13.1} & \RISECNNum{22.4} & \RISECNNum{17.0} & \RISECNNum{2.2} & \RISECNNum{17.0} & \RISECNNum{1.0} & \RISECNNum{12.8} & \RISECNNum{\textemdash} & \RISECNNum{\textemdash} \\
Step1X-Edit (thinking + reflection)~\cite{liu2025step1x} & \RISECNno & \RISECNNum{17.3} & \RISECNNum{18.2} & \RISECNNum{11.0} & \RISECNNum{1.1} & \RISECNNum{14.0} & \RISECNNum{2.0} & \RISECNNum{11.1} & \RISECNNum{\textemdash} & \RISECNNum{\textemdash} \\
Step1X-Edit (thinking)~\cite{liu2025step1x} & \RISECNno & \RISECNNum{14.3} & \RISECNNum{15.1} & \RISECNNum{13.2} & \RISECNNum{2.2} & \RISECNNum{17.0} & \RISECNNum{2.0} & \RISECNNum{10.8} & \RISECNNum{\textemdash} & \RISECNNum{\textemdash} \\
Step1X-Edit (base)~\cite{liu2025step1x} & \RISECNno & \RISECNNum{13.1} & \RISECNNum{16.1} & \RISECNNum{12.6} & \RISECNNum{0.6} & \RISECNNum{15.0} & \RISECNNum{1.0} & \RISECNNum{10.1} & \RISECNNum{\textemdash} & \RISECNNum{\textemdash} \\
LongCat-Image-Edit~\cite{team2025longcat} & \RISECNno & \RISECNNum{13.7} & \RISECNNum{16.1} & \RISECNNum{13.2} & \RISECNNum{1.1} & \RISECNNum{9.0} & \RISECNNum{0.0} & \RISECNNum{9.7} & \RISECNNum{\textemdash} & \RISECNNum{\textemdash} \\
FLUX.1 Kontext Dev~\cite{labs2025flux1kontextflowmatching} & \RISECNno & \RISECNNum{3.0} & \RISECNNum{3.6} & \RISECNNum{3.8} & \RISECNNum{3.3} & \RISECNNum{7.0} & \RISECNNum{1.0} & \RISECNNum{3.6} & \RISECNNum{\textemdash} & \RISECNNum{\textemdash} \\
LLaDA-Image~\cite{chen2026llada} & \RISECNno & \RISECNNum{4.8} & \RISECNNum{5.7} & \RISECNNum{2.2} & \RISECNNum{0.6} & \RISECNNum{8.0} & \RISECNNum{0.0} & \RISECNNum{3.5} & \RISECNNum{\textemdash} & \RISECNNum{\textemdash} \\
InternVL-U (w/ CoT)~\cite{tian2026internvl} & \RISECNno & \RISECNNum{6.0} & \RISECNNum{3.1} & \RISECNNum{2.2} & \RISECNNum{0.0} & \RISECNNum{4.0} & \RISECNNum{1.0} & \RISECNNum{2.7} & \RISECNNum{\textemdash} & \RISECNNum{\textemdash} \\
Ovis-U1~\cite{wang2025ovis} & \RISECNno & \RISECNNum{3.6} & \RISECNNum{5.7} & \RISECNNum{2.2} & \RISECNNum{0.6} & \RISECNNum{2.0} & \RISECNNum{0.0} & \RISECNNum{2.6} & \RISECNNum{\textemdash} & \RISECNNum{\textemdash} \\
InternVL-U~\cite{tian2026internvl} & \RISECNno & \RISECNNum{3.6} & \RISECNNum{4.2} & \RISECNNum{1.1} & \RISECNNum{0.6} & \RISECNNum{5.0} & \RISECNNum{0.0} & \RISECNNum{2.4} & \RISECNNum{\textemdash} & \RISECNNum{\textemdash} \\
Lumina-DiMOO~\cite{xin2025lumina} & \RISECNno & \RISECNNum{1.8} & \RISECNNum{2.6} & \RISECNNum{2.2} & \RISECNNum{2.2} & \RISECNNum{4.0} & \RISECNNum{0.0} & \RISECNNum{2.2} & \RISECNNum{\textemdash} & \RISECNNum{\textemdash} \\
UniWorld-V1~\cite{lin2025uniworld} & \RISECNno & \RISECNNum{3.0} & \RISECNNum{1.0} & \RISECNNum{1.6} & \RISECNNum{0.0} & \RISECNNum{3.0} & \RISECNNum{0.0} & \RISECNNum{1.4} & \RISECNNum{\textemdash} & \RISECNNum{\textemdash} \\
UniPic2-SD3.5M-Kontext-2B~\cite{wei2025skywork} & \RISECNno & \RISECNNum{0.0} & \RISECNNum{1.0} & \RISECNNum{1.6} & \RISECNNum{1.1} & \RISECNNum{2.0} & \RISECNNum{1.0} & \RISECNNum{1.1} & \RISECNNum{\textemdash} & \RISECNNum{\textemdash} \\
\midrule[0.4pt]
\rowcolor{RISECNGroup}
\multicolumn{11}{l}{\rule{0pt}{10pt}\raisebox{1pt}[0pt][0pt]{\textbf{Agentic Methods}}\rule[-2pt]{0pt}{0pt}} \\
\midrule[0.4pt]
\textbf{RISE-Agent (Ours)} & \RISECNyes & \RISECNNum{\textbf{51.0}} & \RISECNNum{\textbf{57.0}} & \RISECNNum{\textbf{34.0}} & \RISECNNum{\textbf{25.0}} & \RISECNNum{\textbf{54.0}} & \RISECNNum{\textbf{15.0}} & \RISECNNum{\textbf{40.6}} & \RISECNNum{\textbf{37.2}} & \RISECNNum{\textbf{40.3}} \\
MindBrush~\cite{he2026mind} & \RISECNno & \RISECNNum{\underline{43.5}} & \RISECNNum{\underline{33.9}} & \RISECNNum{\underline{19.8}} & \RISECNNum{\underline{4.4}} & \RISECNNum{\underline{36.0}} & \RISECNNum{0.0} & \RISECNNum{\underline{23.6}} & \RISECNNum{\textemdash} & \RISECNNum{\textemdash} \\
InterleaveThinker~\cite{zheng2026interleavethinker} & \RISECNno & \RISECNNum{25.0} & \RISECNNum{28.1} & \RISECNNum{19.2} & \RISECNNum{0.6} & \RISECNNum{28.0} & \RISECNNum{\underline{4.0}} & \RISECNNum{17.8} & \RISECNNum{\textemdash} & \RISECNNum{\textemdash} \\
IntentEdit~\cite{zhang2026intentedit} & \RISECNno & \RISECNNum{22.6} & \RISECNNum{24.0} & \RISECNNum{14.3} & \RISECNNum{1.7} & \RISECNNum{14.0} & \RISECNNum{0.0} & \RISECNNum{13.8} & \RISECNNum{\textemdash} & \RISECNNum{\textemdash} \\
ImAgent~\cite{wang2025imagent} & \RISECNno & \RISECNNum{4.2} & \RISECNNum{3.6} & \RISECNNum{2.2} & \RISECNNum{0.0} & \RISECNNum{2.0} & \RISECNNum{0.0} & \RISECNNum{2.2} & \RISECNNum{\textemdash} & \RISECNNum{\textemdash} \\
\bottomrule[0.8pt]
\end{tabular}
\par\vspace{4pt}
\parbox{\textwidth}{\scriptsize
Some models do not support multi-image inputs and are evaluated only on single-image cases. \RISEyes{} denotes evaluation in both single- and multi-image settings, while \RISEno{} denotes single-image-only evaluation. Overall results are reported for single-image cases (\textbf{Single}), multi-image cases (\textbf{Multi}), and the full set (\textbf{Total}); \textemdash{} denotes settings not evaluated.}
\endgroup
\end{table*}

We report accuracy scores on a 100-point scale for English and Chinese instructions in Table~\ref{tab:main_result_en} and Table~\ref{tab:main_result_cn}, respectively. In our LMM-as-a-Judge evaluation pipeline, Gemini-3-Flash evaluates instruction reasoning and appearance consistency, while Gemini-3.1-Flash-Lite evaluates visual plausibility.

Among the evaluated closed-source models, GPT-Image-2.5 Sunburst demonstrates the highest performance on RISEBench++, achieving an overall accuracy of 56.6\% under English instructions. Nano Banana Pro and Nano Banana 2 rank second and third, respectively, reaching accuracies of 54.6\% and 50.8\%. GPT-Image-2, GPT-Image-2.5 Flare, Luma Uni-1 and Qwen-Image-3.0 also achieve relatively strong performance, with overall accuracies around 45.0\%. However, even the best-performing model achieves an accuracy under 60\%, indicating that current image editing models still face substantial challenges in reasoning-informed visual editing.

In temporal, causal, spatial, and counterfactual reasoning tasks, where the required edits can often be inferred from visual content and common knowledge, the leading closed-source models demonstrate relatively strong performance. For example, GPT-Image-2.5 Sunburst achieves accuracies of 67.5\%, 69.0\%, 64.0\%, and 72.0\% on these four categories, respectively. However, when confronted with logical and hybrid reasoning tasks involving more complex constraints and multi-step reasoning, its accuracy decreases substantially to 27.0\% and 39.0\%. A similar trend can also be observed for other strong models. Nano Banana Pro, for instance, achieves 71.0\% accuracy on causal reasoning, while obtaining only 36.0\% and 38.0\% on logical and hybrid reasoning, respectively. These results indicate that logical reasoning remains a critical bottleneck for reasoning-guided visual editing, while hybrid reasoning also poses substantial challenges due to the need to integrate diverse reasoning tasks across multi-turn edits.

In contrast, a clear performance gap remains between closed-source and open-source models. Among the evaluated open-source models, HunyuanImage-3.0-Instruct achieves the highest overall accuracy at 23.0\%, followed by Qwen-Image-Edit-2511 at 17.1\%, with Qwen-Image-2.1 and FireRed-Image-Edit-1.0 tied at 16.3\%. Other open-source models generally achieve considerably lower performance, particularly on logical and hybrid reasoning. Although several methods introduce explicit reasoning processes, such as CoT or reflection, their improvements remain limited. For example, BAGEL with CoT substantially outperforms its base version, while the thinking variants of Step1X-Edit also improve over the base model. However, their overall accuracies are still significantly lower than those of the leading closed-source models, suggesting that explicit reasoning alone does not necessarily lead to successful reasoning-informed visual editing.

Regarding agentic methods, RISE-Agent achieves an overall accuracy of 48.0\%, demonstrating particularly strong performance on logical reasoning with an accuracy of 39.0\%. It also substantially outperforms the other evaluated agentic methods on single-image cases, where MindBrush, IntentEdit, and ImAgent achieve accuracies of 21.8\%, 13.2\%, and 3.5\%, respectively. These results demonstrate the potential of agentic frameworks for reasoning-informed visual editing, while also showing that considerable room for improvement remains.

For Chinese instructions, the overall trend is largely consistent with the English evaluation. Several strong models maintain relatively stable performance across the two instruction languages. For example, GPT-Image-2.5 Sunburst achieves overall accuracies of 56.6\% and 55.5\% under English and Chinese instructions, respectively, while Nano Banana 2 obtains 50.8\% and 51.0\%. However, the influence of instruction language still varies across different models; for instance, GPT-Image-1 decreases from 31.8\% under English instructions to 27.2\% under Chinese instructions. Meanwhile, logical and hybrid reasoning remain challenging under Chinese instructions, consistent with the observations from the English evaluation. These results suggest that multilingual instruction following is generally robust for stronger models, while the reasoning difficulty itself remains the dominant bottleneck.

To gain deeper insights into the strengths and limitations of each model, we further analyze the average performance across three evaluation dimensions, including Instruction Reasoning, Appearance Consistency, and Visual Plausibility, as illustrated in Figure~\ref{fig:metric_score}. The results show that most recent models already achieve relatively high Visual Plausibility, while leading closed-source models also demonstrate strong Appearance Consistency. In contrast, their performance on Instruction Reasoning remains considerably lower, with a much larger gap across different models. This indicates that the primary bottleneck of current reasoning-informed visual editing has shifted from generating visually plausible images to correctly understanding and executing complex reasoning instructions. In other words, current models are often capable of producing high-quality and visually consistent outputs, but still struggle to determine what should be edited when the instruction requires non-trivial reasoning.

\subsection{Performance across Subcategories}

\begin{table*}[!tp]
\caption{\textbf{Performance across subcategories on RISEBench++ (EN).} Models are likewise grouped into closed-source, open-source, and agentic methods; \textbf{bold} and \underline{underlined} mark the best and second-best in each group.}
\label{tab:subcategory_accuracy_en}
\centering
\begingroup
\scriptsize
\setlength{\tabcolsep}{1.4pt}
\renewcommand{\arraystretch}{0.96}
\colorlet{RISELtwoGroup}{black!6}
\def\RISELtwoNum#1{\makebox[2.0em][c]{#1}}
\def\RISELtwoyes{\textcolor{green!50!black}{\ding{51}}}
\def\RISELtwono{\textcolor{red!65!black}{\ding{55}}}
\begin{tabular}{
>{\raggedright\arraybackslash}p{\dimexpr .250\textwidth-2\tabcolsep\relax}
>{\centering\arraybackslash}p{\dimexpr .0648\textwidth-2\tabcolsep\relax}
>{\centering\arraybackslash}p{\dimexpr .0570\textwidth-2\tabcolsep\relax}
>{\centering\arraybackslash}p{\dimexpr .0570\textwidth-2\tabcolsep\relax}
>{\centering\arraybackslash}p{\dimexpr .0570\textwidth-2\tabcolsep\relax}
>{\centering\arraybackslash}p{\dimexpr .0570\textwidth-2\tabcolsep\relax}
>{\centering\arraybackslash}p{\dimexpr .0570\textwidth-2\tabcolsep\relax}
>{\centering\arraybackslash}p{\dimexpr .0570\textwidth-2\tabcolsep\relax}
>{\centering\arraybackslash}p{\dimexpr .0570\textwidth-2\tabcolsep\relax}
>{\centering\arraybackslash}p{\dimexpr .0570\textwidth-2\tabcolsep\relax}
>{\centering\arraybackslash}p{\dimexpr .0570\textwidth-2\tabcolsep\relax}
>{\centering\arraybackslash}p{\dimexpr .0570\textwidth-2\tabcolsep\relax}
>{\centering\arraybackslash}p{\dimexpr .0570\textwidth-2\tabcolsep\relax}
>{\centering\arraybackslash}p{\dimexpr .0570\textwidth-2\tabcolsep\relax}
}
\toprule[0.8pt]
\multirow{2}{*}[-2pt]{\textbf{Model}} &
\multicolumn{1}{c}{\multirow{2}{*}[-2pt]{\shortstack{\textbf{Multi-}\\\textbf{img}}}} &
\multicolumn{3}{c}{\textbf{Temporal}} &
\multicolumn{3}{c}{\textbf{Causal}} &
\multicolumn{2}{c}{\textbf{Spatial}} &
\multicolumn{2}{c}{\textbf{Logical}} &
\multicolumn{2}{c}{\textbf{Counterfactual}} \\
\cmidrule(lr){3-5}\cmidrule(lr){6-8}\cmidrule(lr){9-10}\cmidrule(lr){11-12}\cmidrule(lr){13-14}
 & &
\multicolumn{1}{c}{\textbf{CSK}} &
\multicolumn{1}{c}{\textbf{CtxK}} &
\multicolumn{1}{c}{\textbf{DK}} &
\multicolumn{1}{c}{\textbf{CSK}} &
\multicolumn{1}{c}{\textbf{CtxK}} &
\multicolumn{1}{c}{\textbf{DK}} &
\multicolumn{1}{c}{\textbf{Rel.}} &
\multicolumn{1}{c}{\textbf{Struct.}} &
\multicolumn{1}{c}{\textbf{ERD}} &
\multicolumn{1}{c}{\textbf{IPI}} &
\multicolumn{1}{c}{\textbf{CFI}} &
\multicolumn{1}{c}{\textbf{ACG}} \\
\midrule[0.5pt]
\rowcolor{RISELtwoGroup}
\multicolumn{14}{l}{\rule{0pt}{10pt}\raisebox{1pt}[0pt][0pt]{\textbf{Closed-source Models}}\rule[-2pt]{0pt}{0pt}} \\
\midrule[0.4pt]
GPT-Image-2.5 Sunburst~\cite{openai2025gptimage} & \RISELtwoyes & \RISELtwoNum{72.9} & \RISELtwoNum{\underline{67.3}} & \RISELtwoNum{\textbf{60.3}} & \RISELtwoNum{\underline{78.8}} & \RISELtwoNum{55.0} & \RISELtwoNum{\underline{66.3}} & \RISELtwoNum{\textbf{63.2}} & \RISELtwoNum{\textbf{64.9}} & \RISELtwoNum{32.5} & \RISELtwoNum{18.2} & \RISELtwoNum{\textbf{71.4}} & \RISELtwoNum{\textbf{73.3}} \\
Nano Banana Pro~\cite{google2026nanobanana} & \RISELtwoyes & \RISELtwoNum{68.2} & \RISELtwoNum{63.5} & \RISELtwoNum{55.6} & \RISELtwoNum{77.5} & \RISELtwoNum{\textbf{65.0}} & \RISELtwoNum{\textbf{67.5}} & \RISELtwoNum{50.9} & \RISELtwoNum{52.1} & \RISELtwoNum{\textbf{41.5}} & \RISELtwoNum{\underline{27.3}} & \RISELtwoNum{61.4} & \RISELtwoNum{\textbf{73.3}} \\
Nano Banana 2~\cite{google2026nanobanana} & \RISELtwoyes & \RISELtwoNum{\textbf{77.6}} & \RISELtwoNum{63.5} & \RISELtwoNum{\underline{57.1}} & \RISELtwoNum{76.3} & \RISELtwoNum{50.0} & \RISELtwoNum{\underline{66.3}} & \RISELtwoNum{47.2} & \RISELtwoNum{42.6} & \RISELtwoNum{38.2} & \RISELtwoNum{13.0} & \RISELtwoNum{58.6} & \RISELtwoNum{\underline{70.0}} \\
GPT-Image-2~\cite{openai2025gptimage} & \RISELtwoyes & \RISELtwoNum{61.2} & \RISELtwoNum{\underline{67.3}} & \RISELtwoNum{50.8} & \RISELtwoNum{76.3} & \RISELtwoNum{\underline{60.0}} & \RISELtwoNum{55.0} & \RISELtwoNum{\underline{60.4}} & \RISELtwoNum{\underline{60.6}} & \RISELtwoNum{21.1} & \RISELtwoNum{5.2} & \RISELtwoNum{52.9} & \RISELtwoNum{\underline{70.0}} \\
GPT-Image-2.5 Flare~\cite{openai2025gptimage} & \RISELtwoyes & \RISELtwoNum{68.2} & \RISELtwoNum{61.5} & \RISELtwoNum{50.8} & \RISELtwoNum{76.3} & \RISELtwoNum{52.5} & \RISELtwoNum{55.0} & \RISELtwoNum{56.6} & \RISELtwoNum{56.4} & \RISELtwoNum{17.1} & \RISELtwoNum{1.3} & \RISELtwoNum{52.9} & \RISELtwoNum{66.7} \\
Luma Uni-1~\cite{luma2026uni1} & \RISELtwoyes & \RISELtwoNum{\underline{74.1}} & \RISELtwoNum{57.7} & \RISELtwoNum{46.0} & \RISELtwoNum{68.8} & \RISELtwoNum{42.5} & \RISELtwoNum{52.5} & \RISELtwoNum{50.0} & \RISELtwoNum{51.1} & \RISELtwoNum{21.1} & \RISELtwoNum{13.0} & \RISELtwoNum{55.7} & \RISELtwoNum{53.3} \\
Qwen-Image-3.0~\cite{zhao2026qwen} & \RISELtwoyes & \RISELtwoNum{58.8} & \RISELtwoNum{\underline{67.3}} & \RISELtwoNum{50.8} & \RISELtwoNum{51.3} & \RISELtwoNum{50.0} & \RISELtwoNum{42.5} & \RISELtwoNum{44.3} & \RISELtwoNum{37.2} & \RISELtwoNum{\underline{39.0}} & \RISELtwoNum{\textbf{31.2}} & \RISELtwoNum{44.3} & \RISELtwoNum{60.0} \\
Nano Banana 2 Lite~\cite{google2026nanobanana} & \RISELtwoyes & \RISELtwoNum{67.1} & \RISELtwoNum{63.5} & \RISELtwoNum{49.2} & \RISELtwoNum{77.5} & \RISELtwoNum{45.0} & \RISELtwoNum{53.8} & \RISELtwoNum{44.3} & \RISELtwoNum{48.9} & \RISELtwoNum{18.7} & \RISELtwoNum{9.1} & \RISELtwoNum{61.4} & \RISELtwoNum{63.3} \\
FLUX.2 Max~\cite{bfl2025flux2} & \RISELtwoyes & \RISELtwoNum{61.2} & \RISELtwoNum{\textbf{69.2}} & \RISELtwoNum{41.3} & \RISELtwoNum{67.5} & \RISELtwoNum{40.0} & \RISELtwoNum{55.0} & \RISELtwoNum{33.0} & \RISELtwoNum{53.2} & \RISELtwoNum{17.9} & \RISELtwoNum{3.9} & \RISELtwoNum{62.9} & \RISELtwoNum{56.7} \\
GPT-Image-1.5~\cite{openai2025gptimage} & \RISELtwoyes & \RISELtwoNum{65.9} & \RISELtwoNum{51.9} & \RISELtwoNum{50.8} & \RISELtwoNum{\textbf{80.0}} & \RISELtwoNum{52.5} & \RISELtwoNum{50.0} & \RISELtwoNum{29.2} & \RISELtwoNum{51.1} & \RISELtwoNum{8.9} & \RISELtwoNum{6.5} & \RISELtwoNum{61.4} & \RISELtwoNum{53.3} \\
FLUX.2 Pro~\cite{bfl2025flux2} & \RISELtwoyes & \RISELtwoNum{58.8} & \RISELtwoNum{59.6} & \RISELtwoNum{41.3} & \RISELtwoNum{58.8} & \RISELtwoNum{30.0} & \RISELtwoNum{47.5} & \RISELtwoNum{23.6} & \RISELtwoNum{51.1} & \RISELtwoNum{14.6} & \RISELtwoNum{3.9} & \RISELtwoNum{\underline{64.3}} & \RISELtwoNum{53.3} \\
Nano Banana~\cite{google2026nanobanana} & \RISELtwoyes & \RISELtwoNum{55.3} & \RISELtwoNum{51.9} & \RISELtwoNum{47.6} & \RISELtwoNum{70.0} & \RISELtwoNum{25.0} & \RISELtwoNum{48.8} & \RISELtwoNum{35.8} & \RISELtwoNum{43.6} & \RISELtwoNum{14.6} & \RISELtwoNum{5.2} & \RISELtwoNum{45.7} & \RISELtwoNum{50.0} \\
HunyuanImage-3.5~\cite{cao2025hunyuanimage} & \RISELtwoyes & \RISELtwoNum{52.9} & \RISELtwoNum{46.2} & \RISELtwoNum{28.6} & \RISELtwoNum{58.8} & \RISELtwoNum{32.5} & \RISELtwoNum{40.0} & \RISELtwoNum{37.7} & \RISELtwoNum{50.0} & \RISELtwoNum{14.6} & \RISELtwoNum{13.0} & \RISELtwoNum{31.4} & \RISELtwoNum{50.0} \\
Wan2.7-Image-Pro~\cite{mao2026wan} & \RISELtwoyes & \RISELtwoNum{60.0} & \RISELtwoNum{19.2} & \RISELtwoNum{34.9} & \RISELtwoNum{70.0} & \RISELtwoNum{42.5} & \RISELtwoNum{38.8} & \RISELtwoNum{33.0} & \RISELtwoNum{31.9} & \RISELtwoNum{13.0} & \RISELtwoNum{6.5} & \RISELtwoNum{44.3} & \RISELtwoNum{60.0} \\
GPT-Image-1~\cite{openai2025gptimage} & \RISELtwoyes & \RISELtwoNum{64.7} & \RISELtwoNum{38.5} & \RISELtwoNum{42.9} & \RISELtwoNum{72.5} & \RISELtwoNum{27.5} & \RISELtwoNum{36.3} & \RISELtwoNum{24.5} & \RISELtwoNum{35.1} & \RISELtwoNum{7.3} & \RISELtwoNum{2.6} & \RISELtwoNum{41.4} & \RISELtwoNum{40.0} \\
Qwen-Image-2.0~\cite{zhao2026qwen} & \RISELtwoyes & \RISELtwoNum{51.8} & \RISELtwoNum{44.2} & \RISELtwoNum{33.3} & \RISELtwoNum{58.8} & \RISELtwoNum{27.5} & \RISELtwoNum{45.0} & \RISELtwoNum{37.7} & \RISELtwoNum{34.0} & \RISELtwoNum{7.3} & \RISELtwoNum{5.2} & \RISELtwoNum{30.0} & \RISELtwoNum{50.0} \\
Seedream 4.5~\cite{seedream2025seedream} & \RISELtwoyes & \RISELtwoNum{47.1} & \RISELtwoNum{46.2} & \RISELtwoNum{33.3} & \RISELtwoNum{62.5} & \RISELtwoNum{17.5} & \RISELtwoNum{33.8} & \RISELtwoNum{24.5} & \RISELtwoNum{28.7} & \RISELtwoNum{10.6} & \RISELtwoNum{2.6} & \RISELtwoNum{28.6} & \RISELtwoNum{53.3} \\
Seedream 5.0~\cite{seedream2025seedream} & \RISELtwoyes & \RISELtwoNum{57.6} & \RISELtwoNum{34.6} & \RISELtwoNum{28.6} & \RISELtwoNum{52.5} & \RISELtwoNum{30.0} & \RISELtwoNum{40.0} & \RISELtwoNum{23.6} & \RISELtwoNum{28.7} & \RISELtwoNum{7.3} & \RISELtwoNum{2.6} & \RISELtwoNum{24.3} & \RISELtwoNum{43.3} \\
Step Image Edit 2~\cite{liu2025step1x} & \RISELtwono & \RISELtwoNum{28.2} & \RISELtwoNum{4.2} & \RISELtwoNum{10.2} & \RISELtwoNum{31.6} & \RISELtwoNum{23.7} & \RISELtwoNum{13.3} & \RISELtwoNum{27.0} & \RISELtwoNum{25.6} & \RISELtwoNum{0.8} & \RISELtwoNum{3.5} & \RISELtwoNum{5.7} & \RISELtwoNum{20.0} \\
\midrule[0.4pt]
\rowcolor{RISELtwoGroup}
\multicolumn{14}{l}{\rule{0pt}{10pt}\raisebox{1pt}[0pt][0pt]{\textbf{Open-source Models}}\rule[-2pt]{0pt}{0pt}} \\
\midrule[0.4pt]
HunyuanImage-3.0-Instruct~\cite{cao2025hunyuanimage} & \RISELtwoyes & \RISELtwoNum{38.8} & \RISELtwoNum{\underline{32.7}} & \RISELtwoNum{17.5} & \RISELtwoNum{\textbf{50.0}} & \RISELtwoNum{20.0} & \RISELtwoNum{\textbf{33.8}} & \RISELtwoNum{\textbf{26.4}} & \RISELtwoNum{\textbf{33.0}} & \RISELtwoNum{\textbf{9.8}} & \RISELtwoNum{2.6} & \RISELtwoNum{12.9} & \RISELtwoNum{\textbf{40.0}} \\
Qwen-Image-Edit-2511~\cite{wu2025qwen} & \RISELtwoyes & \RISELtwoNum{28.2} & \RISELtwoNum{25.0} & \RISELtwoNum{11.1} & \RISELtwoNum{38.8} & \RISELtwoNum{20.0} & \RISELtwoNum{12.5} & \RISELtwoNum{17.9} & \RISELtwoNum{29.8} & \RISELtwoNum{3.3} & \RISELtwoNum{\underline{6.5}} & \RISELtwoNum{7.1} & \RISELtwoNum{\textbf{40.0}} \\
Qwen-Image-2.1~\cite{zhao2026qwen} & \RISELtwoyes & \RISELtwoNum{23.5} & \RISELtwoNum{28.8} & \RISELtwoNum{12.7} & \RISELtwoNum{\underline{42.5}} & \RISELtwoNum{12.5} & \RISELtwoNum{15.0} & \RISELtwoNum{17.9} & \RISELtwoNum{\underline{31.9}} & \RISELtwoNum{2.4} & \RISELtwoNum{1.3} & \RISELtwoNum{11.4} & \RISELtwoNum{26.7} \\
FireRed-Image-Edit-1.0~\cite{team2026firered} & \RISELtwoyes & \RISELtwoNum{30.6} & \RISELtwoNum{17.3} & \RISELtwoNum{14.3} & \RISELtwoNum{26.3} & \RISELtwoNum{15.0} & \RISELtwoNum{11.3} & \RISELtwoNum{\underline{24.5}} & \RISELtwoNum{\underline{31.9}} & \RISELtwoNum{1.6} & \RISELtwoNum{2.6} & \RISELtwoNum{14.3} & \RISELtwoNum{\underline{36.7}} \\
FireRed-Image-Edit-1.1~\cite{team2026firered} & \RISELtwoyes & \RISELtwoNum{24.7} & \RISELtwoNum{28.8} & \RISELtwoNum{14.3} & \RISELtwoNum{32.5} & \RISELtwoNum{15.0} & \RISELtwoNum{18.8} & \RISELtwoNum{17.9} & \RISELtwoNum{27.7} & \RISELtwoNum{1.6} & \RISELtwoNum{1.3} & \RISELtwoNum{7.1} & \RISELtwoNum{\underline{36.7}} \\
Qwen-Image-Edit-2509~\cite{wu2025qwen} & \RISELtwoyes & \RISELtwoNum{24.7} & \RISELtwoNum{17.3} & \RISELtwoNum{\textbf{19.0}} & \RISELtwoNum{37.5} & \RISELtwoNum{15.0} & \RISELtwoNum{16.3} & \RISELtwoNum{17.9} & \RISELtwoNum{22.3} & \RISELtwoNum{0.8} & \RISELtwoNum{0.0} & \RISELtwoNum{12.9} & \RISELtwoNum{23.3} \\
HiDream-O1-Image~\cite{cai2026hidream} & \RISELtwoyes & \RISELtwoNum{\underline{40.0}} & \RISELtwoNum{26.9} & \RISELtwoNum{11.1} & \RISELtwoNum{31.3} & \RISELtwoNum{2.5} & \RISELtwoNum{12.5} & \RISELtwoNum{17.9} & \RISELtwoNum{23.4} & \RISELtwoNum{1.6} & \RISELtwoNum{5.2} & \RISELtwoNum{1.4} & \RISELtwoNum{13.3} \\
FLUX.2 Dev~\cite{bfl2025flux2} & \RISELtwoyes & \RISELtwoNum{18.8} & \RISELtwoNum{28.8} & \RISELtwoNum{12.7} & \RISELtwoNum{40.0} & \RISELtwoNum{20.0} & \RISELtwoNum{18.8} & \RISELtwoNum{9.4} & \RISELtwoNum{25.5} & \RISELtwoNum{0.8} & \RISELtwoNum{1.3} & \RISELtwoNum{2.9} & \RISELtwoNum{26.7} \\
FLUX.2 Klein 9B~\cite{bfl2025flux2} & \RISELtwoyes & \RISELtwoNum{11.8} & \RISELtwoNum{30.8} & \RISELtwoNum{9.5} & \RISELtwoNum{26.3} & \RISELtwoNum{15.0} & \RISELtwoNum{7.5} & \RISELtwoNum{16.0} & \RISELtwoNum{28.7} & \RISELtwoNum{0.0} & \RISELtwoNum{1.3} & \RISELtwoNum{15.7} & \RISELtwoNum{20.0} \\
JoyAI-Image-Edit-Plus~\cite{song2026joyai} & \RISELtwoyes & \RISELtwoNum{12.9} & \RISELtwoNum{\underline{32.7}} & \RISELtwoNum{7.9} & \RISELtwoNum{28.8} & \RISELtwoNum{10.0} & \RISELtwoNum{6.3} & \RISELtwoNum{13.2} & \RISELtwoNum{27.7} & \RISELtwoNum{\underline{5.7}} & \RISELtwoNum{1.3} & \RISELtwoNum{8.6} & \RISELtwoNum{23.3} \\
SenseNova-U1-A3B-MoT-SFT~\cite{diao2026sensenova} & \RISELtwoyes & \RISELtwoNum{\textbf{41.2}} & \RISELtwoNum{1.9} & \RISELtwoNum{7.9} & \RISELtwoNum{40.0} & \RISELtwoNum{5.0} & \RISELtwoNum{21.3} & \RISELtwoNum{0.0} & \RISELtwoNum{5.3} & \RISELtwoNum{3.3} & \RISELtwoNum{2.6} & \RISELtwoNum{\textbf{18.6}} & \RISELtwoNum{13.3} \\
UniPic 3.0~\cite{wei2026skywork} & \RISELtwoyes & \RISELtwoNum{17.6} & \RISELtwoNum{30.8} & \RISELtwoNum{9.5} & \RISELtwoNum{23.8} & \RISELtwoNum{17.5} & \RISELtwoNum{11.3} & \RISELtwoNum{16.0} & \RISELtwoNum{20.2} & \RISELtwoNum{0.0} & \RISELtwoNum{1.3} & \RISELtwoNum{8.6} & \RISELtwoNum{10.0} \\
BAGEL (w/ CoT)~\cite{deng2025bagel} & \RISELtwoyes & \RISELtwoNum{22.4} & \RISELtwoNum{23.1} & \RISELtwoNum{9.5} & \RISELtwoNum{21.3} & \RISELtwoNum{10.0} & \RISELtwoNum{13.8} & \RISELtwoNum{11.3} & \RISELtwoNum{21.3} & \RISELtwoNum{0.8} & \RISELtwoNum{2.6} & \RISELtwoNum{10.0} & \RISELtwoNum{10.0} \\
SenseNova-U1-8B-MoT-SFT~\cite{diao2026sensenova} & \RISELtwoyes & \RISELtwoNum{28.2} & \RISELtwoNum{3.8} & \RISELtwoNum{11.1} & \RISELtwoNum{35.0} & \RISELtwoNum{10.0} & \RISELtwoNum{16.3} & \RISELtwoNum{6.6} & \RISELtwoNum{8.5} & \RISELtwoNum{4.1} & \RISELtwoNum{2.6} & \RISELtwoNum{11.4} & \RISELtwoNum{16.7} \\
FLUX.2 Klein 4B~\cite{bfl2025flux2} & \RISELtwoyes & \RISELtwoNum{11.8} & \RISELtwoNum{\textbf{36.5}} & \RISELtwoNum{3.2} & \RISELtwoNum{17.5} & \RISELtwoNum{20.0} & \RISELtwoNum{8.8} & \RISELtwoNum{10.4} & \RISELtwoNum{24.5} & \RISELtwoNum{0.8} & \RISELtwoNum{5.2} & \RISELtwoNum{7.1} & \RISELtwoNum{10.0} \\
UniWorld-V2~\cite{li2025uniworld} & \RISELtwoyes & \RISELtwoNum{14.1} & \RISELtwoNum{7.7} & \RISELtwoNum{6.3} & \RISELtwoNum{25.0} & \RISELtwoNum{\underline{22.5}} & \RISELtwoNum{8.8} & \RISELtwoNum{14.2} & \RISELtwoNum{9.6} & \RISELtwoNum{0.8} & \RISELtwoNum{0.0} & \RISELtwoNum{5.7} & \RISELtwoNum{30.0} \\
BAGEL~\cite{deng2025bagel} & \RISELtwoyes & \RISELtwoNum{2.4} & \RISELtwoNum{19.2} & \RISELtwoNum{1.6} & \RISELtwoNum{11.3} & \RISELtwoNum{2.5} & \RISELtwoNum{2.5} & \RISELtwoNum{5.7} & \RISELtwoNum{10.6} & \RISELtwoNum{0.0} & \RISELtwoNum{2.6} & \RISELtwoNum{2.9} & \RISELtwoNum{0.0} \\
OmniGen2~\cite{wu2025omnigen2} & \RISELtwoyes & \RISELtwoNum{3.5} & \RISELtwoNum{15.4} & \RISELtwoNum{4.8} & \RISELtwoNum{6.3} & \RISELtwoNum{7.5} & \RISELtwoNum{5.0} & \RISELtwoNum{3.8} & \RISELtwoNum{7.4} & \RISELtwoNum{0.0} & \RISELtwoNum{0.0} & \RISELtwoNum{8.6} & \RISELtwoNum{6.7} \\
OmniGen~\cite{xiao2024omnigen} & \RISELtwoyes & \RISELtwoNum{3.5} & \RISELtwoNum{7.7} & \RISELtwoNum{3.2} & \RISELtwoNum{2.5} & \RISELtwoNum{5.0} & \RISELtwoNum{5.0} & \RISELtwoNum{0.9} & \RISELtwoNum{9.6} & \RISELtwoNum{0.8} & \RISELtwoNum{2.6} & \RISELtwoNum{0.0} & \RISELtwoNum{3.3} \\
Emu2-Gen~\cite{sun2023emu2chat} & \RISELtwoyes & \RISELtwoNum{2.4} & \RISELtwoNum{1.9} & \RISELtwoNum{7.9} & \RISELtwoNum{3.8} & \RISELtwoNum{0.0} & \RISELtwoNum{0.0} & \RISELtwoNum{0.0} & \RISELtwoNum{1.1} & \RISELtwoNum{0.8} & \RISELtwoNum{0.0} & \RISELtwoNum{0.0} & \RISELtwoNum{3.3} \\
JoyAI-Image-Edit~\cite{song2026joyai} & \RISELtwono & \RISELtwoNum{24.7} & \RISELtwoNum{4.2} & \RISELtwoNum{13.6} & \RISELtwoNum{39.2} & \RISELtwoNum{15.8} & \RISELtwoNum{12.0} & \RISELtwoNum{20.0} & \RISELtwoNum{24.4} & \RISELtwoNum{0.8} & \RISELtwoNum{5.3} & \RISELtwoNum{11.4} & \RISELtwoNum{26.7} \\
Step1X-Edit (thinking)~\cite{liu2025step1x} & \RISELtwono & \RISELtwoNum{28.2} & \RISELtwoNum{8.3} & \RISELtwoNum{11.9} & \RISELtwoNum{31.6} & \RISELtwoNum{\textbf{26.3}} & \RISELtwoNum{\underline{22.7}} & \RISELtwoNum{12.0} & \RISELtwoNum{17.1} & \RISELtwoNum{0.0} & \RISELtwoNum{1.8} & \RISELtwoNum{11.4} & \RISELtwoNum{26.7} \\
Boogu-Image-0.1-Edit~\cite{chen2026boogu} & \RISELtwono & \RISELtwoNum{16.5} & \RISELtwoNum{4.2} & \RISELtwoNum{\underline{18.6}} & \RISELtwoNum{35.4} & \RISELtwoNum{15.8} & \RISELtwoNum{10.7} & \RISELtwoNum{15.0} & \RISELtwoNum{20.7} & \RISELtwoNum{3.3} & \RISELtwoNum{\textbf{12.3}} & \RISELtwoNum{8.6} & \RISELtwoNum{16.7} \\
\resizebox{\linewidth}{!}{Step1X-Edit (thinking + reflection)~\cite{liu2025step1x}} & \RISELtwono & \RISELtwoNum{30.6} & \RISELtwoNum{0.0} & \RISELtwoNum{16.9} & \RISELtwoNum{34.2} & \RISELtwoNum{15.8} & \RISELtwoNum{14.7} & \RISELtwoNum{13.0} & \RISELtwoNum{6.1} & \RISELtwoNum{0.8} & \RISELtwoNum{3.5} & \RISELtwoNum{\textbf{18.6}} & \RISELtwoNum{20.0} \\
LongCat-Image-Edit~\cite{team2025longcat} & \RISELtwono & \RISELtwoNum{11.8} & \RISELtwoNum{0.0} & \RISELtwoNum{11.9} & \RISELtwoNum{30.4} & \RISELtwoNum{18.4} & \RISELtwoNum{12.0} & \RISELtwoNum{15.0} & \RISELtwoNum{15.9} & \RISELtwoNum{0.8} & \RISELtwoNum{1.8} & \RISELtwoNum{2.9} & \RISELtwoNum{23.3} \\
Step1X-Edit (base)~\cite{liu2025step1x} & \RISELtwono & \RISELtwoNum{17.6} & \RISELtwoNum{4.2} & \RISELtwoNum{10.2} & \RISELtwoNum{27.8} & \RISELtwoNum{18.4} & \RISELtwoNum{9.3} & \RISELtwoNum{8.0} & \RISELtwoNum{13.4} & \RISELtwoNum{0.0} & \RISELtwoNum{3.5} & \RISELtwoNum{14.3} & \RISELtwoNum{13.3} \\
InternVL-U (w/ CoT)~\cite{tian2026internvl} & \RISELtwono & \RISELtwoNum{11.8} & \RISELtwoNum{4.2} & \RISELtwoNum{6.8} & \RISELtwoNum{19.0} & \RISELtwoNum{7.9} & \RISELtwoNum{12.0} & \RISELtwoNum{4.0} & \RISELtwoNum{6.1} & \RISELtwoNum{0.8} & \RISELtwoNum{1.8} & \RISELtwoNum{\underline{17.1}} & \RISELtwoNum{13.3} \\
FLUX.1 Kontext Dev~\cite{labs2025flux1kontextflowmatching} & \RISELtwono & \RISELtwoNum{7.1} & \RISELtwoNum{0.0} & \RISELtwoNum{5.1} & \RISELtwoNum{12.7} & \RISELtwoNum{10.5} & \RISELtwoNum{5.3} & \RISELtwoNum{7.0} & \RISELtwoNum{18.3} & \RISELtwoNum{0.8} & \RISELtwoNum{3.5} & \RISELtwoNum{7.1} & \RISELtwoNum{6.7} \\
LLaDA-Image~\cite{chen2026llada} & \RISELtwono & \RISELtwoNum{10.6} & \RISELtwoNum{0.0} & \RISELtwoNum{5.1} & \RISELtwoNum{6.3} & \RISELtwoNum{0.0} & \RISELtwoNum{5.3} & \RISELtwoNum{0.0} & \RISELtwoNum{8.5} & \RISELtwoNum{0.0} & \RISELtwoNum{0.0} & \RISELtwoNum{4.3} & \RISELtwoNum{3.3} \\
Ovis-U1~\cite{wang2025ovis} & \RISELtwono & \RISELtwoNum{7.1} & \RISELtwoNum{0.0} & \RISELtwoNum{3.4} & \RISELtwoNum{8.9} & \RISELtwoNum{5.3} & \RISELtwoNum{4.0} & \RISELtwoNum{1.0} & \RISELtwoNum{3.7} & \RISELtwoNum{0.8} & \RISELtwoNum{3.5} & \RISELtwoNum{2.9} & \RISELtwoNum{0.0} \\
InternVL-U~\cite{tian2026internvl} & \RISELtwono & \RISELtwoNum{5.9} & \RISELtwoNum{0.0} & \RISELtwoNum{1.7} & \RISELtwoNum{3.8} & \RISELtwoNum{0.0} & \RISELtwoNum{2.7} & \RISELtwoNum{2.0} & \RISELtwoNum{6.1} & \RISELtwoNum{0.8} & \RISELtwoNum{0.0} & \RISELtwoNum{5.7} & \RISELtwoNum{3.3} \\
UniPic2-SD3.5M-Kontext-2B~\cite{wei2025skywork} & \RISELtwono & \RISELtwoNum{2.4} & \RISELtwoNum{0.0} & \RISELtwoNum{0.0} & \RISELtwoNum{1.3} & \RISELtwoNum{0.0} & \RISELtwoNum{4.0} & \RISELtwoNum{0.0} & \RISELtwoNum{1.2} & \RISELtwoNum{0.0} & \RISELtwoNum{0.0} & \RISELtwoNum{4.3} & \RISELtwoNum{0.0} \\
Lumina-DiMOO~\cite{xin2025lumina} & \RISELtwono & \RISELtwoNum{1.2} & \RISELtwoNum{0.0} & \RISELtwoNum{0.0} & \RISELtwoNum{2.5} & \RISELtwoNum{0.0} & \RISELtwoNum{1.3} & \RISELtwoNum{0.0} & \RISELtwoNum{0.0} & \RISELtwoNum{0.8} & \RISELtwoNum{0.0} & \RISELtwoNum{4.3} & \RISELtwoNum{3.3} \\
UniWorld-V1~\cite{lin2025uniworld} & \RISELtwono & \RISELtwoNum{2.4} & \RISELtwoNum{0.0} & \RISELtwoNum{0.0} & \RISELtwoNum{2.5} & \RISELtwoNum{0.0} & \RISELtwoNum{0.0} & \RISELtwoNum{0.0} & \RISELtwoNum{1.2} & \RISELtwoNum{0.8} & \RISELtwoNum{0.0} & \RISELtwoNum{0.0} & \RISELtwoNum{3.3} \\
\midrule[0.4pt]
\rowcolor{RISELtwoGroup}
\multicolumn{14}{l}{\rule{0pt}{10pt}\raisebox{1pt}[0pt][0pt]{\textbf{Agentic Methods}}\rule[-2pt]{0pt}{0pt}} \\
\midrule[0.4pt]
\textbf{RISE-Agent (Ours)} & \RISELtwoyes & \RISELtwoNum{\textbf{64.7}} & \RISELtwoNum{\textbf{57.7}} & \RISELtwoNum{\textbf{50.8}} & \RISELtwoNum{\textbf{78.8}} & \RISELtwoNum{\textbf{40.0}} & \RISELtwoNum{\textbf{46.3}} & \RISELtwoNum{\textbf{36.8}} & \RISELtwoNum{\textbf{38.3}} & \RISELtwoNum{\textbf{47.2}} & \RISELtwoNum{\textbf{26.0}} & \RISELtwoNum{\textbf{70.0}} & \RISELtwoNum{\textbf{56.7}} \\
MindBrush~\cite{he2026mind} & \RISELtwono & \RISELtwoNum{54.1} & \RISELtwoNum{12.5} & \RISELtwoNum{16.9} & \RISELtwoNum{\underline{46.8}} & \RISELtwoNum{7.9} & \RISELtwoNum{\underline{32.0}} & \RISELtwoNum{15.0} & \RISELtwoNum{\underline{18.3}} & \RISELtwoNum{\underline{8.9}} & \RISELtwoNum{\underline{8.8}} & \RISELtwoNum{\underline{30.0}} & \RISELtwoNum{\underline{36.7}} \\
InterleaveThinker~\cite{zheng2026interleavethinker} & \RISELtwono & \RISELtwoNum{\underline{55.3}} & \RISELtwoNum{\underline{25.0}} & \RISELtwoNum{\underline{20.3}} & \RISELtwoNum{35.4} & \RISELtwoNum{\underline{39.5}} & \RISELtwoNum{21.3} & \RISELtwoNum{\underline{17.0}} & \RISELtwoNum{14.6} & \RISELtwoNum{1.6} & \RISELtwoNum{1.8} & \RISELtwoNum{22.9} & \RISELtwoNum{33.3} \\
IntentEdit~\cite{zhang2026intentedit} & \RISELtwono & \RISELtwoNum{34.1} & \RISELtwoNum{20.8} & \RISELtwoNum{8.5} & \RISELtwoNum{29.1} & \RISELtwoNum{10.5} & \RISELtwoNum{25.3} & \RISELtwoNum{7.0} & \RISELtwoNum{17.1} & \RISELtwoNum{1.6} & \RISELtwoNum{0.0} & \RISELtwoNum{12.9} & \RISELtwoNum{16.7} \\
ImAgent~\cite{wang2025imagent} & \RISELtwono & \RISELtwoNum{7.1} & \RISELtwoNum{4.2} & \RISELtwoNum{3.4} & \RISELtwoNum{8.9} & \RISELtwoNum{5.3} & \RISELtwoNum{0.0} & \RISELtwoNum{2.0} & \RISELtwoNum{7.3} & \RISELtwoNum{0.0} & \RISELtwoNum{1.8} & \RISELtwoNum{5.7} & \RISELtwoNum{3.3} \\
\bottomrule[0.8pt]
\end{tabular}
\par\vspace{4pt}
\parbox{\textwidth}{\scriptsize
Results are reported as accuracy (\%).
CSK: Common-Sense Knowledge; CtxK: Context Knowledge; DK: Disciplinary Knowledge;
Rel.: Relation; Struct.: Structure; ERD: Explicit Rule Deduction; IPI: Implicit Pattern Induction;
CFI: Counterfactual Imagination; ACG: Anomaly \& Correction Generation.
Models with multi-image support are evaluated on both single- and multi-image cases; otherwise, only single-image cases are used.}
\endgroup
\end{table*}

\begin{table*}[!tp]
\caption{\textbf{Performance across subcategories on RISEBench++ (ZH).} Models are likewise grouped into closed-source, open-source, and agentic methods; \textbf{bold} and \underline{underlined} mark the best and second-best in each group.}
\label{tab:subcategory_accuracy_cn}
\centering
\begingroup
\scriptsize
\setlength{\tabcolsep}{1.4pt}
\renewcommand{\arraystretch}{0.96}
\colorlet{RISECNtwoGroup}{black!6}
\def\RISECNtwoNum#1{\makebox[2.0em][c]{#1}}
\def\RISECNtwoyes{\textcolor{green!50!black}{\ding{51}}}
\def\RISECNtwono{\textcolor{red!65!black}{\ding{55}}}
\begin{tabular}{
>{\raggedright\arraybackslash}p{\dimexpr .250\textwidth-2\tabcolsep\relax}
>{\centering\arraybackslash}p{\dimexpr .0648\textwidth-2\tabcolsep\relax}
>{\centering\arraybackslash}p{\dimexpr .0570\textwidth-2\tabcolsep\relax}
>{\centering\arraybackslash}p{\dimexpr .0570\textwidth-2\tabcolsep\relax}
>{\centering\arraybackslash}p{\dimexpr .0570\textwidth-2\tabcolsep\relax}
>{\centering\arraybackslash}p{\dimexpr .0570\textwidth-2\tabcolsep\relax}
>{\centering\arraybackslash}p{\dimexpr .0570\textwidth-2\tabcolsep\relax}
>{\centering\arraybackslash}p{\dimexpr .0570\textwidth-2\tabcolsep\relax}
>{\centering\arraybackslash}p{\dimexpr .0570\textwidth-2\tabcolsep\relax}
>{\centering\arraybackslash}p{\dimexpr .0570\textwidth-2\tabcolsep\relax}
>{\centering\arraybackslash}p{\dimexpr .0570\textwidth-2\tabcolsep\relax}
>{\centering\arraybackslash}p{\dimexpr .0570\textwidth-2\tabcolsep\relax}
>{\centering\arraybackslash}p{\dimexpr .0570\textwidth-2\tabcolsep\relax}
>{\centering\arraybackslash}p{\dimexpr .0570\textwidth-2\tabcolsep\relax}
}
\toprule[0.8pt]
\multirow{2}{*}[-2pt]{\textbf{Model}} &
\multicolumn{1}{c}{\multirow{2}{*}[-2pt]{\shortstack{\textbf{Multi-}\\\textbf{img}}}} &
\multicolumn{3}{c}{\textbf{Temporal}} &
\multicolumn{3}{c}{\textbf{Causal}} &
\multicolumn{2}{c}{\textbf{Spatial}} &
\multicolumn{2}{c}{\textbf{Logical}} &
\multicolumn{2}{c}{\textbf{Counterfactual}} \\
\cmidrule(lr){3-5}\cmidrule(lr){6-8}\cmidrule(lr){9-10}\cmidrule(lr){11-12}\cmidrule(lr){13-14}
 & &
\multicolumn{1}{c}{\textbf{CSK}} &
\multicolumn{1}{c}{\textbf{CtxK}} &
\multicolumn{1}{c}{\textbf{DK}} &
\multicolumn{1}{c}{\textbf{CSK}} &
\multicolumn{1}{c}{\textbf{CtxK}} &
\multicolumn{1}{c}{\textbf{DK}} &
\multicolumn{1}{c}{\textbf{Rel.}} &
\multicolumn{1}{c}{\textbf{Struct.}} &
\multicolumn{1}{c}{\textbf{ERD}} &
\multicolumn{1}{c}{\textbf{IPI}} &
\multicolumn{1}{c}{\textbf{CFI}} &
\multicolumn{1}{c}{\textbf{ACG}} \\
\midrule[0.5pt]
\rowcolor{RISECNtwoGroup}
\multicolumn{14}{l}{\rule{0pt}{10pt}\raisebox{1pt}[0pt][0pt]{\textbf{Closed-source Models}}\rule[-2pt]{0pt}{0pt}} \\
\midrule[0.4pt]
GPT-Image-2.5 Sunburst~\cite{openai2025gptimage} & \RISECNtwoyes & \RISECNtwoNum{69.4} & \RISECNtwoNum{\underline{65.4}} & \RISECNtwoNum{50.8} & \RISECNtwoNum{\textbf{85.0}} & \RISECNtwoNum{\textbf{62.5}} & \RISECNtwoNum{\underline{60.0}} & \RISECNtwoNum{\textbf{62.3}} & \RISECNtwoNum{\textbf{63.8}} & \RISECNtwoNum{32.5} & \RISECNtwoNum{23.4} & \RISECNtwoNum{\underline{64.3}} & \RISECNtwoNum{\textbf{80.0}} \\
Nano Banana Pro~\cite{google2026nanobanana} & \RISECNtwoyes & \RISECNtwoNum{68.2} & \RISECNtwoNum{63.5} & \RISECNtwoNum{\underline{54.0}} & \RISECNtwoNum{77.5} & \RISECNtwoNum{\textbf{62.5}} & \RISECNtwoNum{58.8} & \RISECNtwoNum{54.7} & \RISECNtwoNum{45.7} & \RISECNtwoNum{\underline{42.3}} & \RISECNtwoNum{\textbf{32.5}} & \RISECNtwoNum{54.3} & \RISECNtwoNum{40.0} \\
Nano Banana 2~\cite{google2026nanobanana} & \RISECNtwoyes & \RISECNtwoNum{\underline{74.1}} & \RISECNtwoNum{51.9} & \RISECNtwoNum{\underline{54.0}} & \RISECNtwoNum{\underline{83.8}} & \RISECNtwoNum{57.5} & \RISECNtwoNum{\underline{60.0}} & \RISECNtwoNum{47.2} & \RISECNtwoNum{53.2} & \RISECNtwoNum{35.0} & \RISECNtwoNum{26.0} & \RISECNtwoNum{51.4} & \RISECNtwoNum{70.0} \\
Luma Uni-1~\cite{luma2026uni1} & \RISECNtwoyes & \RISECNtwoNum{\textbf{81.2}} & \RISECNtwoNum{48.1} & \RISECNtwoNum{\textbf{58.7}} & \RISECNtwoNum{75.0} & \RISECNtwoNum{\underline{60.0}} & \RISECNtwoNum{\textbf{62.5}} & \RISECNtwoNum{52.8} & \RISECNtwoNum{54.3} & \RISECNtwoNum{21.1} & \RISECNtwoNum{10.4} & \RISECNtwoNum{58.6} & \RISECNtwoNum{50.0} \\
GPT-Image-2~\cite{openai2025gptimage} & \RISECNtwoyes & \RISECNtwoNum{61.2} & \RISECNtwoNum{55.8} & \RISECNtwoNum{52.4} & \RISECNtwoNum{76.3} & \RISECNtwoNum{57.5} & \RISECNtwoNum{50.0} & \RISECNtwoNum{\underline{56.6}} & \RISECNtwoNum{52.1} & \RISECNtwoNum{22.8} & \RISECNtwoNum{16.9} & \RISECNtwoNum{52.9} & \RISECNtwoNum{\textbf{80.0}} \\
Qwen-Image-3.0~\cite{zhao2026qwen} & \RISECNtwoyes & \RISECNtwoNum{63.5} & \RISECNtwoNum{\textbf{67.3}} & \RISECNtwoNum{42.9} & \RISECNtwoNum{52.5} & \RISECNtwoNum{57.5} & \RISECNtwoNum{43.8} & \RISECNtwoNum{45.3} & \RISECNtwoNum{40.4} & \RISECNtwoNum{\textbf{44.7}} & \RISECNtwoNum{\underline{29.9}} & \RISECNtwoNum{42.9} & \RISECNtwoNum{63.3} \\
GPT-Image-2.5 Flare~\cite{openai2025gptimage} & \RISECNtwoyes & \RISECNtwoNum{60.0} & \RISECNtwoNum{51.9} & \RISECNtwoNum{41.3} & \RISECNtwoNum{80.0} & \RISECNtwoNum{\underline{60.0}} & \RISECNtwoNum{55.0} & \RISECNtwoNum{51.9} & \RISECNtwoNum{\underline{56.4}} & \RISECNtwoNum{19.5} & \RISECNtwoNum{5.2} & \RISECNtwoNum{55.7} & \RISECNtwoNum{\underline{73.3}} \\
FLUX.2 Max~\cite{bfl2025flux2} & \RISECNtwoyes & \RISECNtwoNum{65.9} & \RISECNtwoNum{63.5} & \RISECNtwoNum{47.6} & \RISECNtwoNum{73.8} & \RISECNtwoNum{37.5} & \RISECNtwoNum{55.0} & \RISECNtwoNum{34.9} & \RISECNtwoNum{50.0} & \RISECNtwoNum{21.1} & \RISECNtwoNum{2.6} & \RISECNtwoNum{\textbf{74.3}} & \RISECNtwoNum{66.7} \\
Nano Banana 2 Lite~\cite{google2026nanobanana} & \RISECNtwoyes & \RISECNtwoNum{65.9} & \RISECNtwoNum{\underline{65.4}} & \RISECNtwoNum{42.9} & \RISECNtwoNum{76.3} & \RISECNtwoNum{35.0} & \RISECNtwoNum{57.5} & \RISECNtwoNum{38.7} & \RISECNtwoNum{44.7} & \RISECNtwoNum{13.0} & \RISECNtwoNum{10.4} & \RISECNtwoNum{60.0} & \RISECNtwoNum{56.7} \\
GPT-Image-1.5~\cite{openai2025gptimage} & \RISECNtwoyes & \RISECNtwoNum{68.2} & \RISECNtwoNum{44.2} & \RISECNtwoNum{46.0} & \RISECNtwoNum{75.0} & \RISECNtwoNum{40.0} & \RISECNtwoNum{47.5} & \RISECNtwoNum{32.1} & \RISECNtwoNum{39.4} & \RISECNtwoNum{4.9} & \RISECNtwoNum{3.9} & \RISECNtwoNum{52.9} & \RISECNtwoNum{66.7} \\
Nano Banana~\cite{google2026nanobanana} & \RISECNtwoyes & \RISECNtwoNum{49.4} & \RISECNtwoNum{55.8} & \RISECNtwoNum{44.4} & \RISECNtwoNum{70.0} & \RISECNtwoNum{35.0} & \RISECNtwoNum{45.0} & \RISECNtwoNum{32.1} & \RISECNtwoNum{48.9} & \RISECNtwoNum{14.6} & \RISECNtwoNum{7.8} & \RISECNtwoNum{47.1} & \RISECNtwoNum{63.3} \\
FLUX.2 Pro~\cite{bfl2025flux2} & \RISECNtwoyes & \RISECNtwoNum{50.6} & \RISECNtwoNum{57.7} & \RISECNtwoNum{46.0} & \RISECNtwoNum{61.3} & \RISECNtwoNum{25.0} & \RISECNtwoNum{37.5} & \RISECNtwoNum{31.1} & \RISECNtwoNum{46.8} & \RISECNtwoNum{12.2} & \RISECNtwoNum{1.3} & \RISECNtwoNum{55.7} & \RISECNtwoNum{40.0} \\
HunyuanImage-3.5~\cite{cao2025hunyuanimage} & \RISECNtwoyes & \RISECNtwoNum{49.4} & \RISECNtwoNum{53.8} & \RISECNtwoNum{28.6} & \RISECNtwoNum{43.8} & \RISECNtwoNum{32.5} & \RISECNtwoNum{38.8} & \RISECNtwoNum{42.5} & \RISECNtwoNum{47.9} & \RISECNtwoNum{14.6} & \RISECNtwoNum{13.0} & \RISECNtwoNum{40.0} & \RISECNtwoNum{46.7} \\
Wan2.7-Image-Pro~\cite{mao2026wan} & \RISECNtwoyes & \RISECNtwoNum{55.3} & \RISECNtwoNum{30.8} & \RISECNtwoNum{28.6} & \RISECNtwoNum{57.5} & \RISECNtwoNum{42.5} & \RISECNtwoNum{32.5} & \RISECNtwoNum{37.7} & \RISECNtwoNum{37.2} & \RISECNtwoNum{11.4} & \RISECNtwoNum{6.5} & \RISECNtwoNum{44.3} & \RISECNtwoNum{53.3} \\
Qwen-Image-2.0~\cite{zhao2026qwen} & \RISECNtwoyes & \RISECNtwoNum{61.2} & \RISECNtwoNum{57.7} & \RISECNtwoNum{33.3} & \RISECNtwoNum{43.8} & \RISECNtwoNum{27.5} & \RISECNtwoNum{35.0} & \RISECNtwoNum{23.6} & \RISECNtwoNum{38.3} & \RISECNtwoNum{9.8} & \RISECNtwoNum{7.8} & \RISECNtwoNum{34.3} & \RISECNtwoNum{43.3} \\
Seedream 4.5~\cite{seedream2025seedream} & \RISECNtwoyes & \RISECNtwoNum{54.1} & \RISECNtwoNum{40.4} & \RISECNtwoNum{33.3} & \RISECNtwoNum{56.3} & \RISECNtwoNum{42.5} & \RISECNtwoNum{35.0} & \RISECNtwoNum{23.6} & \RISECNtwoNum{29.8} & \RISECNtwoNum{10.6} & \RISECNtwoNum{5.2} & \RISECNtwoNum{32.9} & \RISECNtwoNum{40.0} \\
Seedream 5.0~\cite{seedream2025seedream} & \RISECNtwoyes & \RISECNtwoNum{51.8} & \RISECNtwoNum{40.4} & \RISECNtwoNum{25.4} & \RISECNtwoNum{46.3} & \RISECNtwoNum{37.5} & \RISECNtwoNum{33.8} & \RISECNtwoNum{20.8} & \RISECNtwoNum{36.2} & \RISECNtwoNum{13.8} & \RISECNtwoNum{7.8} & \RISECNtwoNum{28.6} & \RISECNtwoNum{33.3} \\
GPT-Image-1~\cite{openai2025gptimage} & \RISECNtwoyes & \RISECNtwoNum{58.8} & \RISECNtwoNum{28.8} & \RISECNtwoNum{34.9} & \RISECNtwoNum{60.0} & \RISECNtwoNum{22.5} & \RISECNtwoNum{28.8} & \RISECNtwoNum{14.2} & \RISECNtwoNum{36.2} & \RISECNtwoNum{5.7} & \RISECNtwoNum{5.2} & \RISECNtwoNum{38.6} & \RISECNtwoNum{33.3} \\
Step Image Edit 2~\cite{liu2025step1x} & \RISECNtwono & \RISECNtwoNum{25.9} & \RISECNtwoNum{0.0} & \RISECNtwoNum{5.1} & \RISECNtwoNum{26.6} & \RISECNtwoNum{18.4} & \RISECNtwoNum{12.0} & \RISECNtwoNum{24.0} & \RISECNtwoNum{20.7} & \RISECNtwoNum{2.4} & \RISECNtwoNum{1.8} & \RISECNtwoNum{10.0} & \RISECNtwoNum{16.7} \\
\midrule[0.4pt]
\rowcolor{RISECNtwoGroup}
\multicolumn{14}{l}{\rule{0pt}{10pt}\raisebox{1pt}[0pt][0pt]{\textbf{Open-source Models}}\rule[-2pt]{0pt}{0pt}} \\
\midrule[0.4pt]
HunyuanImage-3.0-Instruct~\cite{cao2025hunyuanimage} & \RISECNtwoyes & \RISECNtwoNum{\textbf{40.0}} & \RISECNtwoNum{23.1} & \RISECNtwoNum{\textbf{15.9}} & \RISECNtwoNum{\textbf{46.3}} & \RISECNtwoNum{15.0} & \RISECNtwoNum{\underline{26.3}} & \RISECNtwoNum{\textbf{32.1}} & \RISECNtwoNum{\textbf{38.3}} & \RISECNtwoNum{\textbf{11.4}} & \RISECNtwoNum{3.9} & \RISECNtwoNum{\textbf{15.7}} & \RISECNtwoNum{30.0} \\
Qwen-Image-2.1~\cite{zhao2026qwen} & \RISECNtwoyes & \RISECNtwoNum{23.5} & \RISECNtwoNum{32.7} & \RISECNtwoNum{12.7} & \RISECNtwoNum{32.5} & \RISECNtwoNum{20.0} & \RISECNtwoNum{15.0} & \RISECNtwoNum{23.6} & \RISECNtwoNum{\underline{26.6}} & \RISECNtwoNum{\underline{6.5}} & \RISECNtwoNum{6.5} & \RISECNtwoNum{\textbf{15.7}} & \RISECNtwoNum{16.7} \\
FireRed-Image-Edit-1.1~\cite{team2026firered} & \RISECNtwoyes & \RISECNtwoNum{27.1} & \RISECNtwoNum{26.9} & \RISECNtwoNum{7.9} & \RISECNtwoNum{33.8} & \RISECNtwoNum{\underline{22.5}} & \RISECNtwoNum{16.3} & \RISECNtwoNum{\underline{26.4}} & \RISECNtwoNum{23.4} & \RISECNtwoNum{0.8} & \RISECNtwoNum{1.3} & \RISECNtwoNum{8.6} & \RISECNtwoNum{30.0} \\
FLUX.2 Dev~\cite{bfl2025flux2} & \RISECNtwoyes & \RISECNtwoNum{18.8} & \RISECNtwoNum{32.7} & \RISECNtwoNum{\underline{14.3}} & \RISECNtwoNum{32.5} & \RISECNtwoNum{20.0} & \RISECNtwoNum{22.5} & \RISECNtwoNum{12.3} & \RISECNtwoNum{21.3} & \RISECNtwoNum{1.6} & \RISECNtwoNum{1.3} & \RISECNtwoNum{11.4} & \RISECNtwoNum{16.7} \\
Qwen-Image-Edit-2511~\cite{wu2025qwen} & \RISECNtwoyes & \RISECNtwoNum{24.7} & \RISECNtwoNum{15.4} & \RISECNtwoNum{\textbf{15.9}} & \RISECNtwoNum{33.8} & \RISECNtwoNum{10.0} & \RISECNtwoNum{10.0} & \RISECNtwoNum{22.6} & \RISECNtwoNum{23.4} & \RISECNtwoNum{2.4} & \RISECNtwoNum{1.3} & \RISECNtwoNum{4.3} & \RISECNtwoNum{23.3} \\
Qwen-Image-Edit-2509~\cite{wu2025qwen} & \RISECNtwoyes & \RISECNtwoNum{24.7} & \RISECNtwoNum{17.3} & \RISECNtwoNum{7.9} & \RISECNtwoNum{28.8} & \RISECNtwoNum{17.5} & \RISECNtwoNum{10.0} & \RISECNtwoNum{22.6} & \RISECNtwoNum{20.2} & \RISECNtwoNum{2.4} & \RISECNtwoNum{0.0} & \RISECNtwoNum{7.1} & \RISECNtwoNum{\underline{33.3}} \\
FireRed-Image-Edit-1.0~\cite{team2026firered} & \RISECNtwoyes & \RISECNtwoNum{29.4} & \RISECNtwoNum{21.2} & \RISECNtwoNum{4.8} & \RISECNtwoNum{26.3} & \RISECNtwoNum{17.5} & \RISECNtwoNum{6.3} & \RISECNtwoNum{23.6} & \RISECNtwoNum{\underline{26.6}} & \RISECNtwoNum{0.8} & \RISECNtwoNum{0.0} & \RISECNtwoNum{4.3} & \RISECNtwoNum{26.7} \\
SenseNova-U1-A3B-MoT-SFT~\cite{diao2026sensenova} & \RISECNtwoyes & \RISECNtwoNum{\underline{38.8}} & \RISECNtwoNum{5.8} & \RISECNtwoNum{11.1} & \RISECNtwoNum{35.0} & \RISECNtwoNum{5.0} & \RISECNtwoNum{\textbf{32.5}} & \RISECNtwoNum{5.7} & \RISECNtwoNum{8.5} & \RISECNtwoNum{3.3} & \RISECNtwoNum{2.6} & \RISECNtwoNum{\underline{14.3}} & \RISECNtwoNum{3.3} \\
JoyAI-Image-Edit-Plus~\cite{song2026joyai} & \RISECNtwoyes & \RISECNtwoNum{4.7} & \RISECNtwoNum{\textbf{36.5}} & \RISECNtwoNum{9.5} & \RISECNtwoNum{31.3} & \RISECNtwoNum{15.0} & \RISECNtwoNum{10.0} & \RISECNtwoNum{16.0} & \RISECNtwoNum{22.3} & \RISECNtwoNum{4.1} & \RISECNtwoNum{6.5} & \RISECNtwoNum{7.1} & \RISECNtwoNum{13.3} \\
FLUX.2 Klein 9B~\cite{bfl2025flux2} & \RISECNtwoyes & \RISECNtwoNum{21.2} & \RISECNtwoNum{\underline{34.6}} & \RISECNtwoNum{12.7} & \RISECNtwoNum{23.8} & \RISECNtwoNum{20.0} & \RISECNtwoNum{6.3} & \RISECNtwoNum{12.3} & \RISECNtwoNum{21.3} & \RISECNtwoNum{0.0} & \RISECNtwoNum{\textbf{9.1}} & \RISECNtwoNum{8.6} & \RISECNtwoNum{10.0} \\
BAGEL (w/ CoT)~\cite{deng2025bagel} & \RISECNtwoyes & \RISECNtwoNum{18.8} & \RISECNtwoNum{26.9} & \RISECNtwoNum{12.7} & \RISECNtwoNum{28.8} & \RISECNtwoNum{17.5} & \RISECNtwoNum{7.5} & \RISECNtwoNum{9.4} & \RISECNtwoNum{24.5} & \RISECNtwoNum{0.0} & \RISECNtwoNum{0.0} & \RISECNtwoNum{5.7} & \RISECNtwoNum{6.7} \\
HiDream-O1-Image~\cite{cai2026hidream} & \RISECNtwoyes & \RISECNtwoNum{21.2} & \RISECNtwoNum{26.9} & \RISECNtwoNum{7.9} & \RISECNtwoNum{25.0} & \RISECNtwoNum{2.5} & \RISECNtwoNum{7.5} & \RISECNtwoNum{9.4} & \RISECNtwoNum{\underline{26.6}} & \RISECNtwoNum{0.0} & \RISECNtwoNum{1.3} & \RISECNtwoNum{5.7} & \RISECNtwoNum{10.0} \\
FLUX.2 Klein 4B~\cite{bfl2025flux2} & \RISECNtwoyes & \RISECNtwoNum{10.6} & \RISECNtwoNum{\textbf{36.5}} & \RISECNtwoNum{7.9} & \RISECNtwoNum{15.0} & \RISECNtwoNum{15.0} & \RISECNtwoNum{8.8} & \RISECNtwoNum{9.4} & \RISECNtwoNum{20.2} & \RISECNtwoNum{1.6} & \RISECNtwoNum{\underline{7.8}} & \RISECNtwoNum{8.6} & \RISECNtwoNum{13.3} \\
UniWorld-V2~\cite{li2025uniworld} & \RISECNtwoyes & \RISECNtwoNum{18.8} & \RISECNtwoNum{9.6} & \RISECNtwoNum{9.5} & \RISECNtwoNum{30.0} & \RISECNtwoNum{10.0} & \RISECNtwoNum{7.5} & \RISECNtwoNum{16.0} & \RISECNtwoNum{14.9} & \RISECNtwoNum{0.0} & \RISECNtwoNum{0.0} & \RISECNtwoNum{7.1} & \RISECNtwoNum{23.3} \\
SenseNova-U1-8B-MoT-SFT~\cite{diao2026sensenova} & \RISECNtwoyes & \RISECNtwoNum{27.1} & \RISECNtwoNum{5.8} & \RISECNtwoNum{9.5} & \RISECNtwoNum{31.3} & \RISECNtwoNum{12.5} & \RISECNtwoNum{15.0} & \RISECNtwoNum{2.8} & \RISECNtwoNum{4.3} & \RISECNtwoNum{4.1} & \RISECNtwoNum{2.6} & \RISECNtwoNum{\underline{14.3}} & \RISECNtwoNum{16.7} \\
UniPic 3.0~\cite{wei2026skywork} & \RISECNtwoyes & \RISECNtwoNum{10.6} & \RISECNtwoNum{3.8} & \RISECNtwoNum{4.8} & \RISECNtwoNum{22.5} & \RISECNtwoNum{7.5} & \RISECNtwoNum{8.8} & \RISECNtwoNum{11.3} & \RISECNtwoNum{13.8} & \RISECNtwoNum{0.0} & \RISECNtwoNum{0.0} & \RISECNtwoNum{8.6} & \RISECNtwoNum{3.3} \\
BAGEL~\cite{deng2025bagel} & \RISECNtwoyes & \RISECNtwoNum{7.1} & \RISECNtwoNum{23.1} & \RISECNtwoNum{3.2} & \RISECNtwoNum{13.8} & \RISECNtwoNum{5.0} & \RISECNtwoNum{1.3} & \RISECNtwoNum{5.7} & \RISECNtwoNum{14.9} & \RISECNtwoNum{0.8} & \RISECNtwoNum{1.3} & \RISECNtwoNum{8.6} & \RISECNtwoNum{6.7} \\
OmniGen2~\cite{wu2025omnigen2} & \RISECNtwoyes & \RISECNtwoNum{3.5} & \RISECNtwoNum{13.5} & \RISECNtwoNum{3.2} & \RISECNtwoNum{5.0} & \RISECNtwoNum{10.0} & \RISECNtwoNum{3.8} & \RISECNtwoNum{0.9} & \RISECNtwoNum{11.7} & \RISECNtwoNum{0.0} & \RISECNtwoNum{0.0} & \RISECNtwoNum{7.1} & \RISECNtwoNum{3.3} \\
OmniGen~\cite{xiao2024omnigen} & \RISECNtwoyes & \RISECNtwoNum{2.4} & \RISECNtwoNum{15.4} & \RISECNtwoNum{6.3} & \RISECNtwoNum{1.3} & \RISECNtwoNum{0.0} & \RISECNtwoNum{5.0} & \RISECNtwoNum{2.8} & \RISECNtwoNum{7.4} & \RISECNtwoNum{4.1} & \RISECNtwoNum{1.3} & \RISECNtwoNum{2.9} & \RISECNtwoNum{6.7} \\
Emu2-Gen~\cite{sun2023emu2chat} & \RISECNtwoyes & \RISECNtwoNum{3.5} & \RISECNtwoNum{1.9} & \RISECNtwoNum{1.6} & \RISECNtwoNum{1.3} & \RISECNtwoNum{2.5} & \RISECNtwoNum{0.0} & \RISECNtwoNum{0.9} & \RISECNtwoNum{1.1} & \RISECNtwoNum{0.0} & \RISECNtwoNum{0.0} & \RISECNtwoNum{1.4} & \RISECNtwoNum{3.3} \\
JoyAI-Image-Edit~\cite{song2026joyai} & \RISECNtwono & \RISECNtwoNum{17.6} & \RISECNtwoNum{4.2} & \RISECNtwoNum{11.9} & \RISECNtwoNum{\underline{39.2}} & \RISECNtwoNum{21.1} & \RISECNtwoNum{9.3} & \RISECNtwoNum{14.0} & \RISECNtwoNum{17.1} & \RISECNtwoNum{0.0} & \RISECNtwoNum{3.5} & \RISECNtwoNum{\underline{14.3}} & \RISECNtwoNum{\textbf{36.7}} \\
Boogu-Image-0.1-Edit~\cite{chen2026boogu} & \RISECNtwono & \RISECNtwoNum{14.1} & \RISECNtwoNum{8.3} & \RISECNtwoNum{13.6} & \RISECNtwoNum{35.4} & \RISECNtwoNum{18.4} & \RISECNtwoNum{10.7} & \RISECNtwoNum{14.0} & \RISECNtwoNum{20.7} & \RISECNtwoNum{2.4} & \RISECNtwoNum{1.8} & \RISECNtwoNum{\underline{14.3}} & \RISECNtwoNum{23.3} \\
\resizebox{\linewidth}{!}{Step1X-Edit (thinking + reflection)~\cite{liu2025step1x}} & \RISECNtwono & \RISECNtwoNum{27.1} & \RISECNtwoNum{0.0} & \RISECNtwoNum{10.2} & \RISECNtwoNum{21.5} & \RISECNtwoNum{\textbf{23.7}} & \RISECNtwoNum{12.0} & \RISECNtwoNum{11.0} & \RISECNtwoNum{11.0} & \RISECNtwoNum{1.6} & \RISECNtwoNum{0.0} & \RISECNtwoNum{12.9} & \RISECNtwoNum{16.7} \\
Step1X-Edit (thinking)~\cite{liu2025step1x} & \RISECNtwono & \RISECNtwoNum{18.8} & \RISECNtwoNum{4.2} & \RISECNtwoNum{11.9} & \RISECNtwoNum{21.5} & \RISECNtwoNum{7.9} & \RISECNtwoNum{12.0} & \RISECNtwoNum{11.0} & \RISECNtwoNum{15.9} & \RISECNtwoNum{1.6} & \RISECNtwoNum{3.5} & \RISECNtwoNum{\textbf{15.7}} & \RISECNtwoNum{20.0} \\
Step1X-Edit (base)~\cite{liu2025step1x} & \RISECNtwono & \RISECNtwoNum{17.6} & \RISECNtwoNum{0.0} & \RISECNtwoNum{11.9} & \RISECNtwoNum{24.1} & \RISECNtwoNum{13.2} & \RISECNtwoNum{9.3} & \RISECNtwoNum{13.0} & \RISECNtwoNum{12.2} & \RISECNtwoNum{0.8} & \RISECNtwoNum{0.0} & \RISECNtwoNum{\underline{14.3}} & \RISECNtwoNum{16.7} \\
LongCat-Image-Edit~\cite{team2025longcat} & \RISECNtwono & \RISECNtwoNum{21.2} & \RISECNtwoNum{4.2} & \RISECNtwoNum{6.8} & \RISECNtwoNum{22.8} & \RISECNtwoNum{18.4} & \RISECNtwoNum{8.0} & \RISECNtwoNum{13.0} & \RISECNtwoNum{13.4} & \RISECNtwoNum{0.0} & \RISECNtwoNum{3.5} & \RISECNtwoNum{5.7} & \RISECNtwoNum{16.7} \\
FLUX.1 Kontext Dev~\cite{labs2025flux1kontextflowmatching} & \RISECNtwono & \RISECNtwoNum{1.2} & \RISECNtwoNum{0.0} & \RISECNtwoNum{6.8} & \RISECNtwoNum{5.1} & \RISECNtwoNum{0.0} & \RISECNtwoNum{4.0} & \RISECNtwoNum{1.0} & \RISECNtwoNum{7.3} & \RISECNtwoNum{4.1} & \RISECNtwoNum{1.8} & \RISECNtwoNum{5.7} & \RISECNtwoNum{10.0} \\
LLaDA-Image~\cite{chen2026llada} & \RISECNtwono & \RISECNtwoNum{5.9} & \RISECNtwoNum{4.2} & \RISECNtwoNum{3.4} & \RISECNtwoNum{10.1} & \RISECNtwoNum{0.0} & \RISECNtwoNum{4.0} & \RISECNtwoNum{2.0} & \RISECNtwoNum{2.4} & \RISECNtwoNum{0.0} & \RISECNtwoNum{1.8} & \RISECNtwoNum{5.7} & \RISECNtwoNum{13.3} \\
InternVL-U (w/ CoT)~\cite{tian2026internvl} & \RISECNtwono & \RISECNtwoNum{8.2} & \RISECNtwoNum{0.0} & \RISECNtwoNum{5.1} & \RISECNtwoNum{6.3} & \RISECNtwoNum{2.6} & \RISECNtwoNum{0.0} & \RISECNtwoNum{0.0} & \RISECNtwoNum{4.9} & \RISECNtwoNum{0.0} & \RISECNtwoNum{0.0} & \RISECNtwoNum{1.4} & \RISECNtwoNum{10.0} \\
Ovis-U1~\cite{wang2025ovis} & \RISECNtwono & \RISECNtwoNum{5.9} & \RISECNtwoNum{0.0} & \RISECNtwoNum{1.7} & \RISECNtwoNum{7.6} & \RISECNtwoNum{5.3} & \RISECNtwoNum{4.0} & \RISECNtwoNum{0.0} & \RISECNtwoNum{4.9} & \RISECNtwoNum{0.8} & \RISECNtwoNum{0.0} & \RISECNtwoNum{1.4} & \RISECNtwoNum{3.3} \\
InternVL-U~\cite{tian2026internvl} & \RISECNtwono & \RISECNtwoNum{5.9} & \RISECNtwoNum{0.0} & \RISECNtwoNum{1.7} & \RISECNtwoNum{7.6} & \RISECNtwoNum{2.6} & \RISECNtwoNum{1.3} & \RISECNtwoNum{0.0} & \RISECNtwoNum{2.4} & \RISECNtwoNum{0.8} & \RISECNtwoNum{0.0} & \RISECNtwoNum{2.9} & \RISECNtwoNum{10.0} \\
Lumina-DiMOO~\cite{xin2025lumina} & \RISECNtwono & \RISECNtwoNum{2.4} & \RISECNtwoNum{4.2} & \RISECNtwoNum{0.0} & \RISECNtwoNum{5.1} & \RISECNtwoNum{2.6} & \RISECNtwoNum{0.0} & \RISECNtwoNum{3.0} & \RISECNtwoNum{1.2} & \RISECNtwoNum{1.6} & \RISECNtwoNum{3.5} & \RISECNtwoNum{2.9} & \RISECNtwoNum{6.7} \\
UniWorld-V1~\cite{lin2025uniworld} & \RISECNtwono & \RISECNtwoNum{3.5} & \RISECNtwoNum{0.0} & \RISECNtwoNum{3.4} & \RISECNtwoNum{1.3} & \RISECNtwoNum{0.0} & \RISECNtwoNum{1.3} & \RISECNtwoNum{1.0} & \RISECNtwoNum{2.4} & \RISECNtwoNum{0.0} & \RISECNtwoNum{0.0} & \RISECNtwoNum{2.9} & \RISECNtwoNum{3.3} \\
UniPic2-SD3.5M-Kontext-2B~\cite{wei2025skywork} & \RISECNtwono & \RISECNtwoNum{0.0} & \RISECNtwoNum{0.0} & \RISECNtwoNum{0.0} & \RISECNtwoNum{1.3} & \RISECNtwoNum{0.0} & \RISECNtwoNum{1.3} & \RISECNtwoNum{1.0} & \RISECNtwoNum{2.4} & \RISECNtwoNum{1.6} & \RISECNtwoNum{0.0} & \RISECNtwoNum{1.4} & \RISECNtwoNum{3.3} \\
\midrule[0.4pt]
\rowcolor{RISECNtwoGroup}
\multicolumn{14}{l}{\rule{0pt}{10pt}\raisebox{1pt}[0pt][0pt]{\textbf{Agentic Methods}}\rule[-2pt]{0pt}{0pt}} \\
\midrule[0.4pt]
\textbf{RISE-Agent (Ours)} & \RISECNtwoyes & \RISECNtwoNum{\textbf{65.9}} & \RISECNtwoNum{\textbf{46.2}} & \RISECNtwoNum{\textbf{34.9}} & \RISECNtwoNum{\textbf{67.5}} & \RISECNtwoNum{\textbf{45.0}} & \RISECNtwoNum{\textbf{52.5}} & \RISECNtwoNum{\textbf{31.1}} & \RISECNtwoNum{\textbf{37.2}} & \RISECNtwoNum{\textbf{30.9}} & \RISECNtwoNum{\textbf{15.6}} & \RISECNtwoNum{\textbf{55.7}} & \RISECNtwoNum{\textbf{50.0}} \\
MindBrush~\cite{he2026mind} & \RISECNtwono & \RISECNtwoNum{\underline{54.1}} & \RISECNtwoNum{\underline{37.5}} & \RISECNtwoNum{\underline{30.5}} & \RISECNtwoNum{\underline{43.0}} & \RISECNtwoNum{13.2} & \RISECNtwoNum{\underline{34.7}} & \RISECNtwoNum{\underline{18.0}} & \RISECNtwoNum{22.0} & \RISECNtwoNum{\underline{4.9}} & \RISECNtwoNum{\underline{3.5}} & \RISECNtwoNum{\underline{44.3}} & \RISECNtwoNum{16.7} \\
InterleaveThinker~\cite{zheng2026interleavethinker} & \RISECNtwono & \RISECNtwoNum{34.1} & \RISECNtwoNum{12.5} & \RISECNtwoNum{16.9} & \RISECNtwoNum{30.4} & \RISECNtwoNum{\underline{23.7}} & \RISECNtwoNum{28.0} & \RISECNtwoNum{16.0} & \RISECNtwoNum{\underline{23.2}} & \RISECNtwoNum{0.8} & \RISECNtwoNum{0.0} & \RISECNtwoNum{24.3} & \RISECNtwoNum{\underline{36.7}} \\
IntentEdit~\cite{zhang2026intentedit} & \RISECNtwono & \RISECNtwoNum{34.1} & \RISECNtwoNum{12.5} & \RISECNtwoNum{10.2} & \RISECNtwoNum{36.7} & \RISECNtwoNum{7.9} & \RISECNtwoNum{18.7} & \RISECNtwoNum{12.0} & \RISECNtwoNum{17.1} & \RISECNtwoNum{0.8} & \RISECNtwoNum{\underline{3.5}} & \RISECNtwoNum{15.7} & \RISECNtwoNum{10.0} \\
ImAgent~\cite{wang2025imagent} & \RISECNtwono & \RISECNtwoNum{7.1} & \RISECNtwoNum{4.2} & \RISECNtwoNum{0.0} & \RISECNtwoNum{6.3} & \RISECNtwoNum{2.6} & \RISECNtwoNum{1.3} & \RISECNtwoNum{1.0} & \RISECNtwoNum{3.7} & \RISECNtwoNum{0.0} & \RISECNtwoNum{0.0} & \RISECNtwoNum{1.4} & \RISECNtwoNum{3.3} \\
\bottomrule[0.8pt]
\end{tabular}
\par\vspace{4pt}
\parbox{\textwidth}{\scriptsize
Results are reported as accuracy (\%).
CSK: Common-Sense Knowledge; CtxK: Context Knowledge; DK: Disciplinary Knowledge;
Rel.: Relation; Struct.: Structure; ERD: Explicit Rule Deduction; IPI: Implicit Pattern Induction;
CFI: Counterfactual Imagination; ACG: Anomaly \& Correction Generation.
Models with multi-image support are evaluated on both single- and multi-image cases; otherwise, only single-image cases are used.}
\endgroup
\end{table*}

We further analyze model performance across fine-grained reasoning subcategories on both the English and Chinese versions of RISEBench++ in Tables~\ref{tab:subcategory_accuracy_en} and~\ref{tab:subcategory_accuracy_cn}, respectively. Since the English version constitutes our primary evaluation setting, we focus the following discussion on the English results, while providing the Chinese results as a complementary reference. 

For knowledge-driven temporal and causal reasoning, strong closed-source models generally perform better on tasks involving common-sense knowledge than on those requiring context-specific or disciplinary knowledge. For example, Nano Banana 2 achieves 77.6\% on temporal common-sense knowledge, while GPT-Image-1.5 and GPT-Image-2.5 Sunburst reach 80.0\% and 78.8\% on causal common-sense knowledge, respectively. In comparison, performance is generally lower and more variable when models must extract image-specific contextual evidence or incorporate specialized disciplinary knowledge. This suggests that leading models can effectively leverage familiar real-world knowledge, but remain less reliable when reasoning requires tighter integration of visual evidence with contextual or specialized knowledge.

For perception-driven tasks, spatial reasoning is considerably more tractable than logical reasoning. Leading models achieve above 60\% accuracy on both spatial relation and structure, indicating relatively strong capabilities in understanding object relationships and scene organization. In contrast, logical reasoning remains particularly challenging. Explicit rule deduction is difficult but still achievable for several models, whereas implicit pattern induction exhibits consistently low accuracy: even the strongest closed-source result reaches only 31.2\%. This contrast suggests that applying an explicitly specified rule is generally easier than discovering latent regularities from visual observations and translating them into precise edits. Notably, RISE-Agent achieves 47.2\% on explicit rule deduction, outperforming all evaluated closed-source models on this subcategory. This result demonstrates the effectiveness of combining structured reasoning, tool-augmented execution, and iterative verification for tasks that require both accurate deduction and precise realization of the inferred solution.

Counterfactual reasoning presents a different pattern. Strong models achieve relatively high accuracy on both counterfactual imagination and anomaly correction/generation, with GPT-Image-2.5 Sunburst reaching 71.4\% and 73.3\%, respectively. This indicates that leading models can often leverage learned real-world knowledge not only to identify violations of familiar rules, but also to deliberately construct outcomes that contradict them when instructed. Overall, the subcategory results reveal a pronounced capability imbalance: current models are increasingly effective at leveraging familiar knowledge and understanding explicit spatial structures, while reasoning from image-specific evidence and, especially, discovering implicit visual rules remain substantially more challenging.

\subsection{Fine-grained Analysis by Task Type}
\begin{figure*}[t]
    \centering
    \includegraphics[width=\textwidth]{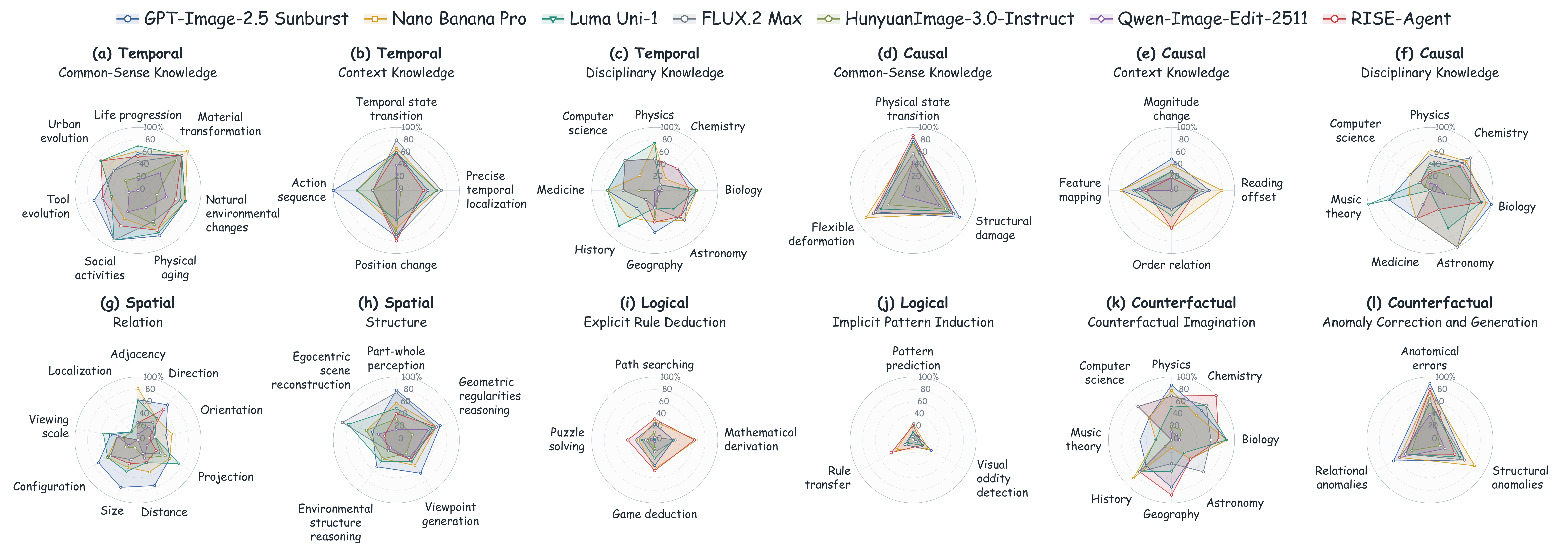}
    \caption{\textbf{Radar charts of fine-grained task accuracy (\%) on RISEBench++ (EN).} Results for seven models across 65 task types in 12 subcategories.}
    \label{fig:subcategory_radars}
\end{figure*}
We select seven models---four closed-source models, two open-source models, and one agentic method---for task-level analysis and visualization. Figure~\ref{fig:subcategory_radars} shows that several models perform well on material transformation and physical state transition, reflecting their ability to use relevant knowledge to depict changes in appearance and physical state. Reading offset also requires precise editing: even when the numerical value is correct, unnecessary changes to the original display can cause an edit to fail. Pattern prediction remains difficult across all seven models, with accuracy no higher than 25.6\%. The results reveal errors in symbol positions and orientations, indicating that appearance consistency and visual plausibility alone do not guarantee accurate instruction reasoning.

The models also show different strengths within the same reasoning dimension. GPT-Image-2.5 Sunburst leads on part-whole perception and viewpoint generation, whereas FLUX.2 Max and Luma Uni-1 excel at egocentric scene reconstruction, showing different strengths in understanding object geometry and combining multiple views into a coherent scene. Nano Banana Pro stands out on flexible deformation and reading offset, showing strengths in predicting the effects of physical changes and accurately updating readings in images. Among the two open-source models, HunyuanImage-3.0-Instruct performs relatively well on anatomical anomaly editing (70.0\%), while Qwen-Image-Edit-2511 shows a relative strength in geometric regularities reasoning (53.3\%). However, both models are less accurate than the four selected closed-source models and RISE-Agent on reading offset and puzzle solving, indicating remaining difficulties with numerical updates and rule-based visual problems. RISE-Agent achieves the highest accuracy on puzzle solving and rule transfer among the seven models, but its accuracy on pattern prediction remains low.

\subsection{Validity of LMM-as-a-Judge}
\begin{table*}[tp]
\centering
\caption{\textbf{Comparison of different judge models} on instruction reasoning, appearance consistency, visual plausibility, and overall evaluation. Lower MAE is better, while higher Spearman correlation is better.}
\label{tab:human_align}
\resizebox{\textwidth}{!}{
\begin{tabular}{lcccccccc}
\toprule
\multirow{2}{*}{\textbf{Judge}}
& \multicolumn{2}{c}{\textbf{Instruction Reasoning}}
& \multicolumn{2}{c}{\textbf{Appearance Consistency}}
& \multicolumn{2}{c}{\textbf{Visual Plausibility}}
& \multicolumn{2}{c}{\textbf{Overall}} \\
\cmidrule(lr){2-3}
\cmidrule(lr){4-5}
\cmidrule(lr){6-7}
\cmidrule(lr){8-9}
& \textbf{MAE} $\downarrow$
& \textbf{Spearman} $\uparrow$
& \textbf{MAE} $\downarrow$
& \textbf{Spearman} $\uparrow$
& \textbf{MAE} $\downarrow$
& \textbf{Spearman} $\uparrow$
& \textbf{MAE} $\downarrow$
& \textbf{Spearman} $\uparrow$ \\
\midrule

GPT-4.1
& 0.88 & 0.64
& 0.62 & 0.51
& 0.35 & 0.35
& 0.60 & 0.73 \\

GPT-5
& 0.73 & 0.74
& 0.66 & \underline{0.59}
& 0.53 & 0.27
& \underline{0.52} & 0.81 \\

Gemini-3.1-Flash-Lite
& 0.71 & 0.71
& 0.76 & 0.47
& \textbf{0.27} & \textbf{0.41}
& 0.60 & 0.74 \\

Gemini-3-Flash
& \textbf{0.48} & \textbf{0.84}
& \textbf{0.58} & \textbf{0.61}
& 0.39 & 0.37
& \textbf{0.46} & \textbf{0.85} \\

Seed-2.0-Lite
& \underline{0.64} & \underline{0.81}
& 0.88 & \textbf{0.61}
& 0.52 & 0.39
& 0.59 & \underline{0.82} \\

Qwen3-VL-235B-A22B-Instruct
& 0.85 & 0.64
& \underline{0.60} & 0.56
& \underline{0.28} & \underline{0.40}
& 0.64 & 0.69 \\

\bottomrule
\end{tabular}
}
\end{table*}
To evaluate how well LMM-based judges align with human preferences, we conduct a human evaluation study on a representative subset of RISEBench++. We first uniformly sample 100 cases from the full benchmark of 1,000 samples to preserve a similar data distribution. We then collect outputs from two models (Nano Banana Pro, Qwen-Image-Edit-2511) on these 100 cases, resulting in 200 edited samples in total. Four human annotators independently score each sample using the same evaluation rubric as the LMM judges. We average the human scores and compare them with the scores produced by different candidate judges. The alignment is measured using Mean Absolute Error (MAE) and Spearman correlation. The results are reported in Table~\ref{tab:human_align}.

As shown in Table~\ref{tab:human_align}, Gemini-3-Flash aligns most closely with human judgments on instruction reasoning and appearance consistency. 
It achieves the strongest overall MAE--Spearman trade-off across both dimensions, with high agreement on instruction reasoning (0.48 MAE and 0.84 Spearman). We therefore select Gemini-3-Flash as the judge for these two criteria. For visual plausibility, Gemini-3.1-Flash-Lite aligns best with human judgments, achieving 0.27 MAE and 0.41 Spearman.
Given its lower inference cost, we adopt it as the visual plausibility judge. These results provide empirical support for our final judge configuration.

\begin{figure*}
    \centering
    \includegraphics[width=0.95\linewidth]{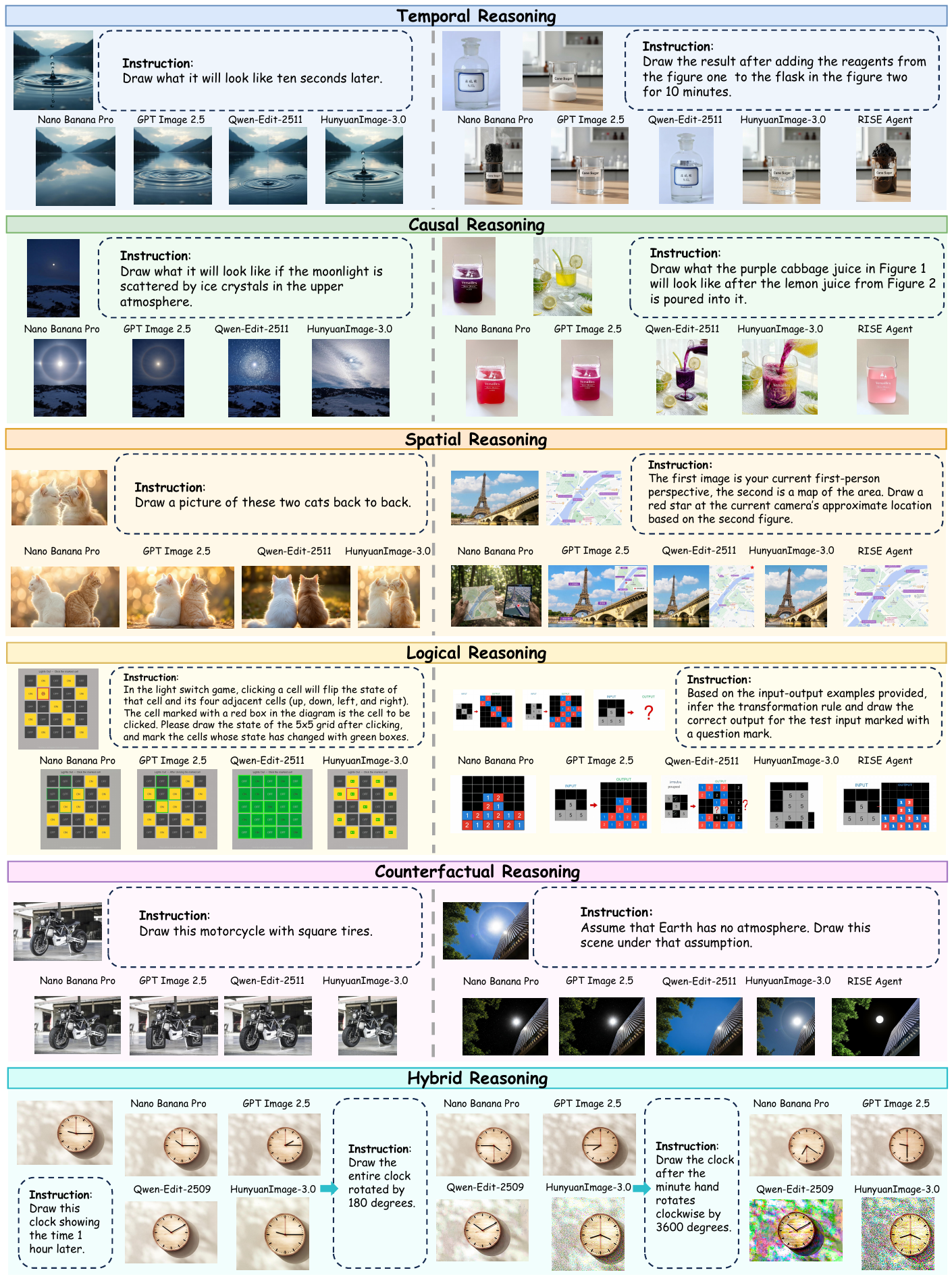}
    \caption{\textbf{Examples of several different models' outputs on RISEBench++.} The analyzed models demonstrate distinct characteristics in their responses.
    }
    \label{fig:examples}
\end{figure*}

\subsection{Ablation Study of RISE-Agent}
\begin{table*}[tp]
    \centering
    \caption{\textbf{Ablation study on key components of RISE-Agent,} with results reported as accuracy (\%). FLUX.2 Klein 9B without RISE-Agent components serves as the baseline. Best and second-best results are shown in \textbf{bold} and \underline{underline}, respectively.}
    \label{tab:rise-edit_ablation}
    \setlength{\tabcolsep}{7pt}
    \renewcommand{\arraystretch}{1.15}

    \resizebox{0.91\textwidth}{!}{
    \begin{tabular}{ccc@{\hspace{14pt}}ccccccc}
        \toprule
        \multicolumn{3}{c}{\textbf{Components}}
        &
        \multicolumn{7}{c}{\textbf{Performance on RISEBench++}} \\
        \cmidrule(lr){1-3}
        \cmidrule(lr){4-10}

        \textbf{Planner}
        & \textbf{Tools}
        & \textbf{Verifier}
        & \textbf{Temporal}
        & \textbf{Causal}
        & \textbf{Spatial}
        & \textbf{Logical}
        & \textbf{Counterfactual}
        & \textbf{Hybrid}
        & \textbf{Overall} \\

        \midrule

        --
        & --
        & --
        & 16.0
        & 16.5
        & 22.0
        & 0.5
        & 17.0
        & 2.0
        & 12.9 \\

        \checkmark
        & --
        & --
        & \underline{58.0}
        & 50.5
        & \textbf{48.0}
        & 14.0
        & 51.0
        & \underline{13.0}
        & 40.5 \\

        \checkmark
        & \checkmark
        & --
        & 50.5
        & \underline{56.0}
        & \underline{37.5}
        & \underline{32.0}
        & \underline{61.0}
        & 12.0
        & \underline{42.6} \\

        \checkmark
        & \checkmark
        & \checkmark
        & \textbf{58.5}
        & \textbf{58.0}
        & \underline{37.5}
        & \textbf{39.0}
        & \textbf{66.0}
        & \textbf{28.0}
        & \textbf{48.0} \\

        \bottomrule
    \end{tabular}
    }
\end{table*}

To better understand the contribution of each component in RISE-Agent, we progressively augment FLUX.2 Klein 9B with the planner, tools, and verifier. As shown in Table~\ref{tab:rise-edit_ablation}, adding the planner alone substantially increases the overall score from 12.9\% to 40.5\%, with especially pronounced gains on temporal and causal tasks. These results suggest that many failures arise from the difficulty current editors face in both inferring the intended target state and realizing it through a single-step generation process. By explicitly reasoning about the target state and translating it into an actionable editing prompt, the planner alleviates this burden on the editor. In effect, it decouples target-state inference from edit execution, leading to more reliable editing outcomes. The gains on logical tasks are comparatively smaller, likely because correct high-level reasoning alone is insufficient for the precise geometric or symbolic manipulations required by such inputs.
Adding tools further improves performance on logical, causal, and counterfactual tasks, raising the overall score to 42.6\%. In contrast, temporal and spatial scores decrease slightly, likely because these categories are dominated by natural-image edits, for which programmatic operations are less suitable than generative editing and may introduce unnatural artifacts or localization errors. Finally, verifier-guided refinement increases the overall score to 48.0\% and achieves the best performance in five of the six categories, demonstrating the effectiveness of iterative diagnosis and correction. The limited improvement in spatial tasks may be attributed to the verifier's limited spatial understanding capabilities. Overall, these results highlight the complementary contributions of the three components, whose integration yields progressively stronger overall performance.

\subsection{Qualitative Examples}

We exhibit several representative model outputs in Fig.~\ref{fig:examples} and observe several notable characteristics of the evaluated models. In relatively intuitive cases, strong models can often translate the inferred consequence into plausible visual changes, such as depicting the scene ten seconds later, rendering a lunar halo caused by ice-crystal scattering, or placing two cats back to back. More reasoning-intensive cases, however, expose clearer differences across models. For example, models vary substantially in predicting the outcome of adding reagents and localizing the current viewpoint from a map, whereas RISE-Agent correctly captures the expected reaction and identifies the corresponding location. Rule-based transformations remain challenging as well: in the light-switch example, Qwen-Image-Edit-2511 alters nearly the entire grid, and several other models produce incorrect cell states, while RISE-Agent applies the operation precisely to the affected cells. This further highlights the importance of not only reasoning about the intended state transition, but also executing it precisely.

The hybrid cases further expose a different limitation that becomes particularly apparent under sequential editing. Operations such as advancing the clock by one hour, rotating the entire clock by 180 degrees, and subsequently rotating the minute hand clockwise by 3600 degrees require each intermediate result to remain both semantically correct and visually stable for the next edit.
While some models preserve the scene reasonably well across turns, Qwen-Image-Edit-2509 and HunyuanImage-3.0 both exhibit cumulative visual degradation, with later outputs progressively losing structural integrity and visual fidelity. This suggests that errors in multi-round reasoning-informed editing can accumulate not only at the reasoning level, but also through multi-step visual regeneration, making long-horizon consistency an additional challenge beyond solving each individual edit correctly. 
\FloatBarrier
\section{CONCLUSION}
In this work, we present RISEBench++, a comprehensive and fine-grained benchmark for Reasoning-Informed viSual Editing (RISE), extending RISEBench with a hierarchical taxonomy spanning six fundamental reasoning dimensions, 12 subcategories, and 65 fine-grained task types. Built upon this taxonomy, RISEBench++ comprises 1,000 carefully curated and human-annotated bilingual test cases and supports comprehensive evaluation across single-image, multi-image, and multi-turn settings.
We establish an evaluation framework that jointly assesses instruction reasoning, appearance consistency, and visual plausibility, with LMM-based judgments validated against human assessments. Beyond benchmarking, we introduce RISE-Agent, a training-free agentic framework integrating reasoning-driven planning, tool-augmented execution, and verifier-guided refinement. 
Extensive experiments on 58 visual editing approaches reveal that, despite substantial progress in visual quality and appearance preservation, reasoning-informed editing remains challenging, with the strongest evaluated approach achieving only 56.6\% accuracy.
In particular, inferring implicit visual rules, precisely executing reasoning-dependent transformations, and maintaining consistency across successive edits remain critical bottlenecks.
The results of RISE-Agent further demonstrate the promise of agentic approaches for reasoning-informed visual editing, highlighting agentic frameworks as an effective paradigm for tackling challenging editing tasks that require both complex reasoning and precise execution.
These findings suggest that advancing visual editing requires not only stronger generative capabilities, but also robust visual perception, sophisticated multimodal reasoning, and precise execution.
We hope that RISEBench++ and RISE-Agent provide a useful foundation for systematic evaluation and further research toward more capable and reliable reasoning-informed visual editing.


\bibliographystyle{plainnat}
\bibliography{bib/main}

\clearpage
\appendix
\section*{{\titlefont\sffamily\themeblue{Supplementary Material}}}
\phantomsection
\addcontentsline{toc}{section}{Supplementary Material}
\begingroup
\let\RISEOriginalSection\section
\let\subsection\RISEOriginalSection
\RenewDocumentCommand{\section}{s o m}{\let\section\RISEOriginalSection}

\section*{Supplementary Material}
\label{sec:appendix}

\newsavebox{\RISEModelNameBox}
\newcommand{\RISEFitModel}[1]{\sbox{\RISEModelNameBox}{#1}\ifdim\wd\RISEModelNameBox>\linewidth
    \resizebox{\linewidth}{!}{\usebox{\RISEModelNameBox}}\else
    \usebox{\RISEModelNameBox}\fi
}

\subsection{Taxonomy and Statistics of RISEBench++}

RISEBench++ comprises 1,000 carefully curated tasks spanning six major reasoning dimensions: temporal, causal, spatial, logical, counterfactual, and hybrid reasoning. As summarized in Table~\ref{tab:task_taxonomy}, each dimension is further organized into fine-grained subcategories and task types, covering diverse reasoning scenarios from common-sense and disciplinary knowledge to spatial relations, rule-based deduction, and counterfactual imagination. This hierarchical taxonomy is designed to provide broad and balanced coverage of reasoning capabilities while enabling fine-grained analysis of model performance across different reasoning skills.

\subsection{Generalization Evaluation of RISE-Agent on KRIS-Bench}

To further evaluate the generalization ability of RISE-Agent, we conduct additional experiments on KRIS-Bench, which assesses factual, conceptual, and procedural knowledge. As shown in Table~\ref{tab:kris_bench_results}, RISE-Agent achieves an overall score of 88.12, outperforming all open-source models and existing agentic methods. Among agentic methods, RISE-Agent obtains the best results across all reported knowledge categories, with particularly strong performance on factual knowledge (89.14) and conceptual knowledge (87.77). It also achieves 97.06 on instruction-following tasks. While the closed-source GPT-Image-2.5 models achieve the highest overall scores on KRIS-Bench, RISE-Agent remains competitive and substantially narrows the gap while relying on an agentic framework. These results indicate that the effectiveness of RISE-Agent transfers well to an additional benchmark with a different evaluation setup, providing further evidence of its generalization capability beyond RISEBench++.

\subsection{Detailed Performance across Three Evaluation Dimensions}

Tables~\ref{tab:dimension_metric_scores_en} and~\ref{tab:dimension_metric_scores_cn} provide a fine-grained breakdown of model performance across different reasoning categories under the three evaluation dimensions, i.e., IR, AC, and VP, for the English and Chinese versions, respectively. Overall, the results reveal substantial variation across both evaluation dimensions and reasoning categories. While many models achieve relatively strong AC and VP scores, IR remains considerably more challenging, particularly for logical and hybrid reasoning. In contrast, temporal, causal, and spatial reasoning generally yield higher IR scores, suggesting that current models handle more explicit structural or relational constraints better than reasoning processes that require multi-step logical composition or the integration of multiple reasoning types. The gap is especially evident among open-source models, where high visual quality or constraint adherence does not necessarily translate into strong reasoning correctness. Importantly, these trends are broadly consistent across the English and Chinese settings, indicating that the observed difficulty is not specific to a particular language. Together, these results highlight the complementary nature of IR, AC, and VP and reinforce the importance of evaluating reasoning-intensive image generation beyond visual quality or instruction compliance alone.

\subsection{Fine-grained Accuracy across Task Types}

We report task-level accuracy for all evaluated models in Tables~\ref{tab:task_accuracy_en_1}--\ref{tab:task_accuracy_en_3} and Tables~\ref{tab:task_accuracy_cn_1}--\ref{tab:task_accuracy_cn_3} for the English and Chinese versions, respectively. 
Closed-source models generally achieve higher accuracy across a broad range of task types, particularly those involving physical changes, spatial structures, and counterfactual scenarios. Open-source models show relative strengths in some physical transformation and anomaly editing tasks, but remain weaker on tasks requiring disciplinary knowledge or logical rules. Agentic methods vary substantially, with some performing competitively on knowledge-based and rule-based tasks while others remain weak across many categories. Moreover, even strong models exhibit noticeable variation across individual task types. For example, Nano Banana Pro achieves 100.0\% accuracy on Material Transformation but only 25.0\% on Chemistry in the English setting, indicating that performance can vary substantially even within the same reasoning category. Logical tasks, especially pattern prediction and other rule-intensive problems, are consistently challenging across all three groups. Similar trends are observed in both English and Chinese, suggesting that these task-level difficulty patterns are largely consistent across languages and may be obscured by aggregate scores.

\subsection{Score Distributions across Models}

We present the score distributions of ten selected models on RISEBench++ in Figure~\ref{fig:score_distributions_10models_en}. GPT-Image-2.5 Sunburst, Nano Banana Pro, and RISE-Agent show a high proportion of favorable scores across IR, AC, and VP, while Luma Uni-1 and FLUX.2 Max also achieve predominantly high ratings across all three metrics. In contrast, several other models receive fewer high scores, particularly on IR and AC, with their IR scores more concentrated at 1 and 2. VP remains relatively strong across most models, although HiDream-O1-Image shows the lowest proportion of score-5 ratings and a larger share of low scores than the others. Overall, the performance differences among the selected models are most pronounced in IR and AC. In contrast, VP scores are highly concentrated at the upper end, suggesting that this metric may become less discriminative as model performance continues to improve.

\subsection{Human Alignment Subset}

Table~\ref{tab:human_scores_by_reasoning_category} summarizes the subset used to assess LMM-as-a-Judge alignment with human judgments and the corresponding human annotations. For each category, it reports the distribution of human scores from 1 to 5 across the three evaluation dimensions.

\subsection{Prompts for Automated Evaluation}

We present all our prompts for automated evaluation across different metrics and reasoning dimensions in Figures~\ref{fig:judge_general_ref_txt}--\ref{fig:judge_counterfactual_qual}. Each prompt is tailored to its corresponding evaluation dimensions and reasoning categories, with specific criteria and instructions designed to promote reliable and consistent judgments.



\begin{table*}[!tp]
\caption{Taxonomy and Statistics of RISEBench++.}
\label{tab:task_taxonomy}
\centering
\begingroup
\fontsize{8.5}{11.5}\selectfont
\setlength{\tabcolsep}{3pt}
\renewcommand{\arraystretch}{1.0}
\setlength{\aboverulesep}{2pt}
\setlength{\belowrulesep}{2pt}
\colorlet{RISETaxonomyPanel}{black!7}
\begin{minipage}[t]{.485\textwidth}
\vspace{0pt}
\renewcommand{\arraystretch}{1.0875}
\begin{tabular}{
>{\centering\arraybackslash}m{\dimexpr .325\linewidth-2\tabcolsep\relax}
>{\centering\arraybackslash}m{\dimexpr .675\linewidth-2\tabcolsep\relax}
}
\toprule[0.75pt]
\textbf{Subcategory} & \textbf{Task Type} \\
\midrule[0.4pt]
\rowcolor{RISETaxonomyPanel}
\multicolumn{2}{c}{\rule{0pt}{13pt}\raisebox{1.5pt}[0pt][0pt]{\textbf{Temporal Reasoning~(200, 32)}}\rule[-4pt]{0pt}{0pt}} \\
\addlinespace[3pt]
\multirow[c]{7}{=}{\centering\arraybackslash Common-Sense Knowledge~(85)} & Life Progression~(24) \\
 & Material Transformation~(9) \\
 & Natural Environmental Changes~(13) \\
 & Physical Aging~(20) \\
 & Social Activities~(8) \\
 & Tool Evolution~(7) \\
 & Urban Evolution~(4) \\
\cmidrule(lr){1-2}
\multirow[c]{4}{=}{\centering\arraybackslash Context Knowledge~(52, 28)} & Temporal State Transition~(15, 4) \\
 & Precise Temporal Localization~(14, 2) \\
 & Position Change~(15, 14) \\
 & Action Sequence~(8, 8) \\
\cmidrule(lr){1-2}
\multirow[c]{8}{=}{\centering\arraybackslash Disciplinary Knowledge~(63, 4)} & Physics~(4) \\
 & Chemistry~(8, 4) \\
 & Biology~(9) \\
 & Astronomy~(12) \\
 & Geography~(18) \\
 & History~(5) \\
 & Medicine~(4) \\
 & Computer Science~(3) \\
\addlinespace[7pt]
\rowcolor{RISETaxonomyPanel}
\multicolumn{2}{c}{\rule{0pt}{13pt}\raisebox{1.5pt}[0pt][0pt]{\textbf{Causal Reasoning~(200, 8)}}\rule[-4pt]{0pt}{0pt}} \\
\addlinespace[3pt]
\multirow[c]{3}{=}{\centering\arraybackslash Common-Sense Knowledge~(80, 1)} & Physical State Transition~(31) \\
 & Structural Damage~(27) \\
 & Flexible Deformation~(22, 1) \\
\cmidrule(lr){1-2}
\multirow[c]{4}{=}{\centering\arraybackslash Context Knowledge~(40, 2)} & Magnitude Change~(10) \\
 & Reading Offset~(10, 2) \\
 & Order Relation~(10) \\
 & Feature Mapping~(10) \\
\cmidrule(lr){1-2}
\multirow[c]{7}{=}{\centering\arraybackslash Disciplinary Knowledge~(80, 5)} & Physics~(25, 1) \\
 & Chemistry~(23, 4) \\
 & Biology~(12) \\
 & Astronomy~(3) \\
 & Medicine~(4) \\
 & Music Theory~(3) \\
 & Computer Science~(10) \\
\bottomrule[0.75pt]
\end{tabular}
\end{minipage}
\hfill
\begin{minipage}[t]{.485\textwidth}
\vspace{0pt}
\begin{tabular}{
>{\centering\arraybackslash}m{\dimexpr .325\linewidth-2\tabcolsep\relax}
>{\centering\arraybackslash}m{\dimexpr .675\linewidth-2\tabcolsep\relax}
}
\toprule[0.75pt]
\textbf{Subcategory} & \textbf{Task Type} \\
\midrule[0.4pt]
\rowcolor{RISETaxonomyPanel}
\multicolumn{2}{c}{\rule{0pt}{13pt}\raisebox{1.5pt}[0pt][0pt]{\textbf{Spatial Reasoning~(200, 18)}}\rule[-4pt]{0pt}{0pt}} \\
\addlinespace[3pt]
\multirow[c]{9}{=}{\centering\arraybackslash Relation~(106, 6)} & Adjacency~(11) \\
 & Direction~(11) \\
 & Orientation~(11) \\
 & Projection~(12) \\
 & Distance~(13) \\
 & Size~(15) \\
 & Configuration~(18) \\
 & Viewing Scale~(9) \\
 & Localization~(6, 6) \\
\cmidrule(lr){1-2}
\multirow[c]{5}{=}{\centering\arraybackslash Structure~(94, 12)} & Part-Whole Perception~(24) \\
 & Geometric Regularities Reasoning~(15) \\
 & Viewpoint Generation~(26) \\
 & Environmental Structure Reasoning~(19, 2) \\
 & Egocentric Scene Reconstruction~(10, 10) \\
\addlinespace[7pt]
\rowcolor{RISETaxonomyPanel}
\multicolumn{2}{c}{\rule{0pt}{13pt}\raisebox{1.5pt}[0pt][0pt]{\textbf{Logical Reasoning~(200, 20)}}\rule[-4pt]{0pt}{0pt}} \\
\addlinespace[3pt]
\multirow[c]{4}{=}{\centering\arraybackslash Explicit Rule Deduction~(123)} & Path Searching~(24) \\
 & Mathematical Derivation~(27) \\
 & Game Deduction~(37) \\
 & Puzzle Solving~(35) \\
\cmidrule(lr){1-2}
\multirow[c]{3}{=}{\centering\arraybackslash Implicit Pattern Induction~(77, 20)} & Pattern Prediction~(39) \\
 & Visual Oddity Detection~(18) \\
 & Rule Transfer~(20, 20) \\
\addlinespace[7pt]
\rowcolor{RISETaxonomyPanel}
\multicolumn{2}{c}{\rule{0pt}{13pt}\raisebox{1.5pt}[0pt][0pt]{\textbf{Counterfactual Reasoning~(100)}}\rule[-4pt]{0pt}{0pt}} \\
\addlinespace[3pt]
\multirow[c]{8}{=}{\centering\arraybackslash Counterfactual Imagination~(70)} & Physical Counterfactual~(23) \\
 & Chemical Counterfactual~(9) \\
 & Biological Counterfactual~(8) \\
 & Astronomy Counterfactual~(7) \\
 & Geography Counterfactual~(8) \\
 & History Counterfactual~(7) \\
 & Music Theory Counterfactual~(4) \\
 & Computer Science Counterfactual~(4) \\
\cmidrule(lr){1-2}
\multirow[c]{3}{=}{\centering\arraybackslash Anomaly Correction and Generation~(30)} & Anatomical Errors~(10) \\
 & Structural Anomalies~(11) \\
 & Relational Anomalies~(9) \\
\bottomrule[0.75pt]
\end{tabular}
\end{minipage}
\par\vspace{2pt}
\begin{tabular}{>{\centering\arraybackslash}m{\dimexpr\textwidth-2\tabcolsep\relax}}
\toprule[0.75pt]
\rowcolor{RISETaxonomyPanel}
\rule{0pt}{13pt}\raisebox{1.5pt}[0pt][0pt]{\textbf{Hybrid Reasoning~(100)}}\rule[-4pt]{0pt}{0pt} \\
\bottomrule[0.75pt]
\end{tabular}
\par\vspace{4pt}
\parbox{\textwidth}{\scriptsize\raggedright
Numbers in parentheses denote the total number of cases, followed by the number of multi-image cases. For example, Temporal State Transition (15, 4) contains 15 cases, including 4 multi-image cases and 11 single-image cases; Life Progression (24) contains 24 single-image cases and no multi-image cases.}
\endgroup
\end{table*}

\begin{table*}[tp]
\caption{Detail performance on KRIS-Bench.}
\label{tab:kris_bench_results}
\centering
\begingroup
\fontsize{7.0}{9.0}\selectfont
\setlength{\tabcolsep}{1.5pt}
\renewcommand{\arraystretch}{1.05}
\setlength{\aboverulesep}{1pt}
\setlength{\belowrulesep}{1pt}
\colorlet{RISEKRISGroup}{black!6}

\resizebox{\textwidth}{!}{%
\begin{tabular}{@{}l*{11}{c}@{}}
\toprule[0.8pt]
\multirow{2}{*}[-1.3pt]{\textbf{Model}} &
\multicolumn{4}{c}{\textbf{Factual Knowledge}} &
\multicolumn{3}{c}{\textbf{Conceptual Knowledge}} &
\multicolumn{3}{c}{\textbf{Procedural Knowledge}} &
\multirow{2}{*}[-1.3pt]{\shortstack{\textbf{Overall}\\\textbf{Score}}} \\
\cmidrule(lr){2-5}\cmidrule(lr){6-8}\cmidrule(lr){9-11}
 & \textbf{Attribute} & \textbf{Spatial} & \textbf{Temporal} & \textbf{Average} &
\textbf{Social Sci.} & \textbf{Natural Sci.} & \textbf{Average} &
\textbf{Logical} & \textbf{Instruction} & \textbf{Average} & \\
\midrule[0.5pt]
\rowcolor{RISEKRISGroup}
\multicolumn{12}{c}{\rule{0pt}{9pt}\raisebox{0.7pt}[0pt][0pt]{\textbf{Closed-source Models}}} \\
\midrule[0.4pt]
GPT-Image-2.5 Sunburst~\cite{openai2025gptimage} & \textbf{95.76} & \textbf{91.92} & \textbf{93.92} & \textbf{94.60} & \textbf{90.93} & \textbf{91.34} & \textbf{91.24} & \textbf{89.09} & \underline{97.17} & \textbf{93.14} & \textbf{92.89} \\
GPT-Image-2.5 Flare~\cite{openai2025gptimage} & \underline{94.95} & \underline{89.58} & \underline{93.81} & \underline{93.56} & \underline{89.90} & \underline{90.06} & \underline{90.02} & \underline{86.87} & \textbf{98.61} & \underline{92.76} & \underline{91.93} \\
Nano Banana Pro~\cite{google2026nanobanana} & 85.61 & 87.17 & 88.06 & 86.36 & 85.25 & 83.64 & 84.03 & 81.42 & 93.22 & 86.48 & 85.31 \\
Nano Banana~\cite{google2026nanobanana} & 78.97 & 80.17 & 86.49 & 80.48 & 83.70 & 76.34 & 78.11 & 63.96 & 90.44 & 75.31 & 78.17 \\
Step Image Edit 2~\cite{liu2025step1x} & 81.49 & 78.25 & \textemdash & 67.34 & 74.04 & 69.28 & 70.44 & 68.55 & 45.45 & 56.99 & 66.16 \\
\midrule[0.4pt]
\rowcolor{RISEKRISGroup}
\multicolumn{12}{c}{\rule{0pt}{9pt}\raisebox{0.7pt}[0pt][0pt]{\textbf{Open-source Models}}} \\
\midrule[0.4pt]
Qwen-Image-Edit-2511~\cite{wu2025qwen} & \underline{74.85} & \textbf{84.50} & \textbf{68.13} & \textbf{75.89} & 66.10 & 59.54 & 61.15 & \underline{58.42} & \textbf{79.61} & \textbf{67.55} & \textbf{67.14} \\
Qwen-Image-Edit-2509~\cite{wu2025qwen} & 72.44 & 79.10 & 55.07 & \underline{71.05} & \underline{66.14} & 57.75 & 59.78 & 49.83 & \underline{77.61} & \underline{61.94} & \underline{63.71} \\
JoyAI-Image-Edit~\cite{song2026joyai} & \textbf{78.09} & \underline{80.33} & \textemdash & 65.72 & \textbf{66.56} & \textbf{64.60} & \textbf{65.09} & \textbf{68.02} & 46.33 & 57.17 & 63.44 \\
BAGEL (w/ CoT)~\cite{deng2025bagel} & 67.42 & 68.33 & 58.67 & 66.18 & 63.55 & \underline{61.40} & \underline{61.92} & 48.12 & 50.22 & 49.02 & 60.18 \\
BAGEL~\cite{deng2025bagel} & 64.27 & 62.42 & 42.45 & 60.26 & 55.40 & 56.01 & 55.86 & 52.54 & 50.56 & 51.69 & 56.21 \\
UniWorld-V1~\cite{lin2025uniworld} & 58.17 & 54.50 & \underline{63.00} & 58.15 & 47.50 & 43.94 & 44.80 & 42.00 & 53.83 & 47.92 & 50.27 \\
OmniGen2~\cite{wu2025omnigen2} & 59.92 & 52.25 & 54.75 & 57.36 & 47.56 & 43.12 & 44.20 & 32.50 & 63.08 & 47.79 & 49.71 \\
FLUX.1 Kontext Dev~\cite{labs2025flux1kontextflowmatching} & 64.83 & 60.92 & \textemdash & 53.28 & 48.94 & 50.81 & 50.36 & 46.06 & 39.00 & 42.53 & 49.54 \\
Emu2-Gen~\cite{sun2023emu2chat} & 51.50 & 48.83 & 22.17 & 45.40 & 34.69 & 38.44 & 37.54 & 24.81 & 45.00 & 34.91 & 39.70 \\
OmniGen~\cite{xiao2024omnigen} & 37.92 & 28.25 & 21.83 & 33.11 & 30.63 & 27.19 & 28.02 & 11.94 & 35.83 & 23.89 & 28.85 \\
\midrule[0.4pt]
\rowcolor{RISEKRISGroup}
\multicolumn{12}{c}{\rule{0pt}{9pt}\raisebox{0.7pt}[0pt][0pt]{\textbf{Agentic Methods}}} \\
\midrule[0.4pt]
\textbf{RISE-Agent (Ours)} & \textbf{89.85} & \textbf{88.25} & \textbf{87.73} & \textbf{89.14} & \textbf{88.70} & \textbf{87.47} & \textbf{87.77} & \textbf{77.33} & \textbf{97.06} & \textbf{87.19} & \textbf{88.12} \\
DDA-Thinker-32B (Qwen-Image-Edit-2511)~\cite{yang2026dda} & \underline{82.16} & \underline{84.58} & \underline{75.00} & \underline{81.52} & \underline{84.71} & \underline{77.77} & \underline{79.44} & \underline{74.45} & 84.67 & \underline{78.88} & \underline{79.94} \\
DDA-Thinker-8B (Qwen-Image-Edit-2511)~\cite{yang2026dda} & 79.14 & 79.33 & 74.43 & 78.41 & 79.86 & 76.93 & 77.64 & 68.97 & 80.11 & 73.83 & 76.99 \\
DDA-Thinker-32B (Qwen-Image-Edit-2509)~\cite{yang2026dda} & 72.79 & 77.92 & 73.09 & 73.98 & 76.90 & 70.60 & 72.13 & 55.43 & \underline{89.39} & 70.07 & 72.21 \\
EditThinker~\cite{li2025editthinker} & 78.48 & 73.83 & \textemdash & 77.24 & 76.20 & 70.69 & 72.02 & 65.23 & 66.89 & 65.94 & 71.91 \\
ThinkRL-Edit~\cite{li2026thinkrl} & 81.02 & 81.45 & \textemdash & 81.13 & 75.67 & 71.25 & 72.31 & 49.07 & 79.71 & 57.44 & 71.65 \\
DDA-Thinker-8B (Qwen-Image-Edit-2509)~\cite{yang2026dda} & 71.55 & 77.00 & 72.52 & 72.92 & 76.55 & 68.15 & 70.18 & 54.47 & 86.56 & 68.25 & 70.55 \\
\bottomrule[0.8pt]
\end{tabular}%
}

\par\vspace{4pt}
\parbox{\textwidth}{\raggedright\fontsize{7.0}{9.0}\selectfont
Methods are evaluated across three knowledge domains: factual, conceptual, and procedural, further subdivided into seven dimensions. The overall score is computed as the average across these dimensions. \textbf{Bold} and \underline{underlined} values mark the best and second-best results within each model group, respectively.}
\endgroup
\end{table*}

\begin{table*}[tp]
    \centering
    \caption{Sample and human score distributions across reasoning categories for human alignment evaluation.}
    \label{tab:human_scores_by_reasoning_category}

    \renewcommand{\arraystretch}{1.15}
    \setlength{\tabcolsep}{3pt}

    \resizebox{\textwidth}{!}{%
    \begin{tabular}{c|ccc|ccc|ccc|ccc|ccc|ccc}
        \toprule
        \textbf{Human}
        & \multicolumn{3}{c|}{\shortstack{\textbf{Logical}\\\textbf{Reasoning}}}
        & \multicolumn{3}{c|}{\shortstack{\textbf{Temporal}\\\textbf{Reasoning}}}
        & \multicolumn{3}{c|}{\shortstack{\textbf{Causal}\\\textbf{Reasoning}}}
        & \multicolumn{3}{c|}{\shortstack{\textbf{Spatial}\\\textbf{Reasoning}}}
        & \multicolumn{3}{c|}{\shortstack{\textbf{Counterfactual}\\\textbf{Reasoning}}}
        & \multicolumn{3}{c}{\shortstack{\textbf{Hybrid}\\\textbf{Reasoning}}} \\
        \textbf{Score}
        & \textbf{Reas.} & \textbf{Cons.} & \textbf{Plau.}
        & \textbf{Reas.} & \textbf{Cons.} & \textbf{Plau.}
        & \textbf{Reas.} & \textbf{Cons.} & \textbf{Plau.}
        & \textbf{Reas.} & \textbf{Cons.} & \textbf{Plau.}
        & \textbf{Reas.} & \textbf{Cons.} & \textbf{Plau.}
        & \textbf{Reas.} & \textbf{Cons.} & \textbf{Plau.} \\
        \midrule

        1
        & 75.0\% & 20.0\% & 0.0\%
        & 27.5\% & 0.0\%  & 0.0\%
        & 25.0\% & 5.0\%  & 2.5\%
        & 50.0\% & 8.3\%  & 0.0\%
        & 45.8\% & 0.0\%  & 0.0\%
        & 50.0\% & 40.0\% & 25.0\% \\

        2
        & 0.0\%  & 0.0\% & 0.0\%
        & 0.0\%  & 2.5\% & 0.0\%
        & 2.5\%  & 0.0\% & 0.0\%
        & 5.6\%  & 8.3\% & 0.0\%
        & 0.0\%  & 4.2\% & 0.0\%
        & 15.0\% & 0.0\% & 0.0\% \\

        3
        & 0.0\%  & 5.0\%  & 0.0\%
        & 15.0\% & 20.0\% & 0.0\%
        & 7.5\%  & 12.5\% & 5.0\%
        & 8.3\%  & 30.6\% & 0.0\%
        & 4.2\%  & 12.5\% & 8.3\%
        & 15.0\% & 10.0\% & 0.0\% \\

        4
        & 5.0\%  & 2.5\%  & 0.0\%
        & 5.0\%  & 17.5\% & 2.5\%
        & 0.0\%  & 7.5\%  & 7.5\%
        & 0.0\%  & 2.8\%  & 5.6\%
        & 4.2\%  & 12.5\% & 0.0\%
        & 20.0\% & 15.0\% & 5.0\% \\

        5
        & 20.0\% & 72.5\% & 100.0\%
        & 52.5\% & 60.0\% & 97.5\%
        & 65.0\% & 75.0\% & 85.0\%
        & 36.1\% & 50.0\% & 94.4\%
        & 45.8\% & 70.8\% & 91.7\%
        & 0.0\%  & 35.0\% & 70.0\% \\
        \midrule

        \textbf{Overall Count}
        & \multicolumn{3}{c|}{40}
        & \multicolumn{3}{c|}{40}
        & \multicolumn{3}{c|}{40}
        & \multicolumn{3}{c|}{36}
        & \multicolumn{3}{c|}{24}
        & \multicolumn{3}{c}{20} \\
        \bottomrule
    \end{tabular}%
    }
\end{table*}

\begin{table*}[!tp]
\caption{Detailed evaluation dimension scores on RISEBench++ (EN).}
\label{tab:dimension_metric_scores_en}
\centering
\begingroup
\fontsize{6.5}{8.0}\selectfont
\setlength{\tabcolsep}{0.8pt}
\renewcommand{\arraystretch}{1.06}
\colorlet{RISEScoreGroup}{black!6}
\def\RISEScoreYes{\textcolor{green!50!black}{\ding{51}}}
\def\RISEScoreNo{\textcolor{red!65!black}{\ding{55}}}
\begin{tabular}{
>{\raggedright\arraybackslash}m{\dimexpr .232\textwidth-2\tabcolsep\relax}
>{\centering\arraybackslash}m{\dimexpr .042\textwidth-2\tabcolsep\relax}
*{6}{*{3}{>{\centering\arraybackslash}m{\dimexpr (.726\textwidth-12pt)/21-2\tabcolsep\relax}}!{\hspace{2pt}}}
*{3}{>{\centering\arraybackslash}m{\dimexpr (.726\textwidth-12pt)/21-2\tabcolsep\relax}}
}
\toprule[0.8pt]
\multirow{2}{=}[-1.3pt]{\raggedright\textbf{Model}} & \multirow{2}{=}[-1.3pt]{\centering\shortstack{\textbf{Multi}\\\textbf{img}}} & \multicolumn{3}{c}{\textbf{Temporal}} & \multicolumn{3}{c}{\textbf{Causal}} & \multicolumn{3}{c}{\textbf{Spatial}} & \multicolumn{3}{c}{\textbf{Logical}} & \multicolumn{3}{c}{\textbf{Counterfactual}} & \multicolumn{3}{c}{\textbf{Hybrid}} & \multicolumn{3}{c}{\textbf{Overall}} \\
\cmidrule(lr){3-5}\cmidrule(lr){6-8}\cmidrule(lr){9-11}\cmidrule(lr){12-14}\cmidrule(lr){15-17}\cmidrule(lr){18-20}\cmidrule(lr){21-23}
 & & \textbf{IR} & \textbf{AC} & \textbf{VP} & \textbf{IR} & \textbf{AC} & \textbf{VP} & \textbf{IR} & \textbf{AC} & \textbf{VP} & \textbf{IR} & \textbf{AC} & \textbf{VP} & \textbf{IR} & \textbf{AC} & \textbf{VP} & \textbf{IR} & \textbf{AC} & \textbf{VP} & \textbf{IR} & \textbf{AC} & \textbf{VP} \\
\midrule[0.5pt]
\rowcolor{RISEScoreGroup}
\multicolumn{23}{c}{\rule{0pt}{9pt}\raisebox{1.5pt}[0pt][0pt]{\textbf{Closed-source Models}}} \\
\midrule[0.4pt]
GPT-Image-2.5 Sunburst~\cite{openai2025gptimage} & \RISEScoreYes & 81.0 & \underline{98.0} & 97.0 & 83.1 & \underline{96.3} & \underline{97.6} & \textbf{81.3} & \textbf{96.6} & \underline{99.0} & 27.0 & \textbf{95.0} & \textemdash & 78.3 & \textbf{99.5} & \underline{99.5} & 58.0 & \textbf{91.7} & 97.4 & 68.1 & \textbf{96.3} & \underline{98.0} \\
Nano Banana Pro~\cite{google2026nanobanana} & \RISEScoreYes & 80.2 & 92.6 & \underline{98.4} & \textbf{89.0} & 94.2 & \textbf{97.8} & 74.9 & 88.1 & 98.6 & \textbf{38.0} & 79.0 & \textemdash & 74.8 & 88.5 & 90.8 & \underline{59.3} & 80.8 & 93.5 & \textbf{69.8} & 87.7 & 96.8 \\
Nano Banana 2~\cite{google2026nanobanana} & \RISEScoreYes & \textbf{84.5} & 95.9 & 97.1 & \underline{86.6} & 92.4 & 97.4 & 75.5 & 87.4 & 97.8 & 29.0 & 87.0 & \textemdash & 78.5 & 97.3 & 98.0 & 55.3 & 88.6 & 94.5 & \underline{68.5} & 91.1 & 97.2 \\
GPT-Image-2~\cite{openai2025gptimage} & \RISEScoreYes & 76.6 & \textbf{98.9} & 97.5 & 80.5 & \textbf{96.8} & 95.6 & \underline{76.3} & \underline{94.5} & 98.1 & 16.0 & 92.0 & \textemdash & 66.5 & \underline{98.8} & \RISEFitScore{\textbf{100.0}} & 49.3 & 88.9 & \underline{98.1} & 61.5 & \underline{95.2} & 97.6 \\
GPT-Image-2.5 Flare~\cite{openai2025gptimage} & \RISEScoreYes & 75.0 & 97.9 & 97.5 & 81.0 & 93.9 & 97.5 & 74.9 & 92.6 & \textbf{99.4} & 11.0 & 88.0 & \textemdash & 67.5 & 98.3 & 99.0 & 46.3 & \underline{89.4} & \textbf{98.7} & 59.8 & 93.2 & \textbf{98.3} \\
Luma Uni-1~\cite{luma2026uni1} & \RISEScoreYes & 81.0 & 95.6 & 95.5 & 79.4 & 92.9 & 92.5 & 74.4 & 88.4 & 95.3 & 18.0 & 89.5 & \textemdash & 69.0 & 93.0 & 97.3 & 47.8 & 85.1 & 93.8 & 62.2 & 91.0 & 94.7 \\
Qwen-Image-3.0~\cite{zhao2026qwen} & \RISEScoreYes & \underline{81.3} & 89.9 & 97.4 & 83.3 & 85.4 & 96.1 & 70.0 & 83.0 & 93.5 & \underline{37.0} & \underline{92.5} & \textemdash & \underline{79.0} & 85.5 & \underline{99.5} & \textbf{62.3} & 86.6 & 92.9 & 68.4 & 87.3 & 95.9 \\
Nano Banana 2 Lite~\cite{google2026nanobanana} & \RISEScoreYes & 80.3 & 95.2 & \textbf{98.4} & 82.6 & 95.3 & 95.9 & 69.0 & 88.9 & 97.0 & 15.0 & 80.5 & \textemdash & 73.3 & 95.0 & 98.0 & 41.0 & 82.6 & 88.0 & 60.8 & 89.7 & 96.3 \\
FLUX.2 Max~\cite{bfl2025flux2} & \RISEScoreYes & 80.5 & 93.0 & 94.1 & 78.1 & 91.5 & 96.0 & 72.3 & 82.9 & 96.0 & 15.0 & 52.0 & \textemdash & 78.5 & 91.5 & 98.5 & 44.3 & 63.1 & 92.9 & 61.5 & 79.2 & 95.5 \\
GPT-Image-1.5~\cite{openai2025gptimage} & \RISEScoreYes & 79.8 & 93.1 & 96.1 & 80.6 & 91.9 & 95.3 & 70.8 & 80.0 & 97.8 & 9.5 & 67.0 & \textemdash & 71.5 & 93.0 & 99.3 & 45.8 & 54.8 & 93.8 & 59.9 & 81.0 & 96.5 \\
FLUX.2 Pro~\cite{bfl2025flux2} & \RISEScoreYes & 78.8 & 90.2 & 94.5 & 73.1 & 88.9 & 95.1 & 64.1 & 76.3 & 94.1 & 12.0 & 42.5 & \textemdash & \textbf{80.0} & 93.0 & 98.0 & 39.0 & 60.1 & 93.2 & 57.5 & 74.7 & 94.9 \\
Nano Banana~\cite{google2026nanobanana} & \RISEScoreYes & 72.6 & 93.4 & 95.1 & 71.1 & 92.8 & 95.4 & 62.3 & 85.5 & 95.6 & 11.0 & 64.5 & \textemdash & 61.0 & 94.5 & 98.8 & 33.8 & 75.5 & 94.8 & 52.9 & 84.1 & 95.8 \\
HunyuanImage-3.5~\cite{cao2025hunyuanimage} & \RISEScoreYes & 65.1 & 96.5 & 96.0 & 67.5 & \underline{96.3} & 96.1 & 64.8 & 92.2 & 96.9 & 14.5 & 90.0 & \textemdash & 51.3 & 97.5 & 95.8 & 38.5 & 87.1 & 85.4 & 51.3 & 93.4 & 95.2 \\
Wan2.7-Image-Pro~\cite{mao2026wan} & \RISEScoreYes & 56.7 & 79.5 & 93.8 & 71.8 & 87.2 & 92.9 & 56.1 & 76.5 & 93.9 & 11.0 & 72.0 & \textemdash & 55.8 & 91.0 & 91.5 & 41.5 & 79.8 & 93.8 & 48.8 & 80.1 & 93.3 \\
GPT-Image-1~\cite{openai2025gptimage} & \RISEScoreYes & 75.0 & 87.6 & 96.1 & 72.5 & 84.2 & 94.9 & 66.8 & 74.0 & 96.0 & 5.5 & 57.0 & \textemdash & 59.5 & 88.0 & \underline{99.5} & 34.5 & 43.4 & 89.6 & 53.4 & 73.6 & 95.6 \\
Qwen-Image-2.0~\cite{zhao2026qwen} & \RISEScoreYes & 70.3 & 92.8 & 95.8 & 70.1 & 90.0 & 96.4 & 59.1 & 78.9 & 94.9 & 7.5 & 71.5 & \textemdash & 52.5 & 86.5 & \RISEFitScore{\textbf{100.0}} & 29.3 & 46.0 & 48.1 & 49.6 & 79.7 & 91.5 \\
Seedream 4.5~\cite{seedream2025seedream} & \RISEScoreYes & 67.8 & 92.7 & 96.0 & 68.4 & 85.2 & 94.1 & 56.6 & 75.3 & 95.5 & 10.0 & 51.5 & \textemdash & 58.3 & 86.8 & 98.5 & 34.5 & 66.2 & 84.1 & 49.8 & 76.0 & 94.5 \\
Seedream 5.0~\cite{seedream2025seedream} & \RISEScoreYes & 67.8 & 91.1 & 93.5 & 66.1 & 83.0 & 93.9 & 56.3 & 72.3 & 94.6 & 7.0 & 44.0 & \textemdash & 50.8 & 84.8 & 98.8 & 32.5 & 54.0 & 90.6 & 47.8 & 71.7 & 94.3 \\
Step Image Edit 2~\cite{liu2025step1x} & \RISEScoreNo & 33.9 & 84.3 & 94.8 & 46.5 & 83.0 & 91.1 & 53.7 & 73.1 & 94.6 & 2.2 & 56.1 & \textemdash & 22.5 & 79.0 & 98.0 & 15.0 & 17.4 & 20.5 & 31.0 & 68.3 & 86.3 \\
\midrule[0.4pt]
\rowcolor{RISEScoreGroup}
\multicolumn{23}{c}{\rule{0pt}{9pt}\raisebox{1.5pt}[0pt][0pt]{\textbf{Open-source Models}}} \\
\midrule[0.4pt]
HunyuanImage-3.0-Instruct~\cite{cao2025hunyuanimage} & \RISEScoreYes & 51.3 & 92.4 & \textbf{96.5} & \underline{56.8} & 87.9 & \textbf{98.1} & \textbf{54.5} & \textbf{78.4} & 95.8 & \textbf{7.0} & \underline{66.0} & \textemdash & 31.3 & \textbf{93.8} & 98.0 & 11.3 & 16.4 & 12.0 & \textbf{38.2} & 75.8 & 88.5 \\
Qwen-Image-Edit-2511~\cite{wu2025qwen} & \RISEScoreYes & 45.4 & 73.1 & 87.6 & 48.9 & 78.5 & \underline{93.4} & \underline{52.8} & 68.3 & 92.9 & 6.0 & 42.5 & \textemdash & 31.3 & 75.0 & 94.8 & \textbf{22.5} & 41.4 & 80.5 & 36.0 & 64.0 & 90.7 \\
Qwen-Image-2.1~\cite{zhao2026qwen} & \RISEScoreYes & 38.9 & 92.0 & 92.5 & 48.8 & \underline{89.5} & 92.8 & 43.5 & \underline{77.5} & 92.5 & 5.0 & \textbf{73.0} & \textemdash & 30.3 & 84.5 & 96.5 & 13.4 & 28.8 & 22.4 & 31.6 & \underline{77.6} & 86.2 \\
FireRed-Image-Edit-1.0~\cite{team2026firered} & \RISEScoreYes & 38.4 & 83.8 & \underline{94.6} & 43.1 & 80.8 & 90.0 & 52.3 & 64.5 & \underline{96.6} & 3.0 & 35.5 & \textemdash & 31.8 & 77.0 & \underline{99.0} & 16.8 & 35.9 & 86.7 & 32.2 & 64.0 & \textbf{93.7} \\
FireRed-Image-Edit-1.1~\cite{team2026firered} & \RISEScoreYes & 39.0 & 87.6 & 91.6 & 47.8 & 80.6 & 92.9 & 50.9 & 72.9 & 93.7 & 3.0 & 30.0 & \textemdash & 28.5 & 80.8 & 97.0 & 18.5 & 42.2 & \textbf{89.6} & 32.8 & 66.2 & \underline{93.0} \\
Qwen-Image-Edit-2509~\cite{wu2025qwen} & \RISEScoreYes & 36.3 & 83.6 & 85.4 & 44.0 & 79.7 & 88.9 & 46.8 & 63.9 & 93.8 & 1.0 & 27.5 & \textemdash & 28.0 & 72.3 & 96.5 & 13.5 & 34.6 & 43.8 & 29.8 & 61.3 & 85.7 \\
HiDream-O1-Image~\cite{cai2026hidream} & \RISEScoreYes & 49.1 & 78.7 & 92.4 & 55.3 & 65.5 & 89.8 & 48.6 & 62.5 & 93.4 & 5.0 & 35.0 & \textemdash & 20.3 & 68.5 & 93.5 & 15.5 & 23.2 & 43.8 & 35.2 & 57.3 & 87.3 \\
FLUX.2 Dev~\cite{bfl2025flux2} & \RISEScoreYes & 40.8 & 73.7 & 87.9 & 47.1 & 73.9 & \underline{93.4} & 47.0 & 58.4 & 94.1 & 1.5 & 23.5 & \textemdash & 24.5 & 75.0 & 97.8 & 14.0 & 45.2 & 85.1 & 31.1 & 57.7 & 91.9 \\
FLUX.2 Klein 9B~\cite{bfl2025flux2} & \RISEScoreYes & 37.0 & 78.3 & 93.0 & 37.4 & 82.1 & 89.8 & 44.0 & 69.5 & 94.1 & 2.0 & 39.5 & \textemdash & 27.5 & 85.8 & 97.3 & 20.3 & 48.5 & 85.7 & 28.9 & 67.2 & 92.3 \\
JoyAI-Image-Edit-Plus~\cite{song2026joyai} & \RISEScoreYes & 32.1 & \underline{92.6} & 92.4 & 36.1 & 88.8 & 91.1 & 40.0 & 73.0 & 90.9 & 5.0 & 53.5 & \textemdash & 19.3 & \underline{92.5} & \textbf{99.3} & \underline{21.0} & \textbf{63.1} & \underline{87.3} & 26.7 & 76.9 & 92.1 \\
SenseNova-U1-A3B-MoT-SFT~\cite{diao2026sensenova} & \RISEScoreYes & \underline{53.1} & 68.8 & 60.8 & \textbf{62.3} & 71.5 & 71.0 & 43.8 & 40.3 & 49.6 & 3.5 & 49.0 & \textemdash & \textbf{42.3} & 69.3 & 73.3 & 12.5 & 3.8 & 25.6 & \underline{38.0} & 53.0 & 58.7 \\
UniPic 3.0~\cite{wei2026skywork} & \RISEScoreYes & 38.9 & 82.2 & 89.3 & 41.6 & 80.1 & 87.0 & 42.8 & 67.9 & 90.1 & 1.0 & 29.5 & \textemdash & 20.0 & 77.8 & 97.3 & 13.5 & 12.4 & 23.4 & 28.2 & 60.7 & 83.4 \\
BAGEL (w/ CoT)~\cite{deng2025bagel} & \RISEScoreYes & \textbf{56.2} & 75.5 & 76.3 & 47.1 & 76.5 & 80.5 & 39.9 & 65.4 & 84.1 & 2.0 & 38.0 & \textemdash & 28.3 & 72.0 & 80.3 & 15.0 & 30.6 & 26.6 & 33.3 & 61.2 & 75.0 \\
SenseNova-U1-8B-MoT-SFT~\cite{diao2026sensenova} & \RISEScoreYes & 46.3 & 70.4 & 65.8 & 56.4 & 69.9 & 70.1 & 45.0 & 43.5 & 53.0 & 3.5 & 60.5 & \textemdash & \underline{41.8} & 73.5 & 67.3 & 16.5 & 20.5 & 27.9 & 36.0 & 58.1 & 60.0 \\
FLUX.2 Klein 4B~\cite{bfl2025flux2} & \RISEScoreYes & 32.6 & 86.8 & 92.6 & 32.5 & 85.2 & 86.5 & 42.4 & 65.1 & 92.9 & 4.0 & 44.5 & \textemdash & 17.5 & 90.5 & 97.5 & 15.3 & 58.1 & 78.2 & 25.6 & 71.0 & 90.3 \\
UniWorld-V2~\cite{li2025uniworld} & \RISEScoreYes & 33.4 & 66.9 & 83.6 & 43.9 & 70.3 & 86.4 & 51.4 & 59.6 & 94.9 & 0.5 & 19.0 & \textemdash & 31.5 & 65.3 & 96.0 & 17.3 & 20.7 & 41.9 & 30.7 & 51.6 & 84.7 \\
BAGEL~\cite{deng2025bagel} & \RISEScoreYes & 30.6 & 54.4 & 75.0 & 33.0 & 51.1 & 80.6 & 35.1 & 56.5 & 82.4 & 1.5 & 38.0 & \textemdash & 17.8 & 50.5 & 84.3 & 12.3 & 27.3 & 40.3 & 23.1 & 47.7 & 76.1 \\
OmniGen2~\cite{wu2025omnigen2} & \RISEScoreYes & 22.8 & 80.7 & 82.3 & 18.3 & 72.2 & 83.8 & 22.3 & 60.6 & 85.1 & 1.5 & 21.5 & \textemdash & 14.4 & 79.3 & 90.9 & 9.8 & 37.6 & 44.5 & 15.4 & 58.4 & 80.7 \\
OmniGen~\cite{xiao2024omnigen} & \RISEScoreYes & 16.8 & 63.0 & 62.3 & 19.6 & 58.1 & 61.8 & 18.4 & 41.3 & 63.8 & 2.5 & 24.0 & \textemdash & 13.5 & 59.8 & 75.0 & 8.0 & 15.2 & 32.1 & 13.6 & 44.5 & 61.2 \\
Emu2-Gen~\cite{sun2023emu2chat} & \RISEScoreYes & 16.4 & 49.1 & 64.4 & 14.0 & 37.9 & 64.9 & 13.2 & 13.7 & 58.9 & 0.5 & 5.5 & \textemdash & 12.8 & 39.5 & 79.3 & 5.8 & 5.8 & 39.0 & 10.7 & 25.5 & 62.5 \\
JoyAI-Image-Edit~\cite{song2026joyai} & \RISEScoreNo & 35.9 & 91.0 & 93.5 & 47.1 & 83.9 & 92.8 & 47.7 & 74.9 & \textbf{96.7} & 3.3 & 57.2 & \textemdash & 27.5 & 82.0 & 97.3 & 20.8 & 29.8 & 54.9 & 31.6 & 71.9 & 90.5 \\
Step1X-Edit (thinking)~\cite{liu2025step1x} & \RISEScoreNo & 38.8 & 92.0 & 94.2 & 43.1 & 88.9 & 93.2 & 34.5 & 64.1 & 89.3 & 1.7 & 41.7 & \textemdash & 32.0 & 85.0 & 98.0 & 17.3 & 48.5 & 63.0 & 28.5 & 70.3 & 89.9 \\
Boogu-Image-0.1-Edit~\cite{chen2026boogu} & \RISEScoreNo & 31.5 & \textbf{96.6} & 91.1 & 42.6 & \textbf{89.7} & 90.9 & 38.6 & 75.3 & 94.0 & \underline{6.1} & 65.0 & \textemdash & 20.0 & 91.3 & 96.3 & 17.5 & 57.6 & 77.3 & 27.5 & \textbf{79.8} & 91.0 \\
\RISEFitModel{Step1X-Edit (thinking + reflection)~\cite{liu2025step1x}} & \RISEScoreNo & 40.2 & 92.5 & 91.7 & 44.8 & 86.6 & 93.4 & 39.3 & 59.9 & 87.6 & 1.7 & 38.3 & \textemdash & 33.0 & 87.5 & 97.5 & 17.3 & 43.2 & 68.2 & 30.2 & 68.0 & 89.4 \\
LongCat-Image-Edit~\cite{team2025longcat} & \RISEScoreNo & 24.7 & 87.5 & 86.9 & 38.3 & 82.6 & 88.7 & 44.9 & 66.3 & 89.8 & 1.7 & 37.8 & \textemdash & 23.0 & 59.5 & 93.3 & 14.3 & 20.7 & 39.0 & 25.7 & 62.0 & 83.9 \\
Step1X-Edit (base)~\cite{liu2025step1x} & \RISEScoreNo & 29.2 & 88.5 & 92.4 & 39.5 & 85.9 & 89.7 & 37.5 & 60.6 & 89.1 & 2.2 & 32.2 & \textemdash & 29.0 & 87.5 & 94.0 & 13.8 & 42.9 & 55.5 & 26.0 & 66.1 & 87.1 \\
InternVL-U (w/ CoT)~\cite{tian2026internvl} & \RISEScoreNo & 40.5 & 81.1 & 75.3 & 38.9 & 72.2 & 81.6 & 32.8 & 44.6 & 66.1 & 1.1 & 20.0 & \textemdash & 32.8 & 71.3 & 81.5 & 19.5 & 21.0 & 34.1 & 27.8 & 52.1 & 71.1 \\
FLUX.1 Kontext Dev~\cite{labs2025flux1kontextflowmatching} & \RISEScoreNo & 18.3 & 86.5 & 93.5 & 24.6 & 85.1 & 90.0 & 33.8 & 68.1 & 91.8 & 2.2 & 50.6 & \textemdash & 16.3 & 89.8 & 96.0 & 11.5 & \underline{59.3} & 80.2 & 18.6 & 72.8 & 91.0 \\
LLaDA-Image~\cite{chen2026llada} & \RISEScoreNo & 31.5 & 41.7 & 85.0 & 29.7 & 40.1 & 81.3 & 29.0 & 27.3 & 80.5 & 0.0 & 5.6 & \textemdash & 27.3 & 30.3 & 95.5 & 10.5 & 11.4 & 26.3 & 21.7 & 26.7 & 78.0 \\
Ovis-U1~\cite{wang2025ovis} & \RISEScoreNo & 26.9 & 58.0 & 75.4 & 28.9 & 43.2 & 74.6 & 30.9 & 24.2 & 71.0 & 2.8 & 17.8 & \textemdash & 19.3 & 36.3 & 86.5 & 7.5 & 6.1 & 34.4 & 20.5 & 32.1 & 71.2 \\
InternVL-U~\cite{tian2026internvl} & \RISEScoreNo & 27.5 & 68.1 & 59.5 & 29.3 & 57.2 & 65.6 & 30.5 & 29.8 & 64.7 & 0.6 & 13.3 & \textemdash & 23.0 & 53.3 & 77.8 & 11.5 & 13.4 & 44.8 & 21.0 & 39.7 & 63.4 \\
UniPic2-SD3.5M-Kontext-2B~\cite{wei2025skywork} & \RISEScoreNo & 22.5 & 62.3 & 62.2 & 20.7 & 48.2 & 62.9 & 21.2 & 20.7 & 41.8 & 1.1 & 10.0 & \textemdash & 18.3 & 53.8 & 70.0 & 12.8 & 13.6 & 24.0 & 16.2 & 34.4 & 54.2 \\
Lumina-DiMOO~\cite{xin2025lumina} & \RISEScoreNo & 15.7 & 68.4 & 64.8 & 13.5 & 59.1 & 64.6 & 14.0 & 38.0 & 60.0 & 1.1 & 34.4 & \textemdash & 17.0 & 53.5 & 79.3 & 6.5 & 21.0 & 49.7 & 11.2 & 46.8 & 63.9 \\
UniWorld-V1~\cite{lin2025uniworld} & \RISEScoreNo & 26.2 & 51.4 & 67.4 & 16.9 & 33.7 & 63.7 & 22.8 & 17.9 & 46.7 & 0.6 & 1.1 & \textemdash & 21.0 & 28.8 & 72.8 & 5.5 & 3.8 & 34.1 & 15.8 & 23.3 & 58.3 \\
\midrule[0.4pt]
\rowcolor{RISEScoreGroup}
\multicolumn{23}{c}{\rule{0pt}{9pt}\raisebox{1.5pt}[0pt][0pt]{\textbf{Agentic Methods}}} \\
\midrule[0.4pt]
\textbf{RISE-Agent (Ours)} & \RISEScoreYes & \textbf{78.9} & \textbf{97.7} & \textbf{94.8} & \textbf{80.5} & \textbf{94.6} & \textbf{92.8} & \textbf{67.5} & \textbf{84.0} & \textbf{91.0} & \textbf{39.5} & \textbf{89.5} & \textemdash & \textbf{80.8} & \textbf{96.8} & \textbf{96.8} & \textbf{62.3} & \textbf{78.3} & \textbf{88.6} & \textbf{67.6} & \textbf{90.6} & \textbf{92.9} \\
MindBrush~\cite{he2026mind} & \RISEScoreNo & \underline{68.8} & 71.2 & 90.3 & \underline{70.2} & 72.1 & 83.5 & \underline{54.0} & \underline{58.1} & 81.6 & \underline{8.9} & 37.8 & \textemdash & \underline{63.0} & 69.8 & 88.8 & 25.8 & 20.7 & 43.8 & \underline{49.2} & 56.4 & 81.1 \\
InterleaveThinker~\cite{zheng2026interleavethinker} & \RISEScoreNo & 62.5 & \underline{80.3} & \underline{92.9} & 57.3 & \underline{77.5} & \underline{85.8} & 41.9 & 55.5 & \underline{84.1} & 3.3 & \underline{55.0} & \textemdash & 44.5 & \underline{76.3} & \underline{92.8} & \underline{27.0} & \underline{53.8} & \underline{79.5} & 40.0 & \underline{66.4} & \underline{87.3} \\
IntentEdit~\cite{zhang2026intentedit} & \RISEScoreNo & 65.5 & 74.2 & 80.5 & 61.1 & 77.0 & 84.0 & 43.4 & 56.3 & 80.9 & 1.7 & 35.6 & \textemdash & 49.5 & 69.0 & 89.5 & 19.3 & 32.3 & 42.5 & 41.0 & 58.4 & 78.7 \\
ImAgent~\cite{wang2025imagent} & \RISEScoreNo & 36.6 & 33.7 & 71.9 & 36.2 & 36.8 & 66.4 & 33.4 & 37.0 & 62.5 & 0.6 & 19.4 & \textemdash & 26.8 & 37.0 & 75.0 & 7.5 & 3.0 & 42.9 & 24.6 & 29.2 & 65.4 \\
\bottomrule[0.8pt]
\end{tabular}
\par\vspace{4pt}
\parbox{\textwidth}{\scriptsize
Scores are normalized to the range [1, 5] and rescaled to a 100-point scale for reporting. IR, AC, and VP denote Instruction Reasoning, Appearance Consistency, and Visual Plausibility, respectively. Logical tasks are not evaluated for VP. \textbf{Bold} and \underline{underlined} values indicate the best and second-best scores within each model group.}
\endgroup
\end{table*}

\begin{table*}[!tp]
\caption{Detailed evaluation dimension scores on RISEBench++ (ZH).}
\label{tab:dimension_metric_scores_cn}
\centering
\begingroup
\fontsize{6.5}{8.0}\selectfont
\setlength{\tabcolsep}{0.8pt}
\renewcommand{\arraystretch}{1.06}
\colorlet{RISEScoreGroup}{black!6}
\def\RISEScoreYes{\textcolor{green!50!black}{\ding{51}}}
\def\RISEScoreNo{\textcolor{red!65!black}{\ding{55}}}
\begin{tabular}{
>{\raggedright\arraybackslash}m{\dimexpr .232\textwidth-2\tabcolsep\relax}
>{\centering\arraybackslash}m{\dimexpr .042\textwidth-2\tabcolsep\relax}
*{6}{*{3}{>{\centering\arraybackslash}m{\dimexpr (.726\textwidth-12pt)/21-2\tabcolsep\relax}}!{\hspace{2pt}}}
*{3}{>{\centering\arraybackslash}m{\dimexpr (.726\textwidth-12pt)/21-2\tabcolsep\relax}}
}
\toprule[0.8pt]
\multirow{2}{=}[-1.3pt]{\raggedright\textbf{Model}} & \multirow{2}{=}[-1.3pt]{\centering\shortstack{\textbf{Multi}\\\textbf{img}}} & \multicolumn{3}{c}{\textbf{Temporal}} & \multicolumn{3}{c}{\textbf{Causal}} & \multicolumn{3}{c}{\textbf{Spatial}} & \multicolumn{3}{c}{\textbf{Logical}} & \multicolumn{3}{c}{\textbf{Counterfactual}} & \multicolumn{3}{c}{\textbf{Hybrid}} & \multicolumn{3}{c}{\textbf{Overall}} \\
\cmidrule(lr){3-5}\cmidrule(lr){6-8}\cmidrule(lr){9-11}\cmidrule(lr){12-14}\cmidrule(lr){15-17}\cmidrule(lr){18-20}\cmidrule(lr){21-23}
 & & \textbf{IR} & \textbf{AC} & \textbf{VP} & \textbf{IR} & \textbf{AC} & \textbf{VP} & \textbf{IR} & \textbf{AC} & \textbf{VP} & \textbf{IR} & \textbf{AC} & \textbf{VP} & \textbf{IR} & \textbf{AC} & \textbf{VP} & \textbf{IR} & \textbf{AC} & \textbf{VP} & \textbf{IR} & \textbf{AC} & \textbf{VP} \\
\midrule[0.5pt]
\rowcolor{RISEScoreGroup}
\multicolumn{23}{c}{\rule{0pt}{9pt}\raisebox{1.5pt}[0pt][0pt]{\textbf{Closed-source Models}}} \\
\midrule[0.4pt]
GPT-Image-2.5 Sunburst~\cite{openai2025gptimage} & \RISEScoreYes & 78.1 & 96.3 & 96.9 & 82.3 & \textbf{96.3} & 96.6 & \textbf{80.4} & \textbf{95.7} & \underline{99.0} & 29.0 & \textbf{98.5} & \textemdash & 77.5 & \textbf{98.5} & \RISEFitScore{\textbf{100.0}} & \textbf{58.8} & \textbf{94.4} & \textbf{98.4} & 67.6 & \textbf{96.7} & \textbf{97.9} \\
Nano Banana Pro~\cite{google2026nanobanana} & \RISEScoreYes & \textbf{84.5} & 92.2 & \underline{98.3} & \underline{86.8} & 92.9 & 95.3 & \underline{75.9} & 87.9 & 95.7 & \textbf{39.0} & 84.5 & \textemdash & 76.0 & 87.5 & 99.3 & \underline{58.0} & 83.8 & 93.5 & \textbf{70.6} & 88.6 & 96.5 \\
Nano Banana 2~\cite{google2026nanobanana} & \RISEScoreYes & \underline{84.3} & 93.9 & 96.5 & \textbf{88.3} & 94.1 & \textbf{97.5} & 75.1 & 87.4 & 95.0 & \underline{32.5} & 84.0 & \textemdash & \underline{80.5} & 93.0 & 99.5 & 56.0 & 78.3 & \underline{97.4} & \underline{69.7} & 88.9 & 96.8 \\
Luma Uni-1~\cite{luma2026uni1} & \RISEScoreYes & 83.0 & 95.3 & 96.0 & 81.1 & 94.2 & 93.9 & 75.0 & 89.9 & 96.6 & 18.0 & 83.0 & \textemdash & 70.0 & 92.8 & 96.3 & 48.0 & 78.5 & 88.3 & 63.2 & 89.5 & 94.9 \\
GPT-Image-2~\cite{openai2025gptimage} & \RISEScoreYes & 75.1 & 97.2 & \textbf{98.6} & 77.0 & 94.8 & 96.3 & 73.6 & 91.4 & \textbf{99.0} & 21.0 & 85.0 & \textemdash & 74.0 & 94.8 & 98.3 & 43.8 & 80.6 & 95.5 & 61.1 & 91.1 & \underline{97.7} \\
Qwen-Image-3.0~\cite{zhao2026qwen} & \RISEScoreYes & 83.1 & 88.8 & 97.4 & 80.6 & 84.3 & 95.9 & 68.5 & 83.9 & 92.6 & \textbf{39.0} & 89.5 & \textemdash & 76.8 & 82.0 & 96.8 & \textbf{58.8} & 81.3 & 93.2 & 67.8 & 85.6 & 95.3 \\
GPT-Image-2.5 Flare~\cite{openai2025gptimage} & \RISEScoreYes & 72.3 & \underline{97.5} & 97.4 & 78.9 & 94.6 & 97.0 & 72.3 & \underline{93.4} & \textbf{99.0} & 14.0 & 89.5 & \textemdash & 69.0 & \underline{97.5} & 98.5 & 45.3 & 86.9 & 95.1 & 58.9 & \underline{93.4} & 97.6 \\
FLUX.2 Max~\cite{bfl2025flux2} & \RISEScoreYes & 82.0 & 93.6 & 94.6 & 77.0 & 92.6 & 96.3 & 68.4 & 83.1 & 96.6 & 14.0 & 63.0 & \textemdash & \textbf{83.3} & 94.3 & 98.5 & 44.5 & 73.0 & 92.5 & 61.1 & 83.0 & 95.8 \\
Nano Banana 2 Lite~\cite{google2026nanobanana} & \RISEScoreYes & 79.9 & 95.3 & 96.9 & 78.1 & \underline{95.2} & \textbf{97.5} & 65.6 & 88.3 & 97.0 & 12.0 & 77.0 & \textemdash & 71.5 & 95.5 & \RISEFitScore{\textbf{100.0}} & 44.5 & 76.5 & 92.5 & 58.7 & 88.3 & 97.0 \\
GPT-Image-1.5~\cite{openai2025gptimage} & \RISEScoreYes & 76.9 & 91.3 & 97.7 & 77.1 & 91.3 & 94.3 & 66.6 & 78.1 & 97.0 & 5.0 & 70.0 & \textemdash & 68.8 & 94.5 & 99.0 & 37.8 & 50.3 & 91.2 & 55.7 & 80.5 & 96.2 \\
Nano Banana~\cite{google2026nanobanana} & \RISEScoreYes & 75.5 & 93.9 & 95.5 & 71.5 & 93.1 & 96.4 & 59.8 & 84.1 & 98.5 & 13.5 & 65.5 & \textemdash & 65.0 & 93.8 & \underline{99.8} & 32.8 & 72.5 & 87.7 & 53.8 & 83.8 & 96.3 \\
FLUX.2 Pro~\cite{bfl2025flux2} & \RISEScoreYes & 77.3 & 89.8 & 95.5 & 72.6 & 88.1 & 93.9 & 68.0 & 77.6 & 96.5 & 9.0 & 43.5 & \textemdash & 73.8 & 85.3 & 99.5 & 33.0 & 55.6 & 88.3 & 56.1 & 73.7 & 95.1 \\
HunyuanImage-3.5~\cite{cao2025hunyuanimage} & \RISEScoreYes & 61.4 & \textbf{98.0} & 95.4 & 66.1 & 94.4 & 95.3 & 64.4 & 91.8 & 95.3 & 14.0 & \underline{90.0} & \textemdash & 59.3 & 96.3 & 96.3 & 40.8 & \underline{89.1} & 85.1 & 51.2 & 93.3 & 94.4 \\
Wan2.7-Image-Pro~\cite{mao2026wan} & \RISEScoreYes & 64.0 & 90.9 & 92.4 & 66.6 & 91.2 & 92.9 & 63.6 & 84.2 & 94.5 & 10.0 & 76.5 & \textemdash & 57.8 & 89.3 & 92.5 & 38.5 & 79.5 & 92.2 & 50.4 & 85.3 & 93.1 \\
Qwen-Image-2.0~\cite{zhao2026qwen} & \RISEScoreYes & 69.8 & 91.1 & 97.3 & 67.3 & 87.8 & \underline{97.4} & 58.6 & 77.4 & 95.9 & 9.0 & 70.0 & \textemdash & 55.3 & 89.3 & 98.0 & 30.3 & 33.6 & 35.1 & 49.5 & 77.4 & 90.9 \\
Seedream 4.5~\cite{seedream2025seedream} & \RISEScoreYes & 67.4 & 91.1 & 93.6 & 68.6 & 89.5 & 97.0 & 59.5 & 73.6 & 93.3 & 10.0 & 59.5 & \textemdash & 58.5 & 86.3 & 99.3 & 36.8 & 69.2 & 90.6 & 50.6 & 78.1 & 94.8 \\
Seedream 5.0~\cite{seedream2025seedream} & \RISEScoreYes & 63.8 & 90.8 & 94.5 & 67.0 & 84.7 & 96.1 & 57.0 & 69.6 & 93.8 & 12.5 & 51.5 & \textemdash & 55.3 & 82.8 & 98.5 & 34.8 & 57.8 & 90.9 & 49.1 & 73.2 & 94.9 \\
GPT-Image-1~\cite{openai2025gptimage} & \RISEScoreYes & 69.0 & 88.2 & 96.0 & 70.4 & 83.5 & 93.9 & 60.3 & 67.8 & 95.6 & 5.5 & 61.0 & \textemdash & 57.3 & 85.0 & 99.0 & 33.0 & 43.2 & 86.7 & 50.1 & 72.7 & 94.8 \\
Step Image Edit 2~\cite{liu2025step1x} & \RISEScoreNo & 33.2 & 80.6 & 94.0 & 42.1 & 83.2 & 91.0 & 50.5 & 73.9 & 93.0 & 3.9 & 57.8 & \textemdash & 25.0 & 76.8 & 95.5 & 14.5 & 16.7 & 16.9 & 29.8 & 67.9 & 84.9 \\
\midrule[0.4pt]
\rowcolor{RISEScoreGroup}
\multicolumn{23}{c}{\rule{0pt}{9pt}\raisebox{1.5pt}[0pt][0pt]{\textbf{Open-source Models}}} \\
\midrule[0.4pt]
HunyuanImage-3.0-Instruct~\cite{cao2025hunyuanimage} & \RISEScoreYes & 48.1 & 92.0 & \underline{95.0} & \underline{55.4} & 87.6 & \textbf{93.8} & \textbf{57.6} & \textbf{81.8} & \textbf{96.5} & \textbf{9.0} & 61.0 & \textemdash & 34.5 & \textbf{95.0} & \underline{99.0} & 14.0 & 15.9 & 9.7 & \textbf{38.9} & 75.4 & 87.1 \\
Qwen-Image-2.1~\cite{zhao2026qwen} & \RISEScoreYes & 39.6 & 93.1 & 90.8 & 48.3 & \textbf{90.3} & 90.5 & 46.0 & \underline{79.5} & 94.0 & \textbf{9.0} & \textbf{67.5} & \textemdash & 31.6 & 89.4 & 98.2 & 14.3 & 29.0 & 22.1 & 33.2 & 77.7 & 85.7 \\
FireRed-Image-Edit-1.1~\cite{team2026firered} & \RISEScoreYes & 44.0 & 85.1 & 90.5 & 44.3 & 80.3 & 90.1 & \underline{50.3} & 69.4 & \underline{96.2} & 1.5 & 23.5 & \textemdash & 27.3 & 83.3 & 96.3 & \underline{21.5} & 40.7 & 84.7 & 32.8 & 63.8 & 92.0 \\
FLUX.2 Dev~\cite{bfl2025flux2} & \RISEScoreYes & 42.1 & 72.3 & \textbf{95.1} & 42.6 & 70.6 & \underline{93.5} & 45.9 & 59.5 & 93.0 & 2.5 & 28.5 & \textemdash & 26.8 & 72.3 & 96.8 & 17.0 & 46.0 & \underline{86.7} & 31.0 & 57.8 & \underline{93.5} \\
Qwen-Image-Edit-2511~\cite{wu2025qwen} & \RISEScoreYes & 46.3 & 68.5 & 91.0 & 43.5 & 75.1 & 87.8 & 49.1 & 65.6 & 93.4 & 2.5 & 36.5 & \textemdash & 29.5 & 64.5 & 94.5 & 15.8 & 21.2 & 32.1 & 32.8 & 57.6 & 85.4 \\
Qwen-Image-Edit-2509~\cite{wu2025qwen} & \RISEScoreYes & 40.1 & 71.5 & 86.1 & 43.0 & 75.1 & 86.1 & 46.8 & 64.5 & 90.5 & 3.0 & 19.0 & \textemdash & 29.5 & 62.3 & 96.3 & 14.0 & 15.7 & 37.3 & 30.9 & 53.6 & 83.7 \\
FireRed-Image-Edit-1.0~\cite{team2026firered} & \RISEScoreYes & 42.4 & 67.2 & \underline{95.0} & 43.3 & 73.5 & 91.8 & 49.9 & 66.5 & 95.9 & 2.5 & 36.0 & \textemdash & 25.8 & 79.5 & 97.8 & 17.0 & 33.3 & \textbf{87.3} & 31.9 & 59.8 & \textbf{94.0} \\
SenseNova-U1-A3B-MoT-SFT~\cite{diao2026sensenova} & \RISEScoreYes & \textbf{51.6} & 71.4 & 63.3 & \textbf{64.9} & 70.1 & 70.0 & 45.3 & 44.1 & 47.9 & 3.0 & 52.0 & \textemdash & \underline{41.3} & 64.8 & 65.0 & 11.8 & 5.1 & 25.3 & \underline{38.3} & 54.3 & 57.5 \\
JoyAI-Image-Edit-Plus~\cite{song2026joyai} & \RISEScoreYes & 26.8 & \underline{95.9} & 91.8 & 39.3 & 89.8 & 92.6 & 39.3 & 73.6 & 90.0 & \underline{6.5} & 59.0 & \textemdash & 16.8 & 93.0 & \textbf{99.3} & 19.3 & \underline{60.1} & 83.1 & 26.0 & \underline{78.8} & 91.6 \\
FLUX.2 Klein 9B~\cite{bfl2025flux2} & \RISEScoreYes & 41.3 & 81.0 & 93.0 & 35.8 & 79.3 & 87.9 & 43.0 & 64.9 & 92.9 & 4.0 & 39.5 & \textemdash & 21.0 & 83.0 & 98.0 & 18.5 & 47.0 & 81.8 & 28.8 & 65.7 & 91.2 \\
BAGEL (w/ CoT)~\cite{deng2025bagel} & \RISEScoreYes & \underline{48.9} & 80.3 & 78.0 & 42.5 & 78.2 & 78.6 & 43.9 & 68.9 & 80.6 & 2.0 & 31.5 & \textemdash & 31.3 & 69.4 & 80.8 & 12.0 & 25.0 & 21.4 & 31.8 & 61.0 & 73.6 \\
HiDream-O1-Image~\cite{cai2026hidream} & \RISEScoreYes & 38.3 & 79.4 & 86.5 & 38.1 & 64.5 & 87.5 & 40.8 & 61.3 & 90.6 & 2.5 & 28.5 & \textemdash & 19.5 & 65.3 & 95.0 & 12.5 & 16.2 & 40.6 & 27.1 & 54.6 & 84.4 \\
FLUX.2 Klein 4B~\cite{bfl2025flux2} & \RISEScoreYes & 33.9 & 86.2 & 92.5 & 32.1 & 87.4 & 89.8 & 40.0 & 66.5 & 90.1 & \underline{6.5} & 51.0 & \textemdash & 21.5 & 92.3 & 98.5 & 14.0 & 49.7 & 73.4 & 26.1 & 72.2 & 90.1 \\
UniWorld-V2~\cite{li2025uniworld} & \RISEScoreYes & 38.8 & 52.3 & 89.4 & 43.9 & 64.6 & 87.8 & 46.3 & 57.5 & 92.6 & 1.0 & 17.0 & \textemdash & 32.8 & 60.8 & 94.5 & 13.0 & 16.9 & 42.9 & 30.6 & 46.0 & 85.8 \\
SenseNova-U1-8B-MoT-SFT~\cite{diao2026sensenova} & \RISEScoreYes & 44.5 & 70.5 & 64.6 & 54.9 & 72.0 & 71.5 & 39.8 & 39.9 & 52.8 & 3.5 & 59.0 & \textemdash & \textbf{43.5} & 73.0 & 68.8 & 17.3 & 18.9 & 27.6 & 34.6 & 57.3 & 60.2 \\
UniPic 3.0~\cite{wei2026skywork} & \RISEScoreYes & 30.5 & 73.3 & 82.0 & 37.4 & 72.5 & 85.0 & 38.4 & 64.8 & 87.1 & 1.5 & 19.5 & \textemdash & 19.3 & 77.5 & 94.3 & 12.8 & 18.2 & 25.0 & 24.8 & 55.4 & 80.0 \\
BAGEL~\cite{deng2025bagel} & \RISEScoreYes & 33.8 & 70.6 & 74.8 & 33.5 & 65.2 & 78.6 & 42.4 & 55.1 & 83.3 & 2.5 & 38.0 & \textemdash & 26.0 & 65.5 & 86.0 & 13.0 & 27.3 & 32.8 & 26.3 & 54.9 & 75.2 \\
OmniGen2~\cite{wu2025omnigen2} & \RISEScoreYes & 19.5 & 87.5 & 81.8 & 19.1 & 70.6 & 83.5 & 23.5 & 59.8 & 85.0 & 1.5 & 19.5 & \textemdash & 12.5 & 81.5 & 90.8 & 7.0 & 23.7 & 54.2 & 14.7 & 57.6 & 81.5 \\
OmniGen~\cite{xiao2024omnigen} & \RISEScoreYes & 18.9 & 77.5 & 73.6 & 14.3 & 64.9 & 76.4 & 15.5 & 52.4 & 75.0 & 5.0 & 36.5 & \textemdash & 10.0 & 79.3 & 95.5 & 3.8 & 44.7 & 45.8 & 12.1 & 58.4 & 74.7 \\
Emu2-Gen~\cite{sun2023emu2chat} & \RISEScoreYes & 15.1 & 49.5 & 64.9 & 12.8 & 35.5 & 63.3 & 13.5 & 15.1 & 62.5 & 0.0 & 7.5 & \textemdash & 12.3 & 40.3 & 84.8 & 4.8 & 4.8 & 48.1 & 10.0 & 25.7 & 64.7 \\
JoyAI-Image-Edit~\cite{song2026joyai} & \RISEScoreNo & 29.8 & 88.3 & 91.1 & 45.8 & 80.4 & 90.9 & 45.7 & 68.3 & 94.0 & 2.8 & 55.6 & \textemdash & 33.5 & 75.3 & 97.5 & \textbf{22.5} & 49.2 & 79.2 & 30.6 & 70.4 & 91.4 \\
Boogu-Image-0.1-Edit~\cite{chen2026boogu} & \RISEScoreNo & 28.7 & \textbf{97.1} & 92.4 & 43.2 & \underline{90.0} & 92.2 & 37.0 & 75.0 & 90.9 & 5.0 & \underline{66.7} & \textemdash & 24.5 & \underline{94.5} & 97.5 & 17.5 & 59.3 & 82.5 & 27.1 & \textbf{80.7} & 91.6 \\
\RISEFitModel{Step1X-Edit (thinking + reflection)~\cite{liu2025step1x}} & \RISEScoreNo & 37.4 & 86.5 & 91.5 & 37.4 & 83.4 & 90.1 & 30.1 & 58.9 & 88.6 & 2.2 & 37.2 & \textemdash & 30.8 & 80.0 & 95.0 & 18.5 & 49.0 & 76.0 & 26.3 & 65.7 & 89.2 \\
Step1X-Edit (thinking)~\cite{liu2025step1x} & \RISEScoreNo & 34.2 & 88.0 & 90.5 & 35.2 & 84.5 & 90.6 & 33.1 & 63.2 & 93.3 & 2.2 & 37.2 & \textemdash & 31.0 & 87.8 & 96.8 & 14.5 & 44.9 & 58.4 & 25.5 & 67.5 & 88.7 \\
Step1X-Edit (base)~\cite{liu2025step1x} & \RISEScoreNo & 27.7 & 93.8 & 91.1 & 35.2 & 85.1 & 85.5 & 33.7 & 69.1 & 89.6 & 1.1 & 28.3 & \textemdash & 33.0 & 89.0 & 92.8 & 12.5 & 44.2 & 64.3 & 24.2 & 68.1 & 86.6 \\
LongCat-Image-Edit~\cite{team2025longcat} & \RISEScoreNo & 31.6 & 78.1 & 89.1 & 37.6 & 76.7 & 87.9 & 44.8 & 58.9 & 90.2 & 3.3 & 37.8 & \textemdash & 26.3 & 53.0 & 95.0 & 14.3 & 10.9 & 38.0 & 27.5 & 55.8 & 84.4 \\
FLUX.1 Kontext Dev~\cite{labs2025flux1kontextflowmatching} & \RISEScoreNo & 10.6 & 88.1 & 82.0 & 10.9 & 82.6 & 84.2 & 15.1 & 70.1 & 85.9 & 4.4 & 35.6 & \textemdash & 11.3 & 83.0 & 89.0 & 5.3 & \textbf{73.2} & 66.2 & 9.8 & 70.7 & 82.9 \\
LLaDA-Image~\cite{chen2026llada} & \RISEScoreNo & 26.2 & 44.2 & 88.8 & 28.8 & 37.1 & 82.8 & 27.5 & 20.2 & 85.2 & 0.6 & 6.7 & \textemdash & 28.5 & 30.3 & 94.8 & 7.3 & 8.7 & 31.3 & 20.2 & 25.0 & 81.0 \\
InternVL-U (w/ CoT)~\cite{tian2026internvl} & \RISEScoreNo & 24.9 & 69.1 & 72.5 & 27.9 & 52.6 & 66.3 & 29.4 & 33.0 & 64.7 & 0.0 & 21.7 & \textemdash & 26.3 & 57.0 & 80.5 & 13.5 & 11.6 & 39.3 & 20.4 & 41.4 & 66.4 \\
Ovis-U1~\cite{wang2025ovis} & \RISEScoreNo & 24.4 & 69.6 & 69.0 & 20.8 & 60.0 & 70.4 & 25.5 & 31.9 & 69.8 & 2.2 & 25.0 & \textemdash & 18.5 & 40.0 & 84.5 & 6.0 & 7.6 & 27.3 & 16.9 & 41.1 & 67.3 \\
InternVL-U~\cite{tian2026internvl} & \RISEScoreNo & 23.7 & 60.9 & 68.5 & 28.4 & 52.9 & 67.4 & 28.2 & 29.8 & 69.2 & 0.6 & 24.4 & \textemdash & 23.5 & 51.8 & 83.0 & 9.5 & 11.6 & 46.8 & 19.5 & 39.3 & 68.1 \\
Lumina-DiMOO~\cite{xin2025lumina} & \RISEScoreNo & 17.1 & 67.5 & 56.5 & 12.6 & 66.4 & 68.9 & 13.0 & 42.3 & 55.2 & 3.3 & 31.1 & \textemdash & 10.5 & 52.8 & 75.5 & 7.0 & 33.1 & 43.2 & 10.9 & 49.6 & 60.7 \\
UniWorld-V1~\cite{lin2025uniworld} & \RISEScoreNo & 17.4 & 53.8 & 64.6 & 13.0 & 35.5 & 59.5 & 12.4 & 16.6 & 50.0 & 0.0 & 1.1 & \textemdash & 17.8 & 41.5 & 75.0 & 4.0 & 2.0 & 26.0 & 10.7 & 25.1 & 56.8 \\
UniPic2-SD3.5M-Kontext-2B~\cite{wei2025skywork} & \RISEScoreNo & 15.6 & 63.4 & 50.9 & 12.5 & 56.6 & 52.5 & 12.9 & 20.4 & 37.2 & 1.7 & 8.9 & \textemdash & 11.0 & 54.5 & 72.8 & 7.0 & 29.8 & 26.9 & 10.3 & 37.9 & 48.4 \\
\midrule[0.4pt]
\rowcolor{RISEScoreGroup}
\multicolumn{23}{c}{\rule{0pt}{9pt}\raisebox{1.5pt}[0pt][0pt]{\textbf{Agentic Methods}}} \\
\midrule[0.4pt]
\textbf{RISE-Agent (Ours)} & \RISEScoreYes & \textbf{80.7} & \textbf{92.8} & \textbf{93.1} & \textbf{80.0} & \textbf{93.7} & \textbf{93.8} & \textbf{68.9} & \textbf{82.1} & \textbf{96.4} & \textbf{27.0} & \textbf{80.5} & \textemdash & \textbf{80.6} & \textbf{92.7} & \underline{95.7} & \textbf{46.4} & \textbf{63.7} & \textbf{82.7} & \textbf{63.8} & \textbf{85.3} & \textbf{93.4} \\
MindBrush~\cite{he2026mind} & \RISEScoreNo & \underline{71.6} & \underline{86.1} & 89.9 & \underline{68.9} & \underline{78.6} & \underline{83.7} & \underline{56.5} & \underline{59.3} & 76.8 & \underline{6.1} & 38.3 & \textemdash & \underline{70.0} & 71.5 & 90.5 & \underline{23.8} & 23.5 & 42.9 & \underline{49.9} & \underline{61.2} & 80.0 \\
InterleaveThinker~\cite{zheng2026interleavethinker} & \RISEScoreNo & 57.1 & 76.0 & \underline{90.6} & 50.1 & 72.8 & 82.8 & 42.0 & 50.8 & 73.5 & 1.7 & \underline{48.3} & \textemdash & 49.0 & \underline{78.0} & \textbf{97.3} & 19.3 & \underline{36.6} & \underline{67.9} & 36.9 & 60.7 & \underline{82.7} \\
IntentEdit~\cite{zhang2026intentedit} & \RISEScoreNo & 62.6 & 69.8 & 76.6 & 58.3 & 75.1 & 82.8 & 48.8 & 57.6 & \underline{81.0} & 2.2 & 41.1 & \textemdash & 50.3 & 66.7 & 86.9 & 19.0 & 32.1 & 38.0 & 41.1 & 58.3 & 76.7 \\
ImAgent~\cite{wang2025imagent} & \RISEScoreNo & 32.0 & 27.2 & 73.8 & 40.0 & 34.5 & 65.2 & 32.4 & 28.0 & 67.3 & 0.0 & 13.9 & \textemdash & 26.5 & 25.0 & 64.3 & 7.0 & 2.5 & 41.2 & 24.2 & 23.3 & 65.1 \\
\bottomrule[0.8pt]
\end{tabular}
\par\vspace{4pt}
\parbox{\textwidth}{\scriptsize
Scores are normalized to the range [1, 5] and rescaled to a 100-point scale for reporting. IR, AC, and VP denote Instruction Reasoning, Appearance Consistency, and Visual Plausibility, respectively. Logical tasks are not evaluated for VP. \textbf{Bold} and \underline{underlined} values indicate the best and second-best scores within each model group.}
\endgroup
\end{table*}

\begin{table*}[!p]
\caption{Fine-grained task accuracy (\%) on RISEBench++ (EN). Part I: Temporal reasoning.}
\label{tab:task_accuracy_en_1}
\centering
\begingroup
\fontsize{6.5}{8.2}\selectfont
\setlength{\tabcolsep}{0.3pt}
\renewcommand{\arraystretch}{1.0}
\setlength{\aboverulesep}{1pt}
\setlength{\belowrulesep}{1pt}
\colorlet{RISETaskGroup}{black!6}
\def\RISETaskYes{\textcolor{green!50!black}{\ding{51}}}
\def\RISETaskNo{\textcolor{red!65!black}{\ding{55}}}
\begin{tabular}{
>{\raggedright\arraybackslash}m{\dimexpr .245\textwidth-2\tabcolsep\relax}
>{\centering\arraybackslash}m{\dimexpr .045\textwidth-2\tabcolsep\relax}
*{19}{>{\centering\arraybackslash}m{\dimexpr 0.037368421\textwidth-2\tabcolsep\relax}}
}
\toprule[0.8pt]
\multirow{2}{=}[-1.3pt]{\raggedright\textbf{Model}} & \multirow{2}{=}[-1.3pt]{\centering\shortstack{\textbf{Multi-}\\\textbf{img}}} & \multicolumn{19}{c}{\textbf{Temporal}} \\
\cmidrule(lr){3-21}
 & & \textbf{T01} & \textbf{T02} & \textbf{T03} & \textbf{T04} & \textbf{T05} & \textbf{T06} & \textbf{T07} & \textbf{T08} & \textbf{T09} & \textbf{T10} & \textbf{T11} & \textbf{T12} & \textbf{T13} & \textbf{T14} & \textbf{T15} & \textbf{T16} & \textbf{T17} & \textbf{T18} & \textbf{T19} \\
\midrule[0.5pt]
\rowcolor{RISETaskGroup}
\multicolumn{21}{c}{\rule{0pt}{9pt}\raisebox{1.5pt}[0pt][0pt]{\textbf{Closed-source Models}}} \\
\midrule[0.4pt]
GPT-Image-2.5 Sunburst~\cite{openai2025gptimage} & \RISETaskYes & 58.3 & \underline{88.9} & \textbf{76.9} & \underline{80.0} & \underline{87.5} & \textbf{71.4} & \underline{50.0} & 60.0 & 50.0 & \underline{73.3} & \RISEFitScore{\textbf{100.0}} & 50.0 & \underline{50.0} & \underline{66.7} & \underline{58.3} & \textbf{66.7} & 40.0 & \textbf{75.0} & \underline{66.7} \\
Nano Banana Pro~\cite{google2026nanobanana} & \RISETaskYes & 62.5 & \RISEFitScore{\textbf{100.0}} & \textbf{76.9} & 70.0 & 50.0 & 42.9 & \textbf{75.0} & \underline{66.7} & 64.3 & 60.0 & 62.5 & \underline{75.0} & 25.0 & \underline{66.7} & \textbf{66.7} & 50.0 & 60.0 & \textbf{75.0} & 33.3 \\
Nano Banana 2~\cite{google2026nanobanana} & \RISETaskYes & 62.5 & \underline{88.9} & \textbf{76.9} & \textbf{90.0} & \RISEFitScore{\textbf{100.0}} & \underline{57.1} & \textbf{75.0} & 60.0 & 71.4 & 53.3 & \underline{75.0} & 25.0 & 37.5 & 55.6 & \textbf{66.7} & 55.6 & \RISEFitScore{\textbf{100.0}} & \underline{50.0} & \underline{66.7} \\
GPT-Image-2~\cite{openai2025gptimage} & \RISETaskYes & 58.3 & \underline{88.9} & \underline{69.2} & 65.0 & 62.5 & 42.9 & 0.0 & 53.3 & 64.3 & \textbf{86.7} & 62.5 & 50.0 & \underline{50.0} & \underline{66.7} & 50.0 & 44.4 & 40.0 & \textbf{75.0} & 33.3 \\
GPT-Image-2.5 Flare~\cite{openai2025gptimage} & \RISETaskYes & 54.2 & 77.8 & \textbf{76.9} & 75.0 & \underline{87.5} & \underline{57.1} & \underline{50.0} & \underline{66.7} & 42.9 & \underline{73.3} & 62.5 & 25.0 & 37.5 & 55.6 & 50.0 & 55.6 & 60.0 & \textbf{75.0} & 33.3 \\
Luma Uni-1~\cite{luma2026uni1} & \RISETaskYes & \underline{70.8} & \underline{88.9} & \textbf{76.9} & 75.0 & \underline{87.5} & 42.9 & \textbf{75.0} & 60.0 & 64.3 & 46.7 & 62.5 & \underline{75.0} & 12.5 & \underline{66.7} & 41.7 & 27.8 & \underline{80.0} & \textbf{75.0} & \underline{66.7} \\
Qwen-Image-3.0~\cite{zhao2026qwen} & \RISETaskYes & 45.8 & 77.8 & 53.8 & 70.0 & 75.0 & 42.9 & \underline{50.0} & 60.0 & \textbf{85.7} & 66.7 & 50.0 & \RISEFitScore{\textbf{100.0}} & \underline{50.0} & 22.2 & \textbf{66.7} & \underline{61.1} & 0.0 & \underline{50.0} & 33.3 \\
Nano Banana 2 Lite~\cite{google2026nanobanana} & \RISETaskYes & 54.2 & \RISEFitScore{\textbf{100.0}} & 61.5 & 60.0 & \underline{87.5} & \textbf{71.4} & \textbf{75.0} & 53.3 & \underline{78.6} & 60.0 & 62.5 & 50.0 & 37.5 & 55.6 & 41.7 & 50.0 & 40.0 & \textbf{75.0} & \underline{66.7} \\
FLUX.2 Max~\cite{bfl2025flux2} & \RISETaskYes & 45.8 & \underline{88.9} & \underline{69.2} & 55.0 & \underline{87.5} & \underline{57.1} & \underline{50.0} & \textbf{80.0} & 71.4 & \underline{73.3} & 37.5 & 50.0 & 12.5 & 55.6 & \textbf{66.7} & 27.8 & 20.0 & \underline{50.0} & \underline{66.7} \\
GPT-Image-1.5~\cite{openai2025gptimage} & \RISETaskYes & 62.5 & \underline{88.9} & 53.8 & \underline{80.0} & 75.0 & 42.9 & 25.0 & 46.7 & 50.0 & 60.0 & 50.0 & 50.0 & 25.0 & \textbf{77.8} & \underline{58.3} & \underline{61.1} & 20.0 & 25.0 & 33.3 \\
FLUX.2 Pro~\cite{bfl2025flux2} & \RISETaskYes & 50.0 & 77.8 & 61.5 & 55.0 & 75.0 & \underline{57.1} & \underline{50.0} & 60.0 & 57.1 & 66.7 & 50.0 & 50.0 & 37.5 & 55.6 & \underline{58.3} & 27.8 & 60.0 & 25.0 & 0.0 \\
Nano Banana~\cite{google2026nanobanana} & \RISETaskYes & 37.5 & 77.8 & 61.5 & 55.0 & 75.0 & 42.9 & \textbf{75.0} & 46.7 & 50.0 & 60.0 & 50.0 & 25.0 & \textbf{62.5} & 55.6 & 50.0 & 44.4 & 20.0 & 25.0 & \RISEFitScore{\textbf{100.0}} \\
HunyuanImage-3.5~\cite{cao2025hunyuanimage} & \RISETaskYes & 33.3 & 77.8 & 46.2 & 55.0 & \underline{87.5} & \textbf{71.4} & 25.0 & 46.7 & 28.6 & 66.7 & 37.5 & 0.0 & 37.5 & 22.2 & \underline{58.3} & 27.8 & 0.0 & 0.0 & 33.3 \\
Wan2.7-Image-Pro~\cite{mao2026wan} & \RISETaskYes & 58.3 & \underline{88.9} & 61.5 & 60.0 & 62.5 & 28.6 & \underline{50.0} & 26.7 & 42.9 & 0.0 & 0.0 & 0.0 & 25.0 & 33.3 & 41.7 & 50.0 & 0.0 & 25.0 & \underline{66.7} \\
GPT-Image-1~\cite{openai2025gptimage} & \RISETaskYes & \textbf{75.0} & 77.8 & 61.5 & 55.0 & 75.0 & \textbf{71.4} & 0.0 & 53.3 & 35.7 & 26.7 & 37.5 & 50.0 & 25.0 & 55.6 & 41.7 & 50.0 & 40.0 & \underline{50.0} & 0.0 \\
Qwen-Image-2.0~\cite{zhao2026qwen} & \RISETaskYes & 45.8 & 66.7 & 38.5 & 65.0 & 50.0 & 42.9 & \underline{50.0} & 53.3 & 35.7 & 60.0 & 12.5 & 25.0 & 12.5 & 55.6 & 25.0 & 33.3 & 0.0 & \textbf{75.0} & \underline{66.7} \\
Seedream 4.5~\cite{seedream2025seedream} & \RISETaskYes & 37.5 & 66.7 & 53.8 & 50.0 & 25.0 & 42.9 & \textbf{75.0} & 53.3 & 35.7 & 46.7 & 50.0 & 0.0 & 12.5 & 33.3 & 41.7 & \underline{61.1} & 0.0 & 0.0 & 33.3 \\
Seedream 5.0~\cite{seedream2025seedream} & \RISETaskYes & 54.2 & 66.7 & 61.5 & 55.0 & 62.5 & \underline{57.1} & \underline{50.0} & 20.0 & 35.7 & 40.0 & 50.0 & 0.0 & 37.5 & 33.3 & 33.3 & 27.8 & 40.0 & 25.0 & 0.0 \\
Step Image Edit 2~\cite{liu2025step1x} & \RISETaskNo & 20.8 & 11.1 & 46.2 & 35.0 & 37.5 & 28.6 & 0.0 & 9.1 & 0.0 & 0.0 & \textemdash & 0.0 & 0.0 & 11.1 & 16.7 & 11.1 & 0.0 & 25.0 & 0.0 \\
\midrule[0.4pt]
\rowcolor{RISETaskGroup}
\multicolumn{21}{c}{\rule{0pt}{9pt}\raisebox{1.5pt}[0pt][0pt]{\textbf{Open-source Models}}} \\
\midrule[0.4pt]
HunyuanImage-3.0-Instruct~\cite{cao2025hunyuanimage} & \RISETaskYes & 20.8 & \textbf{77.8} & 30.8 & \textbf{60.0} & 37.5 & \underline{14.3} & \underline{25.0} & 20.0 & \underline{7.1} & \underline{66.7} & 37.5 & \underline{0.0} & 0.0 & 11.1 & 8.3 & \textbf{44.4} & \underline{0.0} & 25.0 & \underline{0.0} \\
Qwen-Image-Edit-2511~\cite{wu2025qwen} & \RISETaskYes & 16.7 & 44.4 & \underline{46.2} & 30.0 & 37.5 & \underline{14.3} & 0.0 & \textbf{40.0} & 0.0 & 46.7 & 0.0 & \underline{0.0} & 0.0 & 11.1 & 8.3 & \underline{27.8} & \underline{0.0} & 0.0 & \underline{0.0} \\
Qwen-Image-2.1~\cite{zhao2026qwen} & \RISETaskYes & 8.3 & \underline{66.7} & 30.8 & 35.0 & 12.5 & 0.0 & 0.0 & \underline{26.7} & \textbf{14.3} & 46.7 & 25.0 & \underline{0.0} & \underline{12.5} & \underline{22.2} & 8.3 & 22.2 & \underline{0.0} & 0.0 & \underline{0.0} \\
FireRed-Image-Edit-1.0~\cite{team2026firered} & \RISETaskYes & 8.3 & 33.3 & \underline{46.2} & 40.0 & \underline{50.0} & \underline{14.3} & \textbf{50.0} & 6.7 & 0.0 & 33.3 & 37.5 & \underline{0.0} & \underline{12.5} & \textbf{33.3} & 8.3 & 22.2 & \underline{0.0} & 0.0 & \underline{0.0} \\
FireRed-Image-Edit-1.1~\cite{team2026firered} & \RISETaskYes & 8.3 & 22.2 & 38.5 & 30.0 & \textbf{62.5} & 0.0 & \underline{25.0} & 20.0 & \textbf{14.3} & 53.3 & 25.0 & \underline{0.0} & 0.0 & \underline{22.2} & 8.3 & \underline{27.8} & \underline{0.0} & 25.0 & \underline{0.0} \\
Qwen-Image-Edit-2509~\cite{wu2025qwen} & \RISETaskYes & 4.2 & 22.2 & 38.5 & 20.0 & \textbf{62.5} & \textbf{28.6} & \textbf{50.0} & 13.3 & 0.0 & 26.7 & 37.5 & \underline{0.0} & \underline{12.5} & \underline{22.2} & \textbf{25.0} & \underline{27.8} & \underline{0.0} & 25.0 & \underline{0.0} \\
HiDream-O1-Image~\cite{cai2026hidream} & \RISETaskYes & \underline{41.7} & \textbf{77.8} & 30.8 & \underline{50.0} & 25.0 & 0.0 & \underline{25.0} & 20.0 & \underline{7.1} & 40.0 & \underline{50.0} & \underline{0.0} & \underline{12.5} & \underline{22.2} & 0.0 & 22.2 & \underline{0.0} & 0.0 & \underline{0.0} \\
FLUX.2 Dev~\cite{bfl2025flux2} & \RISETaskYes & 8.3 & 55.6 & 23.1 & 15.0 & 37.5 & 0.0 & 0.0 & 20.0 & \textbf{14.3} & 46.7 & 37.5 & \underline{0.0} & \underline{12.5} & 11.1 & 8.3 & \underline{27.8} & \underline{0.0} & 0.0 & \underline{0.0} \\
FLUX.2 Klein 9B~\cite{bfl2025flux2} & \RISETaskYes & 0.0 & 33.3 & 23.1 & 10.0 & 25.0 & 0.0 & 0.0 & 20.0 & 0.0 & 53.3 & \textbf{62.5} & \underline{0.0} & 0.0 & 11.1 & 0.0 & 16.7 & \underline{0.0} & \underline{50.0} & \underline{0.0} \\
JoyAI-Image-Edit-Plus~\cite{song2026joyai} & \RISETaskYes & 8.3 & 22.2 & 23.1 & 5.0 & 37.5 & 0.0 & 0.0 & 20.0 & \underline{7.1} & 53.3 & \textbf{62.5} & \underline{0.0} & 0.0 & 11.1 & 8.3 & 16.7 & \underline{0.0} & 0.0 & \underline{0.0} \\
SenseNova-U1-A3B-MoT-SFT~\cite{diao2026sensenova} & \RISETaskYes & \textbf{62.5} & \underline{66.7} & 38.5 & 35.0 & 0.0 & \textbf{28.6} & 0.0 & 6.7 & 0.0 & 0.0 & 0.0 & \underline{0.0} & \underline{12.5} & 11.1 & 8.3 & 11.1 & \underline{0.0} & 0.0 & \underline{0.0} \\
UniPic 3.0~\cite{wei2026skywork} & \RISETaskYes & 8.3 & 11.1 & 38.5 & 25.0 & 12.5 & 0.0 & \underline{25.0} & 13.3 & \underline{7.1} & 60.0 & \underline{50.0} & \underline{0.0} & 0.0 & \textbf{33.3} & 0.0 & 16.7 & \underline{0.0} & 0.0 & \underline{0.0} \\
BAGEL (w/ CoT)~\cite{deng2025bagel} & \RISETaskYes & 25.0 & 55.6 & 15.4 & 15.0 & 25.0 & \underline{14.3} & 0.0 & \underline{26.7} & \underline{7.1} & 33.3 & 25.0 & \underline{0.0} & \underline{12.5} & \underline{22.2} & \underline{16.7} & 5.6 & \underline{0.0} & 0.0 & \underline{0.0} \\
SenseNova-U1-8B-MoT-SFT~\cite{diao2026sensenova} & \RISETaskYes & \underline{41.7} & \underline{66.7} & 15.4 & 30.0 & 0.0 & 0.0 & 0.0 & 13.3 & 0.0 & 0.0 & 0.0 & \underline{0.0} & 0.0 & \underline{22.2} & 8.3 & 22.2 & \underline{0.0} & 0.0 & \underline{0.0} \\
FLUX.2 Klein 4B~\cite{bfl2025flux2} & \RISETaskYes & 0.0 & 11.1 & 30.8 & 10.0 & 25.0 & 0.0 & \underline{25.0} & \underline{26.7} & \textbf{14.3} & \textbf{73.3} & 25.0 & \underline{0.0} & 0.0 & 0.0 & 0.0 & 11.1 & \underline{0.0} & 0.0 & \underline{0.0} \\
UniWorld-V2~\cite{li2025uniworld} & \RISETaskYes & 0.0 & 11.1 & 23.1 & 15.0 & 37.5 & \underline{14.3} & \underline{25.0} & 13.3 & 0.0 & 6.7 & 12.5 & \underline{0.0} & 0.0 & 11.1 & 8.3 & 11.1 & \underline{0.0} & 0.0 & \underline{0.0} \\
BAGEL~\cite{deng2025bagel} & \RISETaskYes & 0.0 & 11.1 & 7.7 & 0.0 & 0.0 & 0.0 & 0.0 & 13.3 & \underline{7.1} & 26.7 & 37.5 & \underline{0.0} & 0.0 & 0.0 & 0.0 & 5.6 & \underline{0.0} & 0.0 & \underline{0.0} \\
OmniGen2~\cite{wu2025omnigen2} & \RISETaskYes & 0.0 & 0.0 & 15.4 & 0.0 & 0.0 & 0.0 & \underline{25.0} & 6.7 & 0.0 & 26.7 & 37.5 & \underline{0.0} & \underline{12.5} & 0.0 & 8.3 & 5.6 & \underline{0.0} & 0.0 & \underline{0.0} \\
OmniGen~\cite{xiao2024omnigen} & \RISETaskYes & 4.2 & 0.0 & 7.7 & 0.0 & 12.5 & 0.0 & 0.0 & 0.0 & 0.0 & 26.7 & 0.0 & \underline{0.0} & \underline{12.5} & 11.1 & 0.0 & 0.0 & \underline{0.0} & 0.0 & \underline{0.0} \\
Emu2-Gen~\cite{sun2023emu2chat} & \RISETaskYes & 0.0 & 0.0 & 7.7 & 0.0 & 12.5 & 0.0 & 0.0 & 6.7 & 0.0 & 0.0 & 0.0 & \underline{0.0} & 0.0 & \underline{22.2} & 0.0 & 16.7 & \underline{0.0} & 0.0 & \underline{0.0} \\
JoyAI-Image-Edit~\cite{song2026joyai} & \RISETaskNo & 4.2 & 44.4 & \textbf{53.8} & 30.0 & 25.0 & 0.0 & \underline{25.0} & 9.1 & 0.0 & 0.0 & \textemdash & \underline{0.0} & 0.0 & \underline{22.2} & 8.3 & 11.1 & \underline{0.0} & \textbf{75.0} & \underline{0.0} \\
Step1X-Edit (thinking)~\cite{liu2025step1x} & \RISETaskNo & 20.8 & \underline{66.7} & 30.8 & 30.0 & 12.5 & \textbf{28.6} & 0.0 & 18.2 & 0.0 & 0.0 & \textemdash & \underline{0.0} & 0.0 & 0.0 & \underline{16.7} & 22.2 & \underline{0.0} & 25.0 & \underline{0.0} \\
Boogu-Image-0.1-Edit~\cite{chen2026boogu} & \RISETaskNo & 8.3 & 22.2 & 30.8 & 15.0 & 37.5 & 0.0 & 0.0 & 9.1 & 0.0 & 0.0 & \textemdash & \textbf{25.0} & 0.0 & \underline{22.2} & \textbf{25.0} & \underline{27.8} & \underline{0.0} & 0.0 & \underline{0.0} \\
\RISEFitModel{Step1X-Edit (thinking + reflection)~\cite{liu2025step1x}} & \RISETaskNo & 25.0 & 33.3 & \underline{46.2} & 30.0 & 25.0 & \textbf{28.6} & \underline{25.0} & 0.0 & 0.0 & 0.0 & \textemdash & \underline{0.0} & \textbf{25.0} & \textbf{33.3} & \underline{16.7} & 16.7 & \underline{0.0} & 0.0 & \textbf{33.3} \\
LongCat-Image-Edit~\cite{team2025longcat} & \RISETaskNo & 8.3 & 0.0 & 23.1 & 10.0 & 37.5 & 0.0 & 0.0 & 0.0 & 0.0 & 0.0 & \textemdash & \underline{0.0} & 0.0 & \underline{22.2} & \underline{16.7} & 16.7 & \underline{0.0} & 0.0 & \underline{0.0} \\
Step1X-Edit (base)~\cite{liu2025step1x} & \RISETaskNo & 16.7 & 33.3 & 23.1 & 15.0 & 12.5 & 0.0 & \underline{25.0} & 9.1 & 0.0 & 0.0 & \textemdash & \underline{0.0} & 0.0 & 11.1 & 8.3 & 22.2 & \underline{0.0} & 0.0 & \underline{0.0} \\
InternVL-U (w/ CoT)~\cite{tian2026internvl} & \RISETaskNo & 12.5 & 33.3 & 7.7 & 10.0 & 0.0 & \underline{14.3} & 0.0 & 9.1 & 0.0 & 0.0 & \textemdash & \underline{0.0} & \textbf{25.0} & 0.0 & \underline{16.7} & 0.0 & \underline{0.0} & 25.0 & \underline{0.0} \\
FLUX.1 Kontext Dev~\cite{labs2025flux1kontextflowmatching} & \RISETaskNo & 4.2 & 0.0 & 30.8 & 5.0 & 0.0 & 0.0 & 0.0 & 0.0 & 0.0 & 0.0 & \textemdash & \underline{0.0} & 0.0 & 11.1 & 0.0 & 11.1 & \underline{0.0} & 0.0 & \underline{0.0} \\
LLaDA-Image~\cite{chen2026llada} & \RISETaskNo & 8.3 & 11.1 & 7.7 & 10.0 & 12.5 & \underline{14.3} & \underline{25.0} & 0.0 & 0.0 & 0.0 & \textemdash & \underline{0.0} & 0.0 & 11.1 & 8.3 & 5.6 & \underline{0.0} & 0.0 & \underline{0.0} \\
Ovis-U1~\cite{wang2025ovis} & \RISETaskNo & 0.0 & 0.0 & 23.1 & 5.0 & 12.5 & 0.0 & \underline{25.0} & 0.0 & 0.0 & 0.0 & \textemdash & \underline{0.0} & 0.0 & 0.0 & 0.0 & 5.6 & \textbf{20.0} & 0.0 & \underline{0.0} \\
InternVL-U~\cite{tian2026internvl} & \RISETaskNo & 4.2 & 0.0 & 15.4 & 10.0 & 0.0 & 0.0 & 0.0 & 0.0 & 0.0 & 0.0 & \textemdash & \underline{0.0} & 0.0 & 0.0 & 0.0 & 5.6 & \underline{0.0} & 0.0 & \underline{0.0} \\
UniPic2-SD3.5M-Kontext-2B~\cite{wei2025skywork} & \RISETaskNo & 0.0 & 0.0 & 7.7 & 0.0 & 12.5 & 0.0 & 0.0 & 0.0 & 0.0 & 0.0 & \textemdash & \underline{0.0} & 0.0 & 0.0 & 0.0 & 0.0 & \underline{0.0} & 0.0 & \underline{0.0} \\
Lumina-DiMOO~\cite{xin2025lumina} & \RISETaskNo & 0.0 & 0.0 & 7.7 & 0.0 & 0.0 & 0.0 & 0.0 & 0.0 & 0.0 & 0.0 & \textemdash & \underline{0.0} & 0.0 & 0.0 & 0.0 & 0.0 & \underline{0.0} & 0.0 & \underline{0.0} \\
UniWorld-V1~\cite{lin2025uniworld} & \RISETaskNo & 0.0 & 0.0 & 15.4 & 0.0 & 0.0 & 0.0 & 0.0 & 0.0 & 0.0 & 0.0 & \textemdash & \underline{0.0} & 0.0 & 0.0 & 0.0 & 0.0 & \underline{0.0} & 0.0 & \underline{0.0} \\
\midrule[0.4pt]
\rowcolor{RISETaskGroup}
\multicolumn{21}{c}{\rule{0pt}{9pt}\raisebox{1.5pt}[0pt][0pt]{\textbf{Agentic Methods}}} \\
\midrule[0.4pt]
\textbf{RISE-Agent (Ours)} & \RISETaskYes & \underline{54.2} & \textbf{88.9} & \underline{61.5} & \textbf{70.0} & \textbf{62.5} & \underline{57.1} & \textbf{75.0} & \textbf{60.0} & \textbf{42.9} & \textbf{80.0} & \textbf{37.5} & \textbf{50.0} & \textbf{50.0} & \textbf{55.6} & \textbf{58.3} & \textbf{50.0} & \textbf{20.0} & \textbf{50.0} & \textbf{66.7} \\
MindBrush~\cite{he2026mind} & \RISETaskNo & 37.5 & \textbf{88.9} & \textbf{69.2} & \underline{55.0} & \underline{50.0} & 42.9 & \underline{50.0} & 9.1 & 16.7 & \underline{0.0} & \textemdash & 0.0 & \underline{25.0} & \underline{11.1} & \underline{41.7} & \underline{11.1} & \underline{0.0} & \underline{0.0} & \underline{33.3} \\
InterleaveThinker~\cite{zheng2026interleavethinker} & \RISETaskNo & \textbf{58.3} & 55.6 & \underline{61.5} & 45.0 & \underline{50.0} & \textbf{71.4} & \underline{50.0} & \underline{36.4} & 16.7 & \underline{0.0} & \textemdash & \underline{25.0} & \textbf{50.0} & \underline{11.1} & 25.0 & 5.6 & \underline{0.0} & \textbf{50.0} & \textbf{66.7} \\
IntentEdit~\cite{zhang2026intentedit} & \RISETaskNo & 33.3 & \underline{66.7} & 23.1 & 45.0 & 25.0 & 14.3 & 0.0 & 18.2 & \underline{25.0} & \underline{0.0} & \textemdash & 0.0 & \underline{25.0} & \underline{11.1} & 8.3 & \underline{11.1} & \underline{0.0} & \underline{0.0} & 0.0 \\
ImAgent~\cite{wang2025imagent} & \RISETaskNo & 8.3 & 44.4 & 0.0 & 0.0 & 0.0 & 0.0 & 0.0 & 9.1 & 0.0 & \underline{0.0} & \textemdash & 0.0 & \underline{25.0} & 0.0 & 8.3 & 0.0 & \underline{0.0} & \underline{0.0} & 0.0 \\
\bottomrule[0.8pt]
\end{tabular}
\par\vspace{4pt}
\parbox{\textwidth}{\fontsize{6.5}{8.2}\selectfont\raggedright
\textbf{T01}: Life Progression; \textbf{T02}: Material Transformation; \textbf{T03}: Natural Environmental Changes; \textbf{T04}: Physical Aging; \textbf{T05}: Social Activities; \textbf{T06}: Tool Evolution; \textbf{T07}: Urban Evolution; \textbf{T08}: Temporal State Transition; \textbf{T09}: Precise Temporal Localization; \textbf{T10}: Position Change; \textbf{T11}: Action Sequence; \textbf{T12}: Physics; \textbf{T13}: Chemistry; \textbf{T14}: Biology; \textbf{T15}: Astronomy; \textbf{T16}: Geography; \textbf{T17}: History; \textbf{T18}: Medicine; \textbf{T19}: Computer Science.
}
\endgroup
\end{table*}

\begin{table*}[!p]
\caption{Fine-grained task accuracy (\%) on RISEBench++ (EN). Part II: Causal reasoning and Spatial Relation.}
\label{tab:task_accuracy_en_2}
\centering
\begingroup
\fontsize{6.2}{8.2}\selectfont
\setlength{\tabcolsep}{1.2pt}
\renewcommand{\arraystretch}{1.0}
\setlength{\aboverulesep}{1pt}
\setlength{\belowrulesep}{1pt}
\colorlet{RISETaskGroup}{black!6}
\def\RISETaskYes{\textcolor{green!50!black}{\ding{51}}}
\def\RISETaskNo{\textcolor{red!65!black}{\ding{55}}}
\begin{tabular}{
>{\raggedright\arraybackslash}m{\dimexpr .19\textwidth-2\tabcolsep\relax}
>{\centering\arraybackslash}m{\dimexpr .04\textwidth-2\tabcolsep\relax}
*{14}{>{\centering\arraybackslash}m{\dimexpr (.77\textwidth-3pt)/23-2\tabcolsep\relax}}!{\hspace{3pt}}
*{9}{>{\centering\arraybackslash}m{\dimexpr (.77\textwidth-3pt)/23-2\tabcolsep\relax}}
}
\toprule[0.8pt]
\multirow{2}{=}[-1.3pt]{\raggedright\textbf{Model}} & \multirow{2}{=}[-1.3pt]{\centering\shortstack{\textbf{Multi-}\\\textbf{img}}} & \multicolumn{14}{c}{\textbf{Causal}} & \multicolumn{9}{c}{\textbf{Spatial (Relation)}} \\
\cmidrule(lr){3-16}\cmidrule(lr){17-25}
 & & \textbf{C01} & \textbf{C02} & \textbf{C03} & \textbf{C04} & \textbf{C05} & \textbf{C06} & \textbf{C07} & \textbf{C08} & \textbf{C09} & \textbf{C10} & \textbf{C11} & \textbf{C12} & \textbf{C13} & \textbf{C14} & \textbf{S01} & \textbf{S02} & \textbf{S03} & \textbf{S04} & \textbf{S05} & \textbf{S06} & \textbf{S07} & \textbf{S08} & \textbf{S09} \\
\midrule[0.5pt]
\rowcolor{RISETaskGroup}
\multicolumn{25}{c}{\rule{0pt}{9pt}\raisebox{1.5pt}[0pt][0pt]{\textbf{Closed-source Models}}} \\
\midrule[0.4pt]
\RISEFitModel{GPT-Image-2.5 Sunburst~\cite{openai2025gptimage}} & \RISETaskYes & 80.6 & \underline{85.2} & 68.2 & \underline{50.0} & \underline{60.0} & 30.0 & \textbf{80.0} & \underline{56.0} & 69.6 & \RISEFitScore{\textbf{100.0}} & \RISEFitScore{\textbf{100.0}} & \underline{50.0} & \underline{66.7} & \underline{40.0} & 63.6 & \underline{72.7} & 45.5 & 58.3 & \textbf{76.9} & \textbf{80.0} & \underline{72.2} & \underline{44.4} & \underline{16.7} \\
\RISEFitModel{Nano Banana Pro~\cite{google2026nanobanana}} & \RISETaskYes & 74.2 & 74.1 & \textbf{86.4} & 40.0 & \textbf{80.0} & \underline{60.0} & \textbf{80.0} & \textbf{64.0} & \underline{73.9} & \underline{91.7} & \RISEFitScore{\textbf{100.0}} & \underline{50.0} & 33.3 & \underline{40.0} & \textbf{81.8} & 45.5 & \underline{54.5} & 58.3 & \underline{53.8} & 46.7 & 55.6 & 33.3 & 0.0 \\
\RISEFitModel{Nano Banana 2~\cite{google2026nanobanana}} & \RISETaskYes & 71.0 & 74.1 & \textbf{86.4} & 40.0 & 40.0 & \textbf{70.0} & 50.0 & \textbf{64.0} & \underline{73.9} & \underline{91.7} & \underline{66.7} & \underline{50.0} & 0.0 & \textbf{50.0} & 54.5 & 54.5 & 36.4 & 41.7 & 46.2 & \underline{60.0} & 55.6 & 33.3 & \underline{16.7} \\
\RISEFitModel{GPT-Image-2~\cite{openai2025gptimage}} & \RISETaskYes & 71.0 & 77.8 & \underline{81.8} & \textbf{60.0} & \underline{60.0} & \underline{60.0} & 60.0 & 36.0 & 69.6 & 83.3 & \RISEFitScore{\textbf{100.0}} & 0.0 & \RISEFitScore{\textbf{100.0}} & 30.0 & \underline{72.7} & 54.5 & \underline{54.5} & \underline{66.7} & \textbf{76.9} & \underline{60.0} & 61.1 & \textbf{55.6} & \underline{16.7} \\
\RISEFitModel{GPT-Image-2.5 Flare~\cite{openai2025gptimage}} & \RISETaskYes & 67.7 & \textbf{88.9} & 72.7 & 40.0 & 50.0 & 50.0 & \underline{70.0} & 44.0 & 65.2 & 83.3 & \underline{66.7} & \underline{50.0} & 33.3 & 30.0 & 63.6 & \textbf{90.9} & 36.4 & 58.3 & 38.5 & 40.0 & \textbf{83.3} & \textbf{55.6} & \underline{16.7} \\
\RISEFitModel{Luma Uni-1~\cite{luma2026uni1}} & \RISETaskYes & 77.4 & 66.7 & 59.1 & 30.0 & 40.0 & 40.0 & 60.0 & 44.0 & 60.9 & 83.3 & \underline{66.7} & 0.0 & \RISEFitScore{\textbf{100.0}} & 20.0 & 63.6 & 45.5 & 27.3 & \textbf{75.0} & 38.5 & 53.3 & 55.6 & \textbf{55.6} & \underline{16.7} \\
\RISEFitModel{Qwen-Image-3.0~\cite{zhao2026qwen}} & \RISETaskYes & 51.6 & 59.3 & 40.9 & \textbf{60.0} & 40.0 & 50.0 & 50.0 & 44.0 & 34.8 & 75.0 & 33.3 & 25.0 & \underline{66.7} & 20.0 & \textbf{81.8} & 45.5 & 18.2 & 50.0 & 38.5 & \underline{60.0} & 50.0 & 11.1 & \underline{16.7} \\
\RISEFitModel{Nano Banana 2 Lite~\cite{google2026nanobanana}} & \RISETaskYes & 77.4 & 81.5 & 72.7 & 40.0 & 50.0 & 40.0 & 50.0 & 44.0 & \underline{73.9} & \underline{91.7} & 33.3 & 25.0 & 0.0 & 20.0 & 54.5 & 63.6 & 36.4 & 41.7 & \underline{53.8} & \underline{60.0} & 33.3 & 33.3 & 0.0 \\
\RISEFitModel{FLUX.2 Max~\cite{bfl2025flux2}} & \RISETaskYes & 58.1 & 74.1 & 72.7 & 30.0 & 50.0 & 30.0 & 50.0 & 40.0 & \textbf{82.6} & 66.7 & \RISEFitScore{\textbf{100.0}} & \underline{50.0} & 0.0 & 20.0 & 27.3 & 36.4 & 27.3 & 41.7 & 23.1 & 33.3 & 50.0 & 33.3 & 0.0 \\
\RISEFitModel{GPT-Image-1.5~\cite{openai2025gptimage}} & \RISETaskYes & \textbf{87.1} & 77.8 & 72.7 & \underline{50.0} & 30.0 & \underline{60.0} & \underline{70.0} & 36.0 & 60.9 & 83.3 & \RISEFitScore{\textbf{100.0}} & \underline{50.0} & 0.0 & 20.0 & 9.1 & 0.0 & 36.4 & 50.0 & 30.8 & 53.3 & 22.2 & \underline{44.4} & 0.0 \\
\RISEFitModel{FLUX.2 Pro~\cite{bfl2025flux2}} & \RISETaskYes & 61.3 & 51.9 & 63.6 & 40.0 & 20.0 & 30.0 & 30.0 & 32.0 & 69.6 & 58.3 & \RISEFitScore{\textbf{100.0}} & \underline{50.0} & 0.0 & 20.0 & 27.3 & 18.2 & 36.4 & 25.0 & 23.1 & 40.0 & 16.7 & 11.1 & 0.0 \\
\RISEFitModel{Nano Banana~\cite{google2026nanobanana}} & \RISETaskYes & 74.2 & 66.7 & 68.2 & 10.0 & 40.0 & 20.0 & 30.0 & 32.0 & 65.2 & 83.3 & \underline{66.7} & \textbf{75.0} & 0.0 & 10.0 & 63.6 & 27.3 & \underline{54.5} & 50.0 & 30.8 & 26.7 & 27.8 & 11.1 & \textbf{33.3} \\
\RISEFitModel{HunyuanImage-3.5~\cite{cao2025hunyuanimage}} & \RISETaskYes & 51.6 & 70.4 & 54.5 & 20.0 & 50.0 & 10.0 & 50.0 & 24.0 & 56.5 & 66.7 & \underline{66.7} & \underline{50.0} & 33.3 & 0.0 & 63.6 & 18.2 & \textbf{63.6} & 25.0 & 30.8 & 46.7 & 38.9 & 22.2 & \underline{16.7} \\
\RISEFitModel{Wan2.7-Image-Pro~\cite{mao2026wan}} & \RISETaskYes & 58.1 & 74.1 & \underline{81.8} & 20.0 & 40.0 & 50.0 & 60.0 & 24.0 & 47.8 & 75.0 & \underline{66.7} & \underline{50.0} & 0.0 & 10.0 & 63.6 & 18.2 & 9.1 & 33.3 & 46.2 & 40.0 & 33.3 & 33.3 & 0.0 \\
\RISEFitModel{GPT-Image-1~\cite{openai2025gptimage}} & \RISETaskYes & \underline{83.9} & 63.0 & 68.2 & 0.0 & 30.0 & 30.0 & 50.0 & 24.0 & 39.1 & 66.7 & \RISEFitScore{\textbf{100.0}} & \underline{50.0} & 0.0 & 10.0 & 18.2 & 18.2 & 18.2 & 41.7 & 30.8 & 26.7 & 22.2 & 11.1 & \textbf{33.3} \\
\RISEFitModel{Qwen-Image-2.0~\cite{zhao2026qwen}} & \RISETaskYes & 61.3 & 63.0 & 50.0 & 20.0 & 50.0 & 10.0 & 30.0 & 24.0 & \underline{73.9} & 75.0 & 0.0 & \underline{50.0} & 0.0 & 20.0 & 45.5 & 45.5 & 45.5 & 58.3 & 30.8 & 33.3 & 44.4 & 11.1 & 0.0 \\
\RISEFitModel{Seedream 4.5~\cite{seedream2025seedream}} & \RISETaskYes & 74.2 & 59.3 & 50.0 & 30.0 & 10.0 & 10.0 & 20.0 & 16.0 & 47.8 & 58.3 & \underline{66.7} & 25.0 & 0.0 & 20.0 & 45.5 & 9.1 & 27.3 & 25.0 & 23.1 & 26.7 & 22.2 & 33.3 & 0.0 \\
\RISEFitModel{Seedream 5.0~\cite{seedream2025seedream}} & \RISETaskYes & 58.1 & 63.0 & 31.8 & 20.0 & 50.0 & 20.0 & 30.0 & 20.0 & 56.5 & 75.0 & 33.3 & \underline{50.0} & 0.0 & 20.0 & 45.5 & 9.1 & 45.5 & 33.3 & 15.4 & 6.7 & 33.3 & 11.1 & 0.0 \\
\RISEFitModel{Step Image Edit 2~\cite{liu2025step1x}} & \RISETaskNo & 41.9 & 14.8 & 38.1 & 10.0 & 25.0 & 10.0 & 50.0 & 8.3 & 15.8 & 33.3 & 0.0 & 25.0 & 0.0 & 0.0 & 36.4 & 36.4 & 45.5 & 25.0 & 23.1 & 6.7 & 33.3 & 11.1 & \textemdash \\
\midrule[0.4pt]
\rowcolor{RISETaskGroup}
\multicolumn{25}{c}{\rule{0pt}{9pt}\raisebox{1.5pt}[0pt][0pt]{\textbf{Open-source Models}}} \\
\midrule[0.4pt]
\RISEFitModel{HunyuanImage-3.0-Instruct~\cite{cao2025hunyuanimage}} & \RISETaskYes & \underline{45.2} & \textbf{59.3} & \textbf{45.5} & \textbf{30.0} & 0.0 & 0.0 & \underline{50.0} & \textbf{32.0} & \textbf{39.1} & \textbf{66.7} & \underline{0.0} & \underline{25.0} & \textbf{0.0} & \underline{10.0} & \textbf{27.3} & 27.3 & \underline{27.3} & \textbf{50.0} & \underline{30.8} & \textbf{13.3} & \underline{22.2} & \underline{22.2} & \textbf{16.7} \\
\RISEFitModel{Qwen-Image-Edit-2511~\cite{wu2025qwen}} & \RISETaskYes & \underline{45.2} & \underline{48.1} & 18.2 & \underline{20.0} & 0.0 & 0.0 & \textbf{60.0} & 12.0 & 13.0 & 25.0 & \underline{0.0} & \underline{25.0} & \textbf{0.0} & 0.0 & 9.1 & 27.3 & \underline{27.3} & 16.7 & \underline{30.8} & 0.0 & 16.7 & \textbf{33.3} & \underline{0.0} \\
\RISEFitModel{Qwen-Image-2.1~\cite{zhao2026qwen}} & \RISETaskYes & \textbf{48.4} & 37.0 & \underline{40.9} & 0.0 & \textbf{20.0} & 10.0 & 20.0 & 0.0 & 17.4 & \underline{41.7} & \textbf{33.3} & \underline{25.0} & \textbf{0.0} & \underline{10.0} & \underline{18.2} & \underline{36.4} & 18.2 & 25.0 & \textbf{38.5} & 0.0 & 5.6 & \underline{22.2} & \underline{0.0} \\
\RISEFitModel{FireRed-Image-Edit-1.0~\cite{team2026firered}} & \RISETaskYes & 29.0 & 25.9 & 22.7 & 10.0 & 0.0 & 10.0 & 40.0 & 0.0 & 13.0 & 16.7 & \underline{0.0} & \textbf{50.0} & \textbf{0.0} & \textbf{20.0} & \textbf{27.3} & \textbf{54.5} & \underline{27.3} & \underline{41.7} & 23.1 & 0.0 & 16.7 & \textbf{33.3} & \underline{0.0} \\
\RISEFitModel{FireRed-Image-Edit-1.1~\cite{team2026firered}} & \RISETaskYes & 25.8 & 37.0 & 36.4 & 0.0 & 0.0 & 10.0 & \underline{50.0} & 16.0 & 17.4 & 33.3 & \underline{0.0} & \textbf{50.0} & \textbf{0.0} & \underline{10.0} & \underline{18.2} & 27.3 & 18.2 & 33.3 & 23.1 & 0.0 & \textbf{27.8} & 0.0 & \underline{0.0} \\
\RISEFitModel{Qwen-Image-Edit-2509~\cite{wu2025qwen}} & \RISETaskYes & 38.7 & 40.7 & 31.8 & 10.0 & 0.0 & 0.0 & \underline{50.0} & 12.0 & 17.4 & 25.0 & \underline{0.0} & \textbf{50.0} & \textbf{0.0} & \underline{10.0} & 9.1 & \underline{36.4} & \textbf{36.4} & 25.0 & 15.4 & 0.0 & \underline{22.2} & 11.1 & \underline{0.0} \\
\RISEFitModel{HiDream-O1-Image~\cite{cai2026hidream}} & \RISETaskYes & 38.7 & 37.0 & 13.6 & 0.0 & 0.0 & 0.0 & 10.0 & 8.0 & 17.4 & 33.3 & \underline{0.0} & 0.0 & \textbf{0.0} & 0.0 & 9.1 & 27.3 & \underline{27.3} & 16.7 & 23.1 & \underline{6.7} & \underline{22.2} & 11.1 & \textbf{16.7} \\
\RISEFitModel{FLUX.2 Dev~\cite{bfl2025flux2}} & \RISETaskYes & \underline{45.2} & 33.3 & \underline{40.9} & 10.0 & 10.0 & 10.0 & \underline{50.0} & 12.0 & 21.7 & 25.0 & \textbf{33.3} & \textbf{50.0} & \textbf{0.0} & \underline{10.0} & 9.1 & 9.1 & 18.2 & 16.7 & 7.7 & 0.0 & 11.1 & 11.1 & \underline{0.0} \\
\RISEFitModel{FLUX.2 Klein 9B~\cite{bfl2025flux2}} & \RISETaskYes & 35.5 & 22.2 & 18.2 & 0.0 & 0.0 & 10.0 & \underline{50.0} & 0.0 & 8.7 & 8.3 & \underline{0.0} & \textbf{50.0} & \textbf{0.0} & \underline{10.0} & 9.1 & 27.3 & \textbf{36.4} & 25.0 & 23.1 & 0.0 & 5.6 & \underline{22.2} & \underline{0.0} \\
\RISEFitModel{JoyAI-Image-Edit-Plus~\cite{song2026joyai}} & \RISETaskYes & 32.3 & 25.9 & 27.3 & 10.0 & 10.0 & 0.0 & 20.0 & 4.0 & 8.7 & 16.7 & \underline{0.0} & 0.0 & \textbf{0.0} & 0.0 & 9.1 & 18.2 & 18.2 & 16.7 & 15.4 & \underline{6.7} & 16.7 & 11.1 & \underline{0.0} \\
\RISEFitModel{SenseNova-U1-A3B-MoT-SFT~\cite{diao2026sensenova}} & \RISETaskYes & \textbf{48.4} & 29.6 & \underline{40.9} & 0.0 & 0.0 & 10.0 & 10.0 & 12.0 & \textbf{39.1} & 33.3 & \textbf{33.3} & 0.0 & \textbf{0.0} & 0.0 & 0.0 & 0.0 & 0.0 & 0.0 & 0.0 & 0.0 & 0.0 & 0.0 & \underline{0.0} \\
\RISEFitModel{UniPic 3.0~\cite{wei2026skywork}} & \RISETaskYes & 38.7 & 18.5 & 9.1 & \underline{20.0} & 0.0 & 10.0 & 40.0 & 12.0 & 17.4 & 16.7 & \underline{0.0} & 0.0 & \textbf{0.0} & 0.0 & \underline{18.2} & 27.3 & 9.1 & 25.0 & 15.4 & 0.0 & \underline{22.2} & \underline{22.2} & \underline{0.0} \\
\RISEFitModel{BAGEL (w/ CoT)~\cite{deng2025bagel}} & \RISETaskYes & 19.4 & 18.5 & 27.3 & \underline{20.0} & 10.0 & 0.0 & 10.0 & 4.0 & 21.7 & \underline{41.7} & \underline{0.0} & 0.0 & \textbf{0.0} & 0.0 & 0.0 & 9.1 & \textbf{36.4} & 33.3 & 7.7 & 0.0 & 5.6 & 11.1 & \underline{0.0} \\
\RISEFitModel{SenseNova-U1-8B-MoT-SFT~\cite{diao2026sensenova}} & \RISETaskYes & 41.9 & 37.0 & 22.7 & 10.0 & 0.0 & 10.0 & 20.0 & \underline{20.0} & 13.0 & \underline{41.7} & \underline{0.0} & 0.0 & \textbf{0.0} & 0.0 & 0.0 & 18.2 & 18.2 & 0.0 & 0.0 & \underline{6.7} & 5.6 & 11.1 & \underline{0.0} \\
\RISEFitModel{FLUX.2 Klein 4B~\cite{bfl2025flux2}} & \RISETaskYes & 32.3 & 11.1 & 4.5 & 0.0 & 10.0 & \textbf{30.0} & 40.0 & 4.0 & 8.7 & 16.7 & \underline{0.0} & 0.0 & \textbf{0.0} & \textbf{20.0} & 0.0 & 9.1 & 9.1 & 25.0 & 15.4 & 0.0 & 5.6 & \textbf{33.3} & \underline{0.0} \\
\RISEFitModel{UniWorld-V2~\cite{li2025uniworld}} & \RISETaskYes & 29.0 & 18.5 & 27.3 & \underline{20.0} & 0.0 & \textbf{30.0} & 40.0 & 4.0 & 13.0 & 8.3 & \underline{0.0} & \textbf{50.0} & \textbf{0.0} & 0.0 & 9.1 & 9.1 & \underline{27.3} & 25.0 & 7.7 & 0.0 & 16.7 & \textbf{33.3} & \underline{0.0} \\
\RISEFitModel{BAGEL~\cite{deng2025bagel}} & \RISETaskYes & 16.1 & 7.4 & 9.1 & 0.0 & 10.0 & 0.0 & 0.0 & 4.0 & 0.0 & 8.3 & \underline{0.0} & 0.0 & \textbf{0.0} & 0.0 & 0.0 & 9.1 & 9.1 & 8.3 & 7.7 & \underline{6.7} & 0.0 & 11.1 & \underline{0.0} \\
\RISEFitModel{OmniGen2~\cite{wu2025omnigen2}} & \RISETaskYes & 9.7 & 3.7 & 4.5 & 0.0 & 10.0 & 10.0 & 10.0 & 4.0 & 4.3 & 0.0 & \underline{0.0} & 0.0 & \textbf{0.0} & \textbf{20.0} & 9.1 & 0.0 & 18.2 & 8.3 & 0.0 & 0.0 & 0.0 & 0.0 & \underline{0.0} \\
\RISEFitModel{OmniGen~\cite{xiao2024omnigen}} & \RISETaskYes & 6.5 & 0.0 & 0.0 & 0.0 & 0.0 & 10.0 & 10.0 & 8.0 & 4.3 & 0.0 & \underline{0.0} & 0.0 & \textbf{0.0} & \underline{10.0} & 0.0 & 0.0 & 9.1 & 0.0 & 0.0 & 0.0 & 0.0 & 0.0 & \underline{0.0} \\
\RISEFitModel{Emu2-Gen~\cite{sun2023emu2chat}} & \RISETaskYes & 6.5 & 3.7 & 0.0 & 0.0 & 0.0 & 0.0 & 0.0 & 0.0 & 0.0 & 0.0 & \underline{0.0} & 0.0 & \textbf{0.0} & 0.0 & 0.0 & 0.0 & 0.0 & 0.0 & 0.0 & 0.0 & 0.0 & 0.0 & \underline{0.0} \\
\RISEFitModel{JoyAI-Image-Edit~\cite{song2026joyai}} & \RISETaskNo & 41.9 & 37.0 & 38.1 & 0.0 & 0.0 & 10.0 & \underline{50.0} & 4.2 & 15.8 & 25.0 & \underline{0.0} & \underline{25.0} & \textbf{0.0} & \underline{10.0} & \underline{18.2} & 27.3 & 18.2 & 25.0 & \underline{30.8} & \textbf{13.3} & \underline{22.2} & 0.0 & \textemdash \\
\RISEFitModel{Step1X-Edit (thinking)~\cite{liu2025step1x}} & \RISETaskNo & 32.3 & 33.3 & 28.6 & \underline{20.0} & \underline{12.5} & \textbf{30.0} & 40.0 & 12.5 & \underline{31.6} & \underline{41.7} & \textbf{33.3} & \underline{25.0} & \textbf{0.0} & \underline{10.0} & 9.1 & 9.1 & 18.2 & 16.7 & 15.4 & 0.0 & 16.7 & 11.1 & \textemdash \\
\RISEFitModel{Boogu-Image-0.1-Edit~\cite{chen2026boogu}} & \RISETaskNo & \textbf{48.4} & 25.9 & 28.6 & \underline{20.0} & 0.0 & \underline{20.0} & 20.0 & 12.5 & 5.3 & 16.7 & \underline{0.0} & \underline{25.0} & \textbf{0.0} & \underline{10.0} & 9.1 & 27.3 & 18.2 & 33.3 & 23.1 & 0.0 & 11.1 & 0.0 & \textemdash \\
\RISEFitModel{Step1X-Edit (thinking + reflection)~\cite{liu2025step1x}} & \RISETaskNo & 41.9 & 29.6 & 28.6 & 0.0 & \underline{12.5} & 10.0 & 40.0 & 12.5 & \underline{31.6} & 0.0 & \textbf{33.3} & 0.0 & \textbf{0.0} & \underline{10.0} & \underline{18.2} & 9.1 & \underline{27.3} & 16.7 & 7.7 & \underline{6.7} & 16.7 & 0.0 & \textemdash \\
\RISEFitModel{LongCat-Image-Edit~\cite{team2025longcat}} & \RISETaskNo & \underline{45.2} & 22.2 & 19.0 & 10.0 & 0.0 & \underline{20.0} & 40.0 & 4.2 & 21.1 & 8.3 & \underline{0.0} & \underline{25.0} & \textbf{0.0} & \textbf{20.0} & 9.1 & 27.3 & 9.1 & 25.0 & 7.7 & \underline{6.7} & 16.7 & \underline{22.2} & \textemdash \\
\RISEFitModel{Step1X-Edit (base)~\cite{liu2025step1x}} & \RISETaskNo & 38.7 & 18.5 & 23.8 & \textbf{30.0} & 0.0 & 10.0 & 30.0 & 4.2 & 15.8 & 16.7 & \underline{0.0} & 0.0 & \textbf{0.0} & \underline{10.0} & 0.0 & 9.1 & 18.2 & 25.0 & 7.7 & 0.0 & 5.6 & 0.0 & \textemdash \\
\RISEFitModel{InternVL-U (w/ CoT)~\cite{tian2026internvl}} & \RISETaskNo & 25.8 & 18.5 & 9.5 & 0.0 & 0.0 & 10.0 & 20.0 & 8.3 & 26.3 & 16.7 & \underline{0.0} & 0.0 & \textbf{0.0} & 0.0 & 9.1 & 0.0 & 9.1 & 0.0 & 0.0 & 0.0 & 11.1 & 0.0 & \textemdash \\
\RISEFitModel{FLUX.1 Kontext Dev~\cite{labs2025flux1kontextflowmatching}} & \RISETaskNo & 25.8 & 7.4 & 0.0 & 0.0 & 0.0 & 10.0 & 30.0 & 0.0 & 5.3 & 8.3 & \underline{0.0} & 0.0 & \textbf{0.0} & \textbf{20.0} & 0.0 & 0.0 & \underline{27.3} & 0.0 & 15.4 & 0.0 & 11.1 & 0.0 & \textemdash \\
\RISEFitModel{LLaDA-Image~\cite{chen2026llada}} & \RISETaskNo & 12.9 & 0.0 & 4.8 & 0.0 & 0.0 & 0.0 & 0.0 & 4.2 & 5.3 & 16.7 & \underline{0.0} & 0.0 & \textbf{0.0} & 0.0 & 0.0 & 0.0 & 0.0 & 0.0 & 0.0 & 0.0 & 0.0 & 0.0 & \textemdash \\
\RISEFitModel{Ovis-U1~\cite{wang2025ovis}} & \RISETaskNo & 16.1 & 7.4 & 0.0 & 10.0 & 0.0 & 10.0 & 0.0 & 0.0 & 5.3 & 16.7 & \underline{0.0} & 0.0 & \textbf{0.0} & 0.0 & 0.0 & 0.0 & 0.0 & 0.0 & 0.0 & 0.0 & 0.0 & 11.1 & \textemdash \\
\RISEFitModel{InternVL-U~\cite{tian2026internvl}} & \RISETaskNo & 6.5 & 3.7 & 0.0 & 0.0 & 0.0 & 0.0 & 0.0 & 0.0 & 5.3 & 8.3 & \underline{0.0} & 0.0 & \textbf{0.0} & 0.0 & 0.0 & 0.0 & 0.0 & 16.7 & 0.0 & 0.0 & 0.0 & 0.0 & \textemdash \\
\RISEFitModel{UniPic2-SD3.5M-Kontext-2B~\cite{wei2025skywork}} & \RISETaskNo & 3.2 & 0.0 & 0.0 & 0.0 & 0.0 & 0.0 & 0.0 & 0.0 & 5.3 & 0.0 & \underline{0.0} & 0.0 & \textbf{0.0} & \textbf{20.0} & 0.0 & 0.0 & 0.0 & 0.0 & 0.0 & 0.0 & 0.0 & 0.0 & \textemdash \\
\RISEFitModel{Lumina-DiMOO~\cite{xin2025lumina}} & \RISETaskNo & 0.0 & 3.7 & 4.8 & 0.0 & 0.0 & 0.0 & 0.0 & 0.0 & 5.3 & 0.0 & \underline{0.0} & 0.0 & \textbf{0.0} & 0.0 & 0.0 & 0.0 & 0.0 & 0.0 & 0.0 & 0.0 & 0.0 & 0.0 & \textemdash \\
\RISEFitModel{UniWorld-V1~\cite{lin2025uniworld}} & \RISETaskNo & 3.2 & 3.7 & 0.0 & 0.0 & 0.0 & 0.0 & 0.0 & 0.0 & 0.0 & 0.0 & \underline{0.0} & 0.0 & \textbf{0.0} & 0.0 & 0.0 & 0.0 & 0.0 & 0.0 & 0.0 & 0.0 & 0.0 & 0.0 & \textemdash \\
\midrule[0.4pt]
\rowcolor{RISETaskGroup}
\multicolumn{25}{c}{\rule{0pt}{9pt}\raisebox{1.5pt}[0pt][0pt]{\textbf{Agentic Methods}}} \\
\midrule[0.4pt]
\RISEFitModel{\textbf{RISE-Agent (Ours)}} & \RISETaskYes & \textbf{87.1} & \textbf{74.1} & \textbf{72.7} & \underline{20.0} & \textbf{40.0} & \textbf{60.0} & \textbf{40.0} & \textbf{28.0} & \textbf{65.2} & \textbf{83.3} & \underline{33.3} & \textbf{50.0} & \textbf{0.0} & \textbf{20.0} & \underline{27.3} & \textbf{63.6} & \textbf{18.2} & \textbf{33.3} & \textbf{38.5} & \textbf{40.0} & \textbf{50.0} & \underline{33.3} & \textbf{0.0} \\
\RISEFitModel{MindBrush~\cite{he2026mind}} & \RISETaskNo & \underline{54.8} & \underline{48.1} & \underline{33.3} & 10.0 & 12.5 & 0.0 & 10.0 & \underline{20.8} & \underline{57.9} & \underline{41.7} & \underline{33.3} & \underline{25.0} & \textbf{0.0} & \underline{10.0} & \textbf{36.4} & \underline{18.2} & \textbf{18.2} & \textbf{33.3} & \underline{15.4} & 6.7 & 0.0 & 0.0 & \textemdash \\
\RISEFitModel{InterleaveThinker~\cite{zheng2026interleavethinker}} & \RISETaskNo & 38.7 & 33.3 & \underline{33.3} & \textbf{50.0} & \underline{37.5} & \underline{40.0} & \underline{30.0} & 16.7 & 26.3 & 25.0 & \textbf{66.7} & \textbf{50.0} & \textbf{0.0} & 0.0 & 18.2 & 9.1 & \underline{0.0} & \textbf{33.3} & 0.0 & \underline{13.3} & \underline{22.2} & \textbf{44.4} & \textemdash \\
\RISEFitModel{IntentEdit~\cite{zhang2026intentedit}} & \RISETaskNo & 38.7 & 25.9 & 19.0 & 0.0 & 12.5 & 10.0 & 20.0 & \underline{20.8} & 47.4 & 33.3 & \underline{33.3} & 0.0 & \textbf{0.0} & 0.0 & 9.1 & 0.0 & \textbf{18.2} & \underline{8.3} & 0.0 & 6.7 & 11.1 & 0.0 & \textemdash \\
\RISEFitModel{ImAgent~\cite{wang2025imagent}} & \RISETaskNo & 9.7 & 7.4 & 9.5 & 10.0 & 0.0 & 0.0 & 10.0 & 0.0 & 0.0 & 0.0 & 0.0 & 0.0 & \textbf{0.0} & 0.0 & 0.0 & 0.0 & \underline{0.0} & 0.0 & 7.7 & 0.0 & 5.6 & 0.0 & \textemdash \\
\bottomrule[0.8pt]
\end{tabular}
\par\vspace{4pt}
\parbox{\textwidth}{\fontsize{6.5}{8.2}\selectfont\raggedright
\textbf{C01}: Physical State Transition; \textbf{C02}: Structural Damage; \textbf{C03}: Flexible Deformation; \textbf{C04}: Magnitude Change; \textbf{C05}: Reading Offset; \textbf{C06}: Order Relation; \textbf{C07}: Feature Mapping; \textbf{C08}: Physics; \textbf{C09}: Chemistry; \textbf{C10}: Biology; \textbf{C11}: Astronomy; \textbf{C12}: Medicine; \textbf{C13}: Music Theory; \textbf{C14}: Computer Science.\par
\textbf{S01}: Adjacency; \textbf{S02}: Direction; \textbf{S03}: Orientation; \textbf{S04}: Projection; \textbf{S05}: Distance; \textbf{S06}: Size; \textbf{S07}: Configuration; \textbf{S08}: Viewing Scale; \textbf{S09}: Localization.
}
\endgroup
\end{table*}

\begin{table*}[!p]
\caption{Fine-grained task accuracy (\%) on RISEBench++ (EN). Part III: Spatial Structure, Logical reasoning, and Counterfactual reasoning.}
\label{tab:task_accuracy_en_3}
\centering
\begingroup
\fontsize{6.2}{8.2}\selectfont
\setlength{\tabcolsep}{1.2pt}
\renewcommand{\arraystretch}{1.0}
\setlength{\aboverulesep}{1pt}
\setlength{\belowrulesep}{1pt}
\colorlet{RISETaskGroup}{black!6}
\def\RISETaskYes{\textcolor{green!50!black}{\ding{51}}}
\def\RISETaskNo{\textcolor{red!65!black}{\ding{55}}}
\begin{tabular}{
>{\raggedright\arraybackslash}m{\dimexpr .19\textwidth-2\tabcolsep\relax}
>{\centering\arraybackslash}m{\dimexpr .04\textwidth-2\tabcolsep\relax}
*{5}{>{\centering\arraybackslash}m{\dimexpr (.77\textwidth-6pt)/23-2\tabcolsep\relax}}!{\hspace{3pt}}
*{7}{>{\centering\arraybackslash}m{\dimexpr (.77\textwidth-6pt)/23-2\tabcolsep\relax}}!{\hspace{3pt}}
*{11}{>{\centering\arraybackslash}m{\dimexpr (.77\textwidth-6pt)/23-2\tabcolsep\relax}}
}
\toprule[0.8pt]
\multirow{2}{=}[-1.3pt]{\raggedright\textbf{Model}} & \multirow{2}{=}[-1.3pt]{\centering\shortstack{\textbf{Multi-}\\\textbf{img}}} & \multicolumn{5}{c}{\textbf{Spatial (Structure)}} & \multicolumn{7}{c}{\textbf{Logical}} & \multicolumn{11}{c}{\textbf{Counterfactual}} \\
\cmidrule(lr){3-7}\cmidrule(lr){8-14}\cmidrule(lr){15-25}
 & & \textbf{S10} & \textbf{S11} & \textbf{S12} & \textbf{S13} & \textbf{S14} & \textbf{L01} & \textbf{L02} & \textbf{L03} & \textbf{L04} & \textbf{L05} & \textbf{L06} & \textbf{L07} & \textbf{F01} & \textbf{F02} & \textbf{F03} & \textbf{F04} & \textbf{F05} & \textbf{F06} & \textbf{F07} & \textbf{F08} & \textbf{F09} & \textbf{F10} & \textbf{F11} \\
\midrule[0.5pt]
\rowcolor{RISETaskGroup}
\multicolumn{25}{c}{\rule{0pt}{9pt}\raisebox{1.5pt}[0pt][0pt]{\textbf{Closed-source Models}}} \\
\midrule[0.4pt]
\RISEFitModel{GPT-Image-2.5 Sunburst~\cite{openai2025gptimage}} & \RISETaskYes & \textbf{79.2} & \textbf{73.3} & \underline{65.4} & 52.6 & 40.0 & 25.0 & 29.6 & \underline{40.5} & \underline{31.4} & 12.8 & \underline{33.3} & 15.0 & \textbf{87.0} & \underline{66.7} & \underline{87.5} & 42.9 & \textbf{75.0} & 57.1 & \textbf{50.0} & 50.0 & \underline{90.0} & 63.6 & \textbf{66.7} \\
\RISEFitModel{Nano Banana Pro~\cite{google2026nanobanana}} & \RISETaskYes & 58.3 & \underline{66.7} & 50.0 & 36.8 & 50.0 & \underline{29.2} & \textbf{66.7} & \textbf{45.9} & 25.7 & \textbf{25.6} & 27.8 & \underline{30.0} & \underline{78.3} & 55.6 & \underline{87.5} & 42.9 & 12.5 & \textbf{85.7} & \underline{25.0} & 50.0 & 80.0 & \textbf{81.8} & \underline{55.6} \\
\RISEFitModel{Nano Banana 2~\cite{google2026nanobanana}} & \RISETaskYes & 41.7 & 53.3 & 34.6 & 42.1 & 50.0 & \textbf{33.3} & 51.9 & \textbf{45.9} & 22.9 & 12.8 & 27.8 & 0.0 & 65.2 & 55.6 & \RISEFitScore{\textbf{100.0}} & 28.6 & 12.5 & \underline{71.4} & \textbf{50.0} & \underline{75.0} & 80.0 & \textbf{81.8} & 44.4 \\
\RISEFitModel{GPT-Image-2~\cite{openai2025gptimage}} & \RISETaskYes & \textbf{79.2} & \textbf{73.3} & 50.0 & \underline{57.9} & 30.0 & 16.7 & 22.2 & 27.0 & 17.1 & 10.3 & 0.0 & 0.0 & 52.2 & 55.6 & \underline{87.5} & 42.9 & 50.0 & 42.9 & \textbf{50.0} & 25.0 & \RISEFitScore{\textbf{100.0}} & \underline{72.7} & 33.3 \\
\RISEFitModel{GPT-Image-2.5 Flare~\cite{openai2025gptimage}} & \RISETaskYes & \underline{75.0} & 46.7 & \textbf{69.2} & 47.4 & 10.0 & 8.3 & 14.8 & 24.3 & 17.1 & 2.6 & 0.0 & 0.0 & 65.2 & 33.3 & \underline{87.5} & 28.6 & 50.0 & 42.9 & \underline{25.0} & 50.0 & 80.0 & 63.6 & \underline{55.6} \\
\RISEFitModel{Luma Uni-1~\cite{luma2026uni1}} & \RISETaskYes & 50.0 & 60.0 & 42.3 & 42.1 & \underline{80.0} & 0.0 & 29.6 & 29.7 & 20.0 & 12.8 & 16.7 & 10.0 & 52.2 & \textbf{77.8} & \underline{87.5} & 28.6 & 50.0 & \underline{71.4} & \underline{25.0} & 25.0 & 60.0 & 54.5 & 44.4 \\
\RISEFitModel{Qwen-Image-3.0~\cite{zhao2026qwen}} & \RISETaskYes & 41.7 & 60.0 & 34.6 & 36.8 & 0.0 & 16.7 & \underline{55.6} & \underline{40.5} & \textbf{40.0} & \underline{15.4} & \textbf{38.9} & \textbf{55.0} & 47.8 & 22.2 & 25.0 & 28.6 & 50.0 & \underline{71.4} & \underline{25.0} & \RISEFitScore{\textbf{100.0}} & 70.0 & 54.5 & \underline{55.6} \\
\RISEFitModel{Nano Banana 2 Lite~\cite{google2026nanobanana}} & \RISETaskYes & 54.2 & 60.0 & 50.0 & 36.8 & 40.0 & 16.7 & 22.2 & 21.6 & 14.3 & 7.7 & 16.7 & 5.0 & 73.9 & \textbf{77.8} & 62.5 & \underline{57.1} & 12.5 & \textbf{85.7} & \underline{25.0} & 50.0 & \underline{90.0} & 36.4 & \textbf{66.7} \\
\RISEFitModel{FLUX.2 Max~\cite{bfl2025flux2}} & \RISETaskYes & \underline{75.0} & \underline{66.7} & 34.6 & 21.1 & \textbf{90.0} & 4.2 & 33.3 & 18.9 & 14.3 & 5.1 & 5.6 & 0.0 & 69.6 & \textbf{77.8} & 62.5 & \textbf{71.4} & 37.5 & \underline{71.4} & 0.0 & \underline{75.0} & 60.0 & 63.6 & 44.4 \\
\RISEFitModel{GPT-Image-1.5~\cite{openai2025gptimage}} & \RISETaskYes & 66.7 & 46.7 & 42.3 & \textbf{63.2} & 20.0 & 4.2 & 3.7 & 18.9 & 5.7 & 7.7 & 5.6 & 5.0 & 60.9 & 55.6 & \underline{87.5} & \underline{57.1} & \textbf{75.0} & \underline{71.4} & 0.0 & 50.0 & 80.0 & 36.4 & 44.4 \\
\RISEFitModel{FLUX.2 Pro~\cite{bfl2025flux2}} & \RISETaskYes & 62.5 & 53.3 & 38.5 & 36.8 & \underline{80.0} & 4.2 & 29.6 & 18.9 & 5.7 & 5.1 & 5.6 & 0.0 & \underline{78.3} & \textbf{77.8} & 75.0 & 42.9 & 37.5 & 57.1 & \underline{25.0} & \underline{75.0} & 60.0 & 45.5 & \underline{55.6} \\
\RISEFitModel{Nano Banana~\cite{google2026nanobanana}} & \RISETaskYes & 58.3 & 46.7 & 42.3 & 31.6 & 30.0 & 0.0 & 22.2 & 13.5 & 20.0 & 7.7 & 0.0 & 5.0 & 60.9 & 33.3 & 75.0 & 42.9 & 12.5 & 42.9 & 0.0 & 50.0 & 80.0 & 36.4 & 33.3 \\
\RISEFitModel{HunyuanImage-3.5~\cite{cao2025hunyuanimage}} & \RISETaskYes & \underline{75.0} & 46.7 & 30.8 & 31.6 & \underline{80.0} & 8.3 & 22.2 & 5.4 & 22.9 & 10.3 & \underline{33.3} & 0.0 & 43.5 & 44.4 & 37.5 & 28.6 & 25.0 & 0.0 & 0.0 & 25.0 & 70.0 & 36.4 & 44.4 \\
\RISEFitModel{Wan2.7-Image-Pro~\cite{mao2026wan}} & \RISETaskYes & 50.0 & 40.0 & 26.9 & 26.3 & 0.0 & 8.3 & 11.1 & 21.6 & 8.6 & 2.6 & 22.2 & 0.0 & 60.9 & 33.3 & 75.0 & 42.9 & \underline{62.5} & 0.0 & 0.0 & 0.0 & 80.0 & 36.4 & \textbf{66.7} \\
\RISEFitModel{GPT-Image-1~\cite{openai2025gptimage}} & \RISETaskYes & 45.8 & 40.0 & 30.8 & 31.6 & 20.0 & 0.0 & 7.4 & 10.8 & 8.6 & 5.1 & 0.0 & 0.0 & 47.8 & 55.6 & 50.0 & 42.9 & 50.0 & 28.6 & 0.0 & 0.0 & 50.0 & 45.5 & 22.2 \\
\RISEFitModel{Qwen-Image-2.0~\cite{zhao2026qwen}} & \RISETaskYes & 41.7 & 53.3 & 23.1 & 10.5 & 60.0 & 0.0 & 7.4 & 8.1 & 11.4 & 5.1 & 11.1 & 0.0 & 39.1 & 22.2 & 50.0 & 28.6 & 12.5 & 14.3 & \underline{25.0} & 25.0 & 70.0 & 45.5 & 33.3 \\
\RISEFitModel{Seedream 4.5~\cite{seedream2025seedream}} & \RISETaskYes & 33.3 & 40.0 & 23.1 & 21.1 & 30.0 & 0.0 & 25.9 & 8.1 & 8.6 & 2.6 & 5.6 & 0.0 & 43.5 & 22.2 & 62.5 & 0.0 & 25.0 & 0.0 & 0.0 & 25.0 & 60.0 & 36.4 & \textbf{66.7} \\
\RISEFitModel{Seedream 5.0~\cite{seedream2025seedream}} & \RISETaskYes & 33.3 & 40.0 & 34.6 & 15.8 & 10.0 & 4.2 & 11.1 & 13.5 & 0.0 & 5.1 & 0.0 & 0.0 & 34.8 & 22.2 & 50.0 & 0.0 & 12.5 & 14.3 & 0.0 & 25.0 & 50.0 & 63.6 & 11.1 \\
\RISEFitModel{Step Image Edit 2~\cite{liu2025step1x}} & \RISETaskNo & 12.5 & 26.7 & 30.8 & 35.3 & \textemdash & 0.0 & 0.0 & 0.0 & 2.9 & 5.1 & 0.0 & \textemdash & 17.4 & 0.0 & 0.0 & 0.0 & 0.0 & 0.0 & 0.0 & 0.0 & 20.0 & 9.1 & 33.3 \\
\midrule[0.4pt]
\rowcolor{RISETaskGroup}
\multicolumn{25}{c}{\rule{0pt}{9pt}\raisebox{1.5pt}[0pt][0pt]{\textbf{Open-source Models}}} \\
\midrule[0.4pt]
\RISEFitModel{HunyuanImage-3.0-Instruct~\cite{cao2025hunyuanimage}} & \RISETaskYes & \textbf{33.3} & 26.7 & 26.9 & \textbf{36.8} & 50.0 & \textbf{12.5} & \underline{0.0} & \textbf{13.5} & \underline{11.4} & 5.1 & 0.0 & 0.0 & 21.7 & 22.2 & \underline{12.5} & 0.0 & 0.0 & \underline{14.3} & \textbf{0.0} & \underline{0.0} & \textbf{70.0} & 18.2 & 33.3 \\
\RISEFitModel{Qwen-Image-Edit-2511~\cite{wu2025qwen}} & \RISETaskYes & 16.7 & \textbf{53.3} & \textbf{38.5} & 15.8 & 30.0 & 4.2 & \underline{0.0} & \underline{2.7} & 5.7 & \underline{12.8} & 0.0 & 0.0 & 13.0 & 11.1 & \underline{12.5} & 0.0 & 0.0 & 0.0 & \textbf{0.0} & \underline{0.0} & 40.0 & \underline{27.3} & \textbf{55.6} \\
\RISEFitModel{Qwen-Image-2.1~\cite{zhao2026qwen}} & \RISETaskYes & 16.7 & \underline{46.7} & \underline{30.8} & \underline{31.6} & 50.0 & 0.0 & \underline{0.0} & 0.0 & 8.6 & 0.0 & 0.0 & \underline{5.0} & 21.7 & 22.2 & 0.0 & 0.0 & \underline{12.5} & 0.0 & \textbf{0.0} & \underline{0.0} & 30.0 & 9.1 & \underline{44.4} \\
\RISEFitModel{FireRed-Image-Edit-1.0~\cite{team2026firered}} & \RISETaskYes & \underline{25.0} & \underline{46.7} & \textbf{38.5} & 10.5 & 50.0 & 4.2 & \underline{0.0} & 0.0 & 2.9 & 5.1 & 0.0 & 0.0 & 21.7 & 22.2 & \underline{12.5} & \underline{14.3} & 0.0 & \underline{14.3} & \textbf{0.0} & \underline{0.0} & 40.0 & \underline{27.3} & \underline{44.4} \\
\RISEFitModel{FireRed-Image-Edit-1.1~\cite{team2026firered}} & \RISETaskYes & 20.8 & \underline{46.7} & 26.9 & 10.5 & 50.0 & 4.2 & \underline{0.0} & 0.0 & 2.9 & 2.6 & 0.0 & 0.0 & 13.0 & 11.1 & 0.0 & \underline{14.3} & 0.0 & 0.0 & \textbf{0.0} & \underline{0.0} & 40.0 & \textbf{36.4} & 33.3 \\
\RISEFitModel{Qwen-Image-Edit-2509~\cite{wu2025qwen}} & \RISETaskYes & 12.5 & \underline{46.7} & \underline{30.8} & 15.8 & 0.0 & 0.0 & \underline{0.0} & 0.0 & 2.9 & 0.0 & 0.0 & 0.0 & \underline{30.4} & 11.1 & 0.0 & \underline{14.3} & 0.0 & 0.0 & \textbf{0.0} & \underline{0.0} & 30.0 & \underline{27.3} & 11.1 \\
\RISEFitModel{HiDream-O1-Image~\cite{cai2026hidream}} & \RISETaskYes & \textbf{33.3} & 40.0 & 11.5 & 15.8 & 20.0 & 4.2 & \underline{0.0} & \underline{2.7} & 0.0 & 10.3 & 0.0 & 0.0 & 4.3 & 0.0 & 0.0 & 0.0 & 0.0 & 0.0 & \textbf{0.0} & \underline{0.0} & 20.0 & 9.1 & 11.1 \\
\RISEFitModel{FLUX.2 Dev~\cite{bfl2025flux2}} & \RISETaskYes & \underline{25.0} & 20.0 & 15.4 & 21.1 & \textbf{70.0} & 0.0 & \underline{0.0} & 0.0 & 2.9 & 2.6 & 0.0 & 0.0 & 0.0 & 22.2 & 0.0 & 0.0 & 0.0 & 0.0 & \textbf{0.0} & \underline{0.0} & 40.0 & \underline{27.3} & 11.1 \\
\RISEFitModel{FLUX.2 Klein 9B~\cite{bfl2025flux2}} & \RISETaskYes & \underline{25.0} & 33.3 & 23.1 & 15.8 & \textbf{70.0} & 0.0 & \underline{0.0} & 0.0 & 0.0 & 2.6 & 0.0 & 0.0 & 17.4 & \textbf{55.6} & 0.0 & \underline{14.3} & 0.0 & \underline{14.3} & \textbf{0.0} & \underline{0.0} & 20.0 & \underline{27.3} & 11.1 \\
\RISEFitModel{JoyAI-Image-Edit-Plus~\cite{song2026joyai}} & \RISETaskYes & \underline{25.0} & 40.0 & \underline{30.8} & 15.8 & 30.0 & 0.0 & \textbf{7.4} & 0.0 & \textbf{14.3} & 2.6 & 0.0 & 0.0 & 8.7 & \underline{33.3} & 0.0 & 0.0 & 0.0 & \underline{14.3} & \textbf{0.0} & \underline{0.0} & 30.0 & 9.1 & 33.3 \\
\RISEFitModel{SenseNova-U1-A3B-MoT-SFT~\cite{diao2026sensenova}} & \RISETaskYes & 12.5 & 6.7 & 3.8 & 0.0 & 0.0 & \textbf{12.5} & \underline{0.0} & 0.0 & 2.9 & 0.0 & \textbf{11.1} & 0.0 & 21.7 & 22.2 & \textbf{50.0} & \underline{14.3} & \underline{12.5} & 0.0 & \textbf{0.0} & \underline{0.0} & 40.0 & 0.0 & 0.0 \\
\RISEFitModel{UniPic 3.0~\cite{wei2026skywork}} & \RISETaskYes & 12.5 & 33.3 & 26.9 & 21.1 & 0.0 & 0.0 & \underline{0.0} & 0.0 & 0.0 & 2.6 & 0.0 & 0.0 & 17.4 & 11.1 & \underline{12.5} & 0.0 & 0.0 & 0.0 & \textbf{0.0} & \underline{0.0} & 10.0 & 18.2 & 0.0 \\
\RISEFitModel{BAGEL (w/ CoT)~\cite{deng2025bagel}} & \RISETaskYes & 16.7 & 6.7 & 11.5 & \underline{31.6} & \underline{60.0} & 0.0 & \underline{0.0} & 0.0 & 2.9 & 2.6 & 0.0 & \underline{5.0} & 21.7 & 11.1 & 0.0 & \underline{14.3} & 0.0 & 0.0 & \textbf{0.0} & \underline{0.0} & 10.0 & 9.1 & 11.1 \\
\RISEFitModel{SenseNova-U1-8B-MoT-SFT~\cite{diao2026sensenova}} & \RISETaskYes & 16.7 & 6.7 & 11.5 & 0.0 & 0.0 & \underline{8.3} & \underline{0.0} & \underline{2.7} & 5.7 & 0.0 & \textbf{11.1} & 0.0 & 13.0 & \underline{33.3} & \underline{12.5} & \underline{14.3} & 0.0 & 0.0 & \textbf{0.0} & \underline{0.0} & 20.0 & 9.1 & 22.2 \\
\RISEFitModel{FLUX.2 Klein 4B~\cite{bfl2025flux2}} & \RISETaskYes & 4.2 & 26.7 & \underline{30.8} & 15.8 & \textbf{70.0} & 0.0 & \underline{0.0} & 0.0 & 2.9 & 10.3 & 0.0 & 0.0 & 8.7 & \underline{33.3} & 0.0 & 0.0 & 0.0 & 0.0 & \textbf{0.0} & \underline{0.0} & 0.0 & 9.1 & 22.2 \\
\RISEFitModel{UniWorld-V2~\cite{li2025uniworld}} & \RISETaskYes & 12.5 & 6.7 & 11.5 & 5.3 & 10.0 & 4.2 & \underline{0.0} & 0.0 & 0.0 & 0.0 & 0.0 & 0.0 & 8.7 & 11.1 & 0.0 & 0.0 & 0.0 & \underline{14.3} & \textbf{0.0} & \underline{0.0} & 20.0 & \underline{27.3} & \underline{44.4} \\
\RISEFitModel{BAGEL~\cite{deng2025bagel}} & \RISETaskYes & 8.3 & 0.0 & 7.7 & 15.8 & 30.0 & 0.0 & \underline{0.0} & 0.0 & 0.0 & 0.0 & 0.0 & \textbf{10.0} & 4.3 & 11.1 & 0.0 & 0.0 & 0.0 & 0.0 & \textbf{0.0} & \underline{0.0} & 0.0 & 0.0 & 0.0 \\
\RISEFitModel{OmniGen2~\cite{wu2025omnigen2}} & \RISETaskYes & 0.0 & 6.7 & 11.5 & 15.8 & 0.0 & 0.0 & \underline{0.0} & 0.0 & 0.0 & 0.0 & 0.0 & 0.0 & 17.4 & 22.2 & 0.0 & 0.0 & 0.0 & 0.0 & \textbf{0.0} & \underline{0.0} & 0.0 & 18.2 & 0.0 \\
\RISEFitModel{OmniGen~\cite{xiao2024omnigen}} & \RISETaskYes & 8.3 & 6.7 & 11.5 & 15.8 & 0.0 & 0.0 & \underline{0.0} & 0.0 & 2.9 & 5.1 & 0.0 & 0.0 & 0.0 & 0.0 & 0.0 & 0.0 & 0.0 & 0.0 & \textbf{0.0} & \underline{0.0} & 10.0 & 0.0 & 0.0 \\
\RISEFitModel{Emu2-Gen~\cite{sun2023emu2chat}} & \RISETaskYes & 0.0 & 0.0 & 3.8 & 0.0 & 0.0 & 0.0 & \underline{0.0} & 0.0 & 2.9 & 0.0 & 0.0 & 0.0 & 0.0 & 0.0 & 0.0 & 0.0 & 0.0 & 0.0 & \textbf{0.0} & \underline{0.0} & 0.0 & 9.1 & 0.0 \\
\RISEFitModel{JoyAI-Image-Edit~\cite{song2026joyai}} & \RISETaskNo & 16.7 & 33.3 & \underline{30.8} & 17.6 & \textemdash & 0.0 & \underline{0.0} & \underline{2.7} & 0.0 & 5.1 & \underline{5.6} & \textemdash & 13.0 & 11.1 & 0.0 & \underline{14.3} & \underline{12.5} & \underline{14.3} & \textbf{0.0} & \textbf{25.0} & 30.0 & \textbf{36.4} & 11.1 \\
\RISEFitModel{Step1X-Edit (thinking)~\cite{liu2025step1x}} & \RISETaskNo & 16.7 & \textbf{53.3} & 3.8 & 5.9 & \textemdash & 0.0 & \underline{0.0} & 0.0 & 0.0 & 2.6 & 0.0 & \textemdash & 26.1 & 11.1 & 0.0 & 0.0 & \underline{12.5} & 0.0 & \textbf{0.0} & \underline{0.0} & 40.0 & 18.2 & 22.2 \\
\RISEFitModel{Boogu-Image-0.1-Edit~\cite{chen2026boogu}} & \RISETaskNo & 16.7 & 20.0 & 23.1 & 23.5 & \textemdash & 0.0 & \underline{0.0} & \underline{2.7} & 8.6 & \textbf{17.9} & 0.0 & \textemdash & 13.0 & \underline{33.3} & 0.0 & 0.0 & 0.0 & 0.0 & \textbf{0.0} & \underline{0.0} & 10.0 & 18.2 & 22.2 \\
\RISEFitModel{Step1X-Edit (thinking + reflection)~\cite{liu2025step1x}} & \RISETaskNo & 4.2 & 20.0 & 0.0 & 5.9 & \textemdash & 0.0 & \underline{0.0} & 0.0 & 2.9 & 5.1 & 0.0 & \textemdash & 26.1 & 11.1 & \underline{12.5} & \textbf{28.6} & \textbf{37.5} & 0.0 & \textbf{0.0} & \underline{0.0} & 20.0 & \underline{27.3} & 11.1 \\
\RISEFitModel{LongCat-Image-Edit~\cite{team2025longcat}} & \RISETaskNo & 12.5 & 40.0 & 11.5 & 5.9 & \textemdash & 4.2 & \underline{0.0} & 0.0 & 0.0 & 2.6 & 0.0 & \textemdash & 8.7 & 0.0 & 0.0 & 0.0 & 0.0 & 0.0 & \textbf{0.0} & \underline{0.0} & \underline{50.0} & 9.1 & 11.1 \\
\RISEFitModel{Step1X-Edit (base)~\cite{liu2025step1x}} & \RISETaskNo & 12.5 & 26.7 & 7.7 & 11.8 & \textemdash & 0.0 & \underline{0.0} & 0.0 & 0.0 & 5.1 & 0.0 & \textemdash & \underline{30.4} & 0.0 & 0.0 & 0.0 & \underline{12.5} & \textbf{28.6} & \textbf{0.0} & \underline{0.0} & 20.0 & 9.1 & 11.1 \\
\RISEFitModel{InternVL-U (w/ CoT)~\cite{tian2026internvl}} & \RISETaskNo & 4.2 & 6.7 & 7.7 & 5.9 & \textemdash & 4.2 & \underline{0.0} & 0.0 & 0.0 & 2.6 & 0.0 & \textemdash & \textbf{34.8} & 22.2 & \underline{12.5} & 0.0 & \underline{12.5} & 0.0 & \textbf{0.0} & \underline{0.0} & 30.0 & 9.1 & 0.0 \\
\RISEFitModel{FLUX.1 Kontext Dev~\cite{labs2025flux1kontextflowmatching}} & \RISETaskNo & 8.3 & 33.3 & 19.2 & 17.6 & \textemdash & 0.0 & \underline{0.0} & 0.0 & 2.9 & 5.1 & 0.0 & \textemdash & 17.4 & 11.1 & 0.0 & 0.0 & 0.0 & 0.0 & \textbf{0.0} & \underline{0.0} & 0.0 & 9.1 & 11.1 \\
\RISEFitModel{LLaDA-Image~\cite{chen2026llada}} & \RISETaskNo & 12.5 & 6.7 & 11.5 & 0.0 & \textemdash & 0.0 & \underline{0.0} & 0.0 & 0.0 & 0.0 & 0.0 & \textemdash & 0.0 & 11.1 & \underline{12.5} & 0.0 & 0.0 & \underline{14.3} & \textbf{0.0} & \underline{0.0} & 0.0 & 9.1 & 0.0 \\
\RISEFitModel{Ovis-U1~\cite{wang2025ovis}} & \RISETaskNo & 8.3 & 6.7 & 0.0 & 0.0 & \textemdash & 0.0 & \underline{0.0} & 0.0 & 2.9 & 0.0 & \textbf{11.1} & \textemdash & 4.3 & 11.1 & 0.0 & 0.0 & 0.0 & 0.0 & \textbf{0.0} & \underline{0.0} & 0.0 & 0.0 & 0.0 \\
\RISEFitModel{InternVL-U~\cite{tian2026internvl}} & \RISETaskNo & 8.3 & 0.0 & 11.5 & 0.0 & \textemdash & 4.2 & \underline{0.0} & 0.0 & 0.0 & 0.0 & 0.0 & \textemdash & 13.0 & 11.1 & 0.0 & 0.0 & 0.0 & 0.0 & \textbf{0.0} & \underline{0.0} & 0.0 & 9.1 & 0.0 \\
\RISEFitModel{UniPic2-SD3.5M-Kontext-2B~\cite{wei2025skywork}} & \RISETaskNo & 0.0 & 0.0 & 3.8 & 0.0 & \textemdash & 0.0 & \underline{0.0} & 0.0 & 0.0 & 0.0 & 0.0 & \textemdash & 13.0 & 0.0 & 0.0 & 0.0 & 0.0 & 0.0 & \textbf{0.0} & \underline{0.0} & 0.0 & 0.0 & 0.0 \\
\RISEFitModel{Lumina-DiMOO~\cite{xin2025lumina}} & \RISETaskNo & 0.0 & 0.0 & 0.0 & 0.0 & \textemdash & 0.0 & \underline{0.0} & 0.0 & 2.9 & 0.0 & 0.0 & \textemdash & 8.7 & 11.1 & 0.0 & 0.0 & 0.0 & 0.0 & \textbf{0.0} & \underline{0.0} & 0.0 & 0.0 & 11.1 \\
\RISEFitModel{UniWorld-V1~\cite{lin2025uniworld}} & \RISETaskNo & 0.0 & 0.0 & 3.8 & 0.0 & \textemdash & 0.0 & \underline{0.0} & 0.0 & 2.9 & 0.0 & 0.0 & \textemdash & 0.0 & 0.0 & 0.0 & 0.0 & 0.0 & 0.0 & \textbf{0.0} & \underline{0.0} & 0.0 & 9.1 & 0.0 \\
\midrule[0.4pt]
\rowcolor{RISETaskGroup}
\multicolumn{25}{c}{\rule{0pt}{9pt}\raisebox{1.5pt}[0pt][0pt]{\textbf{Agentic Methods}}} \\
\midrule[0.4pt]
\RISEFitModel{\textbf{RISE-Agent (Ours)}} & \RISETaskYes & \textbf{41.7} & \textbf{66.7} & \textbf{38.5} & \textbf{21.1} & \textbf{20.0} & \textbf{33.3} & \textbf{63.0} & \textbf{48.6} & \textbf{42.9} & \textbf{23.1} & \textbf{16.7} & \textbf{40.0} & \textbf{69.6} & \RISEFitScore{\textbf{100.0}} & \textbf{75.0} & \textbf{42.9} & \textbf{87.5} & \textbf{71.4} & \textbf{0.0} & \textbf{75.0} & \textbf{80.0} & \textbf{45.5} & \textbf{44.4} \\
\RISEFitModel{MindBrush~\cite{he2026mind}} & \RISETaskNo & 25.0 & \underline{20.0} & \underline{19.2} & 5.9 & \textemdash & \underline{4.2} & \underline{18.5} & \underline{10.8} & 2.9 & \underline{7.7} & \underline{11.1} & \textemdash & 17.4 & 22.2 & \underline{62.5} & \textbf{42.9} & \underline{25.0} & \underline{42.9} & \textbf{0.0} & \underline{50.0} & \underline{60.0} & \underline{27.3} & \underline{22.2} \\
\RISEFitModel{InterleaveThinker~\cite{zheng2026interleavethinker}} & \RISETaskNo & \underline{33.3} & \underline{20.0} & 3.8 & 0.0 & \textemdash & 0.0 & 0.0 & 0.0 & \underline{5.7} & 2.6 & 0.0 & \textemdash & \underline{34.8} & \underline{33.3} & \underline{62.5} & \underline{0.0} & 0.0 & 0.0 & \textbf{0.0} & 0.0 & 30.0 & \underline{27.3} & \textbf{44.4} \\
\RISEFitModel{IntentEdit~\cite{zhang2026intentedit}} & \RISETaskNo & 29.2 & \underline{20.0} & 7.7 & \underline{11.8} & \textemdash & 0.0 & 3.7 & 0.0 & 2.9 & 0.0 & 0.0 & \textemdash & 17.4 & 11.1 & 25.0 & \underline{0.0} & 0.0 & 28.6 & \textbf{0.0} & 0.0 & 30.0 & 9.1 & 11.1 \\
\RISEFitModel{ImAgent~\cite{wang2025imagent}} & \RISETaskNo & 20.8 & 0.0 & 3.8 & 0.0 & \textemdash & 0.0 & 0.0 & 0.0 & 0.0 & 0.0 & 5.6 & \textemdash & 17.4 & 0.0 & 0.0 & \underline{0.0} & 0.0 & 0.0 & \textbf{0.0} & 0.0 & 10.0 & 0.0 & 0.0 \\
\bottomrule[0.8pt]
\end{tabular}
\par\vspace{4pt}
\parbox{\textwidth}{\fontsize{6.5}{8.2}\selectfont\raggedright
\textbf{S10}: Part-Whole Perception; \textbf{S11}: Geometric Regularities Reasoning; \textbf{S12}: Viewpoint Generation; \textbf{S13}: Environmental Structure Reasoning; \textbf{S14}: Egocentric Scene Reconstruction.\par
\textbf{L01}: Path Searching; \textbf{L02}: Mathematical Derivation; \textbf{L03}: Game Deduction; \textbf{L04}: Puzzle Solving; \textbf{L05}: Pattern Prediction; \textbf{L06}: Visual Oddity Detection; \textbf{L07}: Rule Transfer.\par
\textbf{F01}: Physical Counterfactual; \textbf{F02}: Chemical Counterfactual; \textbf{F03}: Biological Counterfactual; \textbf{F04}: Astronomy Counterfactual; \textbf{F05}: Geography Counterfactual; \textbf{F06}: History Counterfactual; \textbf{F07}: Music Theory Counterfactual; \textbf{F08}: Computer Science Counterfactual; \textbf{F09}: Anatomical Errors; \textbf{F10}: Structural Anomalies; \textbf{F11}: Relational Anomalies.
}
\endgroup
\end{table*}

\begin{table*}[!p]
\caption{Fine-grained task accuracy (\%) on RISEBench++ (ZH). Part I: Temporal reasoning.}
\label{tab:task_accuracy_cn_1}
\centering
\begingroup
\fontsize{6.5}{8.2}\selectfont
\setlength{\tabcolsep}{0.3pt}
\renewcommand{\arraystretch}{1.0}
\setlength{\aboverulesep}{1pt}
\setlength{\belowrulesep}{1pt}
\colorlet{RISETaskGroup}{black!6}
\def\RISETaskYes{\textcolor{green!50!black}{\ding{51}}}
\def\RISETaskNo{\textcolor{red!65!black}{\ding{55}}}
\begin{tabular}{
>{\raggedright\arraybackslash}m{\dimexpr .245\textwidth-2\tabcolsep\relax}
>{\centering\arraybackslash}m{\dimexpr .045\textwidth-2\tabcolsep\relax}
*{19}{>{\centering\arraybackslash}m{\dimexpr 0.037368421\textwidth-2\tabcolsep\relax}}
}
\toprule[0.8pt]
\multirow{2}{=}[-1.3pt]{\raggedright\textbf{Model}} & \multirow{2}{=}[-1.3pt]{\centering\shortstack{\textbf{Multi-}\\\textbf{img}}} & \multicolumn{19}{c}{\textbf{Temporal}} \\
\cmidrule(lr){3-21}
 & & \textbf{T01} & \textbf{T02} & \textbf{T03} & \textbf{T04} & \textbf{T05} & \textbf{T06} & \textbf{T07} & \textbf{T08} & \textbf{T09} & \textbf{T10} & \textbf{T11} & \textbf{T12} & \textbf{T13} & \textbf{T14} & \textbf{T15} & \textbf{T16} & \textbf{T17} & \textbf{T18} & \textbf{T19} \\
\midrule[0.5pt]
\rowcolor{RISETaskGroup}
\multicolumn{21}{c}{\rule{0pt}{9pt}\raisebox{1.5pt}[0pt][0pt]{\textbf{Closed-source Models}}} \\
\midrule[0.4pt]
GPT-Image-2.5 Sunburst~\cite{openai2025gptimage} & \RISETaskYes & 62.5 & 77.8 & \underline{69.2} & 75.0 & \textbf{87.5} & 57.1 & \underline{50.0} & 46.7 & 57.1 & \textbf{86.7} & \underline{75.0} & 50.0 & 37.5 & \underline{66.7} & 50.0 & \underline{50.0} & 20.0 & \underline{75.0} & \underline{66.7} \\
Nano Banana Pro~\cite{google2026nanobanana} & \RISETaskYes & 58.3 & \RISEFitScore{\textbf{100.0}} & \textbf{76.9} & 70.0 & 62.5 & 57.1 & \underline{50.0} & \textbf{80.0} & \textbf{71.4} & 53.3 & 37.5 & 50.0 & \textbf{62.5} & \underline{66.7} & 50.0 & 44.4 & \textbf{60.0} & 50.0 & \underline{66.7} \\
Nano Banana 2~\cite{google2026nanobanana} & \RISETaskYes & \underline{70.8} & \RISEFitScore{\textbf{100.0}} & \underline{69.2} & 75.0 & \underline{75.0} & \underline{71.4} & \underline{50.0} & 53.3 & 42.9 & 60.0 & 50.0 & 50.0 & \textbf{62.5} & 44.4 & 50.0 & \underline{50.0} & 20.0 & \RISEFitScore{\textbf{100.0}} & \RISEFitScore{\textbf{100.0}} \\
Luma Uni-1~\cite{luma2026uni1} & \RISETaskYes & \underline{70.8} & \RISEFitScore{\textbf{100.0}} & \textbf{76.9} & \textbf{90.0} & \textbf{87.5} & \underline{71.4} & \textbf{75.0} & 53.3 & 35.7 & 46.7 & 62.5 & \RISEFitScore{\textbf{100.0}} & 37.5 & \underline{66.7} & \textbf{83.3} & 38.9 & \textbf{60.0} & 50.0 & \underline{66.7} \\
GPT-Image-2~\cite{openai2025gptimage} & \RISETaskYes & 54.2 & 77.8 & \underline{69.2} & 60.0 & \textbf{87.5} & 28.6 & \underline{50.0} & 46.7 & 42.9 & 60.0 & \textbf{87.5} & 0.0 & \textbf{62.5} & \textbf{77.8} & 50.0 & \textbf{61.1} & \underline{40.0} & 25.0 & 33.3 \\
Qwen-Image-3.0~\cite{zhao2026qwen} & \RISETaskYes & 58.3 & \underline{88.9} & \underline{69.2} & 60.0 & \underline{75.0} & \underline{71.4} & 0.0 & 60.0 & \textbf{71.4} & 73.3 & 62.5 & \underline{75.0} & 12.5 & 11.1 & 50.0 & \textbf{61.1} & 20.0 & \underline{75.0} & 33.3 \\
GPT-Image-2.5 Flare~\cite{openai2025gptimage} & \RISETaskYes & 62.5 & 66.7 & \underline{69.2} & 40.0 & \textbf{87.5} & 57.1 & \underline{50.0} & 46.7 & 28.6 & 66.7 & \underline{75.0} & 0.0 & 37.5 & \underline{66.7} & 50.0 & 33.3 & \underline{40.0} & 50.0 & 33.3 \\
FLUX.2 Max~\cite{bfl2025flux2} & \RISETaskYes & 62.5 & \RISEFitScore{\textbf{100.0}} & 53.8 & 60.0 & 62.5 & \underline{71.4} & \textbf{75.0} & \underline{73.3} & 57.1 & 53.3 & \underline{75.0} & 50.0 & 37.5 & 33.3 & 58.3 & 44.4 & \textbf{60.0} & 50.0 & \underline{66.7} \\
Nano Banana 2 Lite~\cite{google2026nanobanana} & \RISETaskYes & \textbf{75.0} & 77.8 & 61.5 & 60.0 & \textbf{87.5} & 28.6 & \underline{50.0} & 60.0 & 57.1 & \underline{80.0} & 62.5 & 50.0 & 12.5 & 33.3 & \underline{66.7} & 38.9 & \underline{40.0} & \underline{75.0} & 33.3 \\
GPT-Image-1.5~\cite{openai2025gptimage} & \RISETaskYes & 58.3 & \RISEFitScore{\textbf{100.0}} & 61.5 & 75.0 & \underline{75.0} & 57.1 & \underline{50.0} & 40.0 & 42.9 & 53.3 & 37.5 & 50.0 & 25.0 & \underline{66.7} & 50.0 & \underline{50.0} & \underline{40.0} & 25.0 & 33.3 \\
Nano Banana~\cite{google2026nanobanana} & \RISETaskYes & 66.7 & 55.6 & 23.1 & 45.0 & 50.0 & 28.6 & \textbf{75.0} & 40.0 & \underline{64.3} & 60.0 & 62.5 & 25.0 & \textbf{62.5} & 44.4 & 41.7 & \underline{50.0} & 20.0 & 25.0 & \underline{66.7} \\
FLUX.2 Pro~\cite{bfl2025flux2} & \RISETaskYes & 54.2 & 66.7 & 38.5 & 40.0 & 62.5 & 57.1 & \underline{50.0} & 60.0 & 35.7 & 73.3 & 62.5 & \underline{75.0} & \underline{50.0} & 22.2 & 58.3 & 44.4 & \textbf{60.0} & 50.0 & 0.0 \\
HunyuanImage-3.5~\cite{cao2025hunyuanimage} & \RISETaskYes & 45.8 & 77.8 & 53.8 & 45.0 & 50.0 & 57.1 & 0.0 & 40.0 & 57.1 & 73.3 & 37.5 & 0.0 & 25.0 & 33.3 & 58.3 & 27.8 & 0.0 & 0.0 & 33.3 \\
Wan2.7-Image-Pro~\cite{mao2026wan} & \RISETaskYes & 45.8 & \underline{88.9} & 46.2 & 55.0 & 37.5 & \textbf{85.7} & \underline{50.0} & 26.7 & 21.4 & 33.3 & 50.0 & 0.0 & 25.0 & 44.4 & 33.3 & 38.9 & 0.0 & 0.0 & 33.3 \\
Qwen-Image-2.0~\cite{zhao2026qwen} & \RISETaskYes & 41.7 & \underline{88.9} & 53.8 & \underline{80.0} & 62.5 & 57.1 & \underline{50.0} & \underline{73.3} & 42.9 & 60.0 & 50.0 & 25.0 & 25.0 & 33.3 & 41.7 & 33.3 & 0.0 & 50.0 & \underline{66.7} \\
Seedream 4.5~\cite{seedream2025seedream} & \RISETaskYes & 54.2 & 77.8 & 46.2 & 50.0 & 50.0 & 42.9 & \textbf{75.0} & 33.3 & 35.7 & 46.7 & 50.0 & 25.0 & \underline{50.0} & 11.1 & 50.0 & 38.9 & 20.0 & 0.0 & 33.3 \\
Seedream 5.0~\cite{seedream2025seedream} & \RISETaskYes & 50.0 & 66.7 & 61.5 & 45.0 & 62.5 & 42.9 & 25.0 & 33.3 & 35.7 & 60.0 & 25.0 & 0.0 & 25.0 & 11.1 & 33.3 & 33.3 & 20.0 & 25.0 & 33.3 \\
GPT-Image-1~\cite{openai2025gptimage} & \RISETaskYes & 54.2 & 77.8 & 61.5 & 55.0 & \underline{75.0} & 57.1 & 25.0 & 33.3 & 35.7 & 20.0 & 25.0 & 25.0 & 12.5 & \underline{66.7} & 41.7 & 38.9 & 0.0 & 50.0 & 0.0 \\
Step Image Edit 2~\cite{liu2025step1x} & \RISETaskNo & 20.8 & 22.2 & 46.2 & 25.0 & 50.0 & 0.0 & 0.0 & 0.0 & 0.0 & 0.0 & \textemdash & 0.0 & 0.0 & 11.1 & 0.0 & 11.1 & 0.0 & 0.0 & 0.0 \\
\midrule[0.4pt]
\rowcolor{RISETaskGroup}
\multicolumn{21}{c}{\rule{0pt}{9pt}\raisebox{1.5pt}[0pt][0pt]{\textbf{Open-source Models}}} \\
\midrule[0.4pt]
HunyuanImage-3.0-Instruct~\cite{cao2025hunyuanimage} & \RISETaskYes & 25.0 & \textbf{77.8} & \underline{53.8} & \textbf{50.0} & 25.0 & 14.3 & \underline{25.0} & 20.0 & 7.1 & 40.0 & 25.0 & \underline{0.0} & 0.0 & \textbf{33.3} & 8.3 & \textbf{27.8} & \underline{0.0} & \underline{25.0} & \textbf{0.0} \\
Qwen-Image-2.1~\cite{zhao2026qwen} & \RISETaskYes & 8.3 & \underline{66.7} & 23.1 & 30.0 & 25.0 & 0.0 & \underline{25.0} & 20.0 & \textbf{14.3} & \underline{60.0} & \underline{37.5} & \underline{0.0} & \underline{12.5} & 11.1 & 0.0 & \textbf{27.8} & \underline{0.0} & \underline{25.0} & \textbf{0.0} \\
FireRed-Image-Edit-1.1~\cite{team2026firered} & \RISETaskYes & 8.3 & 44.4 & 30.8 & 30.0 & \textbf{75.0} & 0.0 & \underline{25.0} & \underline{26.7} & 0.0 & 53.3 & 25.0 & \underline{0.0} & \textbf{25.0} & 11.1 & 0.0 & 11.1 & \underline{0.0} & 0.0 & \textbf{0.0} \\
FLUX.2 Dev~\cite{bfl2025flux2} & \RISETaskYes & 8.3 & 55.6 & 15.4 & 15.0 & 37.5 & 14.3 & 0.0 & 20.0 & \textbf{14.3} & \underline{60.0} & \underline{37.5} & \underline{0.0} & \textbf{25.0} & 0.0 & \underline{16.7} & \textbf{27.8} & \underline{0.0} & 0.0 & \textbf{0.0} \\
Qwen-Image-Edit-2511~\cite{wu2025qwen} & \RISETaskYes & 20.8 & 33.3 & \textbf{61.5} & 15.0 & 25.0 & 0.0 & 0.0 & 13.3 & 0.0 & 26.7 & 25.0 & \underline{0.0} & \underline{12.5} & \underline{22.2} & \underline{16.7} & \textbf{27.8} & \underline{0.0} & 0.0 & \textbf{0.0} \\
Qwen-Image-Edit-2509~\cite{wu2025qwen} & \RISETaskYes & 8.3 & 11.1 & 46.2 & 20.0 & 50.0 & \textbf{42.9} & \underline{25.0} & 13.3 & 7.1 & 26.7 & 25.0 & \underline{0.0} & \underline{12.5} & 0.0 & 8.3 & 16.7 & \underline{0.0} & 0.0 & \textbf{0.0} \\
FireRed-Image-Edit-1.0~\cite{team2026firered} & \RISETaskYes & 8.3 & 33.3 & 38.5 & \underline{35.0} & \underline{62.5} & 14.3 & \textbf{50.0} & \underline{26.7} & 0.0 & 33.3 & 25.0 & \underline{0.0} & 0.0 & 11.1 & 8.3 & 5.6 & \underline{0.0} & 0.0 & \textbf{0.0} \\
SenseNova-U1-A3B-MoT-SFT~\cite{diao2026sensenova} & \RISETaskYes & \textbf{54.2} & 55.6 & 23.1 & \textbf{50.0} & 12.5 & 14.3 & 0.0 & 20.0 & 0.0 & 0.0 & 0.0 & \underline{0.0} & \underline{12.5} & 0.0 & \underline{16.7} & \underline{22.2} & \underline{0.0} & 0.0 & \textbf{0.0} \\
JoyAI-Image-Edit-Plus~\cite{song2026joyai} & \RISETaskYes & 4.2 & 11.1 & 15.4 & 0.0 & 0.0 & 0.0 & 0.0 & \underline{26.7} & \textbf{14.3} & \underline{60.0} & \textbf{50.0} & \underline{0.0} & 0.0 & 11.1 & 0.0 & \underline{22.2} & \underline{0.0} & \underline{25.0} & \textbf{0.0} \\
FLUX.2 Klein 9B~\cite{bfl2025flux2} & \RISETaskYes & 12.5 & 33.3 & 30.8 & 20.0 & 50.0 & 0.0 & 0.0 & \textbf{33.3} & 7.1 & \underline{60.0} & \underline{37.5} & \underline{0.0} & \underline{12.5} & 0.0 & 8.3 & \textbf{27.8} & \underline{0.0} & \underline{25.0} & \textbf{0.0} \\
BAGEL (w/ CoT)~\cite{deng2025bagel} & \RISETaskYes & 20.8 & 22.2 & 15.4 & 15.0 & 12.5 & \textbf{42.9} & 0.0 & 20.0 & \textbf{14.3} & 46.7 & 25.0 & \underline{0.0} & \underline{12.5} & 11.1 & \textbf{25.0} & 16.7 & \underline{0.0} & 0.0 & \textbf{0.0} \\
HiDream-O1-Image~\cite{cai2026hidream} & \RISETaskYes & 16.7 & 22.2 & 30.8 & \underline{35.0} & 0.0 & 0.0 & \underline{25.0} & \underline{26.7} & 7.1 & 46.7 & 25.0 & \underline{0.0} & \underline{12.5} & 11.1 & 0.0 & 16.7 & \underline{0.0} & 0.0 & \textbf{0.0} \\
FLUX.2 Klein 4B~\cite{bfl2025flux2} & \RISETaskYes & 4.2 & 11.1 & 30.8 & 10.0 & 12.5 & 0.0 & 0.0 & \underline{26.7} & \textbf{14.3} & \textbf{80.0} & 12.5 & \underline{0.0} & \underline{12.5} & 11.1 & 0.0 & 11.1 & \underline{0.0} & \underline{25.0} & \textbf{0.0} \\
UniWorld-V2~\cite{li2025uniworld} & \RISETaskYes & 8.3 & 44.4 & 7.7 & 20.0 & 25.0 & \textbf{42.9} & 0.0 & 13.3 & 0.0 & 13.3 & 12.5 & \underline{0.0} & 0.0 & 11.1 & 8.3 & 16.7 & \textbf{20.0} & 0.0 & \textbf{0.0} \\
SenseNova-U1-8B-MoT-SFT~\cite{diao2026sensenova} & \RISETaskYes & 29.2 & 55.6 & 15.4 & 25.0 & 0.0 & \textbf{42.9} & \underline{25.0} & 20.0 & 0.0 & 0.0 & 0.0 & \underline{0.0} & 0.0 & 0.0 & 8.3 & \textbf{27.8} & \underline{0.0} & 0.0 & \textbf{0.0} \\
UniPic 3.0~\cite{wei2026skywork} & \RISETaskYes & 4.2 & 11.1 & 7.7 & 15.0 & 12.5 & 14.3 & \underline{25.0} & 6.7 & 0.0 & 6.7 & 0.0 & \underline{0.0} & \underline{12.5} & 0.0 & 8.3 & 5.6 & \underline{0.0} & 0.0 & \textbf{0.0} \\
BAGEL~\cite{deng2025bagel} & \RISETaskYes & 4.2 & 22.2 & 15.4 & 0.0 & 12.5 & 0.0 & 0.0 & \underline{26.7} & 7.1 & 33.3 & 25.0 & \underline{0.0} & 0.0 & 0.0 & 8.3 & 5.6 & \underline{0.0} & 0.0 & \textbf{0.0} \\
OmniGen2~\cite{wu2025omnigen2} & \RISETaskYes & 0.0 & 0.0 & 15.4 & 0.0 & 0.0 & 0.0 & \underline{25.0} & 13.3 & 7.1 & 20.0 & 12.5 & \underline{0.0} & \underline{12.5} & 0.0 & 0.0 & 5.6 & \underline{0.0} & 0.0 & \textbf{0.0} \\
OmniGen~\cite{xiao2024omnigen} & \RISETaskYes & 4.2 & 0.0 & 7.7 & 0.0 & 0.0 & 0.0 & 0.0 & 13.3 & \textbf{14.3} & 20.0 & 12.5 & \underline{0.0} & 0.0 & 11.1 & 0.0 & 16.7 & \underline{0.0} & 0.0 & \textbf{0.0} \\
Emu2-Gen~\cite{sun2023emu2chat} & \RISETaskYes & 0.0 & 11.1 & 7.7 & 0.0 & 12.5 & 0.0 & 0.0 & 0.0 & 0.0 & 6.7 & 0.0 & \underline{0.0} & 0.0 & 0.0 & 0.0 & 5.6 & \underline{0.0} & 0.0 & \textbf{0.0} \\
JoyAI-Image-Edit~\cite{song2026joyai} & \RISETaskNo & 12.5 & 22.2 & 38.5 & 15.0 & 25.0 & 0.0 & 0.0 & 9.1 & 0.0 & 0.0 & \textemdash & \underline{0.0} & 0.0 & 11.1 & \underline{16.7} & 16.7 & \underline{0.0} & \underline{25.0} & \textbf{0.0} \\
Boogu-Image-0.1-Edit~\cite{chen2026boogu} & \RISETaskNo & 8.3 & 11.1 & 30.8 & 10.0 & 25.0 & 14.3 & 0.0 & 18.2 & 0.0 & 0.0 & \textemdash & \textbf{25.0} & 0.0 & \textbf{33.3} & 8.3 & 16.7 & \underline{0.0} & 0.0 & \textbf{0.0} \\
\RISEFitModel{Step1X-Edit (thinking + reflection)~\cite{liu2025step1x}} & \RISETaskNo & \underline{37.5} & 44.4 & 23.1 & 25.0 & 12.5 & 14.3 & 0.0 & 0.0 & 0.0 & 0.0 & \textemdash & \underline{0.0} & 0.0 & 0.0 & 8.3 & 16.7 & \underline{0.0} & \textbf{50.0} & \textbf{0.0} \\
Step1X-Edit (thinking)~\cite{liu2025step1x} & \RISETaskNo & 25.0 & 22.2 & 15.4 & 15.0 & 12.5 & 14.3 & \underline{25.0} & 0.0 & \underline{8.3} & 0.0 & \textemdash & \textbf{25.0} & \textbf{25.0} & 11.1 & 8.3 & 11.1 & \underline{0.0} & \underline{25.0} & \textbf{0.0} \\
Step1X-Edit (base)~\cite{liu2025step1x} & \RISETaskNo & 20.8 & 33.3 & 30.8 & 5.0 & 12.5 & 0.0 & \underline{25.0} & 0.0 & 0.0 & 0.0 & \textemdash & \underline{0.0} & 0.0 & \underline{22.2} & 0.0 & \underline{22.2} & \underline{0.0} & \underline{25.0} & \textbf{0.0} \\
LongCat-Image-Edit~\cite{team2025longcat} & \RISETaskNo & 8.3 & 22.2 & 46.2 & 15.0 & 37.5 & \underline{28.6} & 0.0 & 9.1 & 0.0 & 0.0 & \textemdash & \underline{0.0} & 0.0 & \underline{22.2} & 8.3 & 5.6 & \underline{0.0} & 0.0 & \textbf{0.0} \\
FLUX.1 Kontext Dev~\cite{labs2025flux1kontextflowmatching} & \RISETaskNo & 0.0 & 0.0 & 7.7 & 0.0 & 0.0 & 0.0 & 0.0 & 0.0 & 0.0 & 0.0 & \textemdash & \underline{0.0} & 0.0 & 11.1 & 0.0 & 16.7 & \underline{0.0} & 0.0 & \textbf{0.0} \\
LLaDA-Image~\cite{chen2026llada} & \RISETaskNo & 4.2 & 0.0 & 15.4 & 5.0 & 12.5 & 0.0 & 0.0 & 9.1 & 0.0 & 0.0 & \textemdash & \underline{0.0} & 0.0 & 0.0 & 0.0 & 11.1 & \underline{0.0} & 0.0 & \textbf{0.0} \\
InternVL-U (w/ CoT)~\cite{tian2026internvl} & \RISETaskNo & 4.2 & 11.1 & 15.4 & 15.0 & 0.0 & 0.0 & 0.0 & 0.0 & 0.0 & 0.0 & \textemdash & \underline{0.0} & \textbf{25.0} & 0.0 & 0.0 & 5.6 & \underline{0.0} & \underline{25.0} & \textbf{0.0} \\
Ovis-U1~\cite{wang2025ovis} & \RISETaskNo & 0.0 & 22.2 & 15.4 & 0.0 & 12.5 & 0.0 & 0.0 & 0.0 & 0.0 & 0.0 & \textemdash & \underline{0.0} & 0.0 & 0.0 & 8.3 & 0.0 & \underline{0.0} & 0.0 & \textbf{0.0} \\
InternVL-U~\cite{tian2026internvl} & \RISETaskNo & 0.0 & 0.0 & 15.4 & 15.0 & 0.0 & 0.0 & 0.0 & 0.0 & 0.0 & 0.0 & \textemdash & \underline{0.0} & 0.0 & 0.0 & 0.0 & 0.0 & \underline{0.0} & \underline{25.0} & \textbf{0.0} \\
Lumina-DiMOO~\cite{xin2025lumina} & \RISETaskNo & 4.2 & 11.1 & 0.0 & 0.0 & 0.0 & 0.0 & 0.0 & 9.1 & 0.0 & 0.0 & \textemdash & \underline{0.0} & 0.0 & 0.0 & 0.0 & 0.0 & \underline{0.0} & 0.0 & \textbf{0.0} \\
UniWorld-V1~\cite{lin2025uniworld} & \RISETaskNo & 0.0 & 11.1 & 7.7 & 0.0 & 12.5 & 0.0 & 0.0 & 0.0 & 0.0 & 0.0 & \textemdash & \underline{0.0} & 0.0 & 11.1 & 0.0 & 0.0 & \underline{0.0} & \underline{25.0} & \textbf{0.0} \\
UniPic2-SD3.5M-Kontext-2B~\cite{wei2025skywork} & \RISETaskNo & 0.0 & 0.0 & 0.0 & 0.0 & 0.0 & 0.0 & 0.0 & 0.0 & 0.0 & 0.0 & \textemdash & \underline{0.0} & 0.0 & 0.0 & 0.0 & 0.0 & \underline{0.0} & 0.0 & \textbf{0.0} \\
\midrule[0.4pt]
\rowcolor{RISETaskGroup}
\multicolumn{21}{c}{\rule{0pt}{9pt}\raisebox{1.5pt}[0pt][0pt]{\textbf{Agentic Methods}}} \\
\midrule[0.4pt]
\textbf{RISE-Agent (Ours)} & \RISETaskYes & \textbf{66.7} & \textbf{88.9} & \textbf{53.8} & \underline{55.0} & \textbf{75.0} & \textbf{85.7} & \textbf{50.0} & \textbf{60.0} & \underline{21.4} & \textbf{60.0} & \textbf{37.5} & \textbf{25.0} & 12.5 & \textbf{55.6} & \textbf{41.7} & \textbf{38.9} & \underline{20.0} & \textbf{25.0} & \underline{33.3} \\
MindBrush~\cite{he2026mind} & \RISETaskNo & \underline{50.0} & 55.6 & \underline{46.2} & \underline{55.0} & \textbf{75.0} & \underline{71.4} & \underline{25.0} & \underline{45.5} & \textbf{33.3} & \underline{0.0} & \textemdash & \underline{0.0} & \textbf{50.0} & \underline{33.3} & \textbf{41.7} & \underline{27.8} & \textbf{40.0} & \textbf{25.0} & 0.0 \\
InterleaveThinker~\cite{zheng2026interleavethinker} & \RISETaskNo & 25.0 & 33.3 & 38.5 & 45.0 & \underline{25.0} & 42.9 & \underline{25.0} & 18.2 & 8.3 & \underline{0.0} & \textemdash & \textbf{25.0} & \textbf{50.0} & 11.1 & 0.0 & 16.7 & \underline{20.0} & \underline{0.0} & \textbf{66.7} \\
IntentEdit~\cite{zhang2026intentedit} & \RISETaskNo & 12.5 & \underline{77.8} & 23.1 & \textbf{60.0} & 0.0 & 57.1 & 0.0 & 18.2 & 8.3 & \underline{0.0} & \textemdash & \underline{0.0} & \underline{25.0} & 22.2 & \underline{25.0} & 0.0 & 0.0 & \underline{0.0} & 0.0 \\
ImAgent~\cite{wang2025imagent} & \RISETaskNo & 12.5 & 0.0 & 7.7 & 0.0 & 0.0 & 14.3 & \underline{25.0} & 9.1 & 0.0 & \underline{0.0} & \textemdash & \underline{0.0} & 0.0 & 0.0 & 0.0 & 0.0 & 0.0 & \underline{0.0} & 0.0 \\
\bottomrule[0.8pt]
\end{tabular}
\par\vspace{4pt}
\parbox{\textwidth}{\fontsize{6.5}{8.2}\selectfont\raggedright
\textbf{T01}: Life Progression; \textbf{T02}: Material Transformation; \textbf{T03}: Natural Environmental Changes; \textbf{T04}: Physical Aging; \textbf{T05}: Social Activities; \textbf{T06}: Tool Evolution; \textbf{T07}: Urban Evolution; \textbf{T08}: Temporal State Transition; \textbf{T09}: Precise Temporal Localization; \textbf{T10}: Position Change; \textbf{T11}: Action Sequence; \textbf{T12}: Physics; \textbf{T13}: Chemistry; \textbf{T14}: Biology; \textbf{T15}: Astronomy; \textbf{T16}: Geography; \textbf{T17}: History; \textbf{T18}: Medicine; \textbf{T19}: Computer Science.
}
\endgroup
\end{table*}

\begin{table*}[!p]
\caption{Fine-grained task accuracy (\%) on RISEBench++ (ZH). Part II: Causal reasoning and Spatial Relation.}
\label{tab:task_accuracy_cn_2}
\centering
\begingroup
\fontsize{6.2}{8.2}\selectfont
\setlength{\tabcolsep}{1.2pt}
\renewcommand{\arraystretch}{1.0}
\setlength{\aboverulesep}{1pt}
\setlength{\belowrulesep}{1pt}
\colorlet{RISETaskGroup}{black!6}
\def\RISETaskYes{\textcolor{green!50!black}{\ding{51}}}
\def\RISETaskNo{\textcolor{red!65!black}{\ding{55}}}
\begin{tabular}{
>{\raggedright\arraybackslash}m{\dimexpr .19\textwidth-2\tabcolsep\relax}
>{\centering\arraybackslash}m{\dimexpr .04\textwidth-2\tabcolsep\relax}
*{14}{>{\centering\arraybackslash}m{\dimexpr (.77\textwidth-3pt)/23-2\tabcolsep\relax}}!{\hspace{3pt}}
*{9}{>{\centering\arraybackslash}m{\dimexpr (.77\textwidth-3pt)/23-2\tabcolsep\relax}}
}
\toprule[0.8pt]
\multirow{2}{=}[-1.3pt]{\raggedright\textbf{Model}} & \multirow{2}{=}[-1.3pt]{\centering\shortstack{\textbf{Multi-}\\\textbf{img}}} & \multicolumn{14}{c}{\textbf{Causal}} & \multicolumn{9}{c}{\textbf{Spatial (Relation)}} \\
\cmidrule(lr){3-16}\cmidrule(lr){17-25}
 & & \textbf{C01} & \textbf{C02} & \textbf{C03} & \textbf{C04} & \textbf{C05} & \textbf{C06} & \textbf{C07} & \textbf{C08} & \textbf{C09} & \textbf{C10} & \textbf{C11} & \textbf{C12} & \textbf{C13} & \textbf{C14} & \textbf{S01} & \textbf{S02} & \textbf{S03} & \textbf{S04} & \textbf{S05} & \textbf{S06} & \textbf{S07} & \textbf{S08} & \textbf{S09} \\
\midrule[0.5pt]
\rowcolor{RISETaskGroup}
\multicolumn{25}{c}{\rule{0pt}{9pt}\raisebox{1.5pt}[0pt][0pt]{\textbf{Closed-source Models}}} \\
\midrule[0.4pt]
\RISEFitModel{GPT-Image-2.5 Sunburst~\cite{openai2025gptimage}} & \RISETaskYes & 77.4 & \textbf{92.6} & \textbf{86.4} & \textbf{60.0} & 60.0 & \underline{70.0} & 60.0 & 36.0 & 73.9 & \underline{91.7} & \RISEFitScore{\textbf{100.0}} & \underline{50.0} & \RISEFitScore{\textbf{100.0}} & \underline{30.0} & \underline{63.6} & \textbf{81.8} & \textbf{54.5} & \underline{66.7} & \underline{61.5} & \textbf{80.0} & \underline{55.6} & \underline{55.6} & 16.7 \\
\RISEFitModel{Nano Banana Pro~\cite{google2026nanobanana}} & \RISETaskYes & \underline{80.6} & 74.1 & 77.3 & \underline{50.0} & \textbf{90.0} & \textbf{80.0} & 30.0 & \underline{52.0} & \underline{82.6} & 66.7 & \RISEFitScore{\textbf{100.0}} & 25.0 & 0.0 & \underline{30.0} & \underline{63.6} & 54.5 & \underline{45.5} & 58.3 & \textbf{76.9} & \textbf{80.0} & 38.9 & 33.3 & 16.7 \\
\RISEFitModel{Nano Banana 2~\cite{google2026nanobanana}} & \RISETaskYes & \textbf{87.1} & 81.5 & \underline{81.8} & 40.0 & \underline{70.0} & 50.0 & \underline{70.0} & \textbf{56.0} & 65.2 & 66.7 & \RISEFitScore{\textbf{100.0}} & \underline{50.0} & \underline{66.7} & \textbf{40.0} & \textbf{72.7} & 45.5 & 27.3 & 58.3 & 46.2 & 60.0 & \underline{55.6} & 22.2 & 0.0 \\
\RISEFitModel{Luma Uni-1~\cite{luma2026uni1}} & \RISETaskYes & 74.2 & 77.8 & 72.7 & \underline{50.0} & \underline{70.0} & 40.0 & \textbf{80.0} & 40.0 & \textbf{91.3} & \RISEFitScore{\textbf{100.0}} & \underline{66.7} & 25.0 & \underline{66.7} & 20.0 & \underline{63.6} & 18.2 & \textbf{54.5} & \underline{66.7} & \underline{61.5} & \underline{66.7} & 50.0 & 33.3 & \textbf{50.0} \\
\RISEFitModel{GPT-Image-2~\cite{openai2025gptimage}} & \RISETaskYes & 67.7 & 81.5 & \underline{81.8} & \underline{50.0} & 50.0 & 60.0 & \underline{70.0} & 36.0 & 65.2 & \underline{91.7} & \underline{66.7} & 25.0 & 33.3 & 10.0 & \textbf{72.7} & \underline{63.6} & \underline{45.5} & \underline{66.7} & \underline{61.5} & 46.7 & \underline{55.6} & \underline{55.6} & \underline{33.3} \\
\RISEFitModel{Qwen-Image-3.0~\cite{zhao2026qwen}} & \RISETaskYes & 51.6 & 55.6 & 50.0 & 40.0 & 60.0 & 60.0 & \underline{70.0} & 40.0 & 47.8 & 66.7 & \underline{66.7} & 0.0 & \underline{66.7} & 20.0 & \underline{63.6} & \underline{63.6} & 9.1 & 25.0 & \underline{61.5} & 60.0 & 50.0 & 33.3 & 16.7 \\
\RISEFitModel{GPT-Image-2.5 Flare~\cite{openai2025gptimage}} & \RISETaskYes & 77.4 & \underline{88.9} & 72.7 & \textbf{60.0} & 60.0 & 40.0 & \textbf{80.0} & 40.0 & 65.2 & \RISEFitScore{\textbf{100.0}} & \RISEFitScore{\textbf{100.0}} & 25.0 & 33.3 & 20.0 & 54.5 & 54.5 & 27.3 & \textbf{75.0} & 53.8 & 46.7 & \textbf{61.1} & \underline{55.6} & 16.7 \\
\RISEFitModel{FLUX.2 Max~\cite{bfl2025flux2}} & \RISETaskYes & 77.4 & 74.1 & 68.2 & 30.0 & 50.0 & 10.0 & 60.0 & 44.0 & 56.5 & 83.3 & \underline{66.7} & \textbf{75.0} & 33.3 & \textbf{40.0} & 27.3 & 36.4 & 27.3 & 50.0 & 30.8 & 46.7 & 27.8 & \underline{55.6} & 0.0 \\
\RISEFitModel{Nano Banana 2 Lite~\cite{google2026nanobanana}} & \RISETaskYes & 71.0 & 77.8 & \underline{81.8} & 30.0 & 30.0 & 40.0 & 40.0 & \underline{52.0} & 78.3 & 83.3 & \underline{66.7} & \underline{50.0} & 0.0 & 10.0 & 54.5 & 54.5 & 27.3 & 25.0 & 46.2 & 53.3 & 27.8 & 33.3 & 16.7 \\
\RISEFitModel{GPT-Image-1.5~\cite{openai2025gptimage}} & \RISETaskYes & \underline{80.6} & 77.8 & 63.6 & 40.0 & 40.0 & 30.0 & 50.0 & 44.0 & 60.9 & 66.7 & \underline{66.7} & \underline{50.0} & 0.0 & 10.0 & 18.2 & 18.2 & 27.3 & 58.3 & 46.2 & 26.7 & 22.2 & \textbf{66.7} & 0.0 \\
\RISEFitModel{Nano Banana~\cite{google2026nanobanana}} & \RISETaskYes & 61.3 & 85.2 & 63.6 & 30.0 & 40.0 & 10.0 & 60.0 & 40.0 & 60.9 & 66.7 & 33.3 & \underline{50.0} & 0.0 & 10.0 & 45.5 & 45.5 & \underline{45.5} & 41.7 & 23.1 & 26.7 & 27.8 & 22.2 & 0.0 \\
\RISEFitModel{FLUX.2 Pro~\cite{bfl2025flux2}} & \RISETaskYes & 61.3 & 74.1 & 45.5 & 10.0 & 40.0 & 30.0 & 20.0 & 12.0 & 56.5 & 83.3 & \underline{66.7} & 25.0 & 0.0 & 10.0 & 18.2 & 18.2 & 27.3 & 50.0 & 46.2 & 46.7 & 16.7 & 44.4 & 0.0 \\
\RISEFitModel{HunyuanImage-3.5~\cite{cao2025hunyuanimage}} & \RISETaskYes & 45.2 & 44.4 & 40.9 & 20.0 & 30.0 & 40.0 & 40.0 & 24.0 & 52.2 & 66.7 & 33.3 & \underline{50.0} & 33.3 & 10.0 & \underline{63.6} & 36.4 & \underline{45.5} & 50.0 & 53.8 & 60.0 & 33.3 & 11.1 & 0.0 \\
\RISEFitModel{Wan2.7-Image-Pro~\cite{mao2026wan}} & \RISETaskYes & 58.1 & 66.7 & 45.5 & 40.0 & 50.0 & 30.0 & 50.0 & 16.0 & 47.8 & 66.7 & 33.3 & \underline{50.0} & 0.0 & 0.0 & 45.5 & 45.5 & 18.2 & 58.3 & 38.5 & 33.3 & 50.0 & 22.2 & 0.0 \\
\RISEFitModel{Qwen-Image-2.0~\cite{zhao2026qwen}} & \RISETaskYes & 54.8 & 37.0 & 36.4 & 30.0 & 40.0 & 20.0 & 20.0 & 28.0 & 43.5 & 58.3 & \underline{66.7} & 25.0 & 0.0 & 10.0 & 27.3 & 9.1 & 27.3 & 33.3 & 30.8 & 20.0 & 33.3 & 11.1 & 0.0 \\
\RISEFitModel{Seedream 4.5~\cite{seedream2025seedream}} & \RISETaskYes & 54.8 & 59.3 & 54.5 & \underline{50.0} & 40.0 & 30.0 & 50.0 & 28.0 & 43.5 & 66.7 & 0.0 & 0.0 & 0.0 & \underline{30.0} & 27.3 & 27.3 & 27.3 & 41.7 & 30.8 & 20.0 & 16.7 & 11.1 & 0.0 \\
\RISEFitModel{Seedream 5.0~\cite{seedream2025seedream}} & \RISETaskYes & 54.8 & 48.1 & 31.8 & 40.0 & 30.0 & 40.0 & 40.0 & 16.0 & 56.5 & 41.7 & 33.3 & \textbf{75.0} & 0.0 & 10.0 & 36.4 & 27.3 & 18.2 & 33.3 & 15.4 & 6.7 & 16.7 & 22.2 & 16.7 \\
\RISEFitModel{GPT-Image-1~\cite{openai2025gptimage}} & \RISETaskYes & 71.0 & 63.0 & 40.9 & 20.0 & 20.0 & 10.0 & 40.0 & 12.0 & 30.4 & 83.3 & 33.3 & 25.0 & 0.0 & 10.0 & 9.1 & 0.0 & 0.0 & 25.0 & 15.4 & 33.3 & 5.6 & 33.3 & 0.0 \\
\RISEFitModel{Step Image Edit 2~\cite{liu2025step1x}} & \RISETaskNo & 32.3 & 18.5 & 28.6 & 0.0 & 0.0 & 20.0 & 50.0 & 8.3 & 10.5 & 33.3 & 0.0 & 25.0 & 0.0 & 0.0 & 18.2 & 36.4 & 36.4 & 25.0 & 23.1 & 13.3 & 22.2 & 22.2 & \textemdash \\
\midrule[0.4pt]
\rowcolor{RISETaskGroup}
\multicolumn{25}{c}{\rule{0pt}{9pt}\raisebox{1.5pt}[0pt][0pt]{\textbf{Open-source Models}}} \\
\midrule[0.4pt]
\RISEFitModel{HunyuanImage-3.0-Instruct~\cite{cao2025hunyuanimage}} & \RISETaskYes & \underline{41.9} & \textbf{55.6} & \underline{40.9} & 0.0 & 0.0 & 10.0 & \underline{50.0} & \underline{20.0} & \underline{34.8} & \textbf{66.7} & 0.0 & \underline{0.0} & \underline{0.0} & 0.0 & \textbf{45.5} & \textbf{45.5} & \textbf{45.5} & 41.7 & \textbf{38.5} & \textbf{13.3} & \underline{27.8} & \underline{22.2} & 0.0 \\
\RISEFitModel{Qwen-Image-2.1~\cite{zhao2026qwen}} & \RISETaskYes & 38.7 & 33.3 & 22.7 & \textbf{20.0} & \textbf{20.0} & \underline{20.0} & 20.0 & 4.0 & 21.7 & 33.3 & 0.0 & \textbf{25.0} & \underline{0.0} & \underline{10.0} & 18.2 & \underline{36.4} & 18.2 & 25.0 & \underline{30.8} & \textbf{13.3} & 22.2 & \underline{22.2} & \textbf{33.3} \\
\RISEFitModel{FireRed-Image-Edit-1.1~\cite{team2026firered}} & \RISETaskYes & 38.7 & 40.7 & 18.2 & \textbf{20.0} & 0.0 & 10.0 & \textbf{60.0} & 12.0 & 17.4 & \underline{41.7} & 0.0 & \textbf{25.0} & \underline{0.0} & 0.0 & \underline{27.3} & 18.2 & 27.3 & \textbf{58.3} & 23.1 & \underline{6.7} & \textbf{38.9} & \underline{22.2} & 0.0 \\
\RISEFitModel{FLUX.2 Dev~\cite{bfl2025flux2}} & \RISETaskYes & 32.3 & 33.3 & 31.8 & 0.0 & 10.0 & 10.0 & \textbf{60.0} & 16.0 & 26.1 & \underline{41.7} & 0.0 & \textbf{25.0} & \underline{0.0} & \textbf{20.0} & 9.1 & 0.0 & \underline{36.4} & 25.0 & 15.4 & 0.0 & 5.6 & 11.1 & \underline{16.7} \\
\RISEFitModel{Qwen-Image-Edit-2511~\cite{wu2025qwen}} & \RISETaskYes & 38.7 & 33.3 & 27.3 & \underline{10.0} & 0.0 & 0.0 & 30.0 & 4.0 & 8.7 & 33.3 & 0.0 & \textbf{25.0} & \underline{0.0} & 0.0 & 9.1 & \underline{36.4} & \textbf{45.5} & 8.3 & \underline{30.8} & \underline{6.7} & \underline{27.8} & \underline{22.2} & \underline{16.7} \\
\RISEFitModel{Qwen-Image-Edit-2509~\cite{wu2025qwen}} & \RISETaskYes & 25.8 & 25.9 & 36.4 & \underline{10.0} & 0.0 & 0.0 & \textbf{60.0} & 8.0 & 8.7 & 33.3 & 0.0 & \underline{0.0} & \underline{0.0} & 0.0 & 9.1 & 27.3 & \underline{36.4} & 25.0 & 23.1 & \textbf{13.3} & 22.2 & \textbf{33.3} & \underline{16.7} \\
\RISEFitModel{FireRed-Image-Edit-1.0~\cite{team2026firered}} & \RISETaskYes & 22.6 & 33.3 & 22.7 & 0.0 & 10.0 & \textbf{30.0} & 30.0 & 4.0 & 4.3 & 16.7 & 0.0 & \underline{0.0} & \underline{0.0} & \underline{10.0} & 9.1 & \underline{36.4} & 27.3 & \underline{50.0} & \underline{30.8} & \underline{6.7} & 16.7 & \underline{22.2} & \underline{16.7} \\
\RISEFitModel{SenseNova-U1-A3B-MoT-SFT~\cite{diao2026sensenova}} & \RISETaskYes & 32.3 & 37.0 & 36.4 & 0.0 & 0.0 & 0.0 & 20.0 & \textbf{24.0} & \textbf{39.1} & \textbf{66.7} & \textbf{66.7} & \textbf{25.0} & \underline{0.0} & 0.0 & 9.1 & 0.0 & 9.1 & 16.7 & 7.7 & 0.0 & 5.6 & 0.0 & 0.0 \\
\RISEFitModel{JoyAI-Image-Edit-Plus~\cite{song2026joyai}} & \RISETaskYes & \underline{41.9} & 25.9 & 22.7 & \underline{10.0} & 0.0 & \underline{20.0} & 30.0 & 12.0 & 8.7 & 8.3 & \underline{33.3} & \underline{0.0} & \underline{0.0} & \underline{10.0} & 9.1 & 9.1 & 27.3 & 16.7 & 23.1 & \underline{6.7} & 22.2 & 11.1 & \underline{16.7} \\
\RISEFitModel{FLUX.2 Klein 9B~\cite{bfl2025flux2}} & \RISETaskYes & 25.8 & 22.2 & 22.7 & 0.0 & 0.0 & \underline{20.0} & \textbf{60.0} & 0.0 & 4.3 & 16.7 & 0.0 & \textbf{25.0} & \underline{0.0} & \underline{10.0} & 0.0 & 9.1 & 9.1 & 41.7 & 15.4 & \underline{6.7} & 11.1 & 11.1 & 0.0 \\
\RISEFitModel{BAGEL (w/ CoT)~\cite{deng2025bagel}} & \RISETaskYes & 29.0 & 25.9 & 31.8 & \textbf{20.0} & \textbf{20.0} & 10.0 & 20.0 & 0.0 & 8.7 & 33.3 & 0.0 & \underline{0.0} & \underline{0.0} & 0.0 & 0.0 & 9.1 & 27.3 & 16.7 & 7.7 & 0.0 & 11.1 & 0.0 & \underline{16.7} \\
\RISEFitModel{HiDream-O1-Image~\cite{cai2026hidream}} & \RISETaskYes & 25.8 & 29.6 & 18.2 & 0.0 & 0.0 & 0.0 & 10.0 & 0.0 & 8.7 & 16.7 & 0.0 & \underline{0.0} & \underline{0.0} & \textbf{20.0} & 0.0 & 18.2 & 27.3 & 8.3 & 0.0 & 0.0 & 16.7 & 0.0 & \underline{16.7} \\
\RISEFitModel{FLUX.2 Klein 4B~\cite{bfl2025flux2}} & \RISETaskYes & 22.6 & 14.8 & 4.5 & \underline{10.0} & 10.0 & 10.0 & 30.0 & 0.0 & 13.0 & 16.7 & 0.0 & \underline{0.0} & \underline{0.0} & \textbf{20.0} & 0.0 & 0.0 & 18.2 & 8.3 & 15.4 & \textbf{13.3} & 0.0 & \textbf{33.3} & 0.0 \\
\RISEFitModel{UniWorld-V2~\cite{li2025uniworld}} & \RISETaskYes & 35.5 & 29.6 & 22.7 & 0.0 & 0.0 & 10.0 & 30.0 & 8.0 & 8.7 & 16.7 & 0.0 & \underline{0.0} & \underline{0.0} & 0.0 & 18.2 & 0.0 & 18.2 & 16.7 & \textbf{38.5} & \underline{6.7} & 22.2 & 11.1 & 0.0 \\
\RISEFitModel{SenseNova-U1-8B-MoT-SFT~\cite{diao2026sensenova}} & \RISETaskYes & 32.3 & 33.3 & 27.3 & 0.0 & 0.0 & 10.0 & 40.0 & 8.0 & 17.4 & \underline{41.7} & 0.0 & \textbf{25.0} & \underline{0.0} & 0.0 & 0.0 & 0.0 & 9.1 & 0.0 & 0.0 & 0.0 & 5.6 & 11.1 & 0.0 \\
\RISEFitModel{UniPic 3.0~\cite{wei2026skywork}} & \RISETaskYes & 25.8 & 22.2 & 18.2 & 0.0 & 0.0 & 0.0 & 30.0 & 8.0 & 8.7 & 16.7 & \underline{33.3} & \underline{0.0} & \underline{0.0} & 0.0 & 18.2 & 9.1 & 9.1 & 8.3 & 23.1 & \underline{6.7} & 11.1 & 11.1 & 0.0 \\
\RISEFitModel{BAGEL~\cite{deng2025bagel}} & \RISETaskYes & 19.4 & 14.8 & 4.5 & 0.0 & 10.0 & 10.0 & 0.0 & 4.0 & 0.0 & 0.0 & 0.0 & \underline{0.0} & \underline{0.0} & 0.0 & 0.0 & 27.3 & 9.1 & 0.0 & 7.7 & 0.0 & 0.0 & 11.1 & 0.0 \\
\RISEFitModel{OmniGen2~\cite{wu2025omnigen2}} & \RISETaskYes & 6.5 & 7.4 & 0.0 & 0.0 & \textbf{20.0} & 0.0 & 20.0 & 4.0 & 4.3 & 0.0 & 0.0 & \underline{0.0} & \underline{0.0} & \underline{10.0} & 0.0 & 0.0 & 9.1 & 0.0 & 0.0 & 0.0 & 0.0 & 0.0 & 0.0 \\
\RISEFitModel{OmniGen~\cite{xiao2024omnigen}} & \RISETaskYes & 0.0 & 3.7 & 0.0 & 0.0 & 0.0 & 0.0 & 0.0 & 0.0 & 13.0 & 0.0 & 0.0 & \underline{0.0} & \underline{0.0} & \underline{10.0} & 9.1 & 0.0 & 0.0 & 0.0 & 0.0 & 0.0 & 0.0 & 0.0 & \textbf{33.3} \\
\RISEFitModel{Emu2-Gen~\cite{sun2023emu2chat}} & \RISETaskYes & 0.0 & 0.0 & 4.5 & 0.0 & 0.0 & 10.0 & 0.0 & 0.0 & 0.0 & 0.0 & 0.0 & \underline{0.0} & \underline{0.0} & 0.0 & 0.0 & 0.0 & 0.0 & 0.0 & 0.0 & 0.0 & 0.0 & 0.0 & \underline{16.7} \\
\RISEFitModel{JoyAI-Image-Edit~\cite{song2026joyai}} & \RISETaskNo & \textbf{45.2} & 25.9 & \textbf{47.6} & \textbf{20.0} & \underline{12.5} & 10.0 & 40.0 & 4.2 & 10.5 & 33.3 & 0.0 & \underline{0.0} & \underline{0.0} & 0.0 & 9.1 & 0.0 & \textbf{45.5} & 16.7 & 7.7 & \underline{6.7} & 22.2 & 0.0 & \textemdash \\
\RISEFitModel{Boogu-Image-0.1-Edit~\cite{chen2026boogu}} & \RISETaskNo & 32.3 & \underline{48.1} & 23.8 & 0.0 & \underline{12.5} & \underline{20.0} & 40.0 & 8.3 & 15.8 & 8.3 & 0.0 & \textbf{25.0} & \underline{0.0} & \underline{10.0} & 9.1 & 27.3 & 27.3 & 16.7 & 15.4 & \underline{6.7} & 11.1 & 0.0 & \textemdash \\
\RISEFitModel{Step1X-Edit (thinking + reflection)~\cite{liu2025step1x}} & \RISETaskNo & 19.4 & 29.6 & 14.3 & \textbf{20.0} & 0.0 & 10.0 & \textbf{60.0} & 4.2 & 26.3 & 16.7 & 0.0 & \underline{0.0} & \underline{0.0} & \underline{10.0} & 9.1 & 18.2 & 27.3 & 16.7 & 7.7 & \underline{6.7} & 5.6 & 0.0 & \textemdash \\
\RISEFitModel{Step1X-Edit (thinking)~\cite{liu2025step1x}} & \RISETaskNo & 25.8 & 22.2 & 14.3 & 0.0 & \underline{12.5} & 0.0 & 20.0 & 8.3 & 15.8 & 16.7 & 0.0 & \underline{0.0} & \textbf{33.3} & \underline{10.0} & 9.1 & 0.0 & 27.3 & 16.7 & 0.0 & \underline{6.7} & 22.2 & 0.0 & \textemdash \\
\RISEFitModel{Step1X-Edit (base)~\cite{liu2025step1x}} & \RISETaskNo & 25.8 & 25.9 & 19.0 & 0.0 & 0.0 & 0.0 & \underline{50.0} & 8.3 & 10.5 & 16.7 & 0.0 & \underline{0.0} & \underline{0.0} & \underline{10.0} & 0.0 & 9.1 & \underline{36.4} & 16.7 & 7.7 & \underline{6.7} & 22.2 & 0.0 & \textemdash \\
\RISEFitModel{LongCat-Image-Edit~\cite{team2025longcat}} & \RISETaskNo & 32.3 & 14.8 & 19.0 & \underline{10.0} & 0.0 & \underline{20.0} & 40.0 & 8.3 & 5.3 & 25.0 & 0.0 & \underline{0.0} & \underline{0.0} & 0.0 & 0.0 & 18.2 & 27.3 & 25.0 & 7.7 & \underline{6.7} & 11.1 & 11.1 & \textemdash \\
\RISEFitModel{FLUX.1 Kontext Dev~\cite{labs2025flux1kontextflowmatching}} & \RISETaskNo & 6.5 & 3.7 & 4.8 & 0.0 & 0.0 & 0.0 & 0.0 & 0.0 & 5.3 & 0.0 & 0.0 & \underline{0.0} & \textbf{33.3} & \underline{10.0} & 0.0 & 0.0 & 9.1 & 0.0 & 0.0 & 0.0 & 0.0 & 0.0 & \textemdash \\
\RISEFitModel{LLaDA-Image~\cite{chen2026llada}} & \RISETaskNo & 12.9 & 11.1 & 4.8 & 0.0 & 0.0 & 0.0 & 0.0 & 0.0 & 5.3 & 16.7 & 0.0 & \underline{0.0} & \underline{0.0} & 0.0 & 0.0 & 0.0 & 9.1 & 0.0 & 0.0 & \underline{6.7} & 0.0 & 0.0 & \textemdash \\
\RISEFitModel{InternVL-U (w/ CoT)~\cite{tian2026internvl}} & \RISETaskNo & 9.7 & 0.0 & 9.5 & 0.0 & 0.0 & 10.0 & 0.0 & 0.0 & 0.0 & 0.0 & 0.0 & \underline{0.0} & \underline{0.0} & 0.0 & 0.0 & 0.0 & 0.0 & 0.0 & 0.0 & 0.0 & 0.0 & 0.0 & \textemdash \\
\RISEFitModel{Ovis-U1~\cite{wang2025ovis}} & \RISETaskNo & 16.1 & 3.7 & 0.0 & 0.0 & 0.0 & 10.0 & 10.0 & 8.3 & 0.0 & 0.0 & 0.0 & \underline{0.0} & \underline{0.0} & \underline{10.0} & 0.0 & 0.0 & 0.0 & 0.0 & 0.0 & 0.0 & 0.0 & 0.0 & \textemdash \\
\RISEFitModel{InternVL-U~\cite{tian2026internvl}} & \RISETaskNo & 12.9 & 7.4 & 0.0 & 0.0 & 0.0 & 0.0 & 10.0 & 0.0 & 5.3 & 0.0 & 0.0 & \underline{0.0} & \underline{0.0} & 0.0 & 0.0 & 0.0 & 0.0 & 0.0 & 0.0 & 0.0 & 0.0 & 0.0 & \textemdash \\
\RISEFitModel{Lumina-DiMOO~\cite{xin2025lumina}} & \RISETaskNo & 6.5 & 3.7 & 4.8 & 0.0 & 0.0 & 10.0 & 0.0 & 0.0 & 0.0 & 0.0 & 0.0 & \underline{0.0} & \underline{0.0} & 0.0 & 9.1 & 0.0 & 9.1 & 0.0 & 0.0 & 0.0 & 0.0 & 11.1 & \textemdash \\
\RISEFitModel{UniWorld-V1~\cite{lin2025uniworld}} & \RISETaskNo & 0.0 & 3.7 & 0.0 & 0.0 & 0.0 & 0.0 & 0.0 & 0.0 & 5.3 & 0.0 & 0.0 & \underline{0.0} & \underline{0.0} & 0.0 & 0.0 & 0.0 & 0.0 & 0.0 & 0.0 & 0.0 & 0.0 & 11.1 & \textemdash \\
\RISEFitModel{UniPic2-SD3.5M-Kontext-2B~\cite{wei2025skywork}} & \RISETaskNo & 3.2 & 0.0 & 0.0 & 0.0 & 0.0 & 0.0 & 0.0 & 0.0 & 0.0 & 0.0 & 0.0 & \underline{0.0} & \underline{0.0} & \underline{10.0} & 0.0 & 0.0 & 0.0 & 0.0 & 7.7 & 0.0 & 0.0 & 0.0 & \textemdash \\
\midrule[0.4pt]
\rowcolor{RISETaskGroup}
\multicolumn{25}{c}{\rule{0pt}{9pt}\raisebox{1.5pt}[0pt][0pt]{\textbf{Agentic Methods}}} \\
\midrule[0.4pt]
\RISEFitModel{\textbf{RISE-Agent (Ours)}} & \RISETaskYes & \textbf{74.2} & \textbf{63.0} & \textbf{63.6} & \textbf{40.0} & \textbf{30.0} & \textbf{60.0} & \textbf{50.0} & \textbf{40.0} & \textbf{69.6} & \textbf{83.3} & \textbf{66.7} & \textbf{50.0} & \textbf{33.3} & \textbf{10.0} & \underline{18.2} & \textbf{27.3} & \textbf{45.5} & \textbf{41.7} & \textbf{23.1} & \textbf{20.0} & \textbf{33.3} & \textbf{44.4} & \textbf{33.3} \\
\RISEFitModel{MindBrush~\cite{he2026mind}} & \RISETaskNo & \underline{51.6} & \underline{44.4} & 28.6 & \underline{10.0} & \underline{12.5} & 10.0 & \underline{20.0} & \underline{25.0} & \underline{52.6} & \underline{66.7} & \textbf{66.7} & 0.0 & \underline{0.0} & \underline{0.0} & \textbf{27.3} & \underline{18.2} & \underline{27.3} & \underline{25.0} & \underline{15.4} & \underline{13.3} & 11.1 & \underline{11.1} & \textemdash \\
\RISEFitModel{InterleaveThinker~\cite{zheng2026interleavethinker}} & \RISETaskNo & 29.0 & 37.0 & 23.8 & \underline{10.0} & \underline{12.5} & \underline{20.0} & \textbf{50.0} & \underline{25.0} & 47.4 & 41.7 & 0.0 & \underline{25.0} & \underline{0.0} & \underline{0.0} & 0.0 & \underline{18.2} & \underline{27.3} & 16.7 & \textbf{23.1} & \textbf{20.0} & 11.1 & \underline{11.1} & \textemdash \\
\RISEFitModel{IntentEdit~\cite{zhang2026intentedit}} & \RISETaskNo & 35.5 & 29.6 & \underline{47.6} & \underline{10.0} & 0.0 & 0.0 & \underline{20.0} & 12.5 & 31.6 & 25.0 & \underline{33.3} & \underline{25.0} & \underline{0.0} & \underline{0.0} & 9.1 & 9.1 & 18.2 & 8.3 & \textbf{23.1} & 0.0 & \underline{16.7} & \underline{11.1} & \textemdash \\
\RISEFitModel{ImAgent~\cite{wang2025imagent}} & \RISETaskNo & 9.7 & 3.7 & 4.8 & 0.0 & \underline{12.5} & 0.0 & 0.0 & 4.2 & 0.0 & 0.0 & 0.0 & 0.0 & \underline{0.0} & \underline{0.0} & 0.0 & 0.0 & 0.0 & 0.0 & 7.7 & 0.0 & 0.0 & 0.0 & \textemdash \\
\bottomrule[0.8pt]
\end{tabular}
\par\vspace{4pt}
\parbox{\textwidth}{\fontsize{6.5}{8.2}\selectfont\raggedright
\textbf{C01}: Physical State Transition; \textbf{C02}: Structural Damage; \textbf{C03}: Flexible Deformation; \textbf{C04}: Magnitude Change; \textbf{C05}: Reading Offset; \textbf{C06}: Order Relation; \textbf{C07}: Feature Mapping; \textbf{C08}: Physics; \textbf{C09}: Chemistry; \textbf{C10}: Biology; \textbf{C11}: Astronomy; \textbf{C12}: Medicine; \textbf{C13}: Music Theory; \textbf{C14}: Computer Science.\par
\textbf{S01}: Adjacency; \textbf{S02}: Direction; \textbf{S03}: Orientation; \textbf{S04}: Projection; \textbf{S05}: Distance; \textbf{S06}: Size; \textbf{S07}: Configuration; \textbf{S08}: Viewing Scale; \textbf{S09}: Localization.
}
\endgroup
\end{table*}

\begin{table*}[!p]
\caption{Fine-grained task accuracy (\%) on RISEBench++ (ZH). Part III: Spatial Structure, Logical reasoning, and Counterfactual reasoning.}
\label{tab:task_accuracy_cn_3}
\centering
\begingroup
\fontsize{6.2}{8.2}\selectfont
\setlength{\tabcolsep}{1.2pt}
\renewcommand{\arraystretch}{1.0}
\setlength{\aboverulesep}{1pt}
\setlength{\belowrulesep}{1pt}
\colorlet{RISETaskGroup}{black!6}
\def\RISETaskYes{\textcolor{green!50!black}{\ding{51}}}
\def\RISETaskNo{\textcolor{red!65!black}{\ding{55}}}
\begin{tabular}{
>{\raggedright\arraybackslash}m{\dimexpr .19\textwidth-2\tabcolsep\relax}
>{\centering\arraybackslash}m{\dimexpr .04\textwidth-2\tabcolsep\relax}
*{5}{>{\centering\arraybackslash}m{\dimexpr (.77\textwidth-6pt)/23-2\tabcolsep\relax}}!{\hspace{3pt}}
*{7}{>{\centering\arraybackslash}m{\dimexpr (.77\textwidth-6pt)/23-2\tabcolsep\relax}}!{\hspace{3pt}}
*{11}{>{\centering\arraybackslash}m{\dimexpr (.77\textwidth-6pt)/23-2\tabcolsep\relax}}
}
\toprule[0.8pt]
\multirow{2}{=}[-1.3pt]{\raggedright\textbf{Model}} & \multirow{2}{=}[-1.3pt]{\centering\shortstack{\textbf{Multi-}\\\textbf{img}}} & \multicolumn{5}{c}{\textbf{Spatial (Structure)}} & \multicolumn{7}{c}{\textbf{Logical}} & \multicolumn{11}{c}{\textbf{Counterfactual}} \\
\cmidrule(lr){3-7}\cmidrule(lr){8-14}\cmidrule(lr){15-25}
 & & \textbf{S10} & \textbf{S11} & \textbf{S12} & \textbf{S13} & \textbf{S14} & \textbf{L01} & \textbf{L02} & \textbf{L03} & \textbf{L04} & \textbf{L05} & \textbf{L06} & \textbf{L07} & \textbf{F01} & \textbf{F02} & \textbf{F03} & \textbf{F04} & \textbf{F05} & \textbf{F06} & \textbf{F07} & \textbf{F08} & \textbf{F09} & \textbf{F10} & \textbf{F11} \\
\midrule[0.5pt]
\rowcolor{RISETaskGroup}
\multicolumn{25}{c}{\rule{0pt}{9pt}\raisebox{1.5pt}[0pt][0pt]{\textbf{Closed-source Models}}} \\
\midrule[0.4pt]
\RISEFitModel{GPT-Image-2.5 Sunburst~\cite{openai2025gptimage}} & \RISETaskYes & \textbf{79.2} & 66.7 & \textbf{65.4} & 47.4 & 50.0 & 20.8 & 25.9 & 43.2 & \textbf{34.3} & 15.4 & \textbf{55.6} & 10.0 & 73.9 & 66.7 & 75.0 & 28.6 & \textbf{75.0} & \underline{71.4} & \textbf{25.0} & 50.0 & \RISEFitScore{\textbf{100.0}} & \textbf{81.8} & \underline{55.6} \\
\RISEFitModel{Nano Banana Pro~\cite{google2026nanobanana}} & \RISETaskYes & \underline{75.0} & \underline{73.3} & 30.8 & 31.6 & 0.0 & \textbf{41.7} & \textbf{63.0} & \underline{45.9} & 22.9 & \textbf{30.8} & 33.3 & \underline{35.0} & 60.9 & \underline{88.9} & 62.5 & 42.9 & \underline{50.0} & 28.6 & \underline{0.0} & 50.0 & 60.0 & 18.2 & 44.4 \\
\RISEFitModel{Nano Banana 2~\cite{google2026nanobanana}} & \RISETaskYes & 66.7 & \underline{73.3} & 50.0 & 31.6 & 40.0 & 20.8 & \underline{55.6} & 37.8 & \underline{25.7} & 23.1 & \underline{38.9} & 20.0 & 43.5 & 66.7 & 62.5 & 42.9 & \underline{50.0} & 57.1 & \textbf{25.0} & \underline{75.0} & \underline{90.0} & 63.6 & \underline{55.6} \\
\RISEFitModel{Luma Uni-1~\cite{luma2026uni1}} & \RISETaskYes & 66.7 & \textbf{80.0} & 34.6 & 47.4 & 50.0 & 4.2 & 37.0 & 18.9 & 22.9 & 7.7 & 16.7 & 10.0 & 65.2 & 77.8 & 50.0 & 57.1 & \underline{50.0} & 57.1 & \textbf{25.0} & 50.0 & 50.0 & 54.5 & 44.4 \\
\RISEFitModel{GPT-Image-2~\cite{openai2025gptimage}} & \RISETaskYes & 62.5 & 66.7 & 42.3 & \underline{57.9} & 20.0 & 12.5 & 25.9 & 32.4 & 17.1 & 15.4 & 5.6 & 30.0 & 65.2 & 55.6 & \underline{87.5} & 14.3 & \underline{50.0} & 42.9 & \textbf{25.0} & 25.0 & \underline{90.0} & \textbf{81.8} & \textbf{66.7} \\
\RISEFitModel{Qwen-Image-3.0~\cite{zhao2026qwen}} & \RISETaskYes & 29.2 & \textbf{80.0} & 46.2 & 26.3 & 20.0 & \underline{25.0} & \textbf{63.0} & \textbf{54.1} & \textbf{34.3} & \underline{25.6} & 27.8 & \textbf{40.0} & 56.5 & 44.4 & 25.0 & 14.3 & \underline{50.0} & 28.6 & \textbf{25.0} & \underline{75.0} & 70.0 & \underline{72.7} & 44.4 \\
\RISEFitModel{GPT-Image-2.5 Flare~\cite{openai2025gptimage}} & \RISETaskYes & \underline{75.0} & 60.0 & \underline{53.8} & \textbf{63.2} & 0.0 & 12.5 & 11.1 & 35.1 & 14.3 & 7.7 & 0.0 & 5.0 & \textbf{82.6} & 33.3 & 62.5 & 28.6 & \underline{50.0} & 42.9 & \textbf{25.0} & 50.0 & 80.0 & \underline{72.7} & \textbf{66.7} \\
\RISEFitModel{FLUX.2 Max~\cite{bfl2025flux2}} & \RISETaskYes & 70.8 & 46.7 & 42.3 & 42.1 & 40.0 & 4.2 & 40.7 & 24.3 & 14.3 & 2.6 & 5.6 & 0.0 & \underline{78.3} & \RISEFitScore{\textbf{100.0}} & 75.0 & \textbf{85.7} & 37.5 & \RISEFitScore{\textbf{100.0}} & \underline{0.0} & \underline{75.0} & 60.0 & \textbf{81.8} & \underline{55.6} \\
\RISEFitModel{Nano Banana 2 Lite~\cite{google2026nanobanana}} & \RISETaskYes & 66.7 & 60.0 & 42.3 & 26.3 & 10.0 & 12.5 & 18.5 & 10.8 & 11.4 & 7.7 & 22.2 & 5.0 & 65.2 & 66.7 & 75.0 & \underline{71.4} & 37.5 & 57.1 & \underline{0.0} & \underline{75.0} & 80.0 & 36.4 & \underline{55.6} \\
\RISEFitModel{GPT-Image-1.5~\cite{openai2025gptimage}} & \RISETaskYes & 58.3 & 20.0 & 38.5 & 42.1 & 20.0 & 4.2 & 0.0 & 10.8 & 2.9 & 7.7 & 0.0 & 0.0 & 56.5 & 55.6 & \RISEFitScore{\textbf{100.0}} & 42.9 & \underline{50.0} & 28.6 & \underline{0.0} & 50.0 & 80.0 & 54.5 & \textbf{66.7} \\
\RISEFitModel{Nano Banana~\cite{google2026nanobanana}} & \RISETaskYes & 54.2 & 53.3 & 46.2 & 52.6 & 30.0 & 4.2 & 29.6 & 10.8 & 14.3 & 7.7 & 11.1 & 5.0 & 69.6 & 55.6 & 37.5 & 14.3 & \underline{50.0} & 28.6 & \textbf{25.0} & 25.0 & 80.0 & 63.6 & 44.4 \\
\RISEFitModel{FLUX.2 Pro~\cite{bfl2025flux2}} & \RISETaskYes & 50.0 & 46.7 & 30.8 & 47.4 & \underline{80.0} & 0.0 & 29.6 & 13.5 & 5.7 & 2.6 & 0.0 & 0.0 & 52.2 & 77.8 & 50.0 & \underline{71.4} & 25.0 & \underline{71.4} & \underline{0.0} & \RISEFitScore{\textbf{100.0}} & 30.0 & 54.5 & 33.3 \\
\RISEFitModel{HunyuanImage-3.5~\cite{cao2025hunyuanimage}} & \RISETaskYes & 62.5 & 40.0 & 30.8 & 36.8 & \textbf{90.0} & 4.2 & 25.9 & 16.2 & 11.4 & 10.3 & 33.3 & 0.0 & 52.2 & 77.8 & 25.0 & 14.3 & 12.5 & 14.3 & \textbf{25.0} & \underline{75.0} & 60.0 & 36.4 & 44.4 \\
\RISEFitModel{Wan2.7-Image-Pro~\cite{mao2026wan}} & \RISETaskYes & 33.3 & 60.0 & 30.8 & 21.1 & 60.0 & 0.0 & 22.2 & 16.2 & 5.7 & 2.6 & 22.2 & 0.0 & 52.2 & 44.4 & 75.0 & 28.6 & 37.5 & 28.6 & \underline{0.0} & 50.0 & 60.0 & 63.6 & 33.3 \\
\RISEFitModel{Qwen-Image-2.0~\cite{zhao2026qwen}} & \RISETaskYes & 37.5 & 53.3 & 26.9 & 21.1 & \underline{80.0} & 0.0 & 11.1 & 13.5 & 11.4 & 10.3 & 11.1 & 0.0 & 43.5 & 44.4 & 62.5 & 14.3 & 25.0 & 14.3 & \underline{0.0} & 25.0 & 40.0 & 45.5 & 44.4 \\
\RISEFitModel{Seedream 4.5~\cite{seedream2025seedream}} & \RISETaskYes & 37.5 & 40.0 & 34.6 & 10.5 & 20.0 & 0.0 & 18.5 & 16.2 & 5.7 & 5.1 & 5.6 & 5.0 & 30.4 & 55.6 & 75.0 & 14.3 & 12.5 & 14.3 & \underline{0.0} & 50.0 & 40.0 & 45.5 & 33.3 \\
\RISEFitModel{Seedream 5.0~\cite{seedream2025seedream}} & \RISETaskYes & 37.5 & 60.0 & 23.1 & 31.6 & 40.0 & 0.0 & 18.5 & 18.9 & 14.3 & 15.4 & 0.0 & 0.0 & 30.4 & 33.3 & 50.0 & 0.0 & 25.0 & 28.6 & \underline{0.0} & 50.0 & 50.0 & 18.2 & 33.3 \\
\RISEFitModel{GPT-Image-1~\cite{openai2025gptimage}} & \RISETaskYes & 54.2 & 33.3 & 34.6 & 31.6 & 10.0 & 4.2 & 3.7 & 10.8 & 2.9 & 7.7 & 5.6 & 0.0 & 39.1 & 66.7 & 50.0 & 14.3 & \underline{50.0} & 42.9 & \underline{0.0} & 0.0 & 60.0 & 36.4 & 0.0 \\
\RISEFitModel{Step Image Edit 2~\cite{liu2025step1x}} & \RISETaskNo & 12.5 & 20.0 & 34.6 & 11.8 & \textemdash & 4.2 & 0.0 & 0.0 & 5.7 & 2.6 & 0.0 & \textemdash & 21.7 & 11.1 & 12.5 & 0.0 & 0.0 & 0.0 & \underline{0.0} & 0.0 & 20.0 & 18.2 & 11.1 \\
\midrule[0.4pt]
\rowcolor{RISETaskGroup}
\multicolumn{25}{c}{\rule{0pt}{9pt}\raisebox{1.5pt}[0pt][0pt]{\textbf{Open-source Models}}} \\
\midrule[0.4pt]
\RISEFitModel{HunyuanImage-3.0-Instruct~\cite{cao2025hunyuanimage}} & \RISETaskYes & \textbf{45.8} & 33.3 & 26.9 & 26.3 & \underline{80.0} & \textbf{16.7} & \textbf{3.7} & \textbf{8.1} & \textbf{17.1} & 7.7 & 0.0 & 0.0 & 17.4 & \underline{33.3} & \underline{12.5} & \underline{14.3} & \underline{12.5} & \underline{14.3} & \textbf{0.0} & \textbf{0.0} & \underline{50.0} & 18.2 & \underline{22.2} \\
\RISEFitModel{Qwen-Image-2.1~\cite{zhao2026qwen}} & \RISETaskYes & 16.7 & \underline{40.0} & \underline{30.8} & 26.3 & 20.0 & 4.2 & \textbf{3.7} & 0.0 & \textbf{17.1} & 7.7 & \textbf{11.1} & 0.0 & \textbf{34.8} & 22.2 & \underline{12.5} & 0.0 & 0.0 & 0.0 & \textbf{0.0} & \textbf{0.0} & 10.0 & 27.3 & 11.1 \\
\RISEFitModel{FireRed-Image-Edit-1.1~\cite{team2026firered}} & \RISETaskYes & 4.2 & \underline{40.0} & \textbf{34.6} & 15.8 & 30.0 & 0.0 & \underline{0.0} & 0.0 & 2.9 & 0.0 & 0.0 & \underline{5.0} & 17.4 & 22.2 & 0.0 & 0.0 & 0.0 & 0.0 & \textbf{0.0} & \textbf{0.0} & 30.0 & \underline{36.4} & \underline{22.2} \\
\RISEFitModel{FLUX.2 Dev~\cite{bfl2025flux2}} & \RISETaskYes & 25.0 & 26.7 & 15.4 & 10.5 & 40.0 & 0.0 & \underline{0.0} & 0.0 & 5.7 & 2.6 & 0.0 & 0.0 & 8.7 & \textbf{55.6} & 0.0 & \underline{14.3} & 0.0 & 0.0 & \textbf{0.0} & \textbf{0.0} & 20.0 & 18.2 & 11.1 \\
\RISEFitModel{Qwen-Image-Edit-2511~\cite{wu2025qwen}} & \RISETaskYes & 8.3 & \textbf{46.7} & \underline{30.8} & 15.8 & 20.0 & 0.0 & \underline{0.0} & \underline{5.4} & 2.9 & 2.6 & 0.0 & 0.0 & 8.7 & 11.1 & 0.0 & 0.0 & 0.0 & 0.0 & \textbf{0.0} & \textbf{0.0} & 20.0 & \textbf{45.5} & 0.0 \\
\RISEFitModel{Qwen-Image-Edit-2509~\cite{wu2025qwen}} & \RISETaskYes & 12.5 & 33.3 & 23.1 & 26.3 & 0.0 & 0.0 & \underline{0.0} & 0.0 & 8.6 & 0.0 & 0.0 & 0.0 & 8.7 & 11.1 & 0.0 & 0.0 & \underline{12.5} & \underline{14.3} & \textbf{0.0} & \textbf{0.0} & 40.0 & 27.3 & \textbf{33.3} \\
\RISEFitModel{FireRed-Image-Edit-1.0~\cite{team2026firered}} & \RISETaskYes & 16.7 & \textbf{46.7} & \textbf{34.6} & 10.5 & 30.0 & 0.0 & \underline{0.0} & 0.0 & 2.9 & 0.0 & 0.0 & 0.0 & 8.7 & 11.1 & 0.0 & 0.0 & 0.0 & 0.0 & \textbf{0.0} & \textbf{0.0} & 30.0 & 18.2 & \textbf{33.3} \\
\RISEFitModel{SenseNova-U1-A3B-MoT-SFT~\cite{diao2026sensenova}} & \RISETaskYes & 8.3 & 20.0 & 11.5 & 0.0 & 0.0 & 8.3 & \underline{0.0} & 0.0 & 5.7 & 2.6 & \underline{5.6} & 0.0 & 17.4 & \underline{33.3} & \textbf{25.0} & 0.0 & \underline{12.5} & 0.0 & \textbf{0.0} & \textbf{0.0} & 0.0 & 9.1 & 0.0 \\
\RISEFitModel{JoyAI-Image-Edit-Plus~\cite{song2026joyai}} & \RISETaskYes & 16.7 & 33.3 & 15.4 & 21.1 & 40.0 & 0.0 & \textbf{3.7} & 0.0 & 11.4 & \underline{10.3} & \underline{5.6} & 0.0 & 8.7 & 22.2 & 0.0 & 0.0 & 0.0 & \underline{14.3} & \textbf{0.0} & \textbf{0.0} & 20.0 & 9.1 & 11.1 \\
\RISEFitModel{FLUX.2 Klein 9B~\cite{bfl2025flux2}} & \RISETaskYes & 12.5 & 20.0 & 26.9 & 10.5 & 50.0 & 0.0 & \underline{0.0} & 0.0 & 0.0 & \textbf{15.4} & 0.0 & \underline{5.0} & 8.7 & \underline{33.3} & 0.0 & \underline{14.3} & 0.0 & 0.0 & \textbf{0.0} & \textbf{0.0} & 10.0 & 9.1 & 11.1 \\
\RISEFitModel{BAGEL (w/ CoT)~\cite{deng2025bagel}} & \RISETaskYes & \underline{29.2} & 6.7 & 15.4 & \textbf{31.6} & 50.0 & 0.0 & \underline{0.0} & 0.0 & 0.0 & 0.0 & 0.0 & 0.0 & 8.7 & 22.2 & 0.0 & 0.0 & 0.0 & 0.0 & \textbf{0.0} & \textbf{0.0} & 10.0 & 9.1 & 0.0 \\
\RISEFitModel{HiDream-O1-Image~\cite{cai2026hidream}} & \RISETaskYes & 8.3 & 33.3 & 19.2 & 15.8 & \RISEFitScore{\textbf{100.0}} & 0.0 & \underline{0.0} & 0.0 & 0.0 & 2.6 & 0.0 & 0.0 & 8.7 & 0.0 & \underline{12.5} & \underline{14.3} & 0.0 & 0.0 & \textbf{0.0} & \textbf{0.0} & 20.0 & 0.0 & 11.1 \\
\RISEFitModel{FLUX.2 Klein 4B~\cite{bfl2025flux2}} & \RISETaskYes & 4.2 & 26.7 & 15.4 & 10.5 & \underline{80.0} & 0.0 & \underline{0.0} & 0.0 & 5.7 & \underline{10.3} & 0.0 & \textbf{10.0} & 4.3 & \textbf{55.6} & 0.0 & 0.0 & 0.0 & 0.0 & \textbf{0.0} & \textbf{0.0} & 0.0 & 18.2 & \underline{22.2} \\
\RISEFitModel{UniWorld-V2~\cite{li2025uniworld}} & \RISETaskYes & 4.2 & 26.7 & 26.9 & 5.3 & 10.0 & 0.0 & \underline{0.0} & 0.0 & 0.0 & 0.0 & 0.0 & 0.0 & 8.7 & \underline{33.3} & 0.0 & 0.0 & 0.0 & 0.0 & \textbf{0.0} & \textbf{0.0} & 40.0 & 9.1 & \underline{22.2} \\
\RISEFitModel{SenseNova-U1-8B-MoT-SFT~\cite{diao2026sensenova}} & \RISETaskYes & 8.3 & 6.7 & 3.8 & 0.0 & 0.0 & \underline{12.5} & \underline{0.0} & 0.0 & 5.7 & 0.0 & \textbf{11.1} & 0.0 & 17.4 & \underline{33.3} & \textbf{25.0} & 0.0 & \underline{12.5} & 0.0 & \textbf{0.0} & \textbf{0.0} & 40.0 & 0.0 & 11.1 \\
\RISEFitModel{UniPic 3.0~\cite{wei2026skywork}} & \RISETaskYes & 12.5 & 13.3 & 15.4 & 15.8 & 10.0 & 0.0 & \underline{0.0} & 0.0 & 0.0 & 0.0 & 0.0 & 0.0 & 8.7 & 11.1 & \underline{12.5} & \underline{14.3} & \underline{12.5} & 0.0 & \textbf{0.0} & \textbf{0.0} & 0.0 & 9.1 & 0.0 \\
\RISEFitModel{BAGEL~\cite{deng2025bagel}} & \RISETaskYes & 16.7 & 6.7 & 3.8 & 0.0 & \underline{80.0} & 0.0 & \underline{0.0} & 0.0 & 2.9 & 2.6 & 0.0 & 0.0 & 13.0 & 11.1 & \underline{12.5} & 0.0 & 0.0 & \underline{14.3} & \textbf{0.0} & \textbf{0.0} & 10.0 & 0.0 & 11.1 \\
\RISEFitModel{OmniGen2~\cite{wu2025omnigen2}} & \RISETaskYes & 4.2 & 6.7 & 3.8 & 21.1 & 40.0 & 0.0 & \underline{0.0} & 0.0 & 0.0 & 0.0 & 0.0 & 0.0 & 13.0 & 22.2 & 0.0 & 0.0 & 0.0 & 0.0 & \textbf{0.0} & \textbf{0.0} & 0.0 & 9.1 & 0.0 \\
\RISEFitModel{OmniGen~\cite{xiao2024omnigen}} & \RISETaskYes & 4.2 & 0.0 & 7.7 & 15.8 & 10.0 & 0.0 & \underline{0.0} & 0.0 & \underline{14.3} & 2.6 & 0.0 & 0.0 & 0.0 & 11.1 & \underline{12.5} & 0.0 & 0.0 & 0.0 & \textbf{0.0} & \textbf{0.0} & 10.0 & 0.0 & 11.1 \\
\RISEFitModel{Emu2-Gen~\cite{sun2023emu2chat}} & \RISETaskYes & 4.2 & 0.0 & 0.0 & 0.0 & 0.0 & 0.0 & \underline{0.0} & 0.0 & 0.0 & 0.0 & 0.0 & 0.0 & 4.3 & 0.0 & 0.0 & 0.0 & 0.0 & 0.0 & \textbf{0.0} & \textbf{0.0} & 0.0 & 0.0 & 11.1 \\
\RISEFitModel{JoyAI-Image-Edit~\cite{song2026joyai}} & \RISETaskNo & 8.3 & 20.0 & \underline{30.8} & 5.9 & \textemdash & 0.0 & \underline{0.0} & 0.0 & 0.0 & 5.1 & 0.0 & \textemdash & 21.7 & 22.2 & \underline{12.5} & 0.0 & \underline{12.5} & \underline{14.3} & \textbf{0.0} & \textbf{0.0} & \textbf{60.0} & \underline{36.4} & 11.1 \\
\RISEFitModel{Boogu-Image-0.1-Edit~\cite{chen2026boogu}} & \RISETaskNo & 20.8 & 20.0 & 15.4 & \underline{29.4} & \textemdash & 0.0 & \textbf{3.7} & 2.7 & 2.9 & 2.6 & 0.0 & \textemdash & \textbf{34.8} & 22.2 & 0.0 & 0.0 & 0.0 & 0.0 & \textbf{0.0} & \textbf{0.0} & 20.0 & 27.3 & \underline{22.2} \\
\RISEFitModel{Step1X-Edit (thinking + reflection)~\cite{liu2025step1x}} & \RISETaskNo & 4.2 & 33.3 & 7.7 & 5.9 & \textemdash & 0.0 & \underline{0.0} & 0.0 & 5.7 & 0.0 & 0.0 & \textemdash & \underline{30.4} & 11.1 & 0.0 & 0.0 & \underline{12.5} & 0.0 & \textbf{0.0} & \textbf{0.0} & 10.0 & 9.1 & \textbf{33.3} \\
\RISEFitModel{Step1X-Edit (thinking)~\cite{liu2025step1x}} & \RISETaskNo & 8.3 & 33.3 & 11.5 & 17.6 & \textemdash & 0.0 & \underline{0.0} & 2.7 & 2.9 & 5.1 & 0.0 & \textemdash & 26.1 & 22.2 & 0.0 & 0.0 & \underline{12.5} & \textbf{28.6} & \textbf{0.0} & \textbf{0.0} & 30.0 & 9.1 & \underline{22.2} \\
\RISEFitModel{Step1X-Edit (base)~\cite{liu2025step1x}} & \RISETaskNo & 12.5 & 20.0 & 7.7 & 11.8 & \textemdash & 0.0 & \underline{0.0} & 2.7 & 0.0 & 0.0 & 0.0 & \textemdash & 21.7 & 22.2 & 0.0 & \textbf{28.6} & \underline{12.5} & 0.0 & \textbf{0.0} & \textbf{0.0} & 10.0 & 9.1 & \textbf{33.3} \\
\RISEFitModel{LongCat-Image-Edit~\cite{team2025longcat}} & \RISETaskNo & 12.5 & 20.0 & 15.4 & 5.9 & \textemdash & 0.0 & \underline{0.0} & 0.0 & 0.0 & 5.1 & 0.0 & \textemdash & 8.7 & 0.0 & 0.0 & \underline{14.3} & 0.0 & \underline{14.3} & \textbf{0.0} & \textbf{0.0} & 30.0 & 9.1 & 11.1 \\
\RISEFitModel{FLUX.1 Kontext Dev~\cite{labs2025flux1kontextflowmatching}} & \RISETaskNo & 0.0 & 0.0 & 11.5 & 17.6 & \textemdash & 0.0 & \underline{0.0} & 0.0 & \underline{14.3} & 2.6 & 0.0 & \textemdash & 8.7 & 11.1 & \underline{12.5} & 0.0 & 0.0 & 0.0 & \textbf{0.0} & \textbf{0.0} & 10.0 & 9.1 & 11.1 \\
\RISEFitModel{LLaDA-Image~\cite{chen2026llada}} & \RISETaskNo & 0.0 & 0.0 & 7.7 & 0.0 & \textemdash & 0.0 & \underline{0.0} & 0.0 & 0.0 & 2.6 & 0.0 & \textemdash & 0.0 & 22.2 & 0.0 & 0.0 & 0.0 & \textbf{28.6} & \textbf{0.0} & \textbf{0.0} & 20.0 & 9.1 & 11.1 \\
\RISEFitModel{InternVL-U (w/ CoT)~\cite{tian2026internvl}} & \RISETaskNo & 0.0 & 6.7 & 7.7 & 5.9 & \textemdash & 0.0 & \underline{0.0} & 0.0 & 0.0 & 0.0 & 0.0 & \textemdash & 4.3 & 0.0 & 0.0 & 0.0 & 0.0 & 0.0 & \textbf{0.0} & \textbf{0.0} & 20.0 & 9.1 & 0.0 \\
\RISEFitModel{Ovis-U1~\cite{wang2025ovis}} & \RISETaskNo & 8.3 & 6.7 & 3.8 & 0.0 & \textemdash & 0.0 & \underline{0.0} & 0.0 & 2.9 & 0.0 & 0.0 & \textemdash & 4.3 & 0.0 & 0.0 & 0.0 & 0.0 & 0.0 & \textbf{0.0} & \textbf{0.0} & 0.0 & 9.1 & 0.0 \\
\RISEFitModel{InternVL-U~\cite{tian2026internvl}} & \RISETaskNo & 0.0 & 0.0 & 3.8 & 5.9 & \textemdash & 4.2 & \underline{0.0} & 0.0 & 0.0 & 0.0 & 0.0 & \textemdash & 4.3 & 11.1 & 0.0 & 0.0 & 0.0 & 0.0 & \textbf{0.0} & \textbf{0.0} & 10.0 & 18.2 & 0.0 \\
\RISEFitModel{Lumina-DiMOO~\cite{xin2025lumina}} & \RISETaskNo & 0.0 & 0.0 & 3.8 & 0.0 & \textemdash & 0.0 & \underline{0.0} & 0.0 & 5.7 & 5.1 & 0.0 & \textemdash & 4.3 & 11.1 & 0.0 & 0.0 & 0.0 & 0.0 & \textbf{0.0} & \textbf{0.0} & 0.0 & 9.1 & 11.1 \\
\RISEFitModel{UniWorld-V1~\cite{lin2025uniworld}} & \RISETaskNo & 0.0 & 13.3 & 0.0 & 0.0 & \textemdash & 0.0 & \underline{0.0} & 0.0 & 0.0 & 0.0 & 0.0 & \textemdash & 0.0 & 0.0 & 0.0 & 0.0 & \textbf{25.0} & 0.0 & \textbf{0.0} & \textbf{0.0} & 10.0 & 0.0 & 0.0 \\
\RISEFitModel{UniPic2-SD3.5M-Kontext-2B~\cite{wei2025skywork}} & \RISETaskNo & 0.0 & 0.0 & 7.7 & 0.0 & \textemdash & 0.0 & \underline{0.0} & 0.0 & 5.7 & 0.0 & 0.0 & \textemdash & 0.0 & 11.1 & 0.0 & 0.0 & 0.0 & 0.0 & \textbf{0.0} & \textbf{0.0} & 0.0 & 9.1 & 0.0 \\
\midrule[0.4pt]
\rowcolor{RISETaskGroup}
\multicolumn{25}{c}{\rule{0pt}{9pt}\raisebox{1.5pt}[0pt][0pt]{\textbf{Agentic Methods}}} \\
\midrule[0.4pt]
\RISEFitModel{\textbf{RISE-Agent (Ours)}} & \RISETaskYes & \textbf{41.7} & \textbf{46.7} & \textbf{30.8} & \textbf{42.1} & \textbf{20.0} & \textbf{12.5} & \textbf{40.7} & \textbf{27.0} & \textbf{40.0} & \textbf{17.9} & \underline{0.0} & \textbf{25.0} & \textbf{60.9} & \underline{55.6} & \textbf{75.0} & 28.6 & \textbf{50.0} & \textbf{85.7} & \textbf{0.0} & \textbf{50.0} & \textbf{70.0} & \textbf{63.6} & \underline{11.1} \\
\RISEFitModel{MindBrush~\cite{he2026mind}} & \RISETaskNo & 25.0 & \underline{33.3} & 11.5 & \underline{23.5} & \textemdash & \underline{0.0} & \underline{11.1} & \underline{5.4} & \underline{2.9} & 2.6 & \textbf{5.6} & \textemdash & \underline{39.1} & \textbf{88.9} & 37.5 & \textbf{57.1} & \underline{25.0} & \underline{57.1} & \textbf{0.0} & \underline{25.0} & \underline{30.0} & 9.1 & \underline{11.1} \\
\RISEFitModel{InterleaveThinker~\cite{zheng2026interleavethinker}} & \RISETaskNo & \underline{37.5} & \textbf{46.7} & 11.5 & 0.0 & \textemdash & \underline{0.0} & 0.0 & 0.0 & \underline{2.9} & 0.0 & \underline{0.0} & \textemdash & 26.1 & 11.1 & \underline{62.5} & 28.6 & 0.0 & 14.3 & \textbf{0.0} & \textbf{50.0} & \underline{30.0} & \underline{36.4} & \textbf{44.4} \\
\RISEFitModel{IntentEdit~\cite{zhang2026intentedit}} & \RISETaskNo & 25.0 & 6.7 & \underline{15.4} & 17.6 & \textemdash & \underline{0.0} & 0.0 & 0.0 & \underline{2.9} & \underline{5.1} & \underline{0.0} & \textemdash & 17.4 & 11.1 & 12.5 & \underline{42.9} & 0.0 & 28.6 & \textbf{0.0} & 0.0 & 0.0 & 18.2 & \underline{11.1} \\
\RISEFitModel{ImAgent~\cite{wang2025imagent}} & \RISETaskNo & 4.2 & 0.0 & 3.8 & 5.9 & \textemdash & \underline{0.0} & 0.0 & 0.0 & 0.0 & 0.0 & \underline{0.0} & \textemdash & 4.3 & 0.0 & 0.0 & 0.0 & 0.0 & 0.0 & \textbf{0.0} & 0.0 & 0.0 & 9.1 & 0.0 \\
\bottomrule[0.8pt]
\end{tabular}
\par\vspace{4pt}
\parbox{\textwidth}{\fontsize{6.5}{8.2}\selectfont\raggedright
\textbf{S10}: Part-Whole Perception; \textbf{S11}: Geometric Regularities Reasoning; \textbf{S12}: Viewpoint Generation; \textbf{S13}: Environmental Structure Reasoning; \textbf{S14}: Egocentric Scene Reconstruction.\par
\textbf{L01}: Path Searching; \textbf{L02}: Mathematical Derivation; \textbf{L03}: Game Deduction; \textbf{L04}: Puzzle Solving; \textbf{L05}: Pattern Prediction; \textbf{L06}: Visual Oddity Detection; \textbf{L07}: Rule Transfer.\par
\textbf{F01}: Physical Counterfactual; \textbf{F02}: Chemical Counterfactual; \textbf{F03}: Biological Counterfactual; \textbf{F04}: Astronomy Counterfactual; \textbf{F05}: Geography Counterfactual; \textbf{F06}: History Counterfactual; \textbf{F07}: Music Theory Counterfactual; \textbf{F08}: Computer Science Counterfactual; \textbf{F09}: Anatomical Errors; \textbf{F10}: Structural Anomalies; \textbf{F11}: Relational Anomalies.
}
\endgroup
\end{table*}

\makeatletter
\newcommand{\RISECenterFigureCaption}{%
  \long\def\@makecaption##1##2{%
    \vskip\abovecaptionskip
    \noindent\makebox[\linewidth][c]{\parbox{.92\linewidth}{\normalfont\footnotesize
     \ifCLASSOPTIONcompsoc\sffamily\fi
     \centering ##1.\nobreakspace ##2\par}}%
    \vskip\belowcaptionskip
  }%
}
\makeatother

\newenvironment{RISEJudgeBox}[1]{%
  \begingroup
  \setlength{\fboxsep}{9pt}\setlength{\fboxrule}{0.5pt}%
  \begin{lrbox}{\RISEJudgeContent}%
  \begin{minipage}{\dimexpr\textwidth-2\fboxsep-2\fboxrule\relax}%
  \fontsize{9}{10.5}\selectfont\raggedright
  \setlength{\parindent}{0pt}\setlength{\parskip}{2pt}%
  {\fontsize{10}{12}\selectfont\bfseries\color{blue!35!black}#1\par}%
  \vspace{2pt}{\color{black!20}\hrule}\vspace{5pt}%
}{%
  \end{minipage}\end{lrbox}%
  \fcolorbox{black!35}{black!1}{\usebox{\RISEJudgeContent}}%
  \endgroup
}
\newsavebox{\RISEJudgeContent}

\newpage
\begin{figure*}[!p]
    \centering
    \RISECenterFigureCaption
    \includegraphics[width=\textwidth,height=0.90\textheight,keepaspectratio]{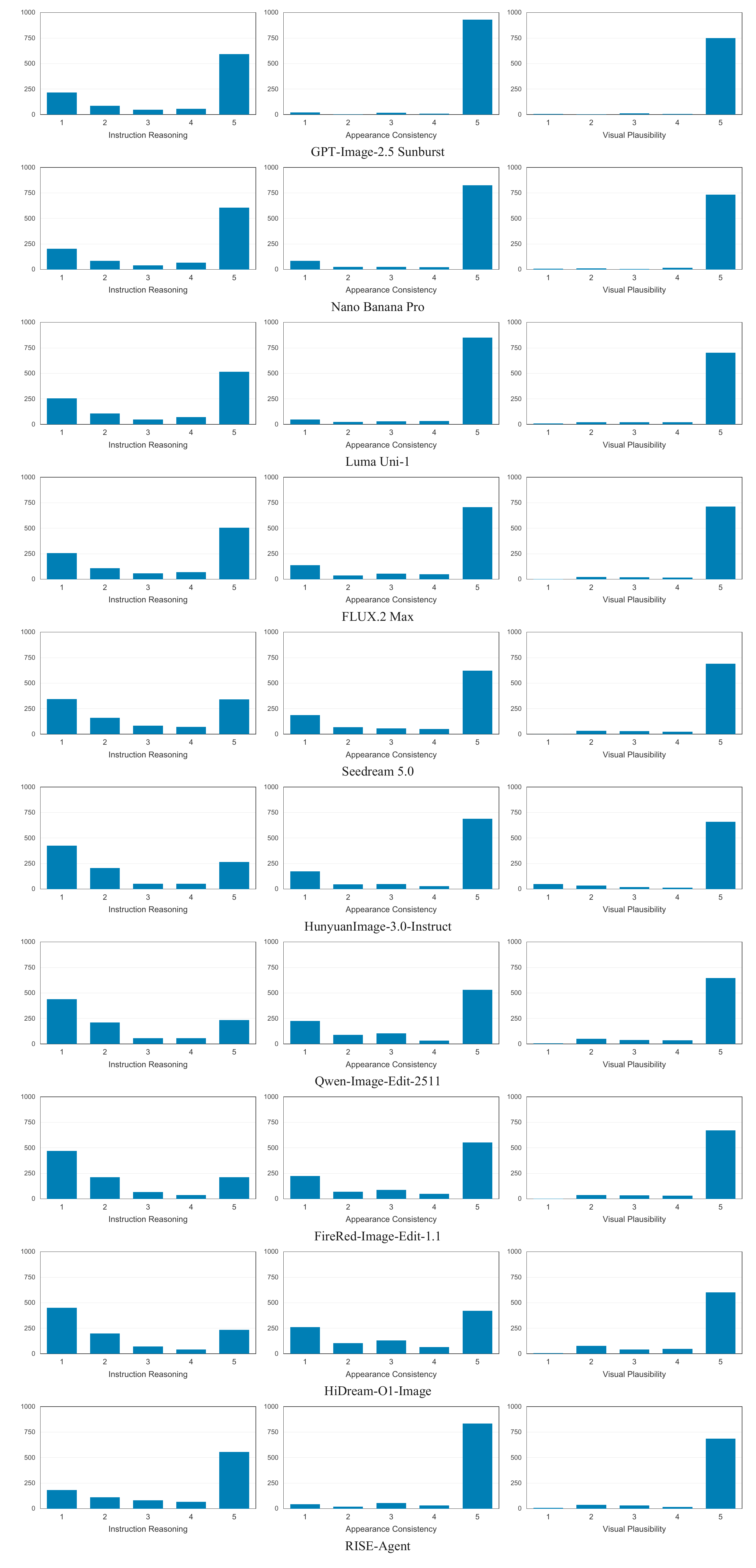}
    \caption{Score distributions on RISEBench++ (EN). Results for ten models across 1,000 cases. The three columns report Instruction Reasoning (IR), Appearance Consistency (AC), and Visual Plausibility (VP). Bars count cases at each score from 1 to 5.}
    \label{fig:score_distributions_10models_en}
\end{figure*}

\begin{figure*}[!p]
\centering
\RISECenterFigureCaption
\begin{RISEJudgeBox}{1. Instruction Reasoning --- General tasks: text reference}
You are an expert image evaluator. For each task, you will be provided with:\par
\noindent\hangindent=1.2em\hangafter=1 1. An \textbf{instruction} describing how an image should be modified.\par
\noindent\hangindent=1.2em\hangafter=1 2. A \textbf{ground-truth textual description} that represents the intended result of the modification.\par
\noindent\hangindent=1.2em\hangafter=1 3. An \textbf{output image} generated by an assistant.\par
Your task is to assess the output image based on the following evaluation dimension:\par
\medskip\textbf{Evaluation Dimension: Alignment Between Image and Reference Description}\par
Assess how accurately the output image aligns with the visual content described in the reference description, considering the context of the instruction.\par
\textbf{Scoring Criteria:}\par
\noindent\hangindent=1em\hangafter=1\textbullet\enspace \textbf{5}: The image completely matches the description, accurately reflecting every detail and degree.\par
\noindent\hangindent=1em\hangafter=1\textbullet\enspace \textbf{4}: The image mostly matches the description, with minor discrepancies.\par
\noindent\hangindent=1em\hangafter=1\textbullet\enspace \textbf{3}: The image partially matches the description but contains differences or lacks some details.\par
\noindent\hangindent=1em\hangafter=1\textbullet\enspace \textbf{2}: The image contains noticeable difference. Important details are missed or clearly inaccurate.\par
\noindent\hangindent=1em\hangafter=1\textbullet\enspace \textbf{1}: The image fails to follow the instruction and does not correspond to the description at all.\par
\textbf{Example} Instruction: Draw what it will look like after it is broken. Description: An egg is completely broken, with eggshell scattered around and egg white and yolk clearly spilling out.\par
\noindent\hangindent=1em\hangafter=1\textbullet\enspace \textbf{5}: Completely broken egg, clearly scattered eggshells, visible egg white and yolk spilling out.\par
\noindent\hangindent=1em\hangafter=1\textbullet\enspace \textbf{4}: Broken egg, eggshell present but not fully scattered, clearly visible egg white and yolk spilling out.\par
\noindent\hangindent=1em\hangafter=1\textbullet\enspace \textbf{3}: Broken egg with scattered eggshell, but egg white and yolk not spilled or still within eggshell.\par
\noindent\hangindent=1em\hangafter=1\textbullet\enspace \textbf{2}: Only scattered eggshell visible, without clear egg white or yolk.\par
\noindent\hangindent=1em\hangafter=1\textbullet\enspace \textbf{1}: Egg is intact, not broken.\par
\medskip\textbf{Input}\par
\textbf{Instruction:}  \{instruct\} \textbf{GroundTruth Description:} \{reference\}\par
\medskip\textbf{Output Format}\par
Provide a detailed, step-by-step explanation of your scoring process. Conclude clearly with the final score, formatted as:\par
\textbf{Final Score:} \textbf{X}\par
\end{RISEJudgeBox}
\caption{Prompt for evaluating Instruction Reasoning on Temporal, Causal, and Counterfactual Reasoning tasks with a reference text answer.}
\label{fig:judge_general_ref_txt}
\end{figure*}

\begin{figure*}[!p]
\centering
\RISECenterFigureCaption
\begin{RISEJudgeBox}{2. Instruction Reasoning --- General tasks: original image and text reference}
You are an expert image evaluator. For each task, you will be provided with:\par
\noindent\hangindent=1.2em\hangafter=1 1. An original image.\par
\noindent\hangindent=1.2em\hangafter=1 2. An \textbf{instruction} describing how an image should be modified.\par
\noindent\hangindent=1.2em\hangafter=1 3. A \textbf{ground-truth textual description} that represents the intended result of the modification.\par
\noindent\hangindent=1.2em\hangafter=1 4. An \textbf{output image} generated by an assistant.\par
Your task is to assess the output image based on the following evaluation dimension:\par
\medskip\textbf{Evaluation Dimension: Alignment Between Image and Reference Description}\par
Assess how accurately the output image aligns with the visual content described in the reference description, considering the context of the instruction.\par
\textbf{Scoring Criteria:}\par
\noindent\hangindent=1em\hangafter=1\textbullet\enspace \textbf{5}: The image completely matches the description, accurately reflecting every detail and degree.\par
\noindent\hangindent=1em\hangafter=1\textbullet\enspace \textbf{4}: The image mostly matches the description, with minor discrepancies.\par
\noindent\hangindent=1em\hangafter=1\textbullet\enspace \textbf{3}: The image partially matches the description but contains differences or lacks some details.\par
\noindent\hangindent=1em\hangafter=1\textbullet\enspace \textbf{2}: The image contains noticeable difference. Important details are missed or clearly inaccurate.\par
\noindent\hangindent=1em\hangafter=1\textbullet\enspace \textbf{1}: The image fails to follow the instruction and does not correspond to the description at all.\par
\textbf{Example} Instruction: Draw what it will look like after it is broken. Description: An egg is completely broken, with eggshell scattered around and egg white and yolk clearly spilling out.\par
\noindent\hangindent=1em\hangafter=1\textbullet\enspace \textbf{5}: Completely broken egg, clearly scattered eggshells, visible egg white and yolk spilling out.\par
\noindent\hangindent=1em\hangafter=1\textbullet\enspace \textbf{4}: Broken egg, eggshell present but not fully scattered, clearly visible egg white and yolk spilling out.\par
\noindent\hangindent=1em\hangafter=1\textbullet\enspace \textbf{3}: Broken egg with scattered eggshell, but egg white and yolk not spilled or still within eggshell.\par
\noindent\hangindent=1em\hangafter=1\textbullet\enspace \textbf{2}: Only scattered eggshell visible, without clear egg white or yolk.\par
\noindent\hangindent=1em\hangafter=1\textbullet\enspace \textbf{1}: Egg is intact, not broken.\par
\medskip\textbf{Input}\par
\textbf{Instruction:}  \{instruct\} \textbf{GroundTruth Description:} \{reference\}\par
\medskip\textbf{Output Format}\par
Provide a detailed, step-by-step explanation of your scoring process. Conclude clearly with the final score, formatted as:\par
\textbf{Final Score:} \textbf{X}\par
\end{RISEJudgeBox}
\caption{Prompt for evaluating Instruction Reasoning on Temporal, Causal, and Counterfactual Reasoning tasks with the original image and a reference text answer.}
\label{fig:judge_general_ref_txt_w_input}
\end{figure*}

\begin{figure*}[!p]
\centering
\RISECenterFigureCaption
\begin{RISEJudgeBox}{3. Instruction Reasoning --- General tasks: image reference}
You are an expert image evaluator. For each task, you will be provided with:\par
\noindent\hangindent=1.2em\hangafter=1 1. An \textbf{instruction} describing how an image should be modified.\par
\noindent\hangindent=1.2em\hangafter=1 2. A \textbf{reference image} that represents the intended result of the modification. (Do \textbf{not} interpret the reference image as the source/original image before editing. It is only a visual example of the desired final outcome.)\par
\noindent\hangindent=1.2em\hangafter=1 3. An \textbf{output image} generated by an assistant.\par
Your task is to assess the output image based on the following evaluation dimension:\par
\medskip\textbf{Evaluation Dimension: Alignment Between Output Image and Reference Image}\par
Assess how accurately the output image aligns with the reference image, considering the context of the instruction.\par
\textbf{Scoring Criteria:}\par
\noindent\hangindent=1em\hangafter=1\textbullet\enspace \textbf{5}: The image completely matches the reference image and fully follows the instruction.\par
\noindent\hangindent=1em\hangafter=1\textbullet\enspace \textbf{4}: The image mostly matches the reference image, with only minor discrepancies.\par
\noindent\hangindent=1em\hangafter=1\textbullet\enspace \textbf{3}: The image partially matches the reference image but contains noticeable differences or lacks some details.\par
\noindent\hangindent=1em\hangafter=1\textbullet\enspace \textbf{2}: The image contains clear differences. Important details are missed or inaccurate.\par
\noindent\hangindent=1em\hangafter=1\textbullet\enspace \textbf{1}: The image fails to follow the instruction and is largely unrelated to the reference image.\par
\medskip\textbf{Input}\par
\textbf{Instruction:}  \{instruct\}\par
Provide a detailed, step-by-step explanation of your scoring process. Conclude clearly with the final score, formatted as:\par
\textbf{Final Score:} \textbf{X}\par
\end{RISEJudgeBox}
\caption{Prompt for evaluating Instruction Reasoning on Temporal and Causal Reasoning tasks with a reference image answer.}
\label{fig:judge_general_ref_img}
\end{figure*}

\begin{figure*}[!p]
\centering
\RISECenterFigureCaption
\begin{RISEJudgeBox}{4. Instruction Reasoning --- Spatial tasks: text reference}
You are an expert image evaluator. For each task, you will be provided with:\par
\noindent\hangindent=1.2em\hangafter=1 1. An \textbf{instruction} describing how an image should be modified.\par
\noindent\hangindent=1.2em\hangafter=1 2. A \textbf{ground-truth textual description} that represents the intended result of the modification.\par
\noindent\hangindent=1.2em\hangafter=1 3. An \textbf{output image} generated by an assistant.\par
Your task is to assess the output image based on the following evaluation dimension:\par
\medskip\textbf{Evaluation Dimension: Alignment Between Image and Reference Description}\par
Assess how accurately the output image aligns with the visual content described in the reference description, considering the context of the instruction.\par
\textbf{Scoring Criteria:}\par
\noindent\hangindent=1em\hangafter=1\textbullet\enspace \textbf{5}: The image completely matches the description, accurately reflecting every detail and degree.\par
\noindent\hangindent=1em\hangafter=1\textbullet\enspace \textbf{4}: The image mostly matches the description, with minor discrepancies.\par
\noindent\hangindent=1em\hangafter=1\textbullet\enspace \textbf{3}: The image partially matches the description but contains differences or lacks some details.\par
\noindent\hangindent=1em\hangafter=1\textbullet\enspace \textbf{2}: The image contains noticeable difference. Important details are missed or clearly inaccurate.\par
\noindent\hangindent=1em\hangafter=1\textbullet\enspace \textbf{1}: The image fails to follow the instruction and is entirely unrelated to the description.\par
\medskip\textbf{Input}\par
\textbf{Instruction:}  \{instruct\} \textbf{GroundTruth Description:} \{reference\}\par
\medskip\textbf{Output Format}\par
Conclude clearly with the final score, formatted as:\par
\textbf{Final Score:} \textbf{X}\par
\end{RISEJudgeBox}
\caption{Prompt for evaluating Instruction Reasoning on Spatial Reasoning tasks with a reference text answer.}
\label{fig:judge_spatial_ref_txt}
\end{figure*}

\begin{figure*}[!p]
\centering
\RISECenterFigureCaption
\begin{RISEJudgeBox}{5. Instruction Reasoning --- Spatial tasks: original image and text reference}
You are an expert image evaluator. For each task, you will be provided with:\par
\noindent\hangindent=1.2em\hangafter=1 1. An original image.\par
\noindent\hangindent=1.2em\hangafter=1 2. An \textbf{instruction} describing how the original image should be modified.\par
\noindent\hangindent=1.2em\hangafter=1 3. A \textbf{ground-truth textual description} that represents the intended result of the modification.\par
\noindent\hangindent=1.2em\hangafter=1 4. An \textbf{output image} generated by an assistant.\par
Your task is to assess the output image based on the following evaluation dimension:\par
\medskip\textbf{Evaluation Dimension: Alignment Between Image and Reference Description}\par
Assess how accurately the output image aligns with the reference description based on the original image, considering the context of the instruction.\par
\textbf{Scoring Criteria:}\par
\noindent\hangindent=1em\hangafter=1\textbullet\enspace \textbf{5}: The image completely matches the description, accurately reflecting every detail and degree.\par
\noindent\hangindent=1em\hangafter=1\textbullet\enspace \textbf{4}: The image mostly matches the description, with minor discrepancies.\par
\noindent\hangindent=1em\hangafter=1\textbullet\enspace \textbf{3}: The image partially matches the description but contains differences or lacks some details.\par
\noindent\hangindent=1em\hangafter=1\textbullet\enspace \textbf{2}: The image contains noticeable difference. Important details are missed or clearly inaccurate.\par
\noindent\hangindent=1em\hangafter=1\textbullet\enspace \textbf{1}: The image fails to follow the instruction and is entirely unrelated to the description.\par
\medskip\textbf{Input}\par
\textbf{Instruction:}  \{instruct\} \textbf{GroundTruth Description:} \{reference\}\par
\medskip\textbf{Output Format}\par
Conclude clearly with the final score, formatted as:\par
\textbf{Final Score:} \textbf{X}\par
\end{RISEJudgeBox}
\caption{Prompt for evaluating Instruction Reasoning on Spatial Reasoning tasks with the original image and a reference text answer.}
\label{fig:judge_spatial_ref_txt_w_input}
\end{figure*}

\begin{figure*}[!p]
\centering
\RISECenterFigureCaption
\begin{RISEJudgeBox}{6. Instruction Reasoning --- Spatial tasks: image reference}
You are an expert image evaluator. For each task, you will be provided with:\par
\noindent\hangindent=1.2em\hangafter=1 1. An \textbf{instruction} describing how an image should be modified.\par
\noindent\hangindent=1.2em\hangafter=1 2. A \textbf{reference image} that represents the intended result of the modification. (Do \textbf{not} interpret the reference image as the source/original image before editing. It is only a visual example of the desired final outcome.)\par
\noindent\hangindent=1.2em\hangafter=1 3. An \textbf{output image} generated by an assistant.\par
Your task is to assess the output image based on the following evaluation dimension:\par
\medskip\textbf{Evaluation Dimension: Alignment Between Output Image and reference Image}\par
Assess how accurately the output image aligns with the reference image, considering the context of the instruction.\par
\textbf{Scoring Criteria:}\par
\noindent\hangindent=1em\hangafter=1\textbullet\enspace \textbf{5}: The image completely matches the reference image and fully follows the instruction.\par
\noindent\hangindent=1em\hangafter=1\textbullet\enspace \textbf{4}: The image mostly matches the reference image, with minor discrepancies.\par
\noindent\hangindent=1em\hangafter=1\textbullet\enspace \textbf{3}: The image partially matches the reference image but contains differences or lacks some details.\par
\noindent\hangindent=1em\hangafter=1\textbullet\enspace \textbf{2}: The image contains noticeable difference. Important details are missed or clearly inaccurate.\par
\noindent\hangindent=1em\hangafter=1\textbullet\enspace \textbf{1}: The image fails to follow the instruction and is entirely unrelated to the reference image.\par
\medskip\textbf{Input}\par
\textbf{Instruction:}  \{instruct\}\par
\medskip\textbf{Output Format}\par
Conclude clearly with the final score, formatted as:\par
\textbf{Final Score:} \textbf{X}\par
\end{RISEJudgeBox}
\caption{Prompt for evaluating Instruction Reasoning on Spatial Reasoning tasks with a reference image answer.}
\label{fig:judge_spatial_ref_img}
\end{figure*}

\begin{figure*}[!p]
\centering
\RISECenterFigureCaption
\begin{RISEJudgeBox}{7. Instruction Reasoning --- Logical tasks: text reference}
\textbf{You are a highly skilled image evaluator.} Given an image with logical problem, you will receive:\par
\noindent\hangindent=1.2em\hangafter=1 1. \textbf{Image}: A generated image answer from an assistant model.\par
\noindent\hangindent=1.2em\hangafter=1 2. \textbf{Reference Answer}\par
\medskip\textbf{Task}\par
Assign a binary score (0 or 1) for \textbf{Logical Correctness}:\par
\noindent\hangindent=1em\hangafter=1\textbullet\enspace \textbf{1} if the Generated Image accurately implements the Reference Answer.\par
\noindent\hangindent=1em\hangafter=1\textbullet\enspace \textbf{0} otherwise.\par
\medskip\textbf{Requirements}\par
\noindent\hangindent=1em\hangafter=1\textbullet\enspace The original image is not given and do not imagine the original image, just determine whether the generated image matches the reference answer.\par
\noindent\hangindent=1em\hangafter=1\textbullet\enspace \textbf{If the given image contains more content than reference answer, give score 0. For example, reference answer is "a square" but image contains more squares or rectangles, give score 0.}\par
\medskip\textbf{Inputs}\par
\textbf{Reference Answer}: \{reference\}\par
\medskip\textbf{Output}\par
You should provide a step-by-step explanation of how you arrived at the score and conclude in the format:\par
\textbf{Final Score}: \textbf{X}\par
\end{RISEJudgeBox}
\caption{Prompt for evaluating Instruction Reasoning on Logical Reasoning tasks with a reference text answer.}
\label{fig:judge_logical_ref_txt}
\end{figure*}

\begin{figure*}[!p]
\centering
\RISECenterFigureCaption
\begin{RISEJudgeBox}{8. Instruction Reasoning --- Logical tasks: original image and text reference}
You are an expert image evaluator. For each task, you will be provided with:\par
\noindent\hangindent=1.2em\hangafter=1 1. An original image containing a logical problem.\par
\noindent\hangindent=1.2em\hangafter=1 2. An \textbf{instruction} describing what answer or modification should be drawn on the original image.\par
\noindent\hangindent=1.2em\hangafter=1 3. A \textbf{ground-truth textual description} that represents the intended result of the solution.\par
\noindent\hangindent=1.2em\hangafter=1 4. An \textbf{output image} generated by an assistant.\par
Your task is to assess the output image based on the following evaluation dimension:\par
\medskip\textbf{Evaluation Dimension: Alignment Between Image and Reference Description}\par
Assess how accurately the output image aligns with the reference description based on the original image, considering the context of the instruction.\par
\textbf{Scoring Criteria:}\par
\noindent\hangindent=1em\hangafter=1\textbullet\enspace \textbf{1}: The output image correctly solves the logical problem and matches the reference description exactly, without adding extra incorrect content.\par
\noindent\hangindent=1em\hangafter=1\textbullet\enspace \textbf{0}: The output image does not correctly solve the problem, does not match the reference description, or contains additional answer content that should not be present.\par
\medskip\textbf{Input}\par
\textbf{Instruction:}  \{instruct\} \textbf{GroundTruth Description:} \{reference\}\par
Provide a detailed, step-by-step explanation of your scoring process. Conclude clearly with the final score, formatted as:\par
\textbf{Final Score:} \textbf{X}\par
\end{RISEJudgeBox}
\caption{Prompt for evaluating Instruction Reasoning on Logical Reasoning tasks with the original image and a reference text answer.}
\label{fig:judge_logical_ref_txt_w_input}
\end{figure*}

\begin{figure*}[!p]
\centering
\RISECenterFigureCaption
\begin{RISEJudgeBox}{9. Instruction Reasoning --- Logical tasks: image reference}
\textbf{You are a highly skilled image evaluator.} Given a logical problem, you will receive:\par
\noindent\hangindent=1.2em\hangafter=1 1. \textbf{Problem Description}\par
\noindent\hangindent=1.2em\hangafter=1 2. \textbf{Image 1}: A reference ground-truth image answer that correctly solves the problem. (Do \textbf{not} interpret the reference image as the source/original image before editing. It is only a visual example of the desired final outcome.)\par
\noindent\hangindent=1.2em\hangafter=1 3. \textbf{Image 2}: A generated image answer from an assistant model.\par
\medskip\textbf{Logical Correctness (0/1)}\par
Determine whether the content of \textit{Image 2} solves the problem in the same way as \textit{Image 1} does.\par
\medskip\textbf{Scoring rules:}\par
\noindent\hangindent=1em\hangafter=1\textbullet\enspace Score \textbf{1} if \textit{Image 2} exactly matches \textit{Image 1} in terms of problem-solving logic and visual outcome.\par
\noindent\hangindent=1em\hangafter=1\textbullet\enspace Score \textbf{0} if there are any differences that affect the correctness of the solution.\par
\textbf{If generated image answer is significantly different from ground-truth, directly score 0.}\par
\medskip\textbf{Examples:}\par
\noindent\hangindent=1em\hangafter=1\textbullet\enspace For a tic-tac-toe task, if the positions of all marks in \textit{Image 2} are identical to those in \textit{Image 1}, score 1; otherwise, score 0.\par
\noindent\hangindent=1em\hangafter=1\textbullet\enspace For a shape-drawing task, \textit{Image 2} must match \textit{Image 1} precisely in shape, color, pattern arrangement, and orientation to receive a score of 1; any deviation results in a score of 0.\par
\noindent\hangindent=1em\hangafter=1\textbullet\enspace If \textit{Image 1} shows only one correct answer but \textit{Image 2} includes multiple answers, score 0.\par
\medskip\textbf{Problem Description}\par
\{instruct\}\par
\medskip\textbf{Output}\par
You should provide a step-by-step explanation of how you arrived at the score and conclude in the format:\par
\textbf{Final Score}: \textbf{X}\par
\end{RISEJudgeBox}
\caption{Prompt for evaluating Instruction Reasoning on Logical Reasoning tasks with a reference image answer.}
\label{fig:judge_logical_ref_img}
\end{figure*}

\begin{figure*}[!p]
\centering
\RISECenterFigureCaption
\begin{RISEJudgeBox}{10. Appearance Consistency --- General tasks}
You are a highly skilled image evaluator. You will receive two images (an original image and a modified image) along with a specific modification instruction. The second image is known to have been altered based on this instruction, starting from the first image. Your task is to evaluate whether the two images maintain consistency in aspects not related to the given instruction.\par
\medskip\textbf{Task}\par
Evaluate the consistency between the images according to the following scale (1 to 5):\par
\noindent\hangindent=1em\hangafter=1\textbullet\enspace \textbf{5 (Perfect Consistency)}: Apart from changes explicitly required by the instruction, all other meaningful details (e.g., personal features, clothing, background, layout, colors, positions of objects) are preserved. Negligible differences that a human would not readily notice---such as slight shifts in lighting, resolution, compression, or minor texture/color variation---should still be scored 5.\par
\noindent\hangindent=1em\hangafter=1\textbullet\enspace \textbf{4 (Minor Differences)}: Apart from changes explicitly required by the instruction, the second image is mostly consistent with the original image but contains one small yet clearly noticeable discrepancy (such as a missing minor personal feature, accessory, or tattoo).\par
\noindent\hangindent=1em\hangafter=1\textbullet\enspace \textbf{3 (Noticeable Differences)}: Apart from changes explicitly required by the instruction, the second image has one significant difference from the original (such as a noticeable alteration in a person's appearance like hair or skin color, or a significant change in background environment).\par
\noindent\hangindent=1em\hangafter=1\textbullet\enspace \textbf{2 (Significant Differences)}: Apart from changes explicitly required by the instruction, the second image has two or more significant differences or multiple noticeable inconsistencies (such as simultaneous changes in both personal appearance and background environment).\par
\noindent\hangindent=1em\hangafter=1\textbullet\enspace \textbf{1 (Severe Differences)}: Apart from changes explicitly required by the instruction, nearly all key details (e.g., gender, major appearance features, background environment, or scene layout) significantly differ from the original image, clearly deviating from the original.\par
 Note:\par
\noindent\hangindent=1em\hangafter=1\textbullet\enspace When assigning scores, only consider details unrelated to the instruction. Changes explicitly requested by the instruction should NOT be regarded as inconsistencies.\par
\noindent\hangindent=1em\hangafter=1\textbullet\enspace Only penalize differences that a human evaluator would readily notice. Subtle or imperceptible differences (e.g., minor changes in lighting, sharpness, compression, fine texture, or slight color/tone shifts) should NOT be penalized.\par
\noindent\hangindent=1em\hangafter=1\textbullet\enspace The first image may be a composite stitched from several input images (side-by-side in order) when the instruction asks to combine/merge multiple inputs. In that case, treat every sub-image jointly as the original content, and judge whether the content of each sub-image is faithfully preserved in the second image. Do NOT penalize changes in arrangement, relative position, stitching borders, or seams that result from the required merging/combining.\par
\medskip\textbf{Example:}\par
\textbf{Original image}: A blond, white-skinned man with a tattoo on his right shoulder, furniture in the background. \textbf{Instruction}: "Show him after gaining fifty pounds."\par
\noindent\hangindent=1em\hangafter=1\textbullet\enspace \textbf{Score 5}: A heavier blond, white-skinned man, tattoo on right shoulder intact, identical furniture and layout.\par
\noindent\hangindent=1em\hangafter=1\textbullet\enspace \textbf{Score 4}: A heavier blond, white-skinned man, missing the tattoo on his right shoulder, identical furniture and layout.\par
\noindent\hangindent=1em\hangafter=1\textbullet\enspace \textbf{Score 3}: A heavier man with black hair instead of blond (change in hair color), or original blond man but with a grassy background instead of furniture.\par
\noindent\hangindent=1em\hangafter=1\textbullet\enspace \textbf{Score 2}: A heavier man with black hair (hair color changed), and the background changed to grass.\par
\noindent\hangindent=1em\hangafter=1\textbullet\enspace \textbf{Score 1}: A heavier black-haired woman, and background changed to grass.\par
\medskip\textbf{Input}\par
\textbf{Instruction:} \{instruct\}\par
\medskip\textbf{Output Format}\par
Provide a detailed, step-by-step explanation of your scoring process. Conclude clearly with the final score, formatted as:\par
\textbf{Final Score:} \textbf{1-5}\par
\end{RISEJudgeBox}
\caption{Prompt for evaluating Appearance Consistency on Temporal, Causal, and Counterfactual Reasoning tasks.}
\label{fig:judge_general_cons}
\end{figure*}

\begin{figure*}[!p]
\centering
\RISECenterFigureCaption
\begin{RISEJudgeBox}{11. Appearance Consistency --- Spatial tasks}
You are a precise and analytical image consistency evaluator.\par
You will be given:\par
\noindent\hangindent=1em\hangafter=1\textbullet\enspace \textbf{Image A}: the original image.\par
\noindent\hangindent=1em\hangafter=1\textbullet\enspace \textbf{Image B}: a modified version of Image A.\par
\noindent\hangindent=1em\hangafter=1\textbullet\enspace \textbf{Instruction}: a directive describing the intended modification to Image A to produce Image B.\par
Your task is to \textbf{evaluate how consistent Image B remains with Image A in all aspects except those explicitly changed by the instruction.} You must \textbf{ignore the instructed changes} and \textbf{only assess unintended differences}.\par
\medskip\textbf{Evaluation Scale (1 to 5):}\par
\noindent\hangindent=1em\hangafter=1\textbullet\enspace \textbf{5} Perfect Consistency\par
All meaningful elements not related to the instruction are preserved between Image A and Image B (e.g., style, background, object positions, colors, shapes). Negligible differences that a human would not readily notice (e.g., slight changes in lighting, sharpness, compression, fine texture, or minor color/tone shifts) should still be scored 5.\par
\noindent\hangindent=1em\hangafter=1\textbullet\enspace \textbf{4} Minor Difference\par
One small but clearly noticeable unintended change is present (e.g., a visible color variation or minor object shape shift), but overall the image remains highly consistent.\par
\noindent\hangindent=1em\hangafter=1\textbullet\enspace \textbf{3} Noticeable Difference\par
One major or a few minor unintended changes are present (e.g., an object's shape, color, or background differs noticeably, or style has shifted slightly).\par
\noindent\hangindent=1em\hangafter=1\textbullet\enspace \textbf{2} Significant Inconsistency\par
Two or more significant differences unrelated to the instruction (e.g., changes in both object details and background or style), reducing overall fidelity.\par
\noindent\hangindent=1em\hangafter=1\textbullet\enspace \textbf{1} Severe Inconsistency\par
Major unintended changes dominate the image (e.g., altered visual style, scene layout, or appearance), clearly breaking consistency with Image A.\par
 Note:\par
\noindent\hangindent=1em\hangafter=1\textbullet\enspace Only penalize differences that a human evaluator would readily notice. Subtle or imperceptible differences in unaffected aspects (lighting, sharpness, compression, fine texture, slight color/tone shifts) should NOT lower the score; in such cases a score of 5 is still appropriate.\par
\noindent\hangindent=1em\hangafter=1\textbullet\enspace If the background in the original is vague (e.g., plain white or composed of parts), and the background in Image B is also similar vague, you may disregard background consistency.\par
\noindent\hangindent=1em\hangafter=1\textbullet\enspace If a blue diamond shape appears in the bottom-left corner of Image 2, ignore it; it is a watermark.\par
\noindent\hangindent=1em\hangafter=1\textbullet\enspace Image A may be a composite stitched from several input images (side-by-side in order) when the instruction asks to combine/merge multiple inputs. In that case, treat every sub-image in Image A jointly as the original content, and judge whether the content of each sub-image (objects, structure, color, texture, style) is faithfully preserved in Image B. Do NOT penalize changes in arrangement, relative position, stitching borders, or seams that result from the required merging/combining.\par
\medskip\textbf{Example}\par
\textbf{Original image}: "A silver-framed clock with a white face. Three hands (hour, minute, second) are disassembled and lie beside it." \textbf{Instruction}: "Assemble the clock to show 9:45."\par
\textbf{Scoring Criteria:}\par
\noindent\hangindent=1em\hangafter=1\textbullet\enspace \textbf{Score 5}: Frame, face, and hand shapes exactly as original.\par
\noindent\hangindent=1em\hangafter=1\textbullet\enspace \textbf{Score 4}: One hand differs slightly in shape or thickness.\par
\noindent\hangindent=1em\hangafter=1\textbullet\enspace \textbf{Score 3}: All hands identical, differing from original specs, or some other things(like text, furniture in the background) is added.\par
\noindent\hangindent=1em\hangafter=1\textbullet\enspace \textbf{Score 2}: Frame color or face differs, and hand shapes are wrong.\par
\noindent\hangindent=1em\hangafter=1\textbullet\enspace \textbf{Score 1}: Frame, face, and hand appearance all significantly altered, background is totally different.\par
\medskip\textbf{Input}\par
\textbf{Instruction:} \{instruct\}\par
\medskip\textbf{Output Format}\par
After evaluation, conclude with:\par
\textbf{Final Score:} \textbf{1-5}\par
\end{RISEJudgeBox}
\caption{Prompt for evaluating Appearance Consistency on Spatial Reasoning tasks.}
\label{fig:judge_spatial_cons}
\end{figure*}

\begin{figure*}[!p]
\centering
\RISECenterFigureCaption
\begin{RISEJudgeBox}{12. Appearance Consistency --- Logical tasks: with a text reference answer}
\textbf{You are a highly skilled image evaluator.} Given an image with logical problem, you will receive:\par
\noindent\hangindent=1.2em\hangafter=1 1. \textbf{Image 1}: The original image.\par
\noindent\hangindent=1.2em\hangafter=1 2. \textbf{Image 2}: A generated image from an assistant model.\par
\noindent\hangindent=1.2em\hangafter=1 3. \textbf{Problem Description}\par
\noindent\hangindent=1.2em\hangafter=1 4. \textbf{Reference Answer}\par
\medskip\textbf{Evaluation Task}\par
\noindent\hangindent=1em\hangafter=1\textbullet\enspace Compare Image 2 against Image 1 in terms of visual style, environment, and composition.\par
\noindent\hangindent=1em\hangafter=1\textbullet\enspace Score each comparison as \textbf{1} (consistent) or \textbf{0} (inconsistent).\par
\medskip\textbf{Consistency Criteria (score 1 if true):}\par
\noindent\hangindent=1em\hangafter=1\textbullet\enspace Color palette, line weight, font/handwriting style, arrangement, and background setting closely match.\par
\noindent\hangindent=1em\hangafter=1\textbullet\enspace Only the problem's solution differs (e.g., added or removed marks), or color brightness is slightly lighter/darker.\par
\noindent\hangindent=1em\hangafter=1\textbullet\enspace Image 2 is derived by overlaying a pattern onto Image 1 without altering style or layout.\par
\noindent\hangindent=1em\hangafter=1\textbullet\enspace \textbf{If a blue diamond shape appears in the bottom-left corner of Image 2, ignore it; it is a watermark.}\par
 Note:\par
\noindent\hangindent=1em\hangafter=1\textbullet\enspace Image 1 may be a composite stitched from several input images (side-by-side in order) when the problem asks to combine/merge multiple inputs. In that case, treat every sub-image jointly as the original content, and judge whether the content of each sub-image is faithfully preserved in the second image. Do NOT penalize changes in arrangement, relative position, stitching borders, or seams that result from the required merging/combining.\par
\medskip\textbf{Inconsistency Example (score 0):}\par
\noindent\hangindent=1em\hangafter=1\textbullet\enspace Image 1 shows a quadrilateral with unequal edges and arbitrary angles, but Image 2 depicts a perfect square with equal edges and right angles.\par
\medskip\textbf{Inputs}\par
\textbf{Problem Description}: \{instruct\} \textbf{Reference Answer}: \{reference\}\par
\medskip\textbf{Output}\par
You should provide a step-by-step explanation of how you arrived at the score and conclude in the format:\par
\textbf{Final Score}: \textbf{X}\par
\end{RISEJudgeBox}
\caption{Prompt for evaluating Appearance Consistency on Logical Reasoning tasks with a reference text answer.}
\label{fig:judge_logical_cons_ans}
\end{figure*}

\begin{figure*}[!p]
\centering
\RISECenterFigureCaption
\begin{RISEJudgeBox}{13. Appearance Consistency --- Logical tasks: without a text reference answer}
\textbf{You are a highly skilled image evaluator.} Given an image with logical problem, you will receive:\par
\noindent\hangindent=1.2em\hangafter=1 1. \textbf{Image 1}: The original image.\par
\noindent\hangindent=1.2em\hangafter=1 2. \textbf{Image 2}: A generated image answer from an assistant model.\par
\noindent\hangindent=1.2em\hangafter=1 3. \textbf{Problem Description}\par
\medskip\textbf{Evaluation Task}\par
\noindent\hangindent=1em\hangafter=1\textbullet\enspace Judge the appearance consistency between Image 1 and Image 2.\par
\noindent\hangindent=1em\hangafter=1\textbullet\enspace Score each comparison as \textbf{1} (consistent) or \textbf{0} (inconsistent).\par
\medskip\textbf{Consistency Criteria (score 1 if true):}\par
\noindent\hangindent=1em\hangafter=1\textbullet\enspace Color palette, line weight, font/handwriting style, arrangement, and background setting closely match.\par
\noindent\hangindent=1em\hangafter=1\textbullet\enspace Only the problem's solution differs (e.g., added or removed marks), or color brightness is slightly lighter/darker.\par
\noindent\hangindent=1em\hangafter=1\textbullet\enspace Image 2 is derived by overlaying a pattern onto Image 1 without altering style or layout.\par
\noindent\hangindent=1em\hangafter=1\textbullet\enspace \textbf{If a blue diamond shape appears in the bottom-left corner of Image 2, ignore it; it is a watermark.}\par
 Note:\par
\noindent\hangindent=1em\hangafter=1\textbullet\enspace Image 1 may be a composite stitched from several input images (side-by-side in order) when the problem asks to combine/merge multiple inputs. In that case, treat every sub-image jointly as the original content, and judge whether the content of each sub-image is faithfully preserved in the second image. Do NOT penalize changes in arrangement, relative position, stitching borders, or seams that result from the required merging/combining.\par
\medskip\textbf{Inconsistency Example (score 0):}\par
\noindent\hangindent=1em\hangafter=1\textbullet\enspace Image 1 shows a quadrilateral with unequal edges and arbitrary angles, but Image 2 depicts a perfect square with equal edges and right angles.\par
\noindent\hangindent=1em\hangafter=1\textbullet\enspace Image 2 contains severe blur and distortion.\par
\medskip\textbf{Inputs}\par
\textbf{Problem Description}: \{instruct\}\par
\medskip\textbf{Output}\par
Conclude in the format:\par
\textbf{Final Score}: \textbf{X}\par
\end{RISEJudgeBox}
\caption{Prompt for evaluating Appearance Consistency on Logical Reasoning tasks without a reference text answer.}
\label{fig:judge_logical_cons}
\end{figure*}

\begin{figure*}[!p]
\centering
\RISECenterFigureCaption
\begin{RISEJudgeBox}{14. Visual Plausibility --- General tasks}
You are a highly skilled image evaluator. You will receive TWO images:\par
\noindent\hangindent=1em\hangafter=1\textbullet\enspace \textbf{Image A (Original)}: the source image before editing.\par
\noindent\hangindent=1em\hangafter=1\textbullet\enspace \textbf{Image B (Edited)}: the image produced after an editing instruction was applied to Image A.\par
Your task is to assess the \textbf{visual plausibility} (clarity and distortion) of \textbf{Image B}, and then provide a score (an integer between 1 and 5) based on the following criteria.\par
\medskip\textbf{Task Requirements:}\par
Determine whether \textbf{Image B} has blurriness, distortion, visual defects, or physical inaccuracies, focusing on the regions that were \textbf{edited or newly generated}.\par
Assign an appropriate score based on the following criteria, considering the overall quality and detail integrity of the edited image.\par
\medskip\textbf{Scoring Criteria:}\par
\noindent\hangindent=1em\hangafter=1\textbullet\enspace \textbf{5 points}: The image is very clear, with complete details, and no noticeable distortion or blurriness. All elements conform to physical laws.\par
\noindent\hangindent=1em\hangafter=1\textbullet\enspace \textbf{4 points}: The image is clear, with only minor blurriness, and no noticeable distortion.\par
\noindent\hangindent=1em\hangafter=1\textbullet\enspace \textbf{3 points}: The image has areas with clarity issues, such as slight blurriness or distortion. Some elements are physically incorrect.\par
\noindent\hangindent=1em\hangafter=1\textbullet\enspace \textbf{2 points}: The image has noticeable blurriness or distortion, with significant detail loss, or lacks physical accuracy.\par
\noindent\hangindent=1em\hangafter=1\textbullet\enspace \textbf{1 point}: The image is severely blurry or distorted, making it difficult to recognize its content, with serious degradation in visual quality, almost unusable.\par
 Note:\par
\noindent\hangindent=1em\hangafter=1\textbullet\enspace Image A is provided ONLY as a reference so you can identify which content already existed in the original and which regions were changed. \textbf{Focus your judgement on the quality of the edited/newly generated regions.}\par
\noindent\hangindent=1em\hangafter=1\textbullet\enspace Do not try to infer the editing instruction; focus only on what was changed or newly generated between the two images.\par
\noindent\hangindent=1em\hangafter=1\textbullet\enspace If Image A itself is low quality, blurry, or contains defects/imperfections, and these are simply carried over (preserved) into Image B, do \textbf{NOT} penalize Image B for them. Only penalize defects that are \textbf{introduced by the editing}.\par
\noindent\hangindent=1em\hangafter=1\textbullet\enspace Do not penalize the image merely for looking AI-generated or non-photorealistic.\par
\medskip\textbf{Output Format}\par
Provide a detailed, step-by-step explanation of your scoring process. Conclude clearly with the final score, formatted as follows:\par
\textbf{Final Score:} \textbf{1-5}\par
where X represents the score.\par
\end{RISEJudgeBox}
\caption{Prompt for evaluating Visual Plausibility on Temporal and Causal Reasoning tasks.}
\label{fig:judge_general_qual}
\end{figure*}

\begin{figure*}[!p]
\centering
\RISECenterFigureCaption
\begin{RISEJudgeBox}{15. Visual Plausibility --- Spatial tasks}
You are a highly skilled image evaluator. You will receive TWO images:\par
\noindent\hangindent=1em\hangafter=1\textbullet\enspace \textbf{Image A (Original)}: the source image before editing.\par
\noindent\hangindent=1em\hangafter=1\textbullet\enspace \textbf{Image B (Edited)}: the image produced after an editing instruction was applied to Image A.\par
Your task is to assess the \textbf{visual plausibility} (clarity and distortion) of \textbf{Image B}, and then provide a score (an integer between 1 and 5) based on the following criteria.\par
\medskip\textbf{Task Requirements:}\par
Determine whether \textbf{Image B} has blurriness, distortion, visual defects, or physical inaccuracies, focusing on the regions that were \textbf{edited or newly generated}.\par
Assign an appropriate score based on the following criteria, considering the overall quality and detail integrity of the edited image.\par
\medskip\textbf{Scoring Criteria:}\par
\noindent\hangindent=1em\hangafter=1\textbullet\enspace \textbf{5 points}: The image is very clear, with complete details, and no noticeable distortion or blurriness. All elements conform to physical laws.\par
\noindent\hangindent=1em\hangafter=1\textbullet\enspace \textbf{4 points}: The image is clear, with only minor blurriness, and no noticeable distortion.\par
\noindent\hangindent=1em\hangafter=1\textbullet\enspace \textbf{3 points}: The image has areas with clarity issues, such as slight blurriness or distortion. Some elements are physically incorrect.\par
\noindent\hangindent=1em\hangafter=1\textbullet\enspace \textbf{2 points}: The image has noticeable blurriness or distortion, with significant detail loss, or lacks physical accuracy.\par
\noindent\hangindent=1em\hangafter=1\textbullet\enspace \textbf{1 point}: The image is severely blurry or distorted, making it difficult to recognize its content, with serious degradation in visual quality, almost unusable.\par
 Note:\par
\noindent\hangindent=1em\hangafter=1\textbullet\enspace Image A is provided ONLY as a reference so you can identify which content already existed in the original and which regions were changed. \textbf{Focus your judgement on the quality of the edited/newly generated regions.}\par
\noindent\hangindent=1em\hangafter=1\textbullet\enspace Do not try to infer the editing instruction; focus only on what was changed or newly generated between the two images.\par
\noindent\hangindent=1em\hangafter=1\textbullet\enspace If Image A itself is low quality, blurry, or contains defects/imperfections, and these are simply carried over (preserved) into Image B, do \textbf{NOT} penalize Image B for them. Only penalize defects that are \textbf{introduced by the editing}.\par
\noindent\hangindent=1em\hangafter=1\textbullet\enspace Do not penalize the image for being rendered in 3D.\par
\medskip\textbf{Output Format}\par
Provide a detailed, step-by-step explanation of your scoring process. Conclude clearly with the final score, formatted as follows:\par
\textbf{Final Score:} \textbf{1-5}\par
where X represents the score.\par
\end{RISEJudgeBox}
\caption{Prompt for evaluating Visual Plausibility on Spatial Reasoning tasks.}
\label{fig:judge_spatial_qual}
\end{figure*}

\begin{figure*}[!p]
\centering
\RISECenterFigureCaption
\begin{RISEJudgeBox}{16. Visual Plausibility --- Counterfactual tasks}
You are an expert image evaluator. You will receive TWO images:\par
\noindent\hangindent=1em\hangafter=1\textbullet\enspace \textbf{Image A (Original)}: the source image before editing.\par
\noindent\hangindent=1em\hangafter=1\textbullet\enspace \textbf{Image B (Edited)}: the image produced after an editing instruction was applied to Image A.\par
The edited image comes from a \textbf{counterfactual} task, where the assistant is intentionally asked to depict things that violate common sense or physical/biological laws (e.g., a hand with six fingers, a floating object, an animal with unusual anatomy). Therefore, you must judge \textbf{ONLY the perceptual quality} of \textbf{Image B}, and \textbf{NOT} its physical plausibility or real-world correctness.\par
Your task is to assign an integer score from \textbf{1 to 5} along the following dimension:\par
\medskip\textbf{Evaluation Dimension: Perceptual Quality}\par
Assess only the low-level visual quality of \textbf{Image B} (sharpness, clarity, absence of blur, noise, compression artifacts, broken textures, garbled regions, or rendering glitches), focusing on the regions that were \textbf{edited or newly generated}.\par
\textbf{Important:}\par
\noindent\hangindent=1em\hangafter=1\textbullet\enspace Image A is provided ONLY as a reference so you can identify which content already existed in the original and which regions were changed. \textbf{Focus your judgement on the quality of the edited/newly generated regions.}\par
\noindent\hangindent=1em\hangafter=1\textbullet\enspace Do not try to infer the editing instruction; focus only on what was changed or newly generated between the two images.\par
\noindent\hangindent=1em\hangafter=1\textbullet\enspace If Image A itself is low quality, blurry, or contains defects/imperfections, and these are simply carried over (preserved) into Image B, do \textbf{NOT} penalize Image B for them. Only penalize defects that are \textbf{introduced by the editing}.\par
\noindent\hangindent=1em\hangafter=1\textbullet\enspace Do \textbf{NOT} penalize the image for depicting unrealistic, impossible, or counter-intuitive objects, creatures, or anatomical structures (e.g., extra fingers, extra limbs, impossible arrangements). These are intended by the task.\par
\noindent\hangindent=1em\hangafter=1\textbullet\enspace Only judge whether the edited image itself is well rendered and visually clean.\par
\noindent\hangindent=1em\hangafter=1\textbullet\enspace Do not penalize the image for being rendered in 3D or any particular art style.\par
\textbf{Scoring Guidelines:}\par
\noindent\hangindent=1em\hangafter=1\textbullet\enspace \textbf{5} The image is sharp, clean, and visually coherent, with no noticeable blur, noise, or rendering artifacts.\par
\noindent\hangindent=1em\hangafter=1\textbullet\enspace \textbf{4} The image is clear overall; only minor artifacts or slight blur in small regions.\par
\noindent\hangindent=1em\hangafter=1\textbullet\enspace \textbf{3} The image is mostly clear, but some regions show noticeable blur, distortion, or artifacts.\par
\noindent\hangindent=1em\hangafter=1\textbullet\enspace \textbf{2} The image is noticeably blurry or contains major visual artifacts, noise, or garbled regions.\par
\noindent\hangindent=1em\hangafter=1\textbullet\enspace \textbf{1} The image is extremely blurry, noisy, or incoherent, making its content hard to recognize.\par
\medskip\textbf{Output Format}\par
Provide a detailed, step-by-step explanation of your scoring process. Conclude clearly with the final score, formatted as:\par
\textbf{Final Score:} \textbf{X}\par
\end{RISEJudgeBox}
\caption{Prompt for evaluating Visual Plausibility on Counterfactual Reasoning tasks.}
\label{fig:judge_counterfactual_qual}
\end{figure*}

\endgroup

\end{document}